\documentclass[journal]{IEEEtran}

\PassOptionsToPackage{table}{xcolor}

\usepackage{amsmath,amsfonts}
\usepackage{algorithmic}
\usepackage{array}
\usepackage[caption=false,font=normalsize,labelfont=sf,textfont=sf]{subfig}
\usepackage{textcomp}
\usepackage{stfloats}
\usepackage{url}
\usepackage{verbatim}
\usepackage{graphicx}
\usepackage{orcidlink}
\usepackage{stmaryrd}
\usepackage{makecell}
\usepackage{dsfont}
\usepackage{mathtools}
\usepackage{listings}
\usepackage{float}
\usepackage{booktabs}
\usepackage{color,soul}
\sethlcolor{gray!30}
\newfloat{prompt}{htbp}{lop}
\floatname{prompt}{Prompt}

\begin{document}

\title{Task Graph Maximum Likelihood Estimation for\\ Procedural Activity Understanding in\\ Egocentric Videos}

\author{Luigi Seminara\orcidlink{0009-0004-2242-1225}, 
Giovanni Maria Farinella\orcidlink{0000-0002-6034-0432}, 
Antonino Furnari\orcidlink{0000-0001-6911-0302}
\\ Department of Mathematics and Computer Science, University of Catania, Italy \\
  \texttt{luigi.seminara@phd.unict.it,\{giovanni.farinella,antonino.furnari\}@unict.it}
}


\maketitle

\begin{abstract}
Humans engage daily in procedural activities such as cooking a recipe or fixing a bike, which can be described as goal-oriented sequences of key-steps following certain ordering constraints.
Task graphs mined from videos or textual descriptions have recently gained popularity as a human-readable, holistic representation of procedural activities encoding a partial ordering over key-steps, and have shown promise in supporting downstream video understanding tasks.
While previous works generally relied on hand-crafted procedures to extract task graphs from videos, this paper introduces an approach based on gradient-based maximum likelihood optimization of edge weights, which can be used to directly estimate an adjacency matrix and can also be naturally plugged into more complex neural network architectures. 
We validate the ability of the proposed approach to generate accurate task graphs on the CaptainCook4D and EgoPER datasets. Moreover, we extend our validation analysis to the EgoProceL dataset, which we manually annotate with task graphs as an additional contribution. The three datasets together constitute a new benchmark for task graph learning, where our approach obtains improvements of +14.5\%, +10.2\% and +13.6\% in $F_1$ score, respectively, over previous approaches.
Thanks to the differentiability of the proposed framework, we also introduce a feature-based approach for predicting task graphs from key-step textual or video embeddings, which exhibits emerging video understanding abilities. 
Beyond that, task graphs learned with our approach obtain top performance in the Ego-Exo4D procedure understanding benchmark including $5$ different downstream tasks, with gains of up to +4.61\%, +0.10\%, +5.02\%, +8.62\%, and +15.16\% in finding Previous Keysteps, Optional Keysteps, Procedural Mistakes, Missing Keysteps, and Future Keysteps, respectively.
We finally show significant enhancements to the challenging task of online mistake detection in procedural egocentric videos, achieving notable gains of +19.8\% and +6.4\% in the Assembly101-O and EPIC-Tent-O datasets, respectively, compared to the state of the art. 
The code for replicating the experiments is available at \url{https://github.com/fpv-iplab/Differentiable-Task-Graph-Learning}.
\end{abstract}


\section{Introduction}

\IEEEPARstart{P}{rocedural} activities are essential for helping humans achieve goals, organize tasks, improve efficiency, and maintain consistency in results. However, mastering and executing procedural activities effectively often demands significant time and effort. This highlights the value of developing artificial intelligence systems capable of assisting humans in performing procedural tasks accurately~\cite{kanade2012first,plizzari2023outlook}.
Developing such capabilities requires constructing a versatile representation of a procedure that captures the partial ordering of key-steps dictated by the specific goal. For instance, a virtual assistant should recognize that breaking eggs must precede mixing them or that releasing a bike's brakes is essential before removing the bike's wheel. Crucially, to ensure scalability, representation of procedural activities should be derived automatically from observations (e.g., repeated video instances of humans following a procedure) rather than manually encoded by an expert.

\begin{figure*}[t]
    \centering
    \includegraphics[width=\linewidth]{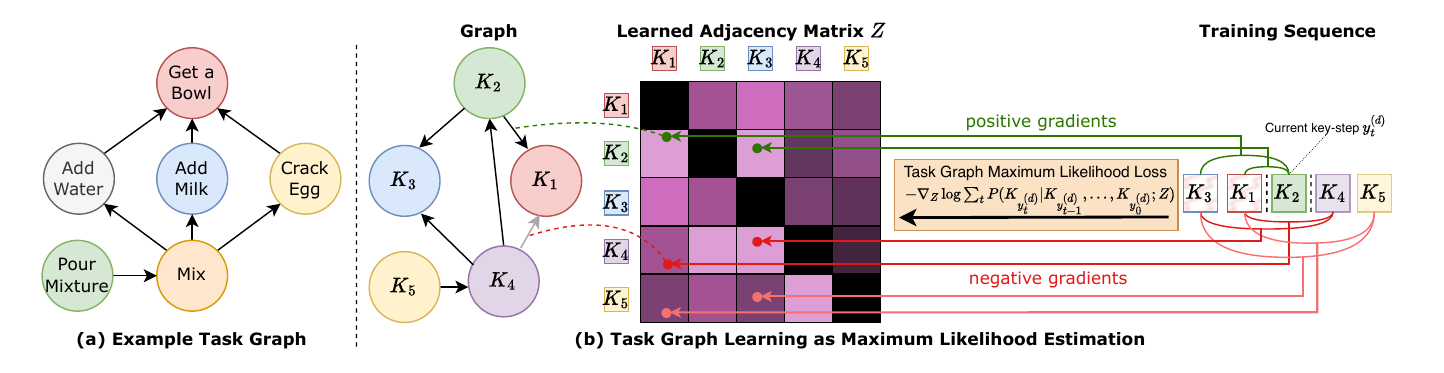}
    \caption{(a) An example task graph encoding dependencies in a ``mix eggs'' procedure. (b) We learn a task graph which encodes a partial ordering between actions (left), represented as an adjacency matrix $Z$ (center), from input action sequences (right). The proposed Task Graph Maximum Likelihood (TGML) loss directly supervises the entries of the adjacency matrix $Z$ generating gradients to maximize the probability of edges from past nodes ($K_3, K_1$) to the current node ($K_2$), while minimizing the probability of edges from past nodes to future nodes ($K_4, K_5$) in a contrastive manner.}
    \label{fig:teaser}
\end{figure*}

Toward this direction, recent works have shown that \textit{task graphs} mined from video or text can serve as a holistic representation of procedures supporting different downstream tasks, including key-step recognition and prediction~\cite{dvornik2022graph2vid,ashutosh2024video,grauman2023ego,zhou2023procedure}.
While different formulations of task graphs have been considered in past works~\cite{dvornik2022graph2vid,ashutosh2024video,grauman2023ego}, we define a task graph as a Directed Acyclic Graph (DAG)~\cite{seminara2024differentiable}, where the nodes denote key-steps, and the directed edges define a partial ordering, capturing the dependencies between these steps.
For instance, the graph in Figure~\ref{fig:teaser}(a) prescribes that ``Add Water'' depends on (and hence should be performed after) ``Get a Bowl'', that ``Add Water'', ``Add Milk'' and ``Crack Egg'' can be performed in any order, provided that ``Get a Bowl'' has been performed, and that ``Mix'' can be performed only after ``Crack Egg'', ``Add Water'', and ``Add Milk''.
Graphs provide an explicit representation which is readily interpretable by humans and easy to incorporate in downstream tasks such as detecting mistakes or validating the execution of a procedure.
Despite the potential of task graphs in procedural video understanding, current methods rely on meticulously crafted graph mining procedures rather than setting graph generation in a learning framework, limiting the inclusion of task graph learning in end-to-end systems.

This work introduces a new method for learning task graphs from demonstrations, where procedures are executed by real users and recorded as sequences of key-steps in videos. Given a task graph represented as an adjacency matrix, along with a set of key-step sequences, the proposed approach estimates the likelihood of observing the sequences under the constraints defined by the graph.
We hence formulate task graph learning under the well-understood framework of Maximum Likelihood (ML) estimation~\cite{seminara2024differentiable} and propose a novel differentiable Task Graph Maximum Likelihood (TGML) loss function which can be used to directly optimize the adjacency matrix through gradient descent. 
The resulting loss function scans each training sequence key-step by key-step, producing positive gradients to reinforce the weights of edges between the currently observed key-step and key-steps previously observed in the same sequence, while reducing the weights directly connecting future key-steps to past key-steps, bypassing the current key-step (see Figure~\ref{fig:teaser}(b)).
Based on the proposed framework, we introduce two approaches to task graph learning. The first one, called ``Direct Optimization (DO)'', directly optimizes the weights of the adjacency matrix, which serve as the sole parameters of the model. 
The second approach, referred to as ``Task Graph Transformer (TGT)'', is a feature-based model that utilizes a transformer encoder and a relation head to predict the adjacency matrix from text or video key-step embeddings. 
This method obtains competitive performance and exhibits emerging video understanding capabilities, showcasing the potential of the proposed loss to guide the optimization of end-to-end architectures.

We evaluate the abilities of the proposed models to generate accurate task graphs on the CaptainCook4D~\cite{peddi2023captaincook4d} and EgoPER~\cite{lee2024error} datasets, which contain egocentric procedural videos paired with ground truth task graphs.
Both datasets have been collected in a scripted scenario in which users were asked to follow action sequences sampled from ground truth graphs. While this approach allows obtaining video sequences aligned to ground truth task graphs, it may introduce a bias as the observed sequences are guaranteed to be a faithful representation of the graph, which is not always the case in complex, real-world videos.
To mitigate this issue, we extend the EgoProceL dataset~\cite{bansal2022my} with manually-labeled task graph annotations, which are sourced independently from the videos, by relying on annotations.
These three datasets together provide a diverse benchmark for task graph generation, on which our best approach achieves improvements of +14.5\%, +10.2\%, and +13.6\%, respectively, over previous methods.

We further assess the usefulness of the proposed representation in $6$ downstream tasks across three datasets by proposing methodologies based on task graphs. On the Ego-Exo4D~\cite{grauman2023ego} procedure understanding benchmark, our method obtains gains of up to +4.61\%, +0.10\%, +5.02\%, +8.62\%, and +15.16\% in the 5 downstream tasks of finding Previous Keysteps, Optional Keysteps, Procedural Mistakes, Missing Keysteps, and Future Keysteps, respectively.
On the online mistake detection benchmark recently introduced in~\cite{flaborea2024prego}, we obtain significant gains of +19.8\% in Assembly101~\cite{sener2022assembly101} and +6.4\% in EPIC-Tent~\cite{jang2019epic} respectively.

In sum, the contributions of this work are as follows:
1) We present a novel framework for learning \textit{task graphs} from action sequences, utilizing maximum likelihood estimation to provide a differentiable loss function that can be integrated into end-to-end models and optimized using gradient descent;
2) We propose two approaches to task graph learning: one based on direct optimization of the adjacency matrix and another one which processes key-step text or video embeddings. These approaches lead to significant improvements over previous methods in task graph generation, and demonstrate emerging video understanding capabilities;
3) To support evaluations and research on task graph generation, we contribute a new dataset based on EgoProceL and equipped with manually labeled task graphs. Differently from previous benchmarks, our task graph annotations are sourced independently from the collected video sequences, relying on annotators;
4) We assess the usefulness of the learned representations on the $5$ downstream procedural video understanding tasks included in the Ego-Exo4D procedure understanding benchmark and on the challenging online mistake detection task on the Assembly101-O and EPIC-Tent-O datasets. These experiments showcase the usefulness of task graphs in diverse downstream tasks, and, in particular, the effectiveness of the proposed graph-based representations;
5) We publicly release the code, EgoProceL annotations and all useful assets to replicate the experiments at~\url{https://github.com/fpv-iplab/Differentiable-Task-Graph-Learning}.

This work builds upon our previous conference paper~\cite{seminara2024differentiable} by extending the validation of the proposed approach to more datasets, tackling more downstream tasks, and providing task graph annotations for EgoProceL.

\section{Related Work} 
Our research is related to previous works on procedural video understanding in general and task-graph learning in particular.

\subsection{Procedural Video Understanding Tasks}
\label{sec:procedure_understanding}
Previous investigations considered different procedural video understanding tasks.
A line of work tackled the task of inferring key-steps from procedural videos relying on subtitles~\cite{zhou2018towards}, fitting individual classifiers for key-steps~\cite{zhukov2019cross}, exploiting self-supervised deep neural networks~\cite{elhamifar2020self}, modeling intra-video and inter-video frame similarities in an unsupervised way~\cite{bansal2022my}, aligning embeddings of identical key-steps~\cite{bansal2024united}, exploiting transformer-based architecture~\cite{dvornik2023stepformer}.
Other methods focused on grounding key-steps in procedural videos using attention-based methods~\cite{lu2022set} or aligning visual and textual features in narrated videos~\cite{miech2020end}. Also, task verification has been explored through learning contextualized step representations~\cite{narasimhan2023learning}, as well as through the development of benchmarks and synthetic datasets~\cite{hazra2023egotv}.
Among the other procedural video understanding tasks, mistake detection has gained increasing attention in recent years. Some methods have approached this task in fully supervised settings, where mistakes are explicitly labeled within videos and detection is performed offline~\cite{peddi2023captaincook4d,sener2022assembly101,wang2023holoassist}. Others have investigated weak supervision, where mistakes are annotated only at the video level rather than at finer spatial and temporal scales~\cite{ghoddoosian2023weakly}. A different approach~\cite{ding2023every} leverages knowledge graphs built from fine-grained spatial and temporal annotations to improve mistake detection.
To advance the field of mistake detection,~\cite{flaborea2024prego} introduced PREGO, an online mistake detection benchmark incorporating videos from the Assembly101~\cite{sener2022assembly101} and EPIC-Tent~\cite{jang2019epic} datasets. The same work proposed a novel method for detecting mistakes in procedural egocentric videos based on large language models.
Notably, these prior works have relied on diverse representations, mostly implicit (e.g., activations of neural network), and hence non-interpretable and not straightforward to generalize across different tasks.

Recently, task graphs, mined from video or external knowledge such as WikiHow articles, have been investigated as a powerful representation of procedures and proved advantageous for learning representations useful for downstream tasks such as key-step recognition and forecasting~\cite{ashutosh2024video,grauman2023ego,zhou2023procedure}, temporal action segmentation~\cite{nagasinghe2024not}, and procedure planning~\cite{shen2024progress}.
Differently from previous works~\cite{narasimhan2023learning,zhong2023learning}, we aim to develop an explicit and human readable representation of the procedure which can be directly exploited to enable downstream tasks~\cite{ashutosh2024video}, rather than an implicit representation obtained with pre-training objective~\cite{zhou2023procedure,narasimhan2023learning}. 
As a departure from previous paradigms which carefully designed task graph construction procedures~\cite{ashutosh2024video,zhou2023procedure,sohn2020meta,jang2023multimodal}, we frame task generation in a general framework, enabling models to learn task graphs directly from input sequences, and propose a differentiable loss function based on maximum likelihood estimation.

\subsection{Task Graph Learning}
\label{sec:task_graph_construction}
Graph-based representations have been historically used to
represent constraints in complex tasks and design optimal sub-tasks scheduling~\cite{skiena1998algorithm}, making them a natural candidate to encode procedural knowledge.
Previous works investigated approaches to construct task graphs from natural language descriptions of procedures (e.g., recipes) using rule-based graph parsing~\cite{dvornik2022graph2vid,schumacher2012extraction}, defining probabilistic models~\cite{kiddon2015mise}, fine-tuning language models~\cite{sakaguchi2021proscript}, or proposing learning-based approaches~\cite{dvornik2022graph2vid} involving parsers and taggers trained on text corpora~\cite{donatelli2021aligning,yamakata2020english}. While these approaches do not require any action sequence as input, they depend on the availability of text corpora including procedural knowledge, such as recipes, which often fail to encapsulate the variety of ways in which the procedure may be executed~\cite{ashutosh2024video}. 
Other works proposed hand-crafted approaches to infer task graphs from sequences of actions depicting task executions~\cite{sohn2020meta,jang2023multimodal}. 
Recent work designed methodologies to mine task graphs from videos and textual descriptions of key-steps~\cite{ashutosh2024video} or cross-referencing visual and textual representations from corpora of procedural text and videos~\cite{zhou2023procedure}.

Differently from previous efforts, we rely on action sequences, grounded in video, rather than natural language descriptions of procedures~\cite{dvornik2022graph2vid,sakaguchi2021proscript} and frame task graph construction as a learning problem, providing a differentiable objective rather than resorting to hand-designed algorithms and task extraction procedures~\cite{ashutosh2024video,zhou2023procedure,sohn2020meta,jang2023multimodal}.

\section{Technical Approach}
\label{sec:tech}
In this section, we present the proposed Task Graph Maximum Likelihood (TGML) framework (Section~\ref{sec:tgml_framework}), the models to learn task graphs based on this framework (Section~\ref{sec:models}), the pre-processing of the input sequences to train the models (Section~\ref{sec:sequences}), the masking strategy applied during the training of the models (Section~\ref{sec:masking}), and the post-processing procedures required to obtain the final graphs from the predicted adjacency matrices (Section~\ref{sec:postprocessing}). More details are reported in the section \textit{Implementation Details} of the supplementary materials.

\subsection{Task Graph Maximum Likelihood Learning Framework}
\label{sec:tgml_framework}
We will first discuss preliminaries and notation (Section~\ref{sec:preliminaries}), then describe how to model the likelihood of a sequence in the simple case of an unweighted graph (Section~\ref{sec:unweighted}) and in the more general case of a weighted graph (Section~\ref{sec:weighted}). We finally derive the proposed loss function in Section~\ref{sec:loss}.

\subsubsection{Preliminaries and notation}
\label{sec:preliminaries}
Let 
\begin{equation}
  \mathcal{K}=\{K_0=S,K_1,\ldots,K_n,K_{n+1}=E\} 
\end{equation}
be the set of key-steps involved in the procedure, where $n$ is the number of key-steps, and symbols $S$ and $E$ are placeholder ``start'' and ``end'' key-steps denoting the \textit{start} and \textit{end} of the procedure. We define the task graph as a weighted directed acyclic graph, i.e., a tuple $G=(\mathcal{K},\mathcal{A},\omega)$, where $\mathcal{K}$ is the set of nodes (the key-steps), $\mathcal{A} = \mathcal{K} \times\mathcal{K}$ is the set of possible directed edges indicating ordering constraints between pairs of key-steps, and $\omega : \mathcal{A} \to [0,1]$ is a function assigning a score to each of the edges in $\mathcal{A}$. 
An edge $(K_i,K_j) \in \mathcal{A}$ (also denoted as $K_i \to K_j$) indicates that $K_j$ is a \textit{pre-condition} of $K_i$ (for instance ``Mix'' $\to$ ``Crack Egg'') with score $\omega(K_i,K_j)$. We assume normalized weights for outgoing edges, i.e., $\sum_j w(K_i,K_j)=1, \forall i$.
We represent the graph $G$ as the adjacency matrix $Z \in [0,1]^{(n+2) \times (n+2)}$, where $Z_{(i,j)} = \omega(K_i,K_j)$. For ease of notation, we will denote the graph $G=(\mathcal{K},\mathcal{A},\omega)$ simply with its adjacency matrix $Z$ in the rest of the paper. 
We assume that a set of $D$ sequences $\mathcal{Y}=\{y^{(d)}\}_{d=1}^D$ showing possible orderings of the key-steps in $\mathcal{K}$ is available, where the generic sequence $y^{(d)} \in \mathcal{Y}$ is defined as a set of indexes to key-steps $\mathcal{K}$, i.e., 
\begin{equation}
y^{(d)} = <y^{(d)}_0,\ldots,y^{(d)}_t,\ldots,y^{(d)}_{m+1}>, \ \ y^{(d)}_t \in \{0,\ldots,n+1\}
\end{equation}
We further assume that each sequence starts with key-step $S$ and ends with key-step $E$, i.e., $y^{(d)}_0=0$ and $y^{(d)}_{m+1}=n+1$\footnote{In practice, we prepend/append $S$ and $E$ to each sequence.} and note that different sequences $y^{(i)}$ and $y^{(j)}$ have in general different lengths. Since we are interested in modeling key-step orderings, we assume that sequences do not contain repetitions (see Section~\ref{sec:sequences} for details). We frame task graph learning as determining an adjacency matrix $\hat Z$ such that sequences in $\mathcal{Y}$ are topological sorts of $\hat Z$ with high probability. A principled way to approach this problem is to provide an estimate of the likelihood $P(\mathcal{Y}|Z)$ and choose the maximum likelihood estimate 
\begin{equation}
\hat Z = \underset{Z}{\arg\max} \text{ } P(\mathcal{Y}|Z).
\end{equation}

\subsubsection{Modeling Sequence Likelihood for an Unweighted Graph}
\label{sec:unweighted}
Let us consider the special case of an unweighted graph, i.e., $\bar Z \in \{0,1\}^{(n+2) \times (n+2)}$. We wish to estimate $P(y^{(d)}|\bar Z)$, the likelihood of the generic sequence $y^{(d)} \in \mathcal{Y}$ given graph $\bar Z$.
Formally, let $Y_t$ be the random variable related to the event ``key-step $K_{y^{(d)}_t}$ appears at position $t$ in sequence $y^{(d)}$''. We can factorize the conditional probability $P(y^{(d)}|\bar Z)$ as:
\begin{equation}
\begin{aligned}
P(y^{(d)}|\bar Z) &= P(Y_0,\ldots,Y_{|y^{(d)}|}|\bar Z) \\
&= P(Y_0|\bar Z) \cdot P(Y_1|Y_0,\bar Z) \cdot \ldots \cdot \\
&\cdot \ldots \cdot P(Y_{|y^{(d)}|}|Y_0,\ldots,Y_{|y^{(d)}|-1},\bar Z).
\end{aligned}
\label{eq:factorization}
\end{equation}
We assume that the probability of observing a given key-step $K_{y^{(d)}_t}$ at position $t$ in $y^{(d)}$ depends on the previously observed key-steps ($K_{y^{(d)}_0},\ldots,K_{y^{(d)}_{t-1}}$), but not on their ordering, i.e., the probability of observing a given key-step depends on whether its pre-conditions are satisfied, regardless of the order in which they have been satisfied. 
Under this assumption, we write $P(Y_t |Y_0, \dots, Y_{t-1},\bar Z)$ simply as $P(K_{y^{(d)}_t} | K_{y^{(d)}_0}, \dots, K_{y^{(d)}_{t-1}},\bar Z)$. 
Without loss of generality, in the following, we denote the current key-step as $K_i=K_{y^{(d)}_t}$, the indexes of key-steps \textit{observed} at time $t$ as $\mathcal{J}^{(d)}_t = \{y^{(d)}_0, \dots, y^{(d)}_{t-1}\}$, and the corresponding set of observed key-steps as $K_{\mathcal{J}^{(d)}_t} = \{K_x|x\in \mathcal{J}^{(d)}_t\}$.
Similarly, we define $\bar{\mathcal{J}}^{(d)}_t = \{0,\ldots,n+1\} \setminus \mathcal{J}^{(d)}_t$ and $K_{\bar{\mathcal{J}}^{(d)}_t}$ as the sets of indexes and corresponding key-steps \textit{unobserved} at position $t$, i.e., those which do not appear before $y^{(d)}_t$ in the sequence.
Given the factorization above, we are hence interested in estimating the general term: 
\begin{equation}
P(K_{y^{(d)}_t} | K_{y^{(d)}_0}, \dots, K_{y^{(d)}_{t-1}}, \bar Z) = P(K_i|K_{\mathcal{J}^{(d)}_t}, \bar Z).
\end{equation}
We can estimate the probability of observing key-step $K_i$ given the set of observed key-steps $K_{\mathcal{J}^{(d)}_t}$ and the constraints imposed by $\bar Z$, following Laplace's classic definition of probability~\cite{marquis1820theorie} as ``the ratio of the number of favorable cases to the number of possible cases''. 
Specifically, if we were to randomly sample a key-step from $\mathcal{K}$ following the constraints of $\bar Z$, and having observed key-steps $K_{\mathcal{J}^{(d)}_t}$, sampling $K_i$ would be a favorable case if all pre-conditions of $K_i$ were satisfied, i.e., if $\sum_{j \in \bar{\mathcal{J}}^{(d)}_t} \bar Z_{(i,j)}=0$ (there are no pre-conditions in unobserved key-steps $K_{\bar{\mathcal{J}}^{(d)}_t}$).
Similarly, sampling a key-steps $K_h$ is a ``possible case'' if $\sum_{j \in \bar{\mathcal{J}}^{(d)}_t} \bar Z_{(h,j)} = 0$. We can hence define the probability of observing key-step $K_i$ after observing all key-steps $K_{\mathcal{J}^{(d)}_t}$ in a sequence as follows:
\begin{equation}
    \begin{aligned}
        P(K_i|K_{\mathcal{J}^{(d)}_t},\bar Z) &= \frac{\text{number of favorable cases}}{\text{number of possible cases}} = \\
        &= \frac{\mathds{1}(\sum_{j \in \bar{\mathcal{J}}^{(d)}_t} \bar Z_{(i,j)} = 0)}{\sum_{h \in \bar{\mathcal{J}}^{(d)}_t} \mathds{1}(\sum_{j \in \bar{\mathcal{J}}^{(d)}_t} \bar Z_{(h,j)} = 0)}.
    \end{aligned}
    \label{eq:binary_factor}
\end{equation}
where $\mathds{1}(\cdot)$ denotes the indicator function, and in the denominator we are counting the number of key-steps that have not appeared yet and hence are considered as ``possible cases'' under the given graph $\bar Z$. The likelihood $P(y^{(d)}|\bar Z)$ can be obtained by plugging Eq.~\eqref{eq:binary_factor} into Eq.~\eqref{eq:factorization}.

\subsubsection{Modeling Sequence Likelihood for a Weighted Graph}
\label{sec:weighted}
To enable gradient-based learning, we consider the general case of a continuous adjacency matrix $Z \in [0,1]^{(n+2) \times (n+2)}$. 
We generalize the concept of ``possible cases'' discussed in the previous section with the concept of ``feasibility of sampling a given key-step $K_i$, having observed a set of key-steps $K_{\mathcal{J}^{(d)}_t}$, given graph $Z$'', which we define as the sum of all weights of edges between observed key-steps 
$K_{\mathcal{J}^{(d)}_t}$ and $K_i$: 
\begin{equation}
 f(K_i|K_{\mathcal{J}^{(d)}_t},Z) = \sum_{j \in \mathcal{J}^{(d)}_t} Z_{(i,j)}.   
\end{equation}
Intuitively, if key-step $K_i$ has many satisfied pre-conditions, we are more likely to sample it as the next key-step. We hence define $P(K_i|K_{\mathcal{J}^{(d)}_t},Z)$ as ``the ratio of the feasibility of sampling $K_i$ to the sum of the feasibilities of sampling any unobserved key-step'':
\begin{equation}
    \begin{aligned}
        P(K_i|K_{\mathcal{J}^{(d)}_t},Z) &= \frac{f(K_i | K_{\mathcal{J}^{(d)}_t},Z)}{\sum_{h \in \bar{\mathcal{J}}^{(d)}_t} f(K_{h}|K_{\mathcal{J}^{(d)}_t},Z)} = \\ 
        &= \frac{\sum_{j \in \mathcal{J}^{(d)}_t}Z_{(i,j)}}{\sum_{h \in \bar{\mathcal{J}}^{(d)}_t} \sum_{j \in \mathcal{J}^{(d)}_t} Z_{(h,j)}}.
        \label{eq:factor}
    \end{aligned}
\end{equation}

\begin{figure*}
    \centering
    \includegraphics[width=\linewidth]{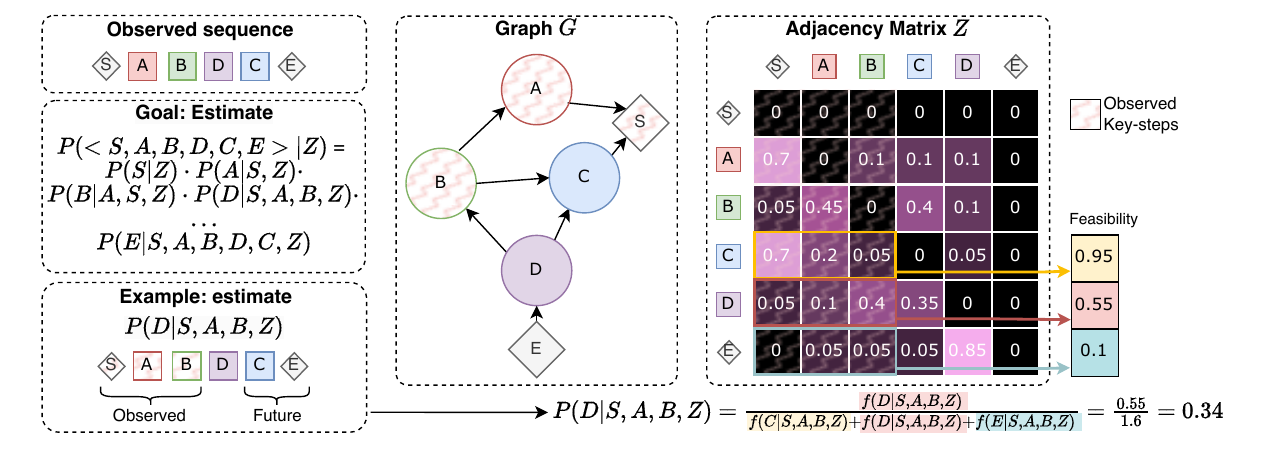}
    \caption{Given a sequence $<S,A,B,D,C,E>$, and a graph $G$ with adjacency matrix $Z$, our goal is to estimate the likelihood $P(<S,A,B,D,C,E>|Z)$, which can be done by factorizing the expression into simpler terms. The figure shows an example of computation of probability $P(D|S,A,B,Z)$ as the ratio of the ``feasibility of sampling key-step D, having observed key-steps S, A, and B'' to the sum of all feasibility scores for unobserved symbols. Feasibility values are computed by summing weights of edges $D \to X$ for all observed key-steps $X$.}
    \label{fig:prob}
\end{figure*}

Figure~\ref{fig:prob} illustrates the computation of the likelihood in Eq.~\eqref{eq:factor}.
Plugging Eq.~\eqref{eq:factor} into Eq.~\eqref{eq:factorization}, we can estimate the likelihood of a sequence $y^{(d)}$ given graph $Z$ as:
\begin{equation}
    \begin{aligned}
        P(y^{(d)}|Z) &= P(S|Z)\prod_{t = 1}^{|y^{(d)}|}{P(K_{y^{(d)}_t}|K_{\mathcal{J}^{(d)}_t},Z)} = \\
        &= \prod_{t = 1}^{|y^{(d)}|} \frac{\sum_{j \in \mathcal{J}^{(d)}_t} Z_{(y^{(d)}_{t},j)}}{\sum_{h \in \bar{\mathcal{J}}^{(d)}_t} \sum_{j \in \mathcal{J}^{(d)}_t} Z_{(h,j)}}.
        \label{eq:factorization_2}
    \end{aligned}
\end{equation}
Where we set $P(K_{y^{(d)}_0}|Z)=P(S|Z)=1$ as sequences always start with the start node $S$.

\subsubsection{Task Graph Maximum Likelihood Loss Function}
\label{sec:loss}
Assuming that sequences $y^{(d)} \in \mathcal{Y}$ are independent and identically distributed, we define the likelihood of $\mathcal{Y}$ given graph $Z$ as follows:
\begin{equation}
    \begin{aligned}
        P(\mathcal{Y}| Z) &= \prod_{d=1}^{|\mathcal{Y}|}{P(y^{(d)}|Z)} = \\
        &= \prod_{d=1}^{|\mathcal{Y}|}\prod_{t = 1}^{|y^{(d)}|}\frac{\sum_{j \in \mathcal{J}^{(d)}_t}Z_{(y^{(d)}_t, j)}}{\sum_{h \in \bar{\mathcal{J}}^{(d)}_t} \sum_{j \in \mathcal{J}^{(d)}_t} Z_{(h,j)}}.
        \label{eq:likelihood}
    \end{aligned}
\end{equation}
We can find the optimal graph $Z$ by maximizing the likelihood in Eq.~\eqref{eq:likelihood}, which is equivalent to minimizing the negative log-likelihood $-\log P(\mathcal{Y},Z)$, leading to the following loss:
\begin{align}
    \mathcal{L}(\mathcal{Y},Z) =  - \sum_{d = 1}^{|Y|} \sum_{t = 1}^{|y^{(d)}|} \big(\textcolor{cyan}{\log{\sum_{\mathclap{j \in \mathcal{J}^{(d)}_t}}Z_{(y^{(d)}_t,j)}}} 
    - \beta \cdot \textcolor{teal}{\log{\sum_{\mathclap{\substack{h \in \bar{\mathcal{J}}^{(d)}_t\\{j \in \mathcal{J}^{(d)}_t}}}}  Z_{(h,j)}}} \big).
    \label{eq:loss}
\end{align}
where $\beta$ is a hyper-parameter. We refer to Eq.~\eqref{eq:loss} as the \textit{Task Graph Maximum Likelihood (TGML)} loss function. Since Eq.~\eqref{eq:loss} is differentiable with respect to all $Z_{(i,j)}$ values, we can learn the adjacency matrix $Z$ by minimizing the loss with gradient descent to find the estimated graph $\hat Z = \arg_Z\max \mathcal{L}(\mathcal{Y},Z)$. 
As illustrated in Figure~\ref{fig:teaser}(b), Eq.~\eqref{eq:loss} works as a contrastive loss in which the \textcolor{cyan}{first logarithmic term} aims to \textit{maximize}, at every step $t$ of each input sequence, the weights $Z_{(y^{(d)}_t,j)}$ of edges $K_{y^{(d)}_t} \to K_j$ going from the current key-step $K_{y^{(d)}_t}$ to all previously observed key-steps $K_j$, while the \textcolor{teal}{second logarithmic term (contrastive term)} aims to \textit{minimize} the weights of edges $K_h \to K_j$ between key-steps yet to appear $K_h$ and already observed key-steps $K_j$.
Intuitively, this encourages future steps to be independent of previous steps or depend on them only through the current step.
The hyper-parameter $\beta$ regulates the influence of the summation in the \textcolor{teal}{contrastive term} which, including many more addends, can dominate gradient updates. As in other contrastive learning frameworks~\cite{oord2018representation, radford2021learning}, our approach only includes positives and negatives and it does not explicitly consider anchor examples.

\begin{figure*}
    \centering
    \includegraphics[width=\linewidth]{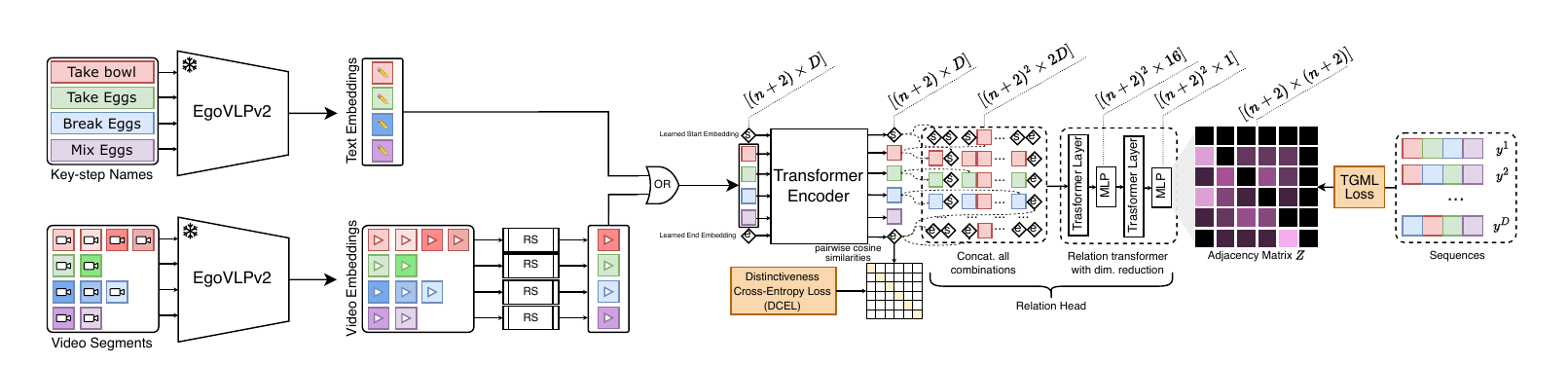}
    \caption{Our Task Graph Transformer (TGT) takes as input either $D$-dimensional text embeddings extracted from key-step names or video embeddings extracted from key-step segments. In both cases, we extract features with a pre-trained EgoVLPv2 model. For video embeddings, multiple embeddings can refer to the same action, so we randomly select one for each key-step (RS blocks). Learnable start (S) and end (E) embeddings are also included. Key-step embeddings are processed using a transformer encoder and regularized with a distinctiveness cross-entropy loss (DCEL) to prevent representation collapse. The output embeddings are processed by our relation head, which concatenates vectors across all $(n + 2)^2$ possible node pairs, producing $(n + 2) \times (n + 2) \times 2D$ relation vectors. These vectors are then processed by a relation transformer, which progressively maps them to an $(n + 2) \times (n + 2)$ adjacency matrix. The model is supervised with input sequences using our proposed Task Graph Maximum Likelihood (TGML) loss.}
    \label{fig:architecture}
\end{figure*}

\subsection{Models}
\label{sec:models}
We propose two models based on the TGML loss function: a ``Direct Optimization'' model, which performs gradient descent directly in graph solution space of the adjacency matrices (Section~\ref{sec:model_do}), and an architecture based on transformers which can predict graphs from video or text embeddings describing key-steps (Section~\ref{sec:model_tgt}).

\subsubsection{Direct Optimization (DO)}
\label{sec:model_do}
The first model aims to directly optimize the parameters of the adjacency matrix by performing gradient descent on the TGML loss (Eq.~\eqref{eq:loss}).
We define the parameters of this model as an edge scoring matrix $A \in \mathbb{R}^{(n+2) \times (n+2)}$, where $n$ is the number of key-steps, plus the placeholder start ($K_0 = S$) and end ($K_{n+1} = E$) nodes, and $A_{(i,j)}$ is a score assigned to edge $K_i \rightarrow K_j$.
To prevent the model from learning edge weights eluding the assumptions of directed acyclic graphs, we mask black cells in Figure~\ref{fig:prob} with $-\infty$ (see Section~\ref{sec:masking} for details).
To obtain the final adjacency matrix $Z$ in the $[0,1]$ range, which represents the predicted task graph, we softmax-normalize the rows of the scoring matrix $A$, i.e., $Z = softmax(A)$. Note that elements masked with $-\infty$ will be automatically mapped to $0$ by the softmax function similarly to~\cite{vaswani2017attention}. We train this model by performing batch gradient descent directly on the score matrix $A$ with the proposed TGML loss. We train a separate model per procedure, as each procedure is associated to a different task graph. %

\subsubsection{Task Graph Transformer (TGT)} 
\label{sec:model_tgt}
Thanks to the differentiable nature of the proposed loss function, we can use it to guide learning of more complex, differentiable architectures. To this aim, we also introduce a transformer-based model which can generate graphs starting from video or text embeddings describing key-steps.
Figure~\ref{fig:architecture} illustrates the proposed model, which is termed Task Graph Transformer (TGT). The proposed model can take as input either $D$-dimensional embeddings of textual descriptions of key-steps or $D$-dimensional video embeddings of key-step segments extracted from video. In the first case, the model takes as input the same set of embeddings at each forward pass, while in the second case, at each forward pass, we randomly sample a video embedding per key-step from the training videos (hence each key-step embedding can be sampled from a different video). We also include two $D$-dimensional learnable embeddings for the $S$ and $E$ nodes. 
All key-step embeddings are processed by a transformer encoder, which outputs $D$-dimensional vectors enriched with information from other embeddings. To prevent representation collapse, we apply a distinctiveness cross-entropy loss (DCEL) encouraging distinctiveness between pairs of different nodes. Let $X$ be the matrix of embeddings produced by the transformer model. 
We L2-normalize features, then compute pairwise cosine similarities $Y = X \cdot X^T \cdot \exp(T)$ as in~\cite{radford2021learning}.
We hence enforce the values outside the diagonal of $Y$ to be smaller than the values in the diagonal by encouraging each row of the matrix $Y$ to be close to a one-hot vector with a cross-entropy loss.
This leads to key-step self-similarities being larger than similarities across key-steps, preventing representation collapse. 
Regularized embeddings are finally passed through a relation transformer head which considers all possible pairs of embeddings and concatenates them in a $(n+2) \times (n+2) \times 2D$ matrix $R$ of relation vectors. For instance, $R[i,j]$ is the concatenation of vectors $X[i]$ and $X[j]$. Relation vectors are passed to a transformer layer which aims to mine relationships among relation vectors, followed by a multilayer perceptron to reduce dimensionality to $16$ units and another pair of transformer layer and multilayer perceptron to map relation vectors to scalar values, which are reshaped to size $(n+2) \times (n+2)$ to form the scoring matrix $A$. We hence softmax-normalize the rows of the scoring matrix $A$, i.e., $Z = softmax(A)$, to obtain the final adjacency matrix representing the predicted task graph.

\subsection{Input Sequence Pre-Processing}
\label{sec:sequences}
Our framework treats key-step sequences as topological sorts of task graphs, which are by definition sequences without repetitions. However, real-world sequences may include repetitions, necessitating specific approaches to handle such cases effectively. Depending on the characteristics of the data, we employ one of the following two approaches to map sequences with repetitions to sequences without repetitions.
\subsubsection{Removing Repeated Key-Steps} In this approach, we retain only the first occurrence of each key-step and eliminate subsequent repetitions. For instance, the sequence $BACAD$ is mapped to $BACD$. The rationale behind this mapping is that repeated occurrences of a key-step do not alter the dependencies established by earlier steps. If we interpret the key-steps as procedural actions, for instance, $A$ as ``Break Egg'' and $B$ as ``Get Bowl'', the second occurrence of ``Break Egg'' ($A$) represents a repetition of the same action, which could occur at any point after ``Get Bowl'' ($B$) is completed. Thus, subsequent repetitions can be safely ignored for topological reasoning. We apply this approach when action sequences are compatible with the dependencies dictated by the ground-truth task graph or when a validation set is available.
\subsubsection{Mapping Multiple Non-Repetitive Sequences} This approach generates multiple sequences by considering all possible orderings of the key-steps, excluding repetitions. For instance, the sequence $BACAD$ would be mapped to $BACD$ and $BCAD$. This method is particularly useful when key-steps can be performed in parallel. For instance, if $A$ is ``Add Milk'', and $C$ is ``Add Water'', these actions can be executed in parallel during the preparation of a cake. By considering multiple non-repetitive sequences, we better capture the flexibility inherent in such parallelizable key-steps. We apply this approach when the ground-truth task graph is unknown and a validation set is not available.

\subsection{Masking Strategy for Directed Acyclic Graphs}
\label{sec:masking}
To ensure that the model complies with the structural constraints of directed acyclic graphs (DAGs), we implement a masking strategy that assigns $-\infty$ to specific elements of the adjacency matrix. The masked elements include: (1) the main diagonal, since no node can have an edge to itself; (2) the row corresponding to the START node, as it has no pre-conditions by definition; and (3) the column corresponding to the END node, as it cannot serve as a pre-condition by definition. This masking strategy effectively prevents the model from learning some edge weights that violate the acyclic properties of the graph. The black cells in Figure~\ref{fig:prob} visually represent the masked regions.

\begin{figure}[t]
    \centering
    \includegraphics[width=0.7\linewidth]{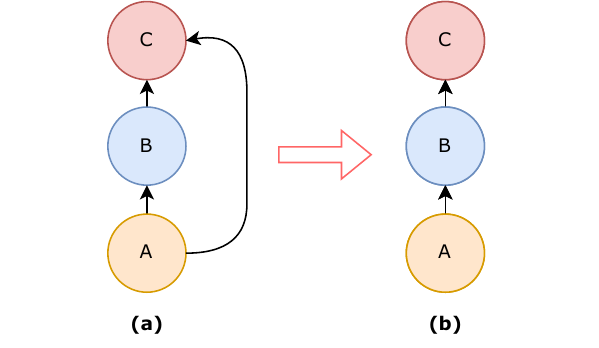}
    \caption{An example of transitive dependency between nodes. In (a) node A depends on B and C, but B depends on C, in this case, we can remove the edge between A and C for transitivity and we obtain the graph in (b).}
    \label{fig:transitive_dependency}
\end{figure}

\subsection{Post-processing of the Output Graph}
\label{sec:postprocessing}
As many applications require an unweighted graph, we binarize the adjacency matrix with the threshold $\frac{1}{n}$, where $n$ is the number of key-steps of the considered procedure. After this thresholding phase, it is possible to encounter situations like the one illustrated in Figure~\ref{fig:transitive_dependency}, where node $A$ depends on nodes $B$ and $C$, and node $B$ depends on node $C$. Due to the transitivity of the pre-conditions, we can remove the edge connecting node $A$ to node $C$, as node $B$ must precede node $A$. Sometimes, it may occur that a node does not serve as a pre-condition for any other node; in such cases, the END node should be directly connected to this node. Conversely, if a node has no pre-conditions, an edge is added from the current node to the START node. 
At the end of the training process, obtaining a graph containing cycles is also possible. In such cases, all cycles within the graph are considered, and the edge with the lowest score within each cycle is removed. This process ensures that the graph remains a Directed Acyclic Graph (DAG). These steps yield the final binary, unweighted task graph $\hat G = (\hat{\mathcal{K}}, \hat{\mathcal{A}})$.

\section{Experiments and Results}
\label{sec:experiments}
In this section, we first introduce our human-annotated task graphs for EgoProceL (see Section~\ref{sec:egoprocel}). Next, we evaluate our models' ability to generate task graphs (Section~\ref{sec:graph_generation_results}) and explore how our TGT model exhibits emerging video understanding abilities (Section~\ref{sec:video_understanding_results}). We further assess the usefulness of the learned representation on the 5 downstream tasks of the Ego-Exo4D procedure understanding benchmark (Section~\ref{sec:pub}) and the online mistake detection task (Section~\ref{sec:omd}). Finally, Section~\ref{sec:ablation} reports ablation studies.

\begin{figure}
    \centering
    \includegraphics[width=1.0\linewidth]{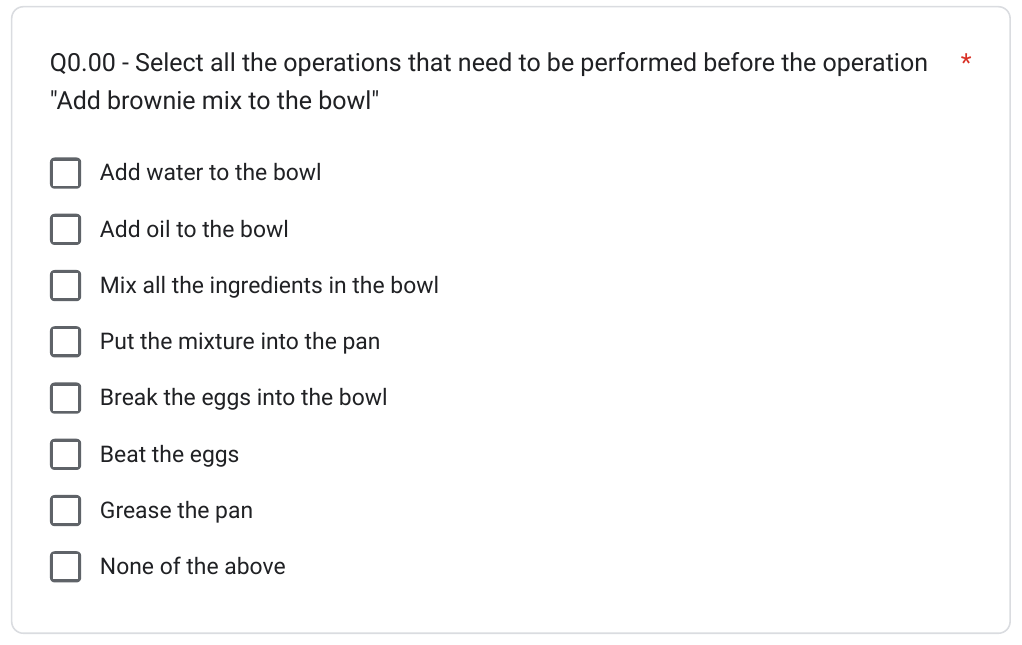}
    \caption{Example of a questionnaire item. Annotators can select multiple options. If annotators determine that a key-step has no pre-conditions, they were instructed to select ``None of the above''.}
    \label{fig:questionnaire}
\end{figure}

\subsection{Human-Annotated Task Graphs for EgoProceL}
\label{sec:egoprocel}
To support our evaluations, we present newly curated human-annotated task graphs for EgoProceL~\cite{bansal2022my} to advance research and evaluation in task graph generation. In contrast to previous publicly available task graph annotations, such as CaptainCook4D~\cite{peddi2023captaincook4d} and EgoPER~\cite{lee2024error}, our annotations are independently derived without direct reference to the video sequences to reduce bias in human-driven labeling. The EgoProceL dataset includes videos and key-step annotations for a diverse set of tasks derived from CMU-MMAC~\cite{de2009guide}, EGTEA Gaze+\cite{li2018eye}, EPIC-Tent\cite{jang2019epic}, MECCANO~\cite{ragusa2021meccano}, as well as PC assembly, and PC disassembly sequences. For our study, we focus specifically on tasks from CMU-MMAC, EGTEA Gaze+, and EPIC-Tent. 
To generate the annotations, we engaged 10 annotators to complete a structured questionnaire (Figure~\ref{fig:questionnaire}). The questionnaire was designed to enable annotators to identify the pre-condition relationships for each key-step without exposing them to video content, ensuring unbiased responses. Annotators were instructed to ensure consistency in their answers. For instance, if step $A$ is marked as a pre-condition for step $B$, step $B$ cannot simultaneously be a pre-condition for step $A$. To enforce consistency, we developed an automated system that analyzed the responses for contradictions. If inconsistencies were detected, the system generated a report highlighting the discrepancies and provided a link to the annotators for them to revise their answers. After all participants submitted their responses, the pre-conditions were finalized based on majority voting, retaining only those relationships with a frequency exceeding a threshold of $0.5$. Also, the resulting graphs were manually reviewed to ensure the absence of cycles, guaranteeing that the extracted dependencies formed a valid directed acyclic graph (DAG).
The resulting task graph annotations are publicly available and can be accessed at \url{https://github.com/fpv-iplab/Differentiable-Task-Graph-Learning}. The dataset includes 13 procedures (i.e., 13 task graphs), encompassing a total of 275 videos and 16.3 hours of video segments. These annotations are provided to facilitate further research and benchmarking in task graph generation\footnote{See section \textit{Qualitative Examples} of the supplementary material for more details.}.

\begin{table*}[t]
  \centering
  \begin{minipage}[t]{0.48\textwidth}
    \centering
    \caption{Task graph generation results on CaptainCook4D. Best results are in \textbf{bold}, second best results are \underline{underlined}, best results among competitors are \hl{highlighted}. Confidence interval bounds computed at $90\%$ conf. for $5$ runs.}
    \label{tab:captaincook}
    \resizebox{\textwidth}{!}{
      \begin{tabular}{llll}
        \toprule
        Method         & Precision     & Recall     & F$_1$          \\
        \midrule
        MSGI \cite{sohn2020meta}                    & 11.3 & 13.3 & 12.2 \\
        Llama-3.1-405B-Instruct \cite{dubey2024llama}                 & 53.0 & 57.4 & 54.9\\
        Count-Based \cite{ashutosh2024video}             & 66.0 & 55.4 & 60.2 \\
        MSG\(^2\) \cite{jang2023multimodal}               & \hl{73.3} & \hl{73.6} & \hl{73.3} \\
        \rowcolor{teal!30}
        {TGT-text (Ours)}               & {\underline{79.9} \scriptsize \(\pm 8.8\)}& {\underline{81.9} \scriptsize \(\pm 6.9\)} & {\underline{80.8} \scriptsize \(\pm 8.0\)}\\
        \rowcolor{teal!30}
        {DO (Ours)}        & {\textbf{86.4} \scriptsize $\pm 1.5$ }& {\textbf{89.7} \scriptsize \(\pm 1.5\) }& {\textbf{87.8} \scriptsize \(\pm 1.5\) }\\
        \rowcolor{green!30}
        Improvement & +13.1 & +16.1 & +14.5 \\
        \bottomrule
      \end{tabular}
    }
  \end{minipage}%
  \hspace{0.02\textwidth}
  \begin{minipage}[t]{0.485\textwidth}
    \centering
    \caption{Task graph generation results on EgoPER. Best results are in \textbf{bold}, second best results are \underline{underlined}, best results among competitors are \hl{highlighted}. Confidence interval bounds computed at $90\%$ conf. for $5$ runs.}
    \label{tab:egoper}
    \resizebox{\textwidth}{!}{
      \begin{tabular}{llll}
        \toprule
        Method         & Precision     & Recall     & F$_1$          \\
        \midrule
        MSGI \cite{sohn2020meta}                    & 48.0 & 54.0 & 50.6 \\
        Llama-3.1-405B-Instruct \cite{dubey2024llama}                 & 47.4 & 54.5 & 50.6\\
        Count-Based \cite{ashutosh2024video}             & \hl{82.5} & \hl{79.5} & \hl{80.8} \\
        MSG\(^2\) \cite{jang2023multimodal}               & 65.0 & 70.5 & 67.5 \\
        \rowcolor{teal!30}
        {TGT-text (Ours)}               & {\underline{82.6} \scriptsize \(\pm 10.3\)}& {\underline{87.7} \scriptsize \(\pm 7.0\)} & {\underline{85.0} \scriptsize \(\pm 8.8\)}\\
        \rowcolor{teal!30}
        {DO (Ours)}        & {\textbf{88.8} \scriptsize $\pm 2.2$ }& {\textbf{93.5} \scriptsize \(\pm 2.0\) }& {\textbf{91.0} \scriptsize \(\pm 2.1\) }\\
        \rowcolor{green!30}
        Improvement & +6.3 & +14.0 & +10.2 \\
        \bottomrule
      \end{tabular}
    }
  \end{minipage}%
\end{table*}

\begin{table}[t]
  \centering
    \caption{Task graph generation results on EgoProceL. Best results are in \textbf{bold}, second best results are \underline{underlined}, best results among competitors are \hl{highlighted}. Confidence interval bounds computed at $90\%$ conf. for $5$ runs.}
    \label{tab:egoprocel}
      \begin{tabular}{llll}
        \toprule
        Method         & Precision     & Recall     & F$_1$          \\
        \midrule
        MSGI \cite{sohn2020meta}                    & 23.4 & 22.9 & 22.9 \\
        Llama-3.1-405B-Instruct \cite{dubey2024llama}                 & \hl{61.8} & \hl{56.6} & \hl{58.7}\\
        Count-Based \cite{ashutosh2024video}             & 56.5 & 44.4 & 49.5 \\
        MSG\(^2\) \cite{jang2023multimodal}               & 55.3 & 56.0 & 55.4 \\
        \rowcolor{teal!30}
        {TGT-text (Ours)}               & {\underline{67.5} \scriptsize \(\pm 2.6\)}& {\underline{66.6} \scriptsize \(\pm 3.5\)} & {\underline{66.9} \scriptsize \(\pm 3.0\)}\\
        \rowcolor{teal!30}
        {DO (Ours)}        & {\textbf{72.3} \scriptsize $\pm 1.9$ }& {\textbf{72.6} \scriptsize \(\pm 2.6\) }& {\textbf{72.3} \scriptsize \(\pm 2.2\) }\\
        \rowcolor{green!30}
        Improvement & +10.5 & +16.0 & +13.6 \\
        \bottomrule
      \end{tabular}
\end{table}

\subsection{Task Graph Generation}
\label{sec:graph_generation_results}
\subsubsection{Datasets}
We evaluate the ability of our approach to learn task graph representations on three datasets of procedural videos: EgoProceL-~\cite{bansal2022my} equipped with the newly introduced graph annotations, CaptainCook4D~\cite{peddi2023captaincook4d}, and EgoPER~\cite{lee2024error}. The CaptainCook4D~\cite{peddi2023captaincook4d} dataset consists of egocentric videos of 24 cooking procedures performed by 8 participants, with each procedure accompanied by a task graph that describes the constraints of the key-steps. Similarly, the EgoPER~\cite{lee2024error} dataset contains egocentric procedural videos of 5 different cooking tasks, accompanied by corresponding task graphs. %

\subsubsection{Problem Setup}
We tackle task graph generation as a weakly supervised learning problem in which models have to generate valid graphs by only observing labeled action sequences (weak supervision) rather than relying on task graph annotations (strong supervision), which are not available at training time. All models are trained on sequences of actions that are free from ordering errors or missing steps to provide a likely representation of procedures. We apply the first approach described in Section~\ref{sec:sequences} to handle repetitions. We then use the two proposed methods in Section~\ref{sec:models} to learn different task graph models, one per procedure, and report average performance across procedures. We trained TGT models using text embeddings derived from key-step names, extracted with EgoVLPv2~\cite{pramanick2023egovlpv2} pre-trained on Ego-Exo4D~\cite{grauman2023ego}. We refer to these models as TGT-text.

\subsubsection{Evaluation Measures}
Task graph generation is evaluated by comparing the binary, unweighted generated graph $\hat G = (\hat{\mathcal{K}},\hat{\mathcal{A}})$ with the corresponding ground truth graph $G= (\mathcal{K},\mathcal{A})$. Since task graphs aim to encode ordering constraints between pairs of nodes, we evaluate task graph generation as the problem of identifying valid pre-conditions (hence valid graph edges) among all possible ones. We therefore adopt the classic detection evaluation measures precision, recall, and $F_1$ score. In this context, we define True Positives (TP) as all edges included in both the predicted and ground truth graph (Eq.~\eqref{eq:tp}), False Positives (FP) as all edges included in the predicted graph, but not in the ground truth graph (Eq.~\eqref{eq:fp}), and False Negatives (FN) as all edges included in the ground truth graph, but not in the predicted one (Eq.~\eqref{eq:tn}). Note that true negatives are not required to compute precision, recall and $F_1$ score.
\\
\begin{minipage}{0.31\linewidth}
\begin{equation}
TP=\hat{\mathcal{A}} \cap \mathcal{A}
\label{eq:tp}
\end{equation}    
\end{minipage}
\begin{minipage}{0.31\linewidth}
\begin{equation}
    FP=\hat{\mathcal{A}} \setminus \mathcal{A}
    \label{eq:fp}
\end{equation}
\end{minipage}
\begin{minipage}{0.31\linewidth}
\begin{equation}
    FN=\mathcal{A} \setminus \hat{\mathcal{A}}
    \label{eq:tn}
\end{equation}
\end{minipage}
\\

\subsubsection{Compared Approaches}
\label{sec:task_graph_generation_approaches}
We compare our methods with respect to previous approaches for task graph generation, and in particular with MSGI \cite{sohn2020meta} and MSG$^2$ \cite{jang2023multimodal}, which are approaches based on Inductive Logic Programming (ILP). 
We also consider the recent approach proposed in~\cite{ashutosh2024video}, which generates a graph by counting co-occurrences of matched video segments. Since we assume labeled actions to be available at training time, we do not perform video matching and use ground truth segment matching provided by the annotations. This approach is referred to as ``Count-Based''. Given the popularity of large language models as reasoning modules, we also consider a baseline which uses Llama-3.1-405B-Instruct~\cite{dubey2024llama} to generate a task graph from key-step descriptions, without any access to key-step sequences.\footnote{See section \textit{Llama-3.1-405B-Instruct Prompts} of the supplementary material for more details.}

\subsubsection{Graph Generation Results}
Results in Table~\ref{tab:captaincook}, Table~\ref{tab:egoper}, and Table~\ref{tab:egoprocel} demonstrate that our proposed framework achieves state-of-the-art results in all considered datasets, outperforming competitive heuristics based methods. The tables highlight the limitations of traditional methods, such as MSGI~\cite{sohn2020meta}, which struggle to generate task graphs directly from action sequences, achieving poor performance across datasets: $12.2$ $F_1$ on CaptainCook4D, $50.6$ $F_1$ on EgoPER and $22.9$ $F_1$ on EgoProceL. Among more advanced heuristic methods, MSG\(^2\)~\cite{jang2023multimodal} achieves the best $F_1$ on CaptainCook4D ($73.3$), while the Count-Based~\cite{ashutosh2024video} approach delivers the best results on EgoPER ($80.8$), and LLaMA-3.1-405B-Instruct~\cite{dubey2024llama} outperforms competitors on EgoProceL ($58.7$). Despite these dataset-specific strengths, these methods fail to generalize effectively across all datasets. MSG\(^2\), which performs well on CaptainCook4D, achieves lower $F_1$ on EgoPER and EgoProceL. Similarly, while excelling on EgoPER, the Count-Based approach performs poorly on CaptainCook4D and EgoProceL. Llama-3.1-405B-Instruct achieves the best $F_1$ on EgoProceL but drops on CaptainCook4D and EgoPER.
In contrast, our Direct Optimization (DO) approach achieves the best performance across all three datasets, with substantial improvements in $F_1$: $+14.5\%$ on CaptainCook4D, $+10.2\%$ on EgoPER, and $+13.6\%$ on EgoProceL compared to the strongest competitors in each case. These results highlight the effectiveness of the proposed framework in learning task graph representations from key-step sequences, especially considering the simplicity of the DO method, which performs gradient descent directly on the adjacency matrix. Across all three datasets, our approach achieves slightly higher recall than precision, indicating its ability to retrieve most ground truth edges while occasionally introducing some hallucinated pre-conditions. This is likely due to the fact that video sequences in datasets typically favor the most common ways of completing a procedure.
Tight confidence intervals for DO highlight the stability of the proposed loss.
Second best results are consistently obtained by our feature-based TGT approach, showing the generality of our learning framework and the potential of integrating it into complex neural architectures. 
The lower performance of TGT, as compared to DO, may be attributed to its feature-based approach, which seeks to generate a more generalized task graph structure. In contrast, DO learns a more specific, data-driven representation. This difference in approach is particularly evident in the experiments on Ego-Exo4D (see Section~\ref{sec:pub}), where TGT's ability to generate a more generalized task graph proves advantageous, outperforming DO.

\begin{table}
    \centering
    \caption{We compare the abilities of our TGT model trained on visual features of CaptainCook4D to generalize to two fundamental video understanding tasks, i.e., pairwise ordering and future prediction. Despite not being explicitly trained for these tasks, our model exhibits video understanding abilities, surpassing the baseline.}
    \label{tab:understanding}
      \begin{tabular}{lll}
        \toprule
        Method         & Ordering     & Fut. Pred.  \\
        \midrule
        Random         & 50.0         & 50.0 \\
        \rowcolor{teal!30}
        TGT-video& \textbf{77.3} & \textbf{74.3} \\
        \rowcolor{green!30}
        Improvement & +27.3 & +24.3 \\
        \bottomrule
      \end{tabular}
\end{table}

\subsection{Video Understanding Abilities of the TGT Model}
\label{sec:video_understanding_results}
We investigate the ability of the TGT model to generalize beyond task graph generation by tackling two key video understanding tasks: pairwise ordering and future prediction~\cite{zhou2015temporal}. The first task, pairwise ordering, involves determining the correct temporal sequence of two short snippets extracted from an egocentric video of an activity. The goal is to infer which snippet occurs first and which follows, requiring a precise understanding of temporal dependencies between the video segments. The second task, future prediction, assesses the model's capability to anticipate the next step in an everyday activity. In this scenario, the model is provided with a longer video depicting part of an activity and two shorter video snippets. The task is to predict which of the two snippets logically and temporally follows the given video, demonstrating the model's ability to project forward in time and infer procedural progression.
\subsubsection{Problem Setup}
We set up the pairwise ordering and future prediction video understanding tasks following~\cite{zhou2015temporal} and evaluate the abilities of our TGT model, trained on visual features of CaptainCook4D~\cite{peddi2023captaincook4d} (TGT-video), to generalize to the two fundamental video understanding tasks despite not being explicitly trained for them. 
For pairwise ordering, models take as input two randomly shuffled video clips and are tasked with recognizing the correct ordering between key-steps. For future prediction, models take as input an anchor video clip and two randomly shuffled video clips and are tasked to select which of the two clips is the correct future of the anchor clip\footnote{See section \textit{Details on Pairwise ordering and future prediction} of the supplementary material for more details.}. We evaluate models using accuracy.

\subsubsection{Dataset}
We employed the subset of the CaptainCook4D dataset designated for task graph generation\footnote{See section \textit{Data Split} of the supplementary material for more details.} which has been divided into training and testing sets. This division was carefully managed to ensure that 50\% of the scenarios were equally represented in both the training and testing sets.

\subsubsection{Model}
We trained our TGT model using video embeddings extracted with a pre-trained EgoVLPv2~\cite{pramanick2023egovlpv2} on Ego-Exo4D~\cite{grauman2023ego}. During the training process, if multiple video embeddings are associated with the same key-step across the training sequences, one embedding per key-step is randomly selected. The model is trained for task graph generation on the training videos and tested for pairwise ordering and future prediction on the test set.

\subsubsection{Results}Table~\ref{tab:understanding} reports the performance of TGT trained on videos on two fundamental video understanding tasks~\cite{zhou2015temporal} of pairwise clip ordering and future prediction. Despite TGT not being explicitly trained for pairwise ordering and future predictions, it exhibits emerging video understanding abilities, surpassing the random baseline. Although we do not aim to directly compete with state-of-the-art methods in this domain, results are promising and suggest that TGT can effectively capture temporal and procedural cues within video data.

\subsection{Performance on the Downstream tasks of the Ego-Exo4D Procedure Understanding Benchmark}
\label{sec:pub}
In the following, we present experiments to show the usefulness of the learned representations in the downstream tasks of the Ego-Exo4D Procedure Understanding Benchmark~\cite{grauman2023ego}.
\subsubsection{Problem Setup}
The recently introduced Ego-Exo4D procedure understanding benchmark~\cite{grauman2023ego} encompasses 5 diverse downstream tasks associated with procedural video comprehension. Given a video segment $s_i$ and its preceding segment history $S_{:i-1} = \{s_1, \dots, s_{i-1}\}$, models are tasked to: (1) identify \textit{previous keysteps}, which refer to key-steps that should be executed before $s_i$; (2) determine whether $s_i$ is \textit{optional}, indicating that it can be skipped without undermining the proper execution of the procedure; (3) detect if $s_i$ constitutes a \textit{procedural mistake}, defined as a key-step performed inappropriately due to unmet pre-conditions; (4) predict \textit{missing keysteps}, which are steps that should have occurred before $s_i$; and (5) determine \textit{next keysteps}, representing key-steps whose dependencies are satisfied and are therefore ready for execution. The benchmark is weakly supervised and is presented in two variants based on the level of supervision: (1) \textit{instance-level}, where video segments and their corresponding key-step labels are provided during both training and inference, akin to an action recognition framework; and (2) \textit{procedure-level}, where training and inference rely on unlabeled video segments and a taxonomy of procedure-specific key-step names.

\subsubsection{Compared approaches}
We evaluate our approach against the baselines defined in~\cite{grauman2023ego}, which include a graph-based and an end-to-end approach. The graph-based baseline relies on a transition graph to perform procedural reasoning, while the end-to-end baseline predicts outcomes directly from video data without utilizing an explicit graph structure. It is important to note that the graph-based baseline is equivalent to the Count-Based method~\cite{ashutosh2024video}. We also compare our approach against all the baselines considered for task graph generation (see Section~\ref{sec:task_graph_generation_approaches}), excluding the MSGI method due to its convergence issues when applied to find large graphs. We apply the first approach described in Section~\ref{sec:sequences} to handle repetitions. Also, we conduct experiments using both instance-level and procedure-level supervision. In the case of instance-level supervision, we generate task graphs using the ground truth labels from the training set. On the other hand, for procedure-level supervision, these annotations cannot be used, thus we adopt two different approaches, as done by the authors of Ego-Exo4D~\cite{grauman2023ego}: \textit{keystep assignment} and \textit{keystep prediction}. \textit{Keystep assignment} involves generating pseudo-labels for video segments based on a pre-trained video-language model. In contrast, \textit{keystep prediction} uses a model specifically trained for key-step recognition to obtain pseudo-labels. The pseudo-labels obtained from both strategies are used by all the compared approaches to generate task graphs. At test time, the generated task graphs are used to perform procedure understanding and address the key-step level questions of the Ego-Exo4D benchmark. 

\subsubsection{DO and TGT} For the task graph generated using either DO or TGT methods, during testing we consider the adjacency matrix $\hat{Z}$ obtained before the post-processing stage (see Section~\ref{sec:postprocessing}) to support procedure understanding. Specifically, given the current key-step $K_i$, we perform the following steps: 

\noindent (1) a key-step $K_{prev}$ is predicted as previous key-step with a confidence score equal to:
\begin{equation}
    P(K_i | K_{prev}, \hat{Z}) = \frac{\hat{Z}_{(i, prev)}}{\sum_{h \in \mathcal{K} - \{K_{prev}\}} \hat{Z}_{(h,prev)}}
\end{equation}
where $\hat{Z}_{(i, prev)}$ is the edge weight from $K_i$ to $K_{prev}$ in the estimated task graph represented by $\hat{Z}$, and the denominator considers all possible key-steps, excluding $K_{prev}$ ($\mathcal{K} - \{K_{prev}\}$), that could have $K_{prev}$ as a potential pre-condition (i.e., as a valid previous key-step). 

\noindent (2) The key-step $K_i$ is classified as optional based on an optionality score $O(K_i)$, which combines \textit{global} and \textit{local} optionality scores. The \textit{global optionality} score $O_g(K_i)$ is derived from training data by analyzing how frequently $K_i$ appears as optional across sequences $y^{(d)} \in \mathcal{Y}$. For each sequence $y^{(d)}$ containing $K_i$, the optionality score for $K_i$ is computed as:
\begin{equation}
    \resizebox{0.91\hsize}{!}{
    $O_{y^{(d)}}(K_i) = \frac{(P(y^{(d)} - \{K_i\} | \hat{Z}) \cdot (1 - fr(K_i))}{[P(y^{(d)} - \{K_i\} | \hat{Z}) \cdot (1 - fr(K_i))] + [P(y^{(d)} | \hat{Z}) \cdot fr(K_i)]}$
    }
\end{equation}
where $fr(K_i)$ represents the frequency of $K_i$ in the training set, $P(y^{(d)} | \hat{Z})$ is the probability of completing sequence $y^{(d)}$ with $K_i$, and $P(y^{(d)} - \{K_i\} | \hat{Z})$ is the probability of completing the sequence without $K_i$. If $O_{y^{(d)}}(K_i)$ is greater than $0.5$, $K_i$ is considered optional for $y^{(d)}$, and a counter $count_o(K_i)$ is incremented. Otherwise, a ``mandatory'' counter $count_m(K_i)$ is incremented. The global optionality score is obtained as:
\begin{equation}
    O_g(K_i) = \frac{count_o(K_i)}{count_o(K_i) + count_m(K_i)}
\end{equation}
The \textit{local optionality} score assesses the optionality of $K_i$ within a specific sequence $y^{(d)}$, and in particular in the sub-sequence $y^{(d)}_{:t} = <y^{(d)}_0 = S, y^{(d)}_1, \dots, y^{(d)}_{t-1}, y^{(d)}_{t} = K_i>$. The score calculates the probability that the procedure can be directly completed skipping the current key-step $K_i$. This is done by removing $K_i$ from the sub-sequence $y^{(d)}_{:t}$ and replacing it with the end key-step $E$, resulting in the modified sub-sequence $\hat{y}^{(d)}_{:t} = <y^{(d)}_0 = S, y^{(d)}_1, \dots, y^{(d)}_{t-1}, y^{(d)}_{t} = E>$. The local score is then computed as:
\begin{equation}
    O_l(K_i) = P(<y^{(d)}_0 = S, y^{(d)}_1, \dots, y^{(d)}_{t-1}, y^{(d)}_{t} = E> | \hat{Z})
\end{equation}
A higher probability indicates that $K_i$ is likely optional.
Finally, the overall optionality score for $K_i$ is a weighted combination of the global and local optionality scores:
\begin{equation}
    O(K_i) = \alpha \cdot O_g(K_i) + (1 - \alpha) \cdot O_l(K_i)
\end{equation}
Here, $\alpha$ is a weighting parameter that balances the influence of global and local optionality scores, we set it to $0.7$.

\noindent (3) The key-step $K_i$ is identified as a procedural mistake if its required pre-condition key-steps $K_{prev}$ are missing from the observed set of key-step ($K_{\mathcal{J}^{(d)}_t}$). The score for this prediction is given by $\sum_{K_{prev}}\mathds{1}(K_{prev} \notin K_{\mathcal{J}^{(d)}_t}) \cdot P(K_i | K_{prev}, \hat{Z})$, where $\mathds{1}(\cdot)$ is the indicator function.

\noindent (4) The key-step $K_m$ is predicted as a possible missing key-step for $K_i$ with probability $\mathds{1}(K_m \notin K_{\mathcal{J}^{(d)}_t}) \cdot P(K_i | K_m, \hat{Z})$. 

\noindent (5) For predicting the future key-steps, the observed history, including the current key-step $K_i$, is utilized ($K_{\mathcal{H}} = K_{\mathcal{J}^{(d)}_t} \cup \{K_i\}$). The probability of a future key-step $K_f$ is calculated as:
\begin{equation}
    P(K_f | K_{\mathcal{H}}, \hat{Z}) = \frac{\sum_{j \in \mathcal{H}} \hat{Z}_{(f,j)}}{\sum_{h \in \bar{\mathcal{H}}} \sum_{j \in \mathcal{H}} \hat{Z}_{(h,j)}}
\end{equation}
where $\mathcal{H}$ denotes the set of indexes corresponding to the observed key-steps including the current one ($K_i$), and $\mathcal{\bar{H}}$ represents the set of indexes for the remaining unseen key-steps.

\begin{table*}[t]
  \caption{Ego-Exo4D procedure understanding benchmark results. Best results are in \textbf{bold}, second best results are \underline{underlined}. Best results of the baselines are \hl{highlighted}.}
  \label{tab:egoexo4d}
  \centering
  \resizebox{\linewidth}{!}{
  \begin{tabular}{lll|c|ccccc}
    \toprule
    Supervision & Method & Keystep Labels & Inf. Set & Prev. Keysteps     & Opt. Keysteps     & Proc. Mistakes & Miss. Keysteps & Fut. Keysteps          \\
    \midrule
    - & Uniform Baseline & - & Val / Test & 59.18 / 59.13 & 56.71 / 56.73 & 60.54 / 60.66 & 65.58 / 65.64 & 65.65 / 65.65 \\ \hline
    Instance-Level & Graph-Based & Ground Truth & Val & \underline{\hl{82.49}} & \hl{58.95} & \hl{73.19} & \hl{84.29} & \hl{63.48} \\
    Instance-Level & End-to-End & Ground Truth & Val & 62.05 & 51.85 & 56.75 & 60.11 & 60.35 \\
    Instance-Level & MSG$^2$~\cite{jang2023multimodal} & Ground Truth & Val & 54.82 & - & 52.88 & 53.87 & 52.03 \\
    Instance-Level & Llama-3.1-405B-Instruct~\cite{dubey2024llama} & Ground Truth & Val & 65.13 & 56.31 & 64.71 & 62.93 & 52.49 \\
    \rowcolor{teal!30}
    Instance-Level & TGT-text (ours) & Ground Truth & Val & 81.77 & \textbf{75.56} & \underline{78.83} & \textbf{88.88} & \textbf{73.56} \\
    \rowcolor{teal!30}
    Instance-Level & DO (ours) & Ground Truth & Val & \textbf{83.25} & \underline{74.52} & \textbf{84.52} & \underline{87.23} & \underline{73.32} \\
    \rowcolor{green!30}
    & Improvement  & & Val & +0.76 & +16.61 & +11.33 & +4.59 & +10.08 \\ \hline
    Procedure-Level & Graph-Based & Keystep Assignment & Val / Test & 54.26 / 53.43 & 49.86 / 52.36 & 56.46 / 57.81 & 60.97 / 53.92 & 52.50 / 53.54 \\
    Procedure-Level & End-to-End & Keystep Assignment & Val / Test & 55.37 / 54.82 & \hl{52.12} / 60.78 & 52.84 / 54.73 & 56.11 / 53.75 & \hl{58.88} / 57.47 \\
    Procedure-Level & Graph-Based & Keystep Prediction & Val / Test & \hl{64.56} / \hl{66.22} & 49.51 / 49.00 & \hl{61.15} / \hl{58.59} & \hl{61.50} / \hl{64.18} & 57.87 / \hl{58.34} \\
    Procedure-Level & End-to-End & Keystep Prediction & Val / Test & 57.43 / 57.92 & 51.54 / \underline{\hl{61.01}} & 51.68 / 54.92 & 54.99 / 55.15 & 57.35 / 56.92 \\ 
    Procedure-Level & MSG$^2$~\cite{jang2023multimodal} & Keystep Assignment & Val / Test & 50.28 / 50.11 & \hspace{3mm}-\hspace{2.5mm} / \hspace{3mm}-\hspace{3mm} & 48.78 / 51.76 & 49.93 / 49.80 & 51.36 / 51.32 \\
    Procedure-Level & MSG$^2$~\cite{jang2023multimodal} & Keystep Prediction & Val / Test & 51.04 / 51.37 & \hspace{3mm}-\hspace{2.5mm} / \hspace{3mm}-\hspace{3mm} & 53.79 / 55.34 & 50.32 / 50.40 & 53.05 / 53.13 \\
    Procedure-Level & Llama-3.1-405B-Instruct~\cite{dubey2024llama} & Keystep Assignment & Val / Test & 55.05 / 54.61 & 50.59 / 50.12 & 48.28 / 55.54 & 53.25 / 55.02 & 51.55 / 51.01 \\
    Procedure-Level & Llama-3.1-405B-Instruct~\cite{dubey2024llama} & Keystep Prediction & Val / Test & 56.37 / 57.61 & 52.92 / 53.38 & 52.25 / 56.92 & 54.13 / 57.07 & 51.65 / 51.88 \\
    \rowcolor{teal!30}
    Procedure-Level & TGT-text (ours) & Keystep Assignment & Val / Test & 67.99 / 63.83 & 54.16 / 56.90 & 57.04 / \underline{61.29} & 66.18 / 64.20 & 67.01 / 66.65\\
    \rowcolor{teal!30}
    Procedure-Level & TGT-text (ours) & Keystep Prediction & Val / Test & \textbf{71.37} / \textbf{70.83} & \underline{63.95} / 60.62 & \underline{60.55} / 59.21 & \textbf{70.47} / \textbf{72.80} & \textbf{73.65} / \textbf{73.50}\\
    \rowcolor{teal!30}
    Procedure-Level & DO (ours) & Keystep Assignment & Val / Test & 62.86 / 62.10 & 54.25 / 55.73 & 53.38 / 59.55 & 63.54 / 63.51 & 66.49 / 65.30\\
    \rowcolor{teal!30}
    Procedure-Level & DO (ours) & Keystep Prediction & Val / Test & \underline{70.25} / \underline{70.53} & \textbf{66.09} / \textbf{61.11} & \textbf{62.25} / \textbf{63.61} & \underline{69.40} / \underline{72.28} & \underline{69.17} / \underline{69.02} \\
    \rowcolor{green!30}
    & Improvement & & Val / Test & +6.81 / +4.61 & +13.97 / +0.10 & +1.00 / +5.02 & +8.97 / +8.62 & +14.77 / +15.16 \\
    \bottomrule
    \end{tabular}
    }
\end{table*}%

\subsubsection{MSG$^2$} \label{sec:ego-MSG}For task graphs generated using MSG$^2$ method~\cite{jang2023multimodal}, only binary adjacency matrices are available. This limitation prevents the use of approaches designed for task graphs generated by the DO and TGT methods, which rely on weighted adjacency matrices to support procedure understanding. Let $\hat G = (\hat{\mathcal{K}}, \hat{\mathcal{A}})$ be the binary, unweighted generated task graph. Given a current key-step $K_i$ we perform the following step: 

\noindent (1) the set of previous key-steps $\mathcal{P}k(K_i; \hat{G})$ consists of all key-steps $K_{prev}$ connected to $K_i$ via outgoing edges: $\mathcal{P}k(K_i; \hat{G}) = \{K_{prev} \in Pred(K_i; \hat{G})\}$ where $Pred(K_i; \hat{G}) = \{K_{j} | (K_i, K_{j}) \in \hat{E}\}$. For each node $K_{j}$, the pre-condition score is computed as follows:
\begin{align}
\label{eq:precondition_score_msg}
    & score_{\mathcal{P}k}(K_i, K_j; \hat{G}) = \begin{cases}
    \frac{1}{|\mathcal{P}k(K_i; \hat{G})|} & \text{if } K_{j} \in \mathcal{P}k(K_i; \hat{G}),\\
    0 & \text{otherwise}.
    \end{cases}
\end{align}

\noindent (2) With MSG$^2$, it is not possible to evaluate whether a key-step $K_i$ is optional. The binary nature of the adjacency matrix does not provide the detailed probabilistic information required for such assessments.

\noindent (3) To determine whether $K_i$ is a procedural mistake, its previous key-steps $\mathcal{P}k(K_i; \hat{G})$ are examined. If any of these pre-conditions are missing from the observed set of key-step ($K_{\mathcal{J}^{(d)}_t}$), a procedural mistake is predicted. The score for predicting a procedural mistake is computed by $\sum_{K_{prev} \in \mathcal{P}k(K_i; \hat{G})} \mathds{1}(K_{prev} \notin K_{\mathcal{J}^{(d)}_t}) \cdot score_{Pk}(K_i, K_{prev}; \hat{G})$, where $score_{Pk}$ represents the previous key-step scores (Eq.~\eqref{eq:precondition_score_msg}), and $\mathds{1}(\cdot)$ is the indicator function. 

\noindent (4) Missing steps ($\mathcal{M}k(K_i; \hat{G}) = \{K_j \in Pred(K_i; \hat{G}) : K_j \notin K_{\mathcal{J}^{(d)}_t}\}$) are identified as those steps that are previous key-steps of the current step ($K_i$), but do not appear in the observed set of key-step indexes ($K_{\mathcal{J}^{(d)}_t}$). For a missing key-step $K_m$, the score is computed as:
\begin{small}
\begin{align}
    & score_{\mathcal{M}k}(K_i, K_m; \hat{G}) = \begin{cases}
        \frac{1}{|\mathcal{M}k(K_i; \hat{G})|} & \text{if } K_m \in \mathcal{M}k(K_i; \hat{G}), \\
        0 & \text{otherwise}.
    \end{cases}
\end{align}
\end{small}

\noindent (5) To predict future key-steps for $K_i$, the future score for a key-step $K_f$ is calculated as follows:
\begin{small}
\begin{equation}
    \begin{aligned}
        & score_{\mathcal{F}k}(K_i, K_f; \hat{G}) = \\ & = \begin{cases}
            \frac{1}{|\mathcal{F}k(K_i; \hat{G})| + |\mathcal{P}k(K_f; \hat{G}) - (K_{\mathcal{J}^{(d)}_t} \cup K_i)|} & \text{if } K_f \in \mathcal{F}k(K_i; \hat{G}),\\
            0 & \text{otherwise}.
        \end{cases}
    \end{aligned}
\end{equation}
\end{small}

\noindent Here, $\mathcal{F}k(K_i; \hat{G})$ is the set of successor for the current key-step $K_i$ taken from the task graph $\hat{G}$, $\mathcal{P}k(K_i; \hat{G})$ is the set of the pre-conditions for $K_i$ taken from the task graph $\hat{G}$, and $(K_{\mathcal{J}^{(d)}_t} 
\cup K_i)$ is the set of observed key-steps including the current key-step $K_i$. The score incorporates both the number of future steps and unmet pre-conditions to balance the prediction scores. The scores are then normalized.

\subsubsection{Llama-3.1-405B-Instruct} For task graph generated using Llama-3.1-405B-Instruct~\cite{dubey2024llama}, only binary adjacency matrices are available, thus we used the same approaches outlined in previous section for MSG$^2$ to perform procedure understanding. The key distinction is that we queried the model regarding optional key-steps and used its responses to determine when a key-step should be classified as optional\footnote{See section \textit{Llama-3.1-405B-Instruct Prompts} of the supplementary material for more details.}.

\subsubsection{Results} Table~\ref{tab:egoexo4d} reports the results of the compared methods on the Ego-Exo4D~\cite{grauman2023ego} procedure understanding benchmark. For instance-level supervision, results are limited to the validation set due to the absence of ground-truth annotations in the test set, a limitation intentionally introduced by the authors as part of the challenge design. Our methods achieve significant improvements with performance gains of up to +0.76\%, +16.61\%, +11.33\%, +4.59\%, and +10.08\% for identifying Previous Keysteps, Optional Keysteps, Procedural Mistakes, Missing Keysteps, and Future Keysteps, respectively in the validation set. Comparing DO and TGT under instance-level supervision, DO achieves superior performance in identifying Previous Keysteps (82.23 vs. 81.77), and Procedural Mistakes (84.52 vs. 78.83). Conversely, TGT excels in detecting Optional Keysteps (75.56 vs. 74.52), Missing Keysteps (88.88 vs. 87.23) and Future Keysteps (73.56 vs. 73.32). These contrasting results can be attributed to the models' differing strengths: TGT's ability to generate more generalizable graphs enhances its effectiveness in predicting optional, missing, and future actions, while DO has an advantage in tasks requiring more Procedure-specific representations. Our methods exhibit notable performance gains even under procedure-level supervision, achieving improvements of up to +4.61\%, +0.10\%, +5.02\%, +8.62\%, and +15.16\% in identifying Previous Keysteps, Optional Keysteps, Procedural Mistakes, Missing Keysteps, and Future Keysteps, respectively. These results are achieved by leveraging Keystep Prediction as pseudo-labels, which has proven to be the most effective approach for generating accurate annotations. In this context, the performance patterns of DO and TGT remain consistent with those observed under instance-level supervision. Specifically, DO outperforms in detecting Procedural Mistakes (63.61 vs. 59.21), while TGT achieves higher scores in identifying Missing Keysteps (72.80 vs. 72.28) and Future Keysteps (73.50 vs. 69.02). This trend mirrors the instance-level results but reveals marginal differences in the detection of Optional Keysteps (61.11 for DO vs. 60.62 for TGT) and Previous Keysteps (70.86 for TGT vs. 70.53 for DO), likely due to noise introduced during the recognition phase.

\subsection{Online Mistake Detection}
\label{sec:omd}
We now show how the proposed representation can tackle the downstream task of online mistake detection. We apply the second approach described in Section~\ref{sec:sequences} to handle repetitions to generate task graphs\footnote{See section \textit{Details on Online Mistake Detection} of the supplementary material for more details.}.

\subsubsection{Problem Setup}
\begin{figure}
    \centering
    \includegraphics[width=1.0\linewidth]{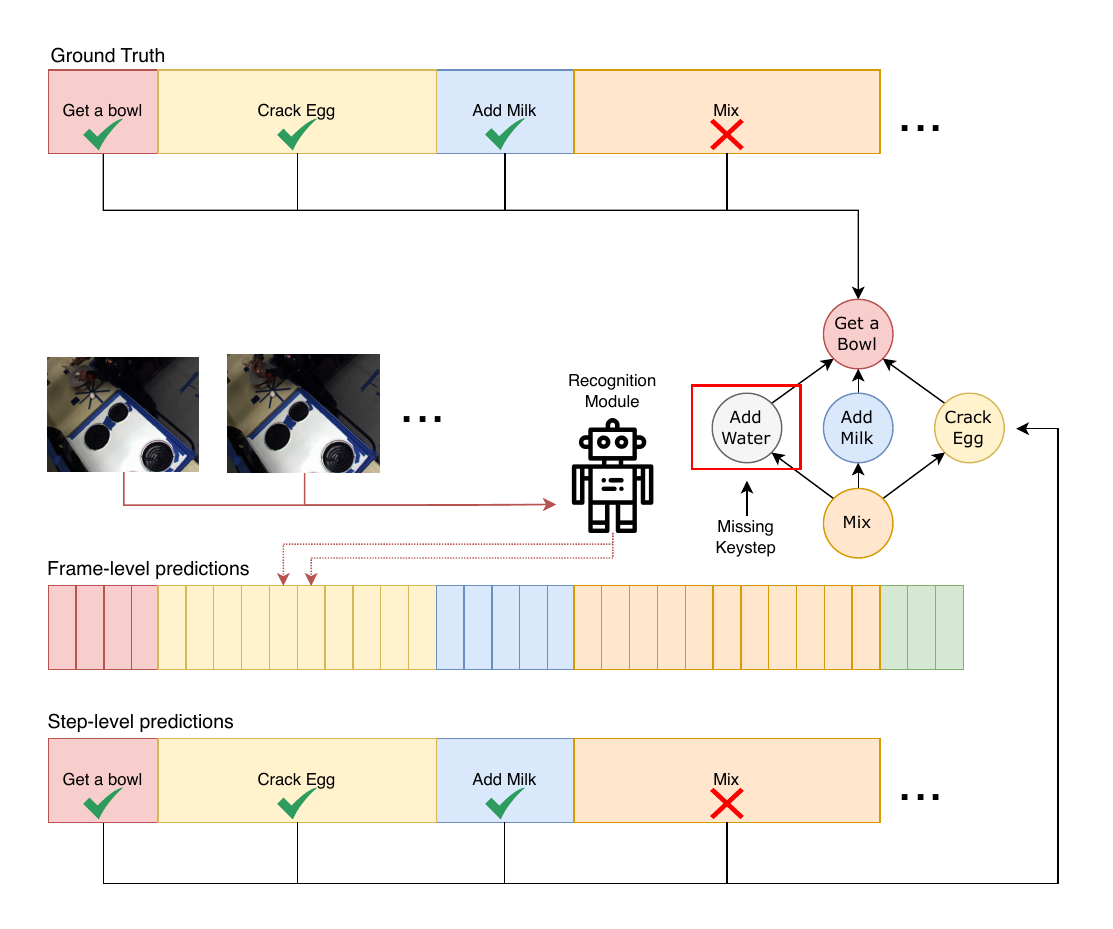}
    \caption{Framework used for online mistake detection. The upper section presents how our framework works when ground truth action sequences are used as input. The lower section shows how our framework works when predicted actions from an online recognition module are used as input. In the example, ``Mix'' is recognized as a mistake because the precondition ``Add Water'' is not satisfied.
    }
    \label{fig:online_mistake_detection}
\end{figure}

We follow the PREGO benchmark~\cite{flaborea2024prego} based on the Assembly101-O and EPIC-Tent-O datasets. In this benchmark, models are tasked to perform online action detection from procedural egocentric videos. To evaluate the usefulness of task graphs on this downstream task, we design a system which flags the current action as a mistake if its pre-conditions in the predicted graph do not appear in previously observed actions (see Figure~\ref{fig:online_mistake_detection}).
Given a video segment $s_i$ and its preceding key-step history $S_{:i-1} = \{s_1, \dots, s_{i-1}\}$, our framework determines whether the current segment $s_i$ constitutes a mistake. We conduct experiments in two configurations: (1) using ground truth action sequences as input, and (2) leveraging an action recognition module to predict the actions, which are then fed into our framework. For both experimental setups the binary task graph obtained after post-processing $\hat G = (\hat{\mathcal{K}}, \hat{\mathcal{A}})$ is used to infer whether $s_i$ is a mistake as illustrated in Figure~\ref{fig:online_mistake_detection}. Specifically, given the ground truth or predicted key-step $K_i$ associated to current segment $s_i$, the task graph is employed to verify whether all the pre-conditions of the current key-step $\mathcal{P}k(K_i; \hat{G}) = \{K_x \in Pred(K_i;  \hat{G})\}$ have been satisfied in the past, meaning that they must exist in the set of previously executed key-steps $K_{\mathcal{J}^{(d)}_t}$. This validation process is formalized as follows:
\begin{align}
    &
    \begin{cases}
    \mathcal{P}k(K_i; \hat{G}) \cap K_{\mathcal{J}^{(d)}_t} \neq \mathcal{P}k(K_i; \hat{G}) & Mistake \\
    \mathcal{P}k(K_i; \hat{G}) \cap K_{\mathcal{J}^{(d)}_t} = \mathcal{P}k(K_i; \hat{G}) & Correct
    \end{cases}
\end{align}

\subsubsection{Compared methods}
We compare our approach against the PREGO model introduced in~\cite{flaborea2024prego}, which identifies mistakes by comparing the currently observed action with a future action predicted by a forecasting module. It is important to highlight that PREGO relies on an implicit representation of the procedure (via the forecasting module), while our approach utilizes an explicit task graph representation, learned using the proposed framework. We also compare our approach with respect to baselines based on all graph generation approaches (see Section~\ref{sec:task_graph_generation_approaches}) to evaluate the impact of accurately predicted graphs on downstream performance. For all evaluated methods, we present results based on both ground-truth action segments and action sequences predicted by a MiniRoad~\cite{an2023miniroad} instance, a state-of-the-art online action detection module trained on each target dataset.

\begin{table*}
  \caption{Online mistake detection results. Results obtained with ground truth action sequences are denoted with $^*$, while results obtained on predicted action sequences are denoted with $^+$.}
  \label{tab:online_mistake_detection_results}
  \centering
  \resizebox{0.85\linewidth}{!}{
  \setlength\tabcolsep{4pt}
      \begin{tabular}{lccccccccclccccccccc}
      
        \toprule
        & \multicolumn{7}{c}{Assembly101-O} & \multicolumn{8}{c}{EPIC-Tent-O} \\
        \cmidrule(lr){2-8} \cmidrule(lr){10-16}
        & \multicolumn{1}{c}{Avg} & \multicolumn{3}{c}{Correct} & \multicolumn{3}{c}{Mistake} & & \multicolumn{1}{c}{Avg} & \multicolumn{3}{c}{Correct} & \multicolumn{3}{c}{Mistake} \\
        \cmidrule(lr){2-2} \cmidrule(lr){3-5} \cmidrule(lr){6-8} \cmidrule(lr){10-10} \cmidrule(lr){11-13} \cmidrule(lr){14-16} 
        Method & F$_1$ & F$_1$ & \scriptsize{Prec} & \scriptsize{Rec} & F$_1$ & \scriptsize{Prec} & \scriptsize{Rec} & & F$_1$ & F$_1$ & \scriptsize{Prec} & \scriptsize{Rec} & F$_1$ & \scriptsize{Prec} & \scriptsize{Rec} \\
        \midrule
        Count-Based$^*$ \cite{ashutosh2024video}  & 26.5 & 9.9 & \scriptsize{5.2} & \scriptsize{89.7} & 43.1 & \scriptsize{98.4} & \scriptsize{27.6} & & \hl{57.7} & \hl{93.2} & \scriptsize{94.1} & \scriptsize{92.3} & \hl{22.2} & \scriptsize{20.0} & \scriptsize{25.0}  \\
        Llama-3.1-405B-Instruct$^*$ \cite{dubey2024llama}  & 31.1 & 37.7 & \scriptsize{28.3} & \scriptsize{56.7} & 24.5 & \scriptsize{41.2} & \scriptsize{17.4} & & 46.0 & 79.4 & \scriptsize{70.6} & \scriptsize{90.8} & 12.5 & \scriptsize{26.7} & \scriptsize{8.2}  \\
        MSGI$^*$ \cite{sohn2020meta}   & 33.4 & 23.3 & \scriptsize{13.5} & \scriptsize{83.8} & 43.4 & \scriptsize{92.9} & \scriptsize{28.3} & & 44.5 & 66.9 & \scriptsize{51.6} & \scriptsize{95.2} & 22.0 & \scriptsize{73.3} & \scriptsize{12.9}  \\
        PREGO$^*$ \cite{flaborea2024prego}   & 39.4 & 32.6 & \scriptsize{89.7} & \scriptsize{19.9} & 46.3 & \scriptsize{30.7} & \scriptsize{94.0} & & 32.1 & 45.0 & \scriptsize{95.7} & \scriptsize{29.4} & 19.1 & \scriptsize{10.7} & \scriptsize{86.7}  \\
        MSG$^{2*}$ \cite{jang2023multimodal}  & \hl{56.1} & \hl{63.9} & \scriptsize{51.5} & \scriptsize{84.2} & \hl{48.2} & \scriptsize{73.6} & \scriptsize{35.8} & & 54.1 & 92.9 & \scriptsize{94.1} & \scriptsize{91.7} & 15.4 & \scriptsize{13.3} & \scriptsize{18.2}  \\
        \rowcolor{teal!30} TGT-text (Ours)$^*$ & \underline{62.8} & \underline{69.8} & \scriptsize{56.8} & \scriptsize{90.6} & \underline{55.7} & \scriptsize{84.1} & \scriptsize{41.7} & & \textbf{64.1} & \textbf{93.8} & \scriptsize{94.1} & \scriptsize{93.5} & \textbf{34.5} & \scriptsize{33.3} & \scriptsize{35.7} \\
        \rowcolor{teal!30} DO (Ours)$^*$ & \textbf{75.9} & \textbf{90.2} & \scriptsize{98.2} & \scriptsize{83.4} & \textbf{61.6} & \scriptsize{46.7} & \scriptsize{90.4} & & \underline{58.3} & \underline{93.5} & \scriptsize{94.8} & \scriptsize{92.4} & \underline{23.1} & \scriptsize{20.0} & \scriptsize{27.3} \\
        \rowcolor{green!30} Improvement$^*$ & +19.8 & +26.3 & &  & +13.4 & & & & +6.4 & +0.6 & & & +12.3 & &\\
        \midrule
        Count-Based$^+$ \cite{ashutosh2024video}  & 23.0 & 2.1 & \scriptsize{1.0} & \scriptsize{62.5} & \textbf{\hl{43.8}} & \scriptsize{98.4} & \scriptsize{28.2} & & 44.8 & 67.0 & \scriptsize{51.7} & \scriptsize{95.0} & 22.7 & \scriptsize{73.3} & \scriptsize{13.4}  \\
        Llama-3.1-405B-Instruct$^+$ \cite{dubey2024llama} & 41.7 & 42.9 & \scriptsize{30.6} & \scriptsize{71.9} & 40.4 & \scriptsize{69.8} & \scriptsize{28.4} & & 40.8 & 59.8 & \scriptsize{43.5} & \scriptsize{95.5} & 21.8 & \scriptsize{80.0} & \scriptsize{12.6}  \\
        MSGI$^+$ \cite{sohn2020meta}   & 28.3 & 14.0 & \scriptsize{7.8} & \scriptsize{66.7} & \underline{42.5} & \scriptsize{90.1} & \scriptsize{27.8} & & 40.4 & 59.2 & \scriptsize{42.9} & \scriptsize{95.5} & 21.6 & \scriptsize{80.0} & \scriptsize{12.5}  \\
        PREGO$^+$ \cite{flaborea2024prego}   & 32.5 & 23.1 & \scriptsize{68.8} & \scriptsize{13.9} & 41.8 & \scriptsize{27.8} & \scriptsize{84.1} & & 29.4 & 41.6 & \scriptsize{97.9} & \scriptsize{26.4} & 17.2 & \scriptsize{9.5} & \scriptsize{93.3}  \\
        MSG$^{2+}$ \cite{jang2023multimodal}  & \hl{46.2} & \hl{59.1} & \scriptsize{51.2} & \scriptsize{70.0} & 33.2 & \scriptsize{44.5} & \scriptsize{26.5} & & \underline{{\hl{45.2}}} & {\hl{67.5}} & \scriptsize{52.4} & \scriptsize{95.1} & \underline{{\hl{22.9}}} & \scriptsize{73.3} & \scriptsize{13.6}  \\
        \rowcolor{teal!30} TGT-text (Ours)$^+$ & \underline{53.0} & \underline{67.8} & \scriptsize{62.3} & \scriptsize{74.5} & 38.2 & \scriptsize{46.2} & \scriptsize{32.6} & & 43.8 & \textbf{69.5} & \scriptsize{55.8} & \scriptsize{92.1} & 18.2 & \scriptsize{53.3} & \scriptsize{11.0} \\
        \rowcolor{teal!30} DO (Ours)$^+$ & \textbf{53.5} & \textbf{78.9} & \scriptsize{85.0} & \scriptsize{73.5} & 28.1 & \scriptsize{22.5} & \scriptsize{37.3} & & \textbf{46.5} & \underline{69.3} & \scriptsize{54.4} & \scriptsize{95.2} & \textbf{23.7} & \scriptsize{73.3} & \scriptsize{14.1} \\
        \rowcolor{green!30} Improvement$^+$ & +7.3 & +19.8 & &  & -5.6 & & & & +1.3 & +1.2 & & & +1.2 & & \\
        
        \bottomrule
      \end{tabular}
  }
\end{table*}

\subsubsection{Results}
The results presented in Table~\ref{tab:online_mistake_detection_results} underscore the effectiveness of the learned task graphs for the downstream application of online mistake detection. The proposed methods demonstrate substantial improvements over prior methods, achieving increases of $+19.8$ and $+6.4$ in average $F_1$ score on the Assembly101-O and EPIC-Tent-O datasets, respectively, when predictions are made using ground-truth action sequences. While TGT ranks as the second-best performer on Assembly101-O, it outperforms other methods on EPIC-Tent-O, achieving an average $F_1$ score of $64.1$ compared to $58.3$. This performance discrepancy can be attributed to the nature of the action annotations in the two datasets. Indeed, key-step names in EPIC-Tent (e.g., ``Place Vent Cover'', ``Open Stake Bag'', or ``Spread Tent'') are more descriptive and distinctive than those in Assembly101 (e.g., ``attach cabin'', ``attach interior'', or ``screw chassis''). This highlights the versatility of the proposed learning framework, which can operate effectively in abstract, symbolic environments with the DO approach, while also leveraging semantics with TGT when advantageous.
Notably, the third-best performing methods are graph-based approaches, with MSG$^2$ achieving an average $F_1$ score of $56.1$ on Assembly101-O, while the simpler Count-Based approach obtaining an average $F_1$ score of $57.7$ on EPIC-Tent-O. In comparison, the PREGO model, which relies on implicit representations, yields significantly lower average $F_1$ scores of $39.4$ and $32.1$ on Assembly101-O and EPIC-Tent-O, respectively. These results highlight the advantages of explicit graph-based representations for mistake detection over implicit approaches like PREGO. Breaking down performance into correct and mistake $F_1$ scores reveals some degree of unbalance of our approaches and the main competitors (MSG$^2$ and Count-Based) towards identifying correct actions rather than mistakes. This suggests that graph-based representations may detect spurious pre-conditions, likely due to the limited number of demonstrations in the videos. Conversely, the implicit PREGO model exhibits a tendency to skew toward detecting mistakes. Further examination of precision and recall values provides insight into the sources of performance discrepancies. For instance, the Count-Based method shows a significant imbalance in Assembly101-O, achieving a high recall of $89.7$, but an extremely low precision of $5.2$ for predicting correct segments. In contrast, the proposed approach obtains balanced precision and recall values in detecting correct segments in Assembly101-O ($98.2$/$83.4$) and EPIC-Tent-O ($94.1$/$93.5$), and detecting mistakes in EPIC-Tent-O ($33.3$/$35.7$), while the prediction of mistakes on Assembly101-O is more skewed ($46.7$/$90.4$).
Results based on action sequences predicted from videos (bottom part of Table~\ref{tab:online_mistake_detection_results}) underscore the difficulty of handling noisy action sequences (see ablation study in Section~\ref{sec:ablation_oad}). While the explicit task graph representation may not accurately reflect the predicted noisy action sequences, our methods still achieve notable gains over prior approaches, with improvements of $+7.3$ and $+1.3$ in average $F_1$ scores for Assembly101-O and EPIC-Tent-O, respectively. Interestingly, the best-performing competitors remain graph-based methods, such as $MSG^2$ and the Count-Based approach, which demonstrate considerable advantages over the implicit representation used by the PREGO model. Indeed, the DO method achieves an average $F_1$ score of $53.5$ and $46.5$ in Assembly101-O and EPIC-tent-O, respectively, significantly outperforming PREGO's scores of $32.5$ and $29.4$. Also, in this case, we observe that graph-based methods tend to be skewed towards detecting correct action sequences. In this context, while the TGT model achieves competitive overall performance, its $F_1$ score for mistake detection is limited to $38.2$, trailing the Count-Based approach by $5.6$ points on Assembly101-O. In contrast, the count-based method only achieves a $F_1$ score of $2.1$ when predicting correct segments.

\begin{figure}[t]
    \centering
    \includegraphics[width=1.0\linewidth]{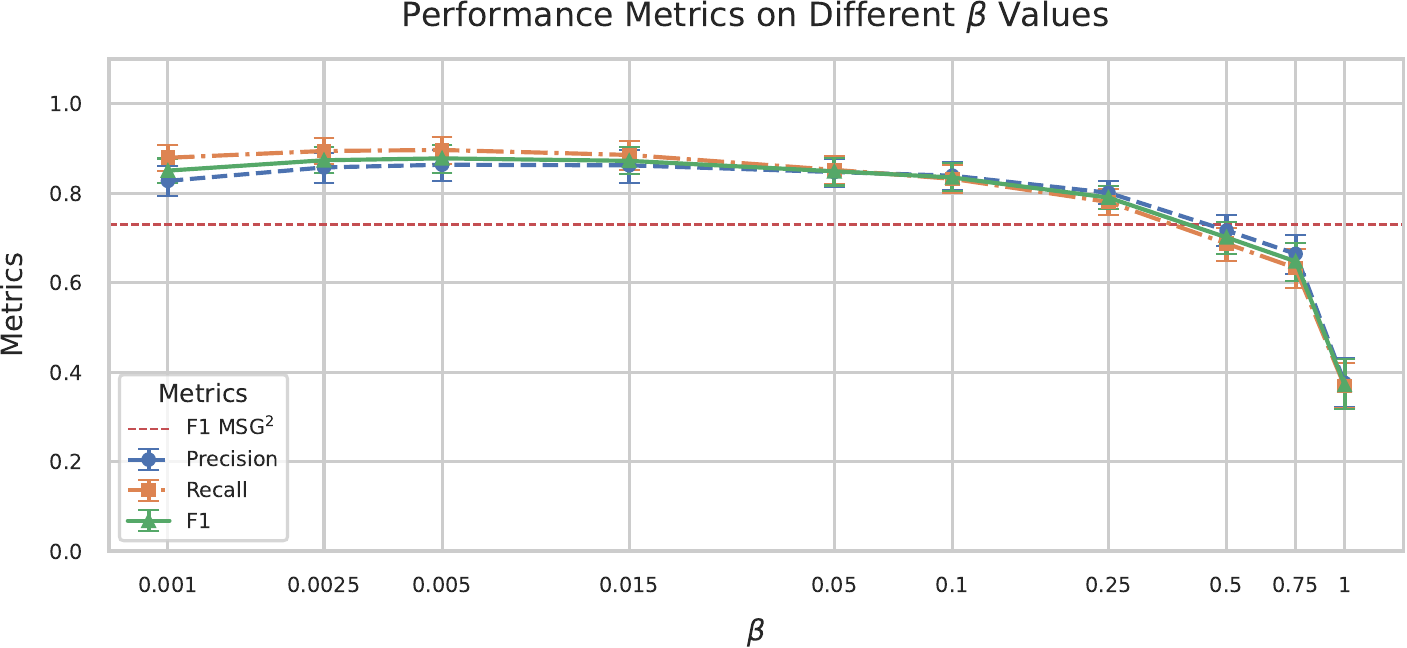}
    \caption{Performance metrics on different $\beta$ values using Direct Optimization (DO) on CaptainCook4D. The dashed line represents the best-performing method among the competitors on this dataset.}
    \label{fig:beta_captaincook}
\end{figure}

\begin{figure*}
\centering
\subfloat[Assembly101]{%
    \includegraphics[width=0.49\textwidth]{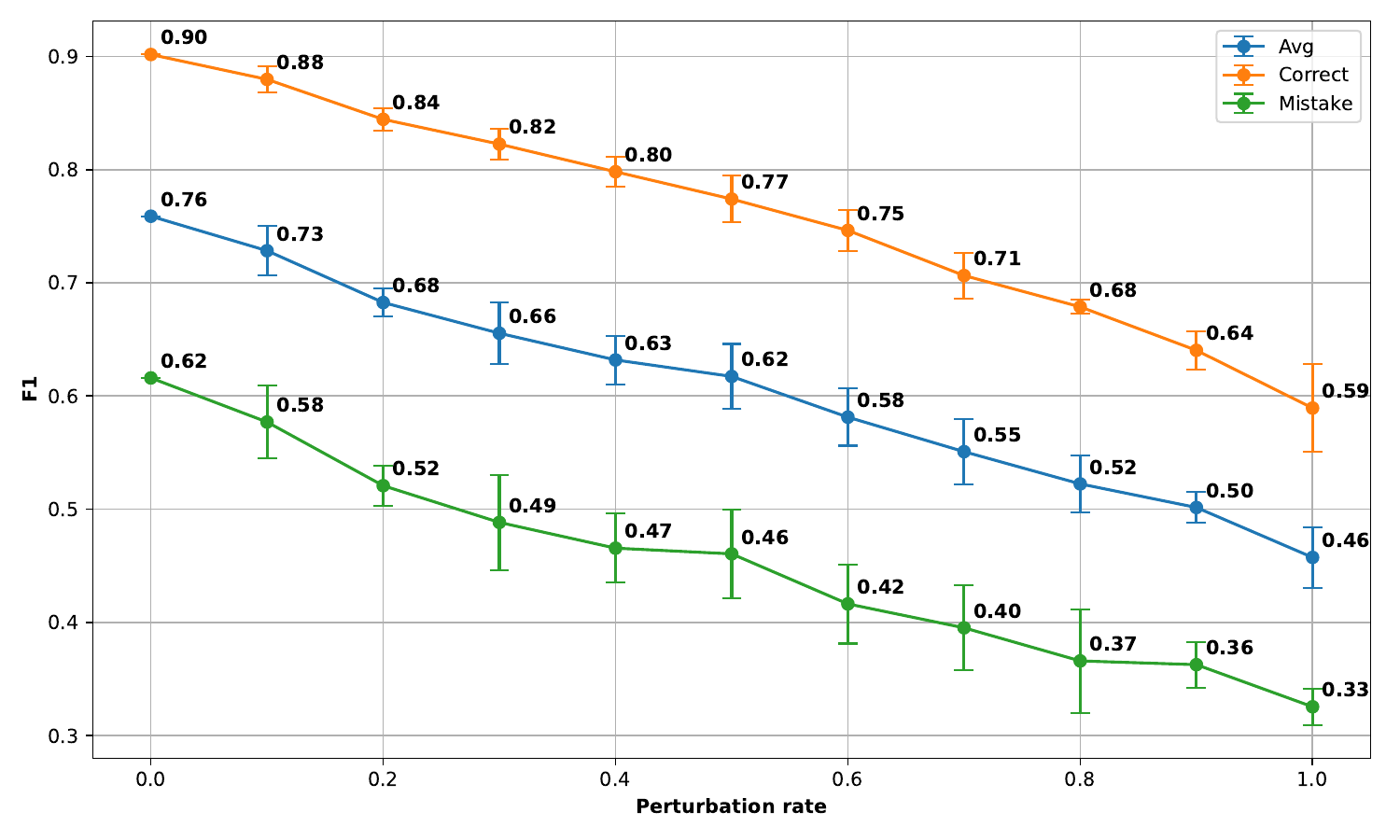}%
    \label{fig:sub1}
}
\hfill
\subfloat[EPIC-Tent]{%
    \includegraphics[width=0.49\textwidth]{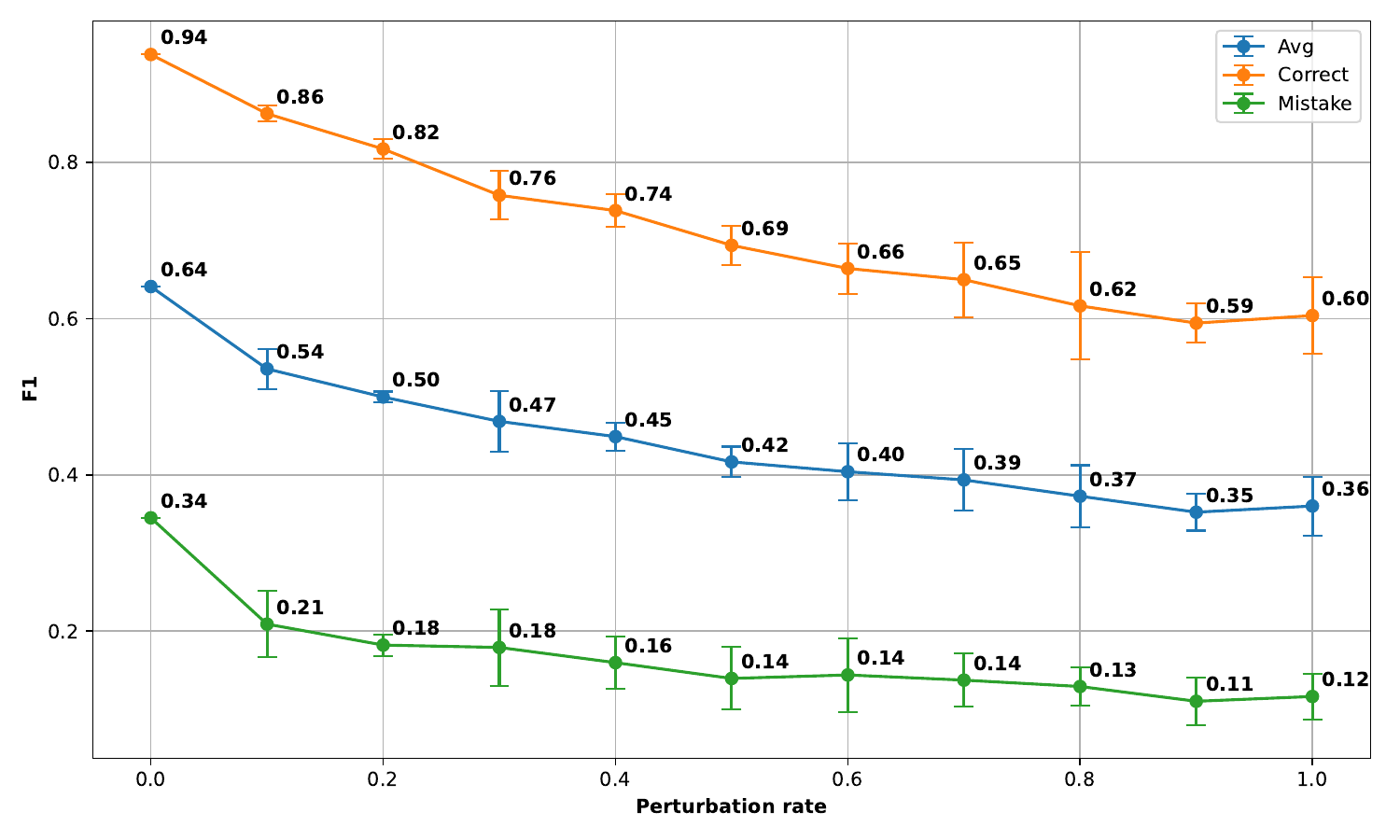}%
    \label{fig:sub2}
}
\caption{To further investigate the effect of noise, we conducted an analysis based on the controlled perturbation of ground truth action sequences, with the aim to simulate noise in the action detection process. At inference, we perturbed each key-step with a probability \(p\) (the ``perturbation rate''), with three kinds of perturbations: insert (inserting a new key-step with a random action class), delete (deleting a key-step), or replace (randomly changing the class of a key-step). The plots show the trend of the F1 score (Average, Correct, and Mistake) as the perturbation rate increases in the case of Assembly101-O (left) and EPIC-Tent-O (right). Results suggest that the proposed approach can still bring benefits even in the presence of imperfect action detections, with the average F1 score dropping down $10-15$ points with a moderate noise level of $20\%$.}
\label{fig:test}
\end{figure*}

\subsection{Ablation Studies}
\label{sec:ablation}
In this section, we first analyze the impact of different \(\beta\) values (see Eq.~\eqref{eq:loss}) on the performance of the Direct Optimization (DO) method (Section~\ref{sec:ablation_beta}). We then evaluate the effectiveness of the Distinctiveness Cross-Entropy Loss (DCEL) in improving task graph generation with TGT trained using text embeddings, highlighting its dataset-dependent impact (Section~\ref{sec:ablation_DCEL}). Finally, we investigate the role of action recognition accuracy in online mistake detection by simulating controlled noise scenarios (Section~\ref{sec:ablation_oad}).

\subsubsection{Performance Metrics on Different \(\beta\) Values}
\label{sec:ablation_beta}
Figures~\ref{fig:beta_captaincook} presents performance metrics across various $\beta$ values for the Direct Optimization (DO) method on the CaptainCook4D~\cite{peddi2023captaincook4d} dataset. The plot includes a comparison with the best performing competitor (the red dotted line), highlighting the range of $\beta$ values where DO outperforms the leading alternative. The experiments on EgoPER~\cite{lee2024error} and EgoProceL~\cite{bansal2022my} are reported in the supplementary material and reveal similar behaviour. Based on these results, setting the $\beta$ value to $0.005$ emerges as a consistently effective choice for experiments utilizing the Direct Optimization (DO) approach, even if results remain stable for a range of choices of $\beta$ values.

\begin{table}
  \caption{(First Row) Effectiveness of the Distinctiveness Cross-Entropy Loss (DCEL). (Second Row) Average accuracy scores of the similarity matrices generated from the textual embeddings across different scenarios in the considered datasets.}
  \label{tab:ablation_DCEL_complete}
  \centering
  \resizebox{\linewidth}{!}{
  \setlength\tabcolsep{4pt}
      \begin{tabular}{lcccccc}
        \toprule
        & \multicolumn{2}{c}{CaptainCook4D} & \multicolumn{2}{c}{EgoPER} & \multicolumn{2}{c}{EgoProceL}  \\
        \cmidrule(lr){2-3} \cmidrule(lr){4-5} \cmidrule(lr){6-7}
        Metrics & \multicolumn{1}{c}{\textit{w/o} DCEL} & \multicolumn{1}{c}{Full} & \multicolumn{1}{c}{\textit{w/o} DCEL} & \multicolumn{1}{c}{Full} & \multicolumn{1}{c}{\textit{w/o} DCEL} & \multicolumn{1}{c}{Full}\\
        \midrule
        F$_1$ & {80.0 \scriptsize \(\pm 8.0\)} & {\textbf{80.8} \scriptsize \(\pm 8.0\)} & {\textbf{85.0} \scriptsize \(\pm 8.8\)} & {\textbf{85.0} \scriptsize \(\pm 8.8\)} & {66.3 \scriptsize \(\pm 3.0\)} & {\textbf{66.9} \scriptsize \(\pm 3.0\)}\\
        Avg. Accuracy & \multicolumn{2}{c}{73.1} & \multicolumn{2}{c}{81.6} & \multicolumn{2}{c}{40.0} \\
        \bottomrule
      \end{tabular}
      }
\end{table}

\subsubsection{Effectiveness of the Distinctiveness Cross-Entropy Loss (DCEL) in TGT}
\label{sec:ablation_DCEL}
In the first row of the Table~\ref{tab:ablation_DCEL_complete}, we evaluate the impact of the DCEL on the TGT model's performance across the CaptainCook4D~\cite{peddi2023captaincook4d}, EgoPER~\cite{lee2024error}, and EgoProceL~\cite{bansal2022my} datasets. Comparisons are made between TGT with and without the DCEL component. For CaptainCook4D (columns 1-2 of the Table~\ref{tab:ablation_DCEL_complete}), incorporating DCEL yields small improvements with gains of $+0.8$ in F$_1$ score. Similarly, in EgoProceL (columns 5-6 of the Table~\ref{tab:ablation_DCEL_complete}), the inclusion of DCEL leads to modest but measurable increases in F$_1$ score ($+0.6$). In contrast, in EgoPER (columns 3-4 of the Table~\ref{tab:ablation_DCEL_complete}), no performance difference is observed considering or not DCEL, suggesting that its contribution is dataset-dependent. To investigate this, we report in the second row of the Table~\ref{tab:ablation_DCEL_complete} the average accuracy scores of the similarity matrices generated from the textual embeddings across different scenarios in the considered datasets. The high similarity average accuracy in EgoPER ($81.6$) indicates that the dataset inherently supports an effective distinction between actions, reducing the need for DCEL to improve performance. In contrast, lower similarity average accuracy in CaptainCook4D ($73.1$) and EgoProceL ($40.0$) underscores the added value of DCEL in these datasets, where the task graph requires more explicit learning of distinctive features. These findings suggest that the effectiveness of DCEL is closely tied to the inherent distinctiveness of action representations within each dataset. While DCEL proves crucial in datasets with less distinctive action embeddings, it has limited impact in scenarios where the underlying action representations are already well-separated.

\subsubsection{Role of Action Recognition Accuracy in Online  Mistake Detection}
\label{sec:ablation_oad}
Table~\ref{tab:online_mistake_detection_results} shows that our model works even in the presence of imperfect predictions. To investigate the effect of noise, we conducted an analysis based on the controlled perturbation of ground truth action sequences, with the aim to simulate noise in the action detection process. At inference, we perturbed each key-step with a probability $p$ (the ``perturbation rate''), with three kinds of perturbations: insert (inserting a new key-step with a random action class), delete (deleting a key-step), or replace (randomly changing the class of a key-step). For each prediction, we perform a replace with probability $p$ on the current key-step, then for each previous key-step we perform either a replace, delete or insert with probability $p$. The results presented in Figure~\ref{fig:test} show how our system is significantly impacted by the quality of the action recognition module, thus failing to detect an action can result in incorrectly signaling a missing pre-condition, while false positives in action detection may prevent the system from identifying actual mistakes. Advancements in online action recognition technology will be critical to improving the robustness and reliability of the proposed method as well as procedure understanding in general.

\section{Conclusion}
We addressed the challenge of learning task graph representations of procedures from video demonstrations. By framing task graph learning as a maximum likelihood estimation problem, we introduced a new differentiable loss function that enables direct optimization of the adjacency matrix via gradient descent and can be integrated into complex neural network architectures. Experiments conducted on six datasets demonstrate that the proposed approach not only learns accurate task graphs, but also enhances video understanding capabilities and improves performance on the downstream task of online mistake detection, surpassing state-of-the-art methods. Furthermore, task graphs generated using our approach achieve top performance in the Ego-Exo4D procedure understanding benchmark. The implementation of our methods is publicly available at \url{https://github.com/fpv-iplab/Differentiable-Task-Graph-Learning}.

\section*{Acknowledgments}
This research is supported in part by the PNRR PhD scholarship ``Digital Innovation: Models, Systems and Applications'' DM 118/2023, by the project Future Artificial Intelligence Research (FAIR) – PNRR MUR Cod. PE0000013 - CUP: E63C22001940006, and by the Research Program PIAno di inCEntivi per la Ricerca di Ateneo 2020/2022 — Linea di Intervento 3 ``Starting Grant'' EVIPORES Project - University of Catania.

\bibliographystyle{IEEEtran}
\bibliography{bibtex}

\begin{thebibliography}{10}
\providecommand{\url}[1]{#1}
\csname url@samestyle\endcsname
\providecommand{\newblock}{\relax}
\providecommand{\bibinfo}[2]{#2}
\providecommand{\BIBentrySTDinterwordspacing}{\spaceskip=0pt\relax}
\providecommand{\BIBentryALTinterwordstretchfactor}{4}
\providecommand{\BIBentryALTinterwordspacing}{\spaceskip=\fontdimen2\font plus
\BIBentryALTinterwordstretchfactor\fontdimen3\font minus \fontdimen4\font\relax}
\providecommand{\BIBforeignlanguage}[2]{{%
\expandafter\ifx\csname l@#1\endcsname\relax
\typeout{** WARNING: IEEEtran.bst: No hyphenation pattern has been}%
\typeout{** loaded for the language `#1'. Using the pattern for}%
\typeout{** the default language instead.}%
\else
\language=\csname l@#1\endcsname
\fi
#2}}
\providecommand{\BIBdecl}{\relax}
\BIBdecl

\bibitem{kanade2012first}
T.~Kanade and M.~Hebert, ``First-person vision,'' \emph{Proceedings of the IEEE}, vol. 100, no.~8, pp. 2442--2453, 2012.

\bibitem{plizzari2023outlook}
C.~Plizzari, G.~Goletto, A.~Furnari, S.~Bansal, F.~Ragusa, G.~M. Farinella, D.~Damen, and T.~Tommasi, ``An outlook into the future of egocentric vision,'' \emph{International Journal fn Computer Vision}, 2023.

\bibitem{dvornik2022graph2vid}
N.~Dvornik, I.~Hadji, H.~Pham, D.~Bhatt, B.~Martinez, A.~Fazly, and A.~D. Jepson, ``Graph2vid: Flow graph to video grounding for weakly-supervised multi-step localization,'' in \emph{Proceedings of the European Conference on Computer Vision (ECCV)}, 2022.

\bibitem{ashutosh2024video}
K.~Ashutosh, S.~K. Ramakrishnan, T.~Afouras, and K.~Grauman, ``Video-mined task graphs for keystep recognition in instructional videos,'' \emph{Advances in Neural Information Processing Systems}, vol.~36, 2024.

\bibitem{grauman2023ego}
K.~Grauman, A.~Westbury, L.~Torresani, K.~Kitani, J.~Malik, T.~Afouras, K.~Ashutosh, V.~Baiyya, S.~Bansal, B.~Boote \emph{et~al.}, ``Ego-exo4d: Understanding skilled human activity from first-and third-person perspectives,'' in \emph{Proceedings of the IEEE/CVF Conference on Computer Vision and Pattern Recognition}, 2024, pp. 19\,383--19\,400.

\bibitem{zhou2023procedure}
H.~Zhou, R.~Mart{\'\i}n-Mart{\'\i}n, M.~Kapadia, S.~Savarese, and J.~C. Niebles, ``Procedure-aware pretraining for instructional video understanding,'' in \emph{Proceedings of the IEEE/CVF Conference on Computer Vision and Pattern Recognition}, 2023, pp. 10\,727--10\,738.

\bibitem{seminara2024differentiable}
\BIBentryALTinterwordspacing
L.~Seminara, G.~M. Farinella, and A.~Furnari, ``Differentiable task graph learning: Procedural activity representation and online mistake detection from egocentric videos,'' in \emph{The Thirty-eighth Annual Conference on Neural Information Processing Systems}, 2024. [Online]. Available: \url{https://openreview.net/forum?id=2HvgvB4aWq}
\BIBentrySTDinterwordspacing

\bibitem{peddi2023captaincook4d}
\BIBentryALTinterwordspacing
R.~Peddi, S.~Arya, B.~Challa, L.~Pallapothula, A.~Vyas, B.~Gouripeddi, Q.~Zhang, J.~Wang, V.~Komaragiri, E.~Ragan, N.~Ruozzi, Y.~Xiang, and V.~Gogate, ``Captaincook4d: A dataset for understanding errors in procedural activities,'' in \emph{The Thirty-eight Conference on Neural Information Processing Systems Datasets and Benchmarks Track}, 2024. [Online]. Available: \url{https://openreview.net/forum?id=YFUp7zMrM9}
\BIBentrySTDinterwordspacing

\bibitem{lee2024error}
S.-P. Lee, Z.~Lu, Z.~Zhang, M.~Hoai, and E.~Elhamifar, ``Error detection in egocentric procedural task videos,'' in \emph{Proceedings of the IEEE/CVF Conference on Computer Vision and Pattern Recognition}, 2024, pp. 18\,655--18\,666.

\bibitem{bansal2022my}
S.~Bansal, C.~Arora, and C.~Jawahar, ``My view is the best view: Procedure learning from egocentric videos,'' in \emph{European Conference on Computer Vision}.\hskip 1em plus 0.5em minus 0.4em\relax Springer, 2022, pp. 657--675.

\bibitem{flaborea2024prego}
A.~Flaborea, G.~M.~D. di~Melendugno, L.~Plini, L.~Scofano, E.~De~Matteis, A.~Furnari, G.~M. Farinella, and F.~Galasso, ``Prego: online mistake detection in procedural egocentric videos,'' in \emph{Proceedings of the IEEE/CVF Conference on Computer Vision and Pattern Recognition}, 2024, pp. 18\,483--18\,492.

\bibitem{sener2022assembly101}
F.~Sener, D.~Chatterjee, D.~Shelepov, K.~He, D.~Singhania, R.~Wang, and A.~Yao, ``Assembly101: A large-scale multi-view video dataset for understanding procedural activities,'' in \emph{Proceedings of the IEEE/CVF Conference on Computer Vision and Pattern Recognition}, 2022, pp. 21\,096--21\,106.

\bibitem{jang2019epic}
Y.~Jang, B.~Sullivan, C.~Ludwig, I.~Gilchrist, D.~Damen, and W.~Mayol-Cuevas, ``Epic-tent: An egocentric video dataset for camping tent assembly,'' in \emph{Proceedings of the IEEE/CVF International Conference on Computer Vision Workshops}, 2019, pp. 0--0.

\bibitem{zhou2018towards}
L.~Zhou, C.~Xu, and J.~Corso, ``Towards automatic learning of procedures from web instructional videos,'' in \emph{Proceedings of the AAAI Conference on Artificial Intelligence}, vol.~32, no.~1, 2018.

\bibitem{zhukov2019cross}
D.~Zhukov, J.-B. Alayrac, R.~G. Cinbis, D.~Fouhey, I.~Laptev, and J.~Sivic, ``Cross-task weakly supervised learning from instructional videos,'' in \emph{Proceedings of the IEEE/CVF Conference on Computer Vision and Pattern Recognition}, 2019, pp. 3537--3545.

\bibitem{elhamifar2020self}
E.~Elhamifar and D.~Huynh, ``Self-supervised multi-task procedure learning from instructional videos,'' in \emph{Computer Vision--ECCV 2020: 16th European Conference, Glasgow, UK, August 23--28, 2020, Proceedings, Part XVII 16}.\hskip 1em plus 0.5em minus 0.4em\relax Springer, 2020, pp. 557--573.

\bibitem{bansal2024united}
S.~Bansal, C.~Arora, and C.~Jawahar, ``United we stand, divided we fall: Unitygraph for unsupervised procedure learning from videos,'' in \emph{2024 IEEE/CVF Winter Conference on Applications of Computer Vision (WACV)}, 2024, pp. 6495--6505.

\bibitem{dvornik2023stepformer}
N.~Dvornik, I.~Hadji, R.~Zhang, K.~G. Derpanis, R.~P. Wildes, and A.~D. Jepson, ``Stepformer: Self-supervised step discovery and localization in instructional videos,'' in \emph{Proceedings of the IEEE/CVF Conference on Computer Vision and Pattern Recognition}, 2023, pp. 18\,952--18\,961.

\bibitem{lu2022set}
Z.~Lu and E.~Elhamifar, ``Set-supervised action learning in procedural task videos via pairwise order consistency,'' in \emph{Proceedings of the IEEE/CVF Conference on Computer Vision and Pattern Recognition}, 2022, pp. 19\,903--19\,913.

\bibitem{miech2020end}
A.~Miech, J.-B. Alayrac, L.~Smaira, I.~Laptev, J.~Sivic, and A.~Zisserman, ``End-to-end learning of visual representations from uncurated instructional videos,'' in \emph{Proceedings of the IEEE/CVF conference on computer vision and pattern recognition}, 2020, pp. 9879--9889.

\bibitem{narasimhan2023learning}
M.~Narasimhan, L.~Yu, S.~Bell, N.~Zhang, and T.~Darrell, ``Learning and verification of task structure in instructional videos,'' \emph{arXiv preprint arXiv:2303.13519}, 2023.

\bibitem{hazra2023egotv}
R.~Hazra, B.~Chen, A.~Rai, N.~Kamra, and R.~Desai, ``Egotv: Egocentric task verification from natural language task descriptions,'' in \emph{Proceedings of the IEEE/CVF International Conference on Computer Vision}, 2023, pp. 15\,417--15\,429.

\bibitem{wang2023holoassist}
X.~Wang, T.~Kwon, M.~Rad, B.~Pan, I.~Chakraborty, S.~Andrist, D.~Bohus, A.~Feniello, B.~Tekin, F.~V. Frujeri \emph{et~al.}, ``Holoassist: an egocentric human interaction dataset for interactive ai assistants in the real world,'' in \emph{Proceedings of the IEEE/CVF International Conference on Computer Vision}, 2023, pp. 20\,270--20\,281.

\bibitem{ghoddoosian2023weakly}
R.~Ghoddoosian, I.~Dwivedi, N.~Agarwal, and B.~Dariush, ``Weakly-supervised action segmentation and unseen error detection in anomalous instructional videos,'' in \emph{Proceedings of the IEEE/CVF International Conference on Computer Vision}, 2023, pp. 10\,128--10\,138.

\bibitem{ding2023every}
G.~Ding, F.~Sener, S.~Ma, and A.~Yao, ``Every mistake counts in assembly,'' \emph{arXiv preprint arXiv:2307.16453}, 2023.

\bibitem{nagasinghe2024not}
K.~R.~Y. Nagasinghe, H.~Zhou, M.~Gunawardhana, M.~R. Min, D.~Harari, and M.~H. Khan, ``Why not use your textbook? knowledge-enhanced procedure planning of instructional videos,'' in \emph{Proceedings of the IEEE/CVF Conference on Computer Vision and Pattern Recognition}, 2024, pp. 18\,816--18\,826.

\bibitem{shen2024progress}
Y.~Shen and E.~Elhamifar, ``Progress-aware online action segmentation for egocentric procedural task videos,'' in \emph{Proceedings of the IEEE/CVF Conference on Computer Vision and Pattern Recognition}, 2024, pp. 18\,186--18\,197.

\bibitem{zhong2023learning}
Y.~Zhong, L.~Yu, Y.~Bai, S.~Li, X.~Yan, and Y.~Li, ``Learning procedure-aware video representation from instructional videos and their narrations,'' in \emph{Proceedings of the IEEE/CVF Conference on Computer Vision and Pattern Recognition}, 2023, pp. 14\,825--14\,835.

\bibitem{sohn2020meta}
S.~Sohn, H.~Woo, J.~Choi, and H.~Lee, ``Meta reinforcement learning with autonomous inference of subtask dependencies,'' \emph{arXiv preprint arXiv:2001.00248}, 2020.

\bibitem{jang2023multimodal}
Y.~Jang, S.~Sohn, L.~Logeswaran, T.~Luo, M.~Lee, and H.~Lee, ``Multimodal subtask graph generation from instructional videos,'' \emph{arXiv preprint arXiv:2302.08672}, 2023.

\bibitem{skiena1998algorithm}
S.~S. Skiena, \emph{The algorithm design manual}.\hskip 1em plus 0.5em minus 0.4em\relax Springer, 1998, vol.~2.

\bibitem{schumacher2012extraction}
P.~Schumacher, M.~Minor, K.~Walter, and R.~Bergmann, ``Extraction of procedural knowledge from the web: A comparison of two workflow extraction approaches,'' in \emph{Proceedings of the 21st International Conference on World Wide Web}, 2012, pp. 739--747.

\bibitem{kiddon2015mise}
C.~Kiddon, G.~T. Ponnuraj, L.~Zettlemoyer, and Y.~Choi, ``Mise en place: Unsupervised interpretation of instructional recipes,'' in \emph{Proceedings of the 2015 Conference on Empirical Methods in Natural Language Processing}, 2015, pp. 982--992.

\bibitem{sakaguchi2021proscript}
\BIBentryALTinterwordspacing
K.~Sakaguchi, C.~Bhagavatula, R.~Le~Bras, N.~Tandon, P.~Clark, and Y.~Choi, ``pro{S}cript: Partially ordered scripts generation,'' in \emph{Findings of the Association for Computational Linguistics: EMNLP 2021}, M.-F. Moens, X.~Huang, L.~Specia, and S.~W.-t. Yih, Eds.\hskip 1em plus 0.5em minus 0.4em\relax Punta Cana, Dominican Republic: Association for Computational Linguistics, Nov. 2021, pp. 2138--2149. [Online]. Available: \url{https://aclanthology.org/2021.findings-emnlp.184}
\BIBentrySTDinterwordspacing

\bibitem{donatelli2021aligning}
L.~Donatelli, T.~Schmidt, D.~Biswas, A.~K{\"o}hn, F.~Zhai, and A.~Koller, ``Aligning actions across recipe graphs,'' in \emph{Proceedings of the 2021 conference on empirical methods in natural language processing}, 2021, pp. 6930--6942.

\bibitem{yamakata2020english}
Y.~Yamakata, S.~Mori, and J.~A. Carroll, ``English recipe flow graph corpus,'' in \emph{Proceedings of the Twelfth Language Resources and Evaluation Conference}, 2020, pp. 5187--5194.

\bibitem{marquis1820theorie}
P.~S. Marquis~de Laplace, \emph{Th{\'e}orie analytique des probabilit{\'e}s}.\hskip 1em plus 0.5em minus 0.4em\relax Courcier, 1820, vol.~7.

\bibitem{oord2018representation}
A.~v.~d. Oord, Y.~Li, and O.~Vinyals, ``Representation learning with contrastive predictive coding,'' \emph{arXiv preprint arXiv:1807.03748}, 2018.

\bibitem{radford2021learning}
A.~Radford, J.~W. Kim, C.~Hallacy, A.~Ramesh, G.~Goh, S.~Agarwal, G.~Sastry, A.~Askell, P.~Mishkin, J.~Clark \emph{et~al.}, ``Learning transferable visual models from natural language supervision,'' in \emph{International conference on machine learning}.\hskip 1em plus 0.5em minus 0.4em\relax PMLR, 2021, pp. 8748--8763.

\bibitem{vaswani2017attention}
A.~Vaswani, N.~Shazeer, N.~Parmar, J.~Uszkoreit, L.~Jones, A.~N. Gomez, {\L}.~Kaiser, and I.~Polosukhin, ``Attention is all you need,'' \emph{Advances in neural information processing systems}, vol.~30, 2017.

\bibitem{de2009guide}
F.~De~la Torre, J.~Hodgins, A.~Bargteil, X.~Martin, J.~Macey, A.~Collado, and P.~Beltran, ``Guide to the carnegie mellon university multimodal activity (cmu-mmac) database,'' 2009.

\bibitem{li2018eye}
Y.~Li, M.~Liu, and J.~M. Rehg, ``In the eye of beholder: Joint learning of gaze and actions in first person video,'' in \emph{Proceedings of the European conference on computer vision (ECCV)}, 2018, pp. 619--635.

\bibitem{ragusa2021meccano}
F.~Ragusa, A.~Furnari, S.~Livatino, and G.~M. Farinella, ``The meccano dataset: Understanding human-object interactions from egocentric videos in an industrial-like domain,'' in \emph{Proceedings of the IEEE/CVF Winter Conference on Applications of Computer Vision}, 2021, pp. 1569--1578.

\bibitem{dubey2024llama}
A.~Dubey, A.~Jauhri, A.~Pandey, A.~Kadian, A.~Al-Dahle, A.~Letman, A.~Mathur, A.~Schelten, A.~Yang, A.~Fan \emph{et~al.}, ``The llama 3 herd of models,'' \emph{arXiv preprint arXiv:2407.21783}, 2024.

\bibitem{pramanick2023egovlpv2}
S.~Pramanick, Y.~Song, S.~Nag, K.~Q. Lin, H.~Shah, M.~Z. Shou, R.~Chellappa, and P.~Zhang, ``Egovlpv2: Egocentric video-language pre-training with fusion in the backbone,'' in \emph{Proceedings of the IEEE/CVF International Conference on Computer Vision}, 2023, pp. 5285--5297.

\bibitem{zhou2015temporal}
Y.~Zhou and T.~L. Berg, ``Temporal perception and prediction in ego-centric video,'' in \emph{Proceedings of the IEEE International Conference on Computer Vision}, 2015, pp. 4498--4506.

\bibitem{an2023miniroad}
J.~An, H.~Kang, S.~H. Han, M.-H. Yang, and S.~J. Kim, ``Miniroad: Minimal rnn framework for online action detection,'' in \emph{Proceedings of the IEEE/CVF International Conference on Computer Vision}, 2023, pp. 10\,341--10\,350.

\end{thebibliography}

\vfill

\newpage

{\appendices

\section{Implementation Details}
In this section, we provide detailed supplementary information. First, we outline the dataset splits used in our experiments (Section~\ref{app:split}). Next, we introduce the early stopping procedure with the Sequence Accuracy score, which helps prevent overfitting and reduces computational costs (Section~\ref{app:sequence_accuracy}). We then present an overview of the hyperparameters used for task graph generation (Section~\ref{app:hyperparameters}). 
We hence provide details on the Pairwise Ordering, Future Prediction, and Online Mistake Detection experiments (Sections~\ref{app:video_understanding} and \ref{app:omd}). Following this, we outline the prompts used for Llama-3.1-405B-Instruct~\cite{dubey2024llama} (Section~\ref{app:llama}). Finally, we discuss the computational resources employed in our experiments (Section~\ref{app:resources}).

\subsection{Data Split}
\label{app:split}
The CaptainCook4D dataset~\cite{peddi2023captaincook4d} includes various types of errors, such as order errors, timing errors, temperature errors, preparation errors, missing steps errors, measurement errors, and technique errors. Among these, missing steps and order errors directly affect the integrity of the action sequences. Therefore, for task graph generation, we selected only those action sequences that were free from these specific errors. Table~\ref{tab:data_split_captaincook4D} provides statistics on the subsets of the CaptainCook4D dataset used in task graph generation. For the EgoPER dataset~\cite{lee2024error} (see Table~\ref{tab:data_split_egoper}), we utilized all the correct/normal video sequences as defined by the authors. In the case of EgoProceL (see Table~\ref{tab:data_split_egoprocel}), we included all sequences from the CMU-MMAC~\cite{de2009guide}, EGTEA Gaze+~\cite{li2018eye}, and EPIC-tent~\cite{jang2019epic} datasets.
In the context of pairwise ordering and forecasting, we employed the subset of the CaptainCook4D dataset designated for task graph generation (refer to Table~\ref{tab:data_split_captaincook4D}) and divided it into training and testing sets. This division was carefully managed to ensure that 50\% of the scenarios were equally represented in both the training and testing sets.
In the Ego-Exo4D~\cite{grauman2023ego} procedure understanding benchmark, we adhered to the official train, validation, and test splits.
For Online Mistake Detection, we considered the datasets defined by the authors of PREGO~\cite{flaborea2024prego}.

\begin{table}
\centering
\caption{A detailed breakdown of the data used from the CaptainCook4D dataset~\cite{peddi2023captaincook4d} for task graph generation. This table categorizes each scenario by the number of videos, segments, and total duration in hours of the video segments. The ``Total'' row aggregates the dataset characteristics.}
\label{tab:data_split_captaincook4D}
\begin{tabular}{lccc}
\toprule
Scenario & Videos & Segments & Duration (h) \\
\midrule
Microwave Egg Sandwich & 5 & 60 & 0.8 \\
Dressed Up Meatballs & 8 & 128 & 2.6 \\
Microwave Mug Pizza & 6 & 84 & 0.9 \\
Ramen & 11 & 165 & 2.0 \\
Coffee & 9 & 144 & 2.2 \\
Breakfast Burritos & 8 & 88 & 1.4 \\
Spiced Hot Chocolate & 7 & 49 & 0.9 \\
Microwave French Toast & 11 & 121 & 1.9 \\
Pinwheels & 5 & 95 & 0.7 \\
Tomato Mozzarella Salad & 13 & 117 & 0.7 \\
Butter Corn Cup & 5 & 60 & 1.0 \\
Tomato Chutney & 5 & 95 & 2.3 \\
Scrambled Eggs & 6 & 138 & 2.3 \\
Cucumber Raita & 12 & 132 & 2.4 \\
Zoodles & 6 & 78 & 1.4 \\
Sauted Mushrooms & 7 & 126 & 3.1 \\
Blender Banana Pancakes & 10 & 140 & 1.8 \\
Herb Omelet with Fried Tomatoes & 8 & 120 & 2.2 \\
Broccoli Stir Fry & 10 & 250 & 4.8 \\
Pan Fried Tofu & 9 & 171 & 2.6 \\
Mug Cake & 9 & 180 & 2.7 \\
Cheese Pimiento & 7 & 77 & 1.3 \\
Spicy Tuna Avocado Wraps & 9 & 153 & 2.5 \\
Caprese Bruschetta & 8 & 88 & 2.1 \\
\rowcolor{teal!30} Total & 194 & 2859 & 46.5 \\
\bottomrule
\end{tabular}
\end{table}

\begin{table}
\centering
\caption{A detailed breakdown of the data used from the EgoPER dataset~\cite{lee2024error} for task graph generation. This table categorizes each scenario by the number of videos, segments, and total duration in hours of the video segments. The ``Total'' row aggregates the dataset characteristics.}
\label{tab:data_split_egoper}
\begin{tabular}{lccc}
\toprule
Scenario & Videos & Segments & Duration (h) \\
\midrule
Coffee & 33 & 583 & 2.6 \\
Pinwheels & 42 & 838 & 3.2 \\
Oatmeal & 45 & 673 & 2.7 \\
Quesadilla & 48 & 384 & 1.0 \\
Tea & 48 & 461 & 1.5 \\
\rowcolor{teal!30} Total & 216 & 2939 & 11.0 \\
\bottomrule
\end{tabular}
\end{table}

\begin{table}
\centering
\caption{A detailed breakdown of the data used from the EgoProceL dataset~\cite{bansal2022my} for task graph generation. This table categorizes each scenario by the number of videos, segments, and total duration in hours of the video segments. The ``Total'' row aggregates the dataset characteristics.}
\label{tab:data_split_egoprocel}
\resizebox{!}{0.27\columnwidth}{
\begin{tabular}{lccc}
\toprule
Scenario & Videos & Segments & Duration (h) \\
\midrule
CMU-MMAC Brownie & 34 & 332 & 1.7 \\
CMU-MMAC Eggs & 33 & 340 & 0.7 \\
CMU-MMAC Pizza & 33 & 223 & 2.5 \\
CMU-MMAC Salad & 29 & 211 & 1.0 \\
CMU-MMAC Sandwich & 31 & 191 & 0.4 \\
EGTEA Gaze+ Bacon and Eggs & 16 & 279 & 0.9 \\
EGTEA Gaze+ Cheeseburger & 10 & 226 & 0.5 \\
EGTEA Gaze+ Continental Breakfast & 12 & 150 & 0.5 \\
EGTEA Gaze+ Greek Salad & 10 & 145 & 0.5 \\
EGTEA Gaze+ Pasta Salad & 19 & 889 & 2.3 \\
EGTEA Gaze+ Pizza & 6 & 110 & 0.4 \\
EGTEA Gaze+ Turkey Sandwich & 13 & 159 & 0.5 \\
EPIC-Tent & 29 & 1089 & 4.4 \\
\rowcolor{teal!30} Total & 275 & 4344 & 16.3 \\
\bottomrule
\end{tabular}
}
\end{table}

\subsection{Sequence Accuracy (SA) Score}
\label{app:sequence_accuracy}
We employed an early stopping strategy to avoid overfitting and reduce training time. Since the task is weakly supervised and lacks a clear metric to maximize, we define a ``Sequence Accuracy (SA)'' score to detect when the model reaches a learning plateau:
\begin{equation}
     \text{SA} = \frac{1}{|\mathcal{Y}|} \sum_{y^{(d)} \in \mathcal{Y}} \frac{1}{|y^{(d)}|} \sum_{i=0}^{|y^{(d)}|-1} c(y^{(d)}_i, y^{(d)}[:i], Pred(y^{(d)}_i, Z))
\end{equation} 
where \( \mathcal{Y} \) defined sequences in the training set, \( y^{(d)} \) is a sequence from \( \mathcal{Y} \), \( y^{(d)}_i \) is the \( i \)-th element of sequence \( y^{(d)} \), \(y^{(d)}[:i]\) are the predecessors of the \(i\)-th element in the sequence $y^{(d)}$, and \( Pred(y^{(d)}_i, Z) \) are the predicted predecessors for \( y^{(d)}_i \) from the current binarized adjacency matrix $Z$. The function \( c \) is defined in Eq.~\eqref{eq:sequence_score}. The SA score measures the compatibility of each sequence with the current task graph based on the ratio of correctly predicted predecessors of the current symbol $y^{(d)}_i$ of the sequence to the total number of predicted predecessors for $y^{(d)}_i$ in the current task graph. This score is primarily applied to the training set and, when available, also to the validation set.

\begin{figure*}[t]
\begin{align}
    \label{eq:sequence_score}
     &c(y^{(d)}_i, y^{(d)}[:i], Pred(y^{(d)}_i, Z)) = \begin{cases} 
       1 & \text{if } |y^{(d)}[:i]| = 0 \text{ and } |Pred(y^{(d)}_i, Z)| = 0 \\
       -\frac{1}{|y^{(d)}[:i]|} & \text{if } |y^{(d)}[:i]| = 0 \text{ and } |Pred(y^{(d)}_i, Z)| > 0 \\
       \frac{|y^{(d)}[:i] \cap Pred(y^{(d)}_i, Z)|}{|Pred(y^{(d)}_i, Z)|} & \text{if } |y^{(d)}[:i]| > 0 \text{ and } |Pred(y^{(d)}_i, Z)| > 0 \\ 
       0 & \text{otherwise}
    \end{cases}
\end{align}
\end{figure*}

\begin{table*}
    \begin{minipage}[t]{0.49\textwidth}
    \centering
    \caption{List of hyperparameters used in the models training process for task graph generation using CaptainCook4D~\cite{peddi2023captaincook4d}, EgoPER~\cite{lee2024error}, and EgoProceL~\cite{bansal2022my}.}
    \label{tab:hyperparameters1}
    \begin{tabular}{lcc}
    \toprule
    & \multicolumn{2}{c}{Value} \\
    \cmidrule(lr){2-3}
    Hyperparameter & DO & TGT \\
    \midrule
    Learning Rate & 0.1 & $1 \times 10^{-6}$ \\
    Max Epochs & 1000 & 3000\\
    Optimizer & Adam & Adam\\
    $\beta$ & 0.50 / 0.005 & 1.0 \(\sim\) 0.50 / 1.0 \(\sim\) 0.05\\
    Dropout Rate & - & 0.25\\
    \bottomrule
    \end{tabular}
    \end{minipage}
    \hspace{1.5mm}
    \begin{minipage}[t]{0.45\textwidth}
    \centering
    \caption{List of hyperparameters used in the models training process for task graph generation using \\Ego-Exo4D~\cite{grauman2023ego}.}
    \label{tab:hyperparameters3}
    \begin{tabular}{lcc}
    \toprule
    & \multicolumn{2}{c}{Value} \\
    \cmidrule(lr){2-3}
    Hyperparameter & DO & TGT \\
    \midrule
    Learning Rate & 0.1 & $1 \times 10^{-6}$ \\
    Max Epochs & 300 & 3000\\
    Optimizer & Adam & Adam\\
    $\beta$ & [0.005, 1.0] & 1.0 \(\sim\) 0.05\\
    Dropout Rate & - & 0.1\\
    \bottomrule
    \end{tabular}
    \end{minipage}
\end{table*}

\begin{table}[t]
    \centering
    \caption{List of hyperparameters used in the models training process for task graph generation using Assembly101-O and EPIC-Tent-O~\cite{flaborea2024prego}.}
    \label{tab:hyperparameters2}
    \begin{tabular}{lcc}
    \toprule
    & \multicolumn{2}{c}{Value} \\
    \cmidrule(lr){2-3}
    Hyperparameter & DO & TGT \\
    \midrule
    Learning Rate & 0.1 & $1 \times 10^{-6}$ / $1.5 \times 10^{-5}$ \\
    Max Epochs & 1200 & 1200\\
    Optimizer & Adam & Adam\\
    $\beta$ & 0.005 & 1.0 \(\sim\) 0.55\\
    Dropout Rate & - & 0.1\\
    \bottomrule
    \end{tabular}
\end{table}

\subsection{Hyperparameters}
\label{app:hyperparameters}
Table~\ref{tab:hyperparameters1} provides an overview of the hyperparameters used in the task graph generation experiments conducted on the CaptainCook4D~\cite{peddi2023captaincook4d}, EgoPER~\cite{lee2024error}, and EgoProceL~\cite{bansal2022my} datasets. For the DO method, a learning rate of $0.1$ was employed with the Adam optimizer, and training was limited to a maximum of 1000 epochs.
The $\beta$ parameter was set to $0.005$ for all scenarios. However, we found it advantageous to increase it to $0.50$ in cases where sequences do not include all taxonomy-defined steps. This adjustment enhances the contrastive term, particularly when complete sequence examples are absent, i.e. those containing all the taxonomy-defined steps that represent a typical way to complete the procedure. 
The training of the DO method was halted early when the SA score reached at least $0.95$ and no improvement in the SA score was observed over $25$ consecutive epochs.
For the training of TGT models, we utilized a pre-trained EgoVLPv2~\cite{pramanick2023egovlpv2} on Ego-Exo4D~\cite{grauman2023ego} to extract text and video embeddings. The temperature value $T$ used in the distinctiveness cross-entropy loss (DCEL) was set to $0.9$ as in~\cite{radford2021learning}. We employed a learning rate of $1 \times 10^{-6}$ with the Adam optimizer, and training was limited to a maximum of 3000 epochs. The $\beta$ parameter was linearly annealed from an initial value of $1.0$ to a final value of either $0.50$ or $0.05$, with updates occurring every 100 epochs. This annealing process follows the warm-up strategy introduced in~\cite{vaswani2017attention}, enabling smoother optimization during the initial training phase and improved convergence in later stages. The final value of $\beta$ is determined by the characteristics of the video sequences in the dataset, consistent with the approach used in the DO method: the final value of $\beta$ is set to $0.50$ when none of the sequences include all taxonomy-defined steps, thereby enhancing the contrastive term throughout the training process.

Table~\ref{tab:hyperparameters3} outlines the hyperparameters used for training the models on Ego-Exo4D~\cite{grauman2023ego}. The DO configuration matches that of Table~\ref{tab:hyperparameters1}, except for two key adjustments: the $\beta$ parameter and the reduction of epochs to $300$.  In line with the validation annotation guidelines from~\cite{grauman2023ego}, the validation set was employed to determine the optimal value of $\beta$. The values tested for $\beta$ were $[0.005, 0.05, 0.1, 0.25, 0.5, 0.75, 1.0]$, with the SA score guiding the selection of the best $\beta$ and the most suitable task graph. This necessitated reducing the number of training epochs to 300 for efficient validation. Similarly, the TGT settings align with those in Table~\ref{tab:hyperparameters1}, with the sole difference being the dropout rate, which was set to $0.1$.

Table~\ref{tab:hyperparameters2} details the hyperparameters employed for training the models on the Assembly101-O and EPIC-Tent-O datasets~\cite{flaborea2024prego}. For the DO method, a learning rate of $0.1$ was used with the Adam optimizer, and training was limited to a maximum of 1200 epochs, more than the previous settings. This change was necessary because on Assembly101-O, even after 1000 epochs, the model continued to exhibit many cycles among its 86 nodes. Extending the number of epochs allows the model additional time to learn and minimize these cycles, which is crucial given the complexity of the graph. 
In the TGT configuration, we set the dropout rate to $0.1$, while the $\beta$ parameter was gradually annealed from an initial value of $1.0$ to $0.55$ to prevent overfitting. For Assembly101-O, a learning rate of $1 \times 10^{-6}$ was employed, as the larger and more complex structure of this dataset required a slower rate for stable convergence. In contrast, EPIC-Tent-O demonstrated better performance with a higher learning rate of $1.5 \times 10^{-5}$, likely due to its comparatively simpler structure, allowing for faster optimization without sacrificing stability.
The reader is referred to the code for additional implementation details.

\begin{figure}
    \centering
    \includegraphics[width=1.0\linewidth]{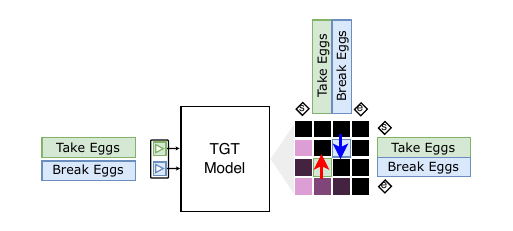}
    \caption{Pairwise ordering example. The ordering between \textit{Take Eggs} and \textit{Break Eggs} is determined by satisfying at least one of the following conditions: (a) If the weight of the edge \textit{Break Eggs} $\rightarrow$ \textit{Take Eggs} (red arrow) exceeds that of \textit{Take Eggs} $\rightarrow$ \textit{Break Eggs} (blue arrow), we infer that \textit{Take Eggs} precedes \textit{Break Eggs}. (b) By evaluating the sequences $<\text{START}, \textit{Take Eggs}, \textit{Break Eggs}, \text{END}>$ and $<\text{START}, \textit{Break Eggs}, \textit{Take Eggs}, \text{END}>$, we compute their probabilities using Eq.~\eqref{eq:factorization_2}. If $P(<\text{START}, \textit{Take Eggs}, \textit{Break Eggs}, \text{END}> \mid Z)$ is greater than $P(<\text{START}, \textit{Break Eggs}, \textit{Take Eggs}, \text{END}> \mid Z)$, we deduce that \textit{Take Eggs} precedes \textit{Break Eggs}. (c) If the weight of the edge \textit{END} $\rightarrow$ \textit{Break Eggs} is greater than that of \textit{END} $\rightarrow$ \textit{Take Eggs}, this indicates that \textit{Break Eggs} is essential for concluding the procedure. Consequently, \textit{Break Eggs} follows \textit{Take Eggs}, implying that \textit{Take Eggs} precedes \textit{Break Eggs}. If none of these conditions are satisfied, we conclude that \textit{Break Eggs} precedes \textit{Take Eggs}.}
    \label{fig:pairwise}
\end{figure}

\begin{figure}
    \centering
    \includegraphics[width=1.0\linewidth]{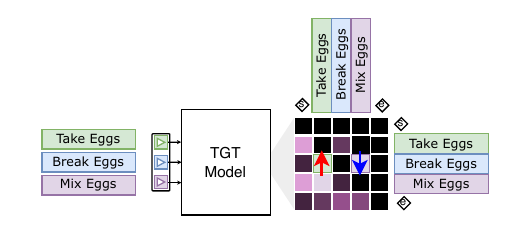}
    \caption{Future prediction example. To determine the future clip, we evaluate the weights of the edges \textit{Break Eggs} $\to$ \textit{Take Eggs} (red arrow) and \textit{Break Eggs} $\to$ \textit{Mix Eggs} (blue arrow). The clip associated with the smaller weight is selected as the future clip, as a lower weight signifies that the corresponding clip is less likely to be a pre-condition. An alternative approach involves analyzing the probabilities of the sequences $<\text{START}, \textit{Take Eggs}, \textit{Break Eggs}, \textit{Mix Eggs}, \text{END}>$ and $<\text{START}, \textit{Mix Eggs}, \textit{Break Eggs}, \textit{Take Eggs}, \text{END}>$ using Eq.~\eqref{eq:factorization_2}. If $P(<\text{START}, \textit{Take Eggs}, \textit{Break Eggs}, \textit{Mix Eggs}, \text{END}> \mid Z)$ exceeds $P(<\text{START}, \textit{Mix Eggs}, \textit{Break Eggs}, \textit{Take Eggs}, \text{END}> \mid Z)$, it implies that \textit{Mix Eggs} is the future clip relative to \textit{Break Eggs}. Conversely, if the latter probability is higher, \textit{Take Eggs} is identified as the future clip for \textit{Break Eggs}.}
    \label{fig:future}
\end{figure}

\subsection{Details on Pairwise ordering and future prediction}
\label{app:video_understanding}
We set up the pairwise ordering and future prediction video understanding tasks following~\cite{zhou2015temporal}. 

\subsubsection{Pairwise Ordering} Our model processes two clips and generates a $4 \times 4$ adjacency matrix, where the nodes correspond to \textit{START}, $A$, $B$, and \textit{END}. The ordering between $A$ and $B$ is determined by satisfying at least one of the following conditions: (a) If the weight of the edge $B \rightarrow A$ exceeds that of $A \rightarrow B$, we infer that $A$ precedes $B$. (b) By evaluating the sequences $<\text{START}, A, B, \text{END}>$ and $<\text{START}, B, A, \text{END}>$, we compute their probabilities using Eq.~\eqref{eq:factorization_2}.
If $P(<\text{START}, A, B, \text{END}> \mid Z)$ is greater than $P(<\text{START}, B, A, \text{END}> \mid Z)$, we deduce that $A$ precedes $B$. (c) If the weight of the edge \textit{END} $\rightarrow B$ is greater than that of \textit{END} $\rightarrow A$, this indicates that $B$ is essential for concluding the procedure. Consequently, $B$ follows $A$, implying that $A$ precedes $B$. If none of these conditions are satisfied, we conclude that $B$ precedes $A$ (see Figure~\ref{fig:pairwise}).

\subsubsection{Future Prediction} Our model processes three clips and generates a $5 \times 5$ adjacency matrix, where the nodes correspond to \textit{START}, $A$, $anchor$, $B$, and \textit{END}. To determine the future clip, we evaluate the weights of the edges $anchor \to A$ and $anchor \to B$. The clip associated with the smaller weight is selected as the future clip, as a lower weight signifies that the corresponding clip is less likely to be a pre-condition. An alternative approach involves analyzing the probabilities of the sequences $<\text{START}, A, anchor, B, \text{END}>$ and $<\text{START}, B, anchor, A, \text{END}>$ using Eq.~\eqref{eq:factorization_2}. If $P(<\text{START}, A, anchor, B, \text{END}> \mid Z)$ exceeds $P(<\text{START}, B, anchor, A, \text{END}> \mid Z)$, it implies that $B$ is the future clip relative to $anchor$. Conversely, if the latter probability is higher, $A$ is identified as the future clip for $anchor$ (see Figure~\ref{fig:future}).

\subsection{Details on Online Mistake Detection}
\label{app:omd}
Due to the noisy sequences in the Assembly101~\cite{sener2022assembly101} and EPIC-Tent~\cite{jang2019epic} datasets, we implemented a tailored approach during the post-processing phase of task graph generation. Specifically, when a key-step in the task graph has exactly two pre-conditions, one of which is the START node, we remove the other pre-condition, irrespective of its score. In all other cases, we apply a reduction in transitivity dependencies. This approach allows for a graph with fewer pre-conditions in the initial steps.

For Assembly101, which consists of multiple procedural tasks, we chose to generate a unified task graph that encompasses all procedures rather than creating separate graphs for each task.

\subsection{Llama-3.1-405B-Instruct Prompts}
\label{app:llama}
Prompt~\ref{prompt:preconditions} was used to guide the model in identifying pre-conditions for specific procedural steps. Similarly, Prompt~\ref{prompt:optionals} was employed to instruct the model on determining whether a key-step is optional. This last prompt was used to construct graphs with optional nodes, aligning with one of the downstream tasks in the Ego-Exo4D~\cite{grauman2023ego} procedure understanding benchmark, which involves recognizing optional keysteps.

\subsection{Experiments Compute Resources}
\label{app:resources}
The experiments conducted with the DO model on symbolic data were highly efficient. We successfully generated all the task graphs from CaptainCook4D, EgoPER, and EgoProceL in about one hour using a Tesla V100S-PCI GPU, which allowed us to run up to 8 training processes concurrently. In comparison, training the TGT models for all scenarios in the CaptainCook4D, EgoPER, and EgoProceL datasets took around 48 hours, with the same GPU supporting the concurrent training of up to 2 models. Moreover, once the task graphs were generated, running the PREGO benchmarks for online mistake detection or executing the Ego-Exo4D procedure understanding benchmark was much faster, as we only needed time to load the task graphs, after which the execution could occur in real-time.

\begin{figure}[t]
    \centering
    \includegraphics[width=1.0\linewidth]{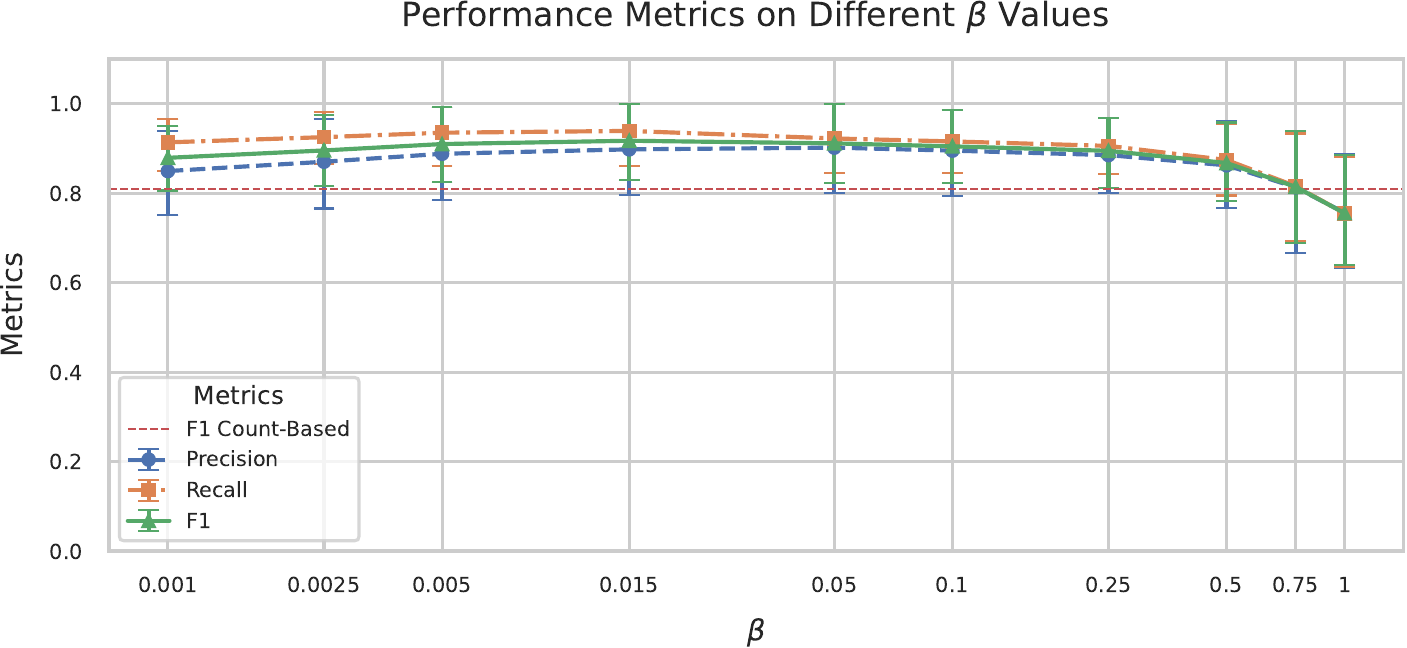}
    \caption{Performance metrics on different $\beta$ values using Direct Optimization (DO) on EgoPER. The dashed line represents the best-performing method among the competitors on this dataset.}
    \label{fig:beta_egoper}
\end{figure}

\begin{figure}[t]
    \centering
    \includegraphics[width=1.0\linewidth]{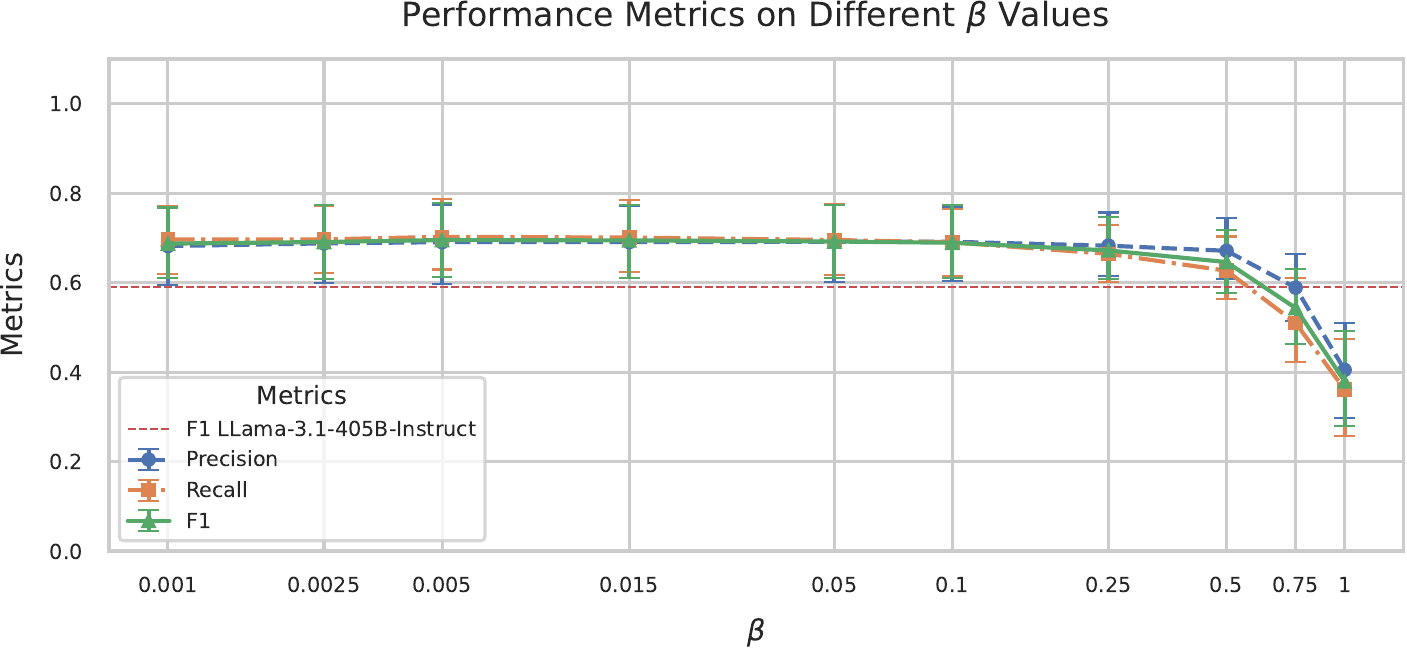}
    \caption{Performance metrics on different $\beta$ values using Direct Optimization (DO) on EgoProceL. The dashed line represents the best-performing method among the competitors on this dataset.}
    \label{fig:beta_egoprocel}
\end{figure}

\section{Ablation Studies}
\subsubsection{Performance Metrics on Different \(\beta\) Values}
Figures~\ref{fig:beta_egoper}, and~\ref{fig:beta_egoprocel} present performance metrics across various $\beta$ values for the Direct Optimization (DO) method on the EgoPER~\cite{lee2024error}, and EgoProceL~\cite{bansal2022my} datasets, respectively. Each graph includes a comparison with the best performing competitor (the red dotted line), highlighting the range of $\beta$ values where DO outperforms the leading alternative in each data set.

\section{Qualitative Examples}
Figures~\ref{fig:qualitative1} -~\ref{fig:qualitative24} report qualitative examples of prediction using our Direct Optimization (DO) method on the CaptainCook4D~\cite{peddi2023captaincook4d} procedures. Figures~\ref{fig:comparison_captaincook4d},~\ref{fig:comparison_egoper}, and~\ref{fig:comparison_egoprocel} present qualitative comparison of the task graphs generated by the considered methods across the CaptainCook4D, EgoPER, and EgoProceL datasets. Figures~\ref{fig:annotated_start} -~\ref{fig:annotated_end} show the annotated task graphs from EgoProceL~\cite{bansal2022my}. The task graphs must be read in a bottom-up manner, where the START node (bottom) is at the lowest position and represents the first node with no pre-conditions, while the END node (up) is the final step of the procedure.

Figure~\ref{fig:qualitative_mistake} reports a qualitative analysis of the generated task graph for detecting the mistakes on EPIC-Tent-O.

\section{Societal Impact}
\label{sec:societal}
Reconstructing task graphs from procedural videos may enable the construction of agents able to assist users during the execution of the task. Learning task graphs from videos may be affected by geographical or cultural biases appearing in the data (e.g., specific ways of performing given tasks), which may limit the quality of the feedback returned to the user, potentially leading to harm. We expect that training data of sufficient quality should limit such risks.
}

\newpage

\begin{prompt*}
\centering
\fbox{
    \begin{minipage}{0.8\linewidth}
    I would like you to learn to answer questions by telling me the steps 
    that need to be performed before a given one.\\
    
    The questions refer to procedural activities and these are of the following type:\\
    
    Q - Which of the following key-steps is a pre-condition for the current key-step 
    ``add brownie mix''?\\
    
    - add oil\\
    - add water\\
    - break eggs\\
    - mix all the contents\\
    - mix eggs\\
    - pour the mixture in the tray\\
    - spray oil on the tray\\
    - None of the above\\
    \\
    Your task is to use your immense knowledge and your immense ability to tell me 
    which preconditions are among those listed that must necessarily be carried out 
    before the keystep indicated in quotes in the question.\\
    
    Provide the correct preconditions answer inside a JSON format like this:\\
    
    \{
        ``add brownie mix'': [``add oil'', ``add water'', ``break eggs'']
    \}
    \\\\
    You must provide only the JSON without any explanations.\\
    You must choose at least one of the proposed answers.\\
    You must avoid loops in the preconditions.\\
    You must not change the name of the keysteps.
    \end{minipage}
}
\caption{Reports the prompt used with Llama-3.1-405B-Instruct~\cite{dubey2024llama} to identify the pre-conditions.}
\label{prompt:preconditions}
\end{prompt*}

\begin{prompt*}
\centering
\fbox{
    \begin{minipage}{0.8\linewidth}
    I would like you to learn to answer questions regarding whether a key-step is optional or not.\\
    
    The questions refer to procedural activities and these are of the following type:\\
    
    Q - Is the step ``add brownie mix'' optional?\\
    
    - true\\
    - false\\

    Your task is to use your knowledge and determine whether the given key-step, indicated in quotes, is optional or not based on the process.\\
    
    Provide the correct answer inside a JSON format like this:\\
    
    \{``add brownie mix'': false\}\\

    You must provide only the JSON without any explanations.\\
    You must choose either ``true'' or ``false'' as the answer.
    \end{minipage}
}
\caption{Reports the prompt used with Llama-3.1-405B-Instruct~\cite{dubey2024llama} to identify optional key-steps.}
\label{prompt:optionals}
\end{prompt*}

\begin{figure*}[t]
    \centering
    \includegraphics[width=1\linewidth]{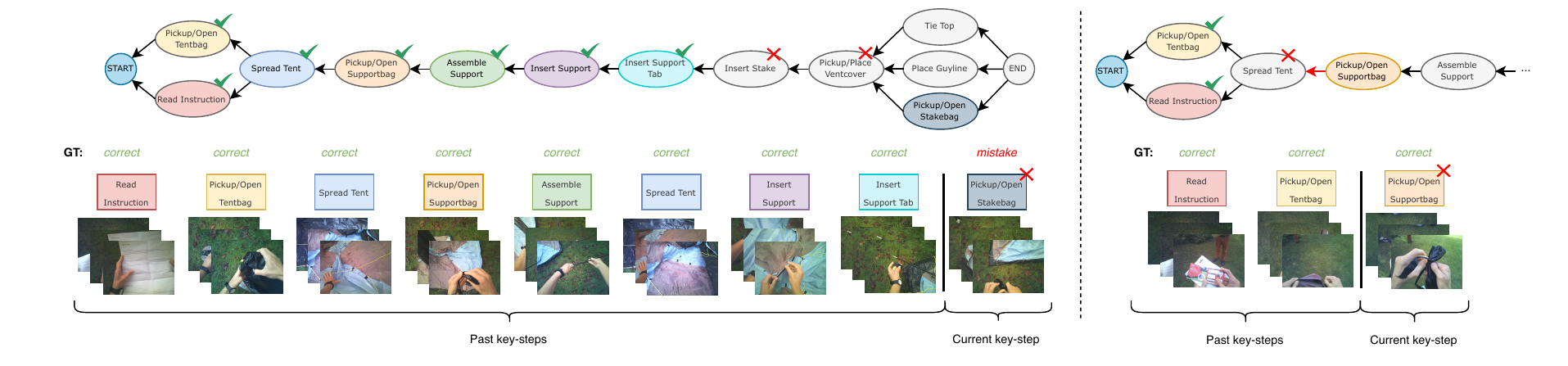}
    \caption{A success (left) and failure (right) case on EPIC-Tent-O. Past key-steps' colors match nodes' colors. On the left, the current key-step ``Pickup/Open Stakebag'' is correctly evaluated as a mistake because the step ``Pickup/Place Ventcover'' is a precondition of the current key-step, but it is not included among the previous key-steps. On the right, ``Pickup/Open Supportbag'' is incorrectly evaluated as mistake because the step ``Spread Tent'' is precondition of the current key-step, but it is not included among the previous key-steps. This is due to the fact that our method wrongly predicted ``Spread Tent'' as a pre-condition of ``Pickup/Open Supportbag'', probably due to the two actions often occurring in this order.}
    \label{fig:qualitative_mistake}
\end{figure*}

\begin{figure*}[t]
    \centering
    \subfloat[]{
        \includegraphics[width=0.45\textwidth]{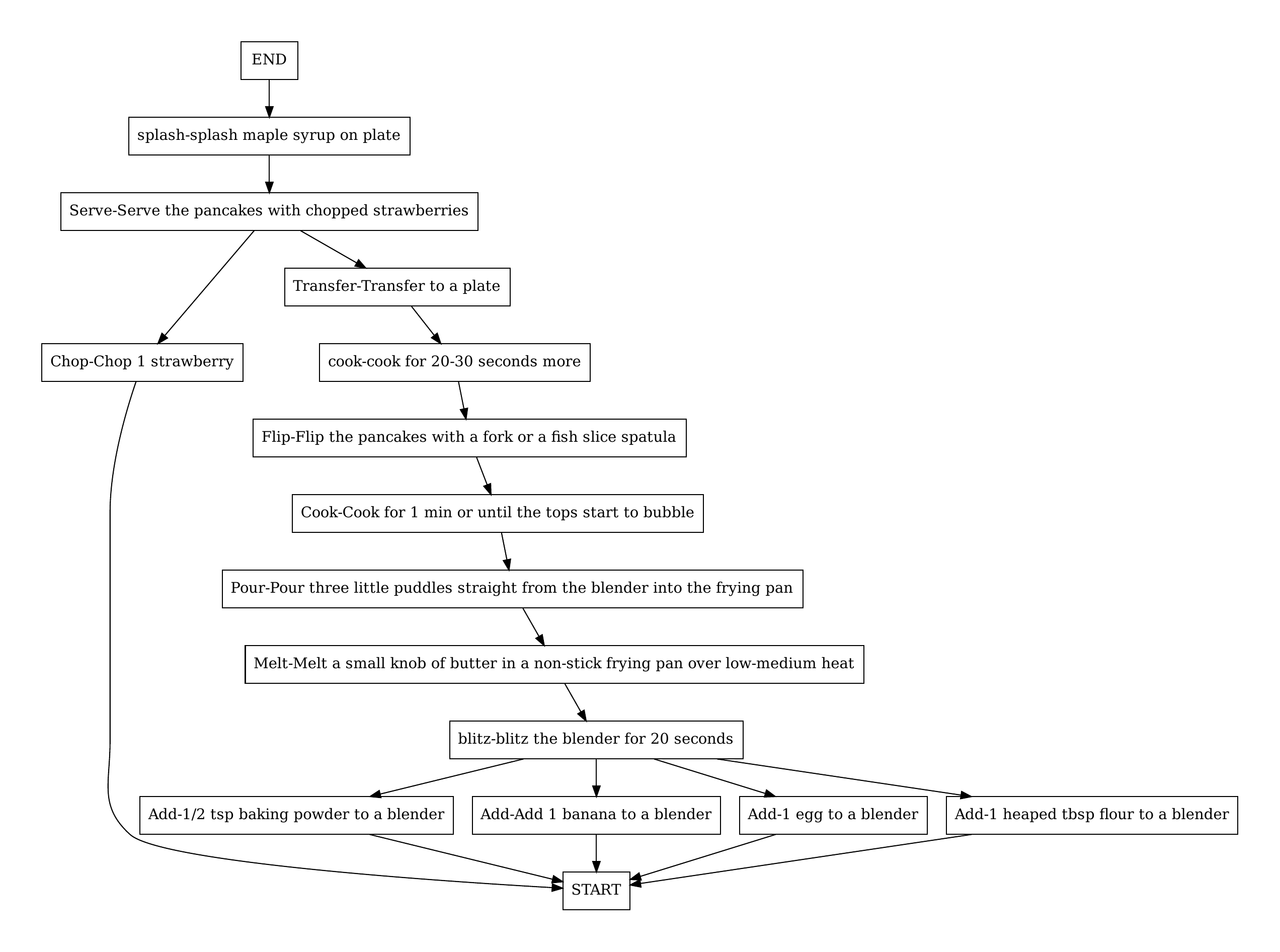}
    }
    \hfill
    \subfloat[]{
        \includegraphics[width=0.45\textwidth]{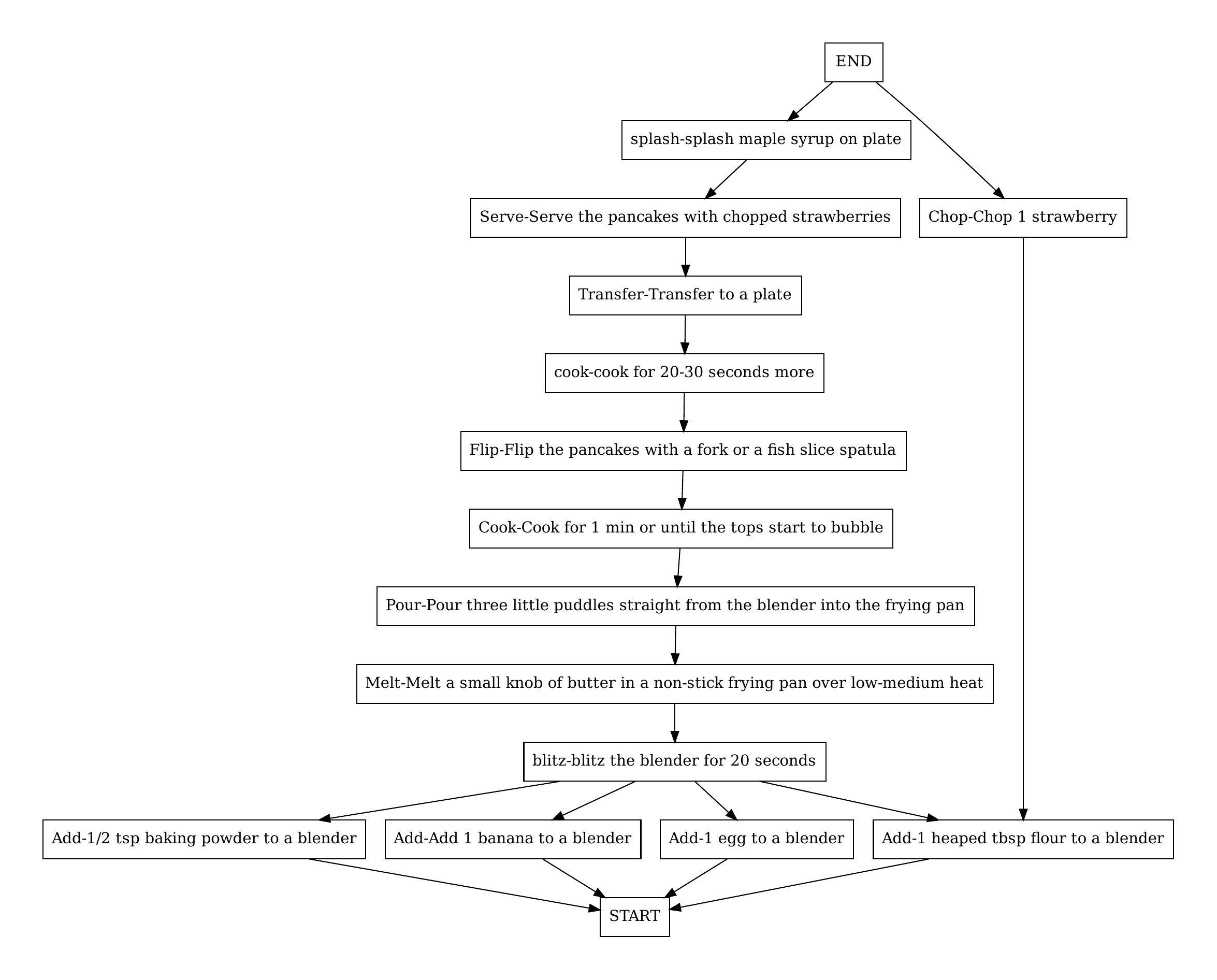}
    }
    \caption{(a) Ground truth task graph and (b) predicted task graph of the scenario Breakfast Burritos.}
    \label{fig:qualitative1}
\end{figure*}

\begin{figure*}[t]
    \centering
    \subfloat[]{
        \includegraphics[width=0.45\textwidth]{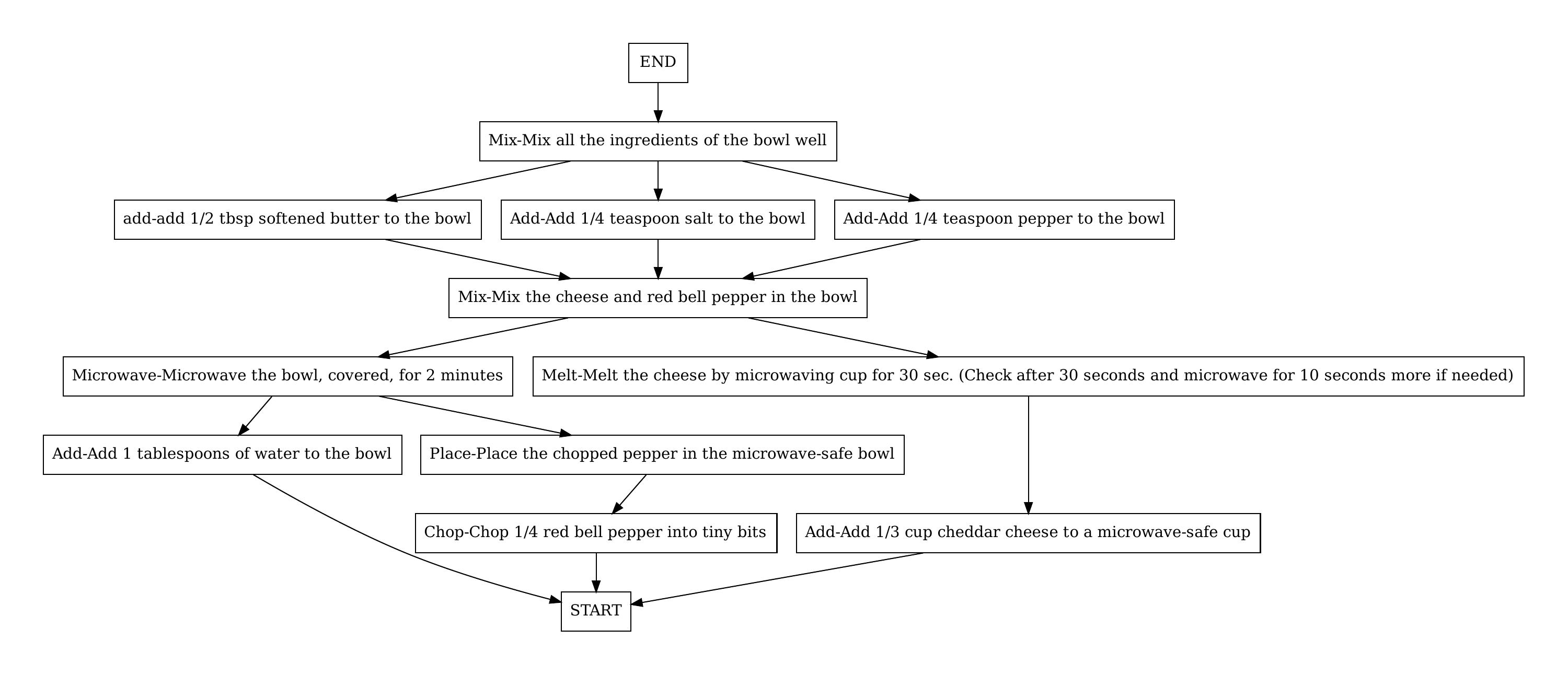}
    }
    \hfill
    \subfloat[]{
        \includegraphics[width=0.45\textwidth]{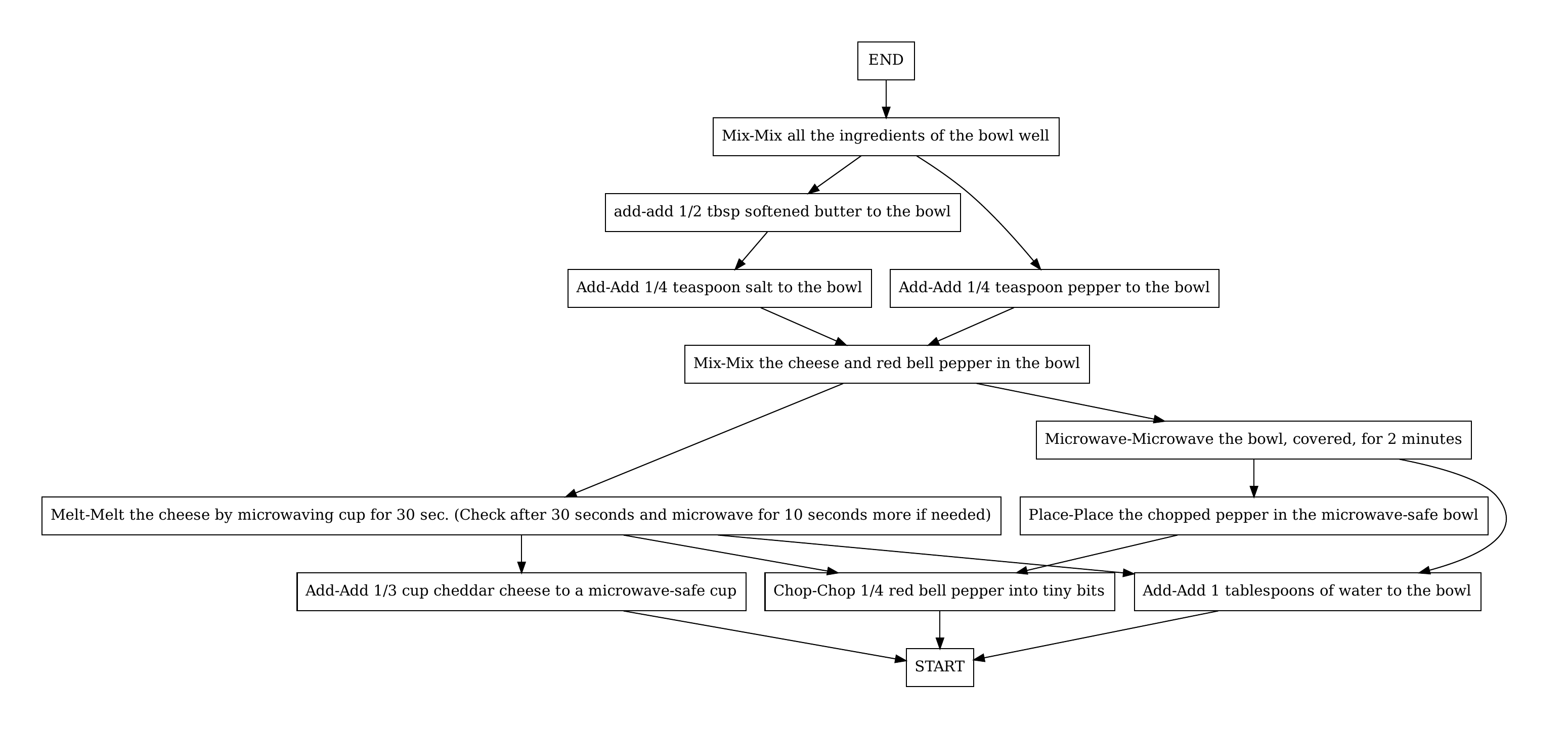}
    }
    \caption{(a) Ground truth task graph and (b) predicted task graph of the scenario Cheese Pimiento.}
\end{figure*}

\begin{figure*}[t]
    \centering
    \subfloat[]{
        \includegraphics[width=0.45\textwidth]{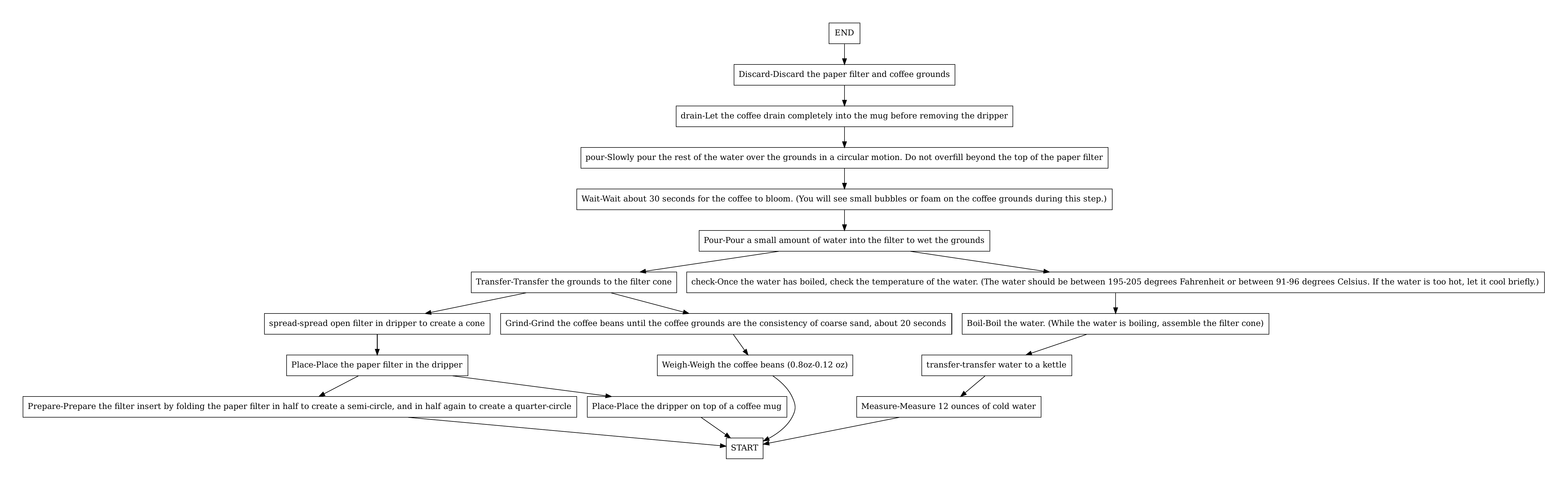}
    }
    \hfill
    \subfloat[]{
        \includegraphics[width=0.45\textwidth]{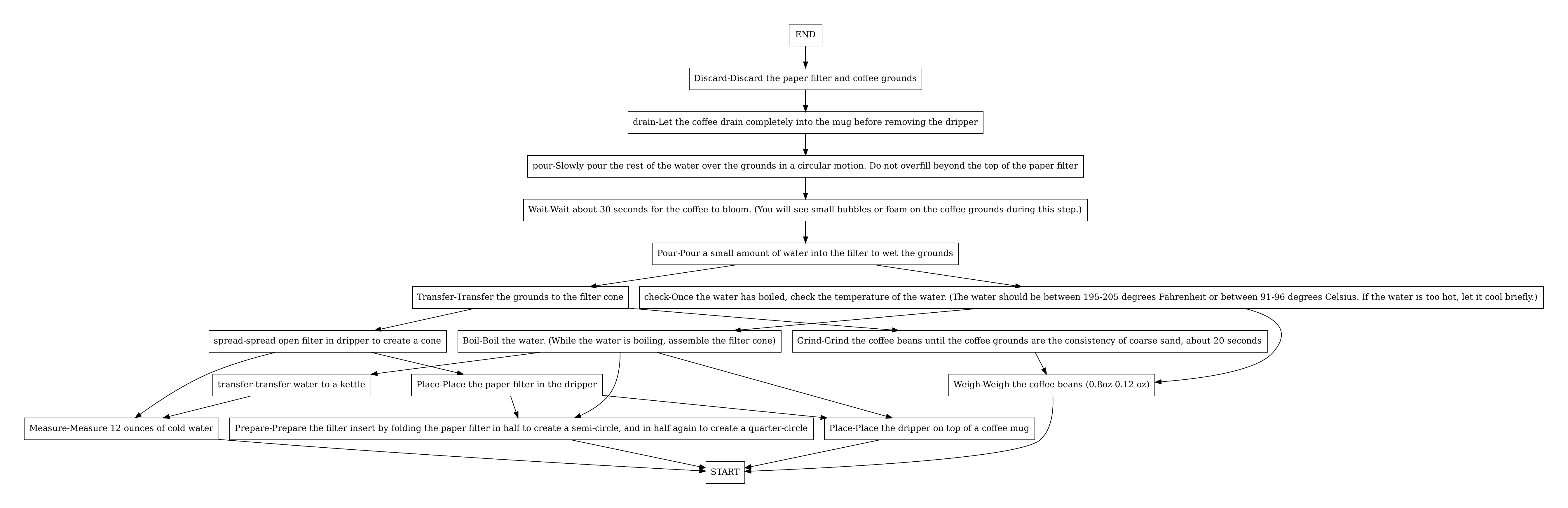}
    }
    \caption{(a) Ground truth task graph and (b) predicted task graph of the scenario Coffee.}
\end{figure*}

\begin{figure*}[t]
    \centering
    \subfloat[]{
        \includegraphics[width=0.45\textwidth]{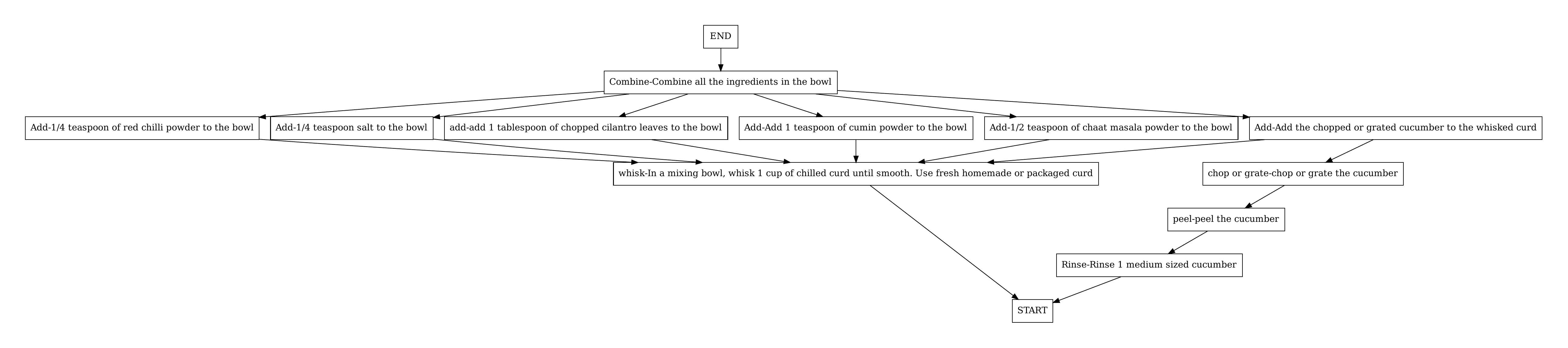}
    }
    \hfill
    \subfloat[]{
        \includegraphics[width=0.45\textwidth]{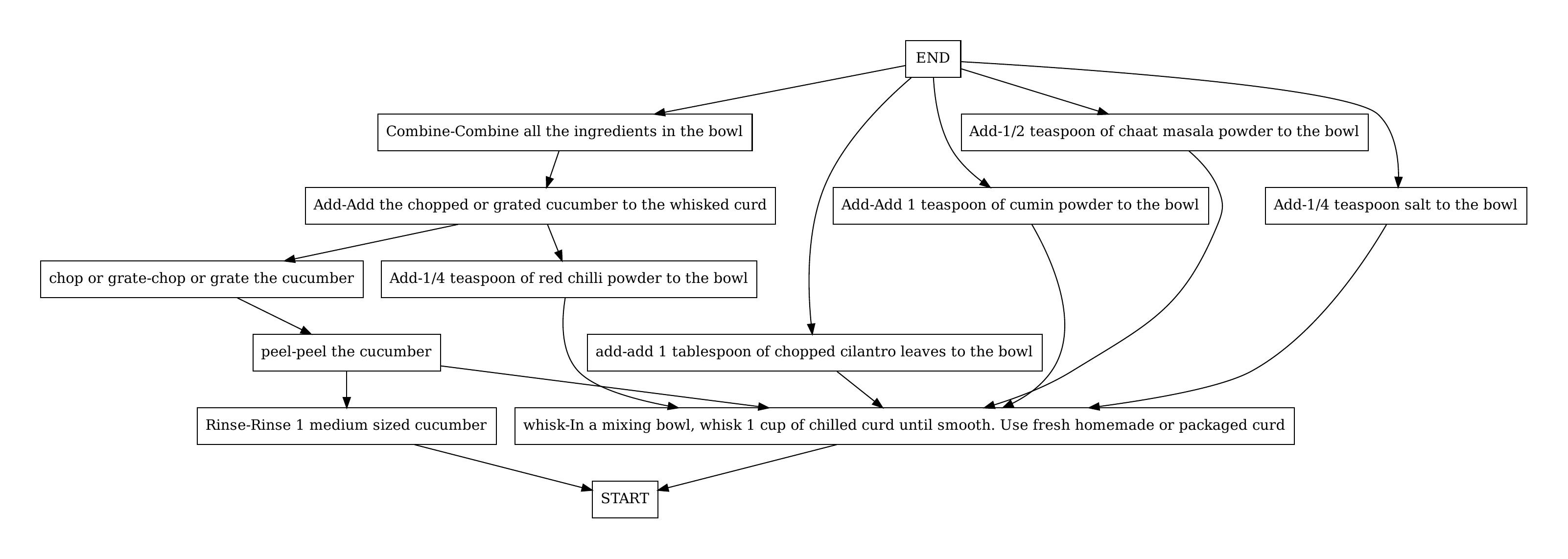}
    }
    \caption{(a) Ground truth task graph and (b) predicted task graph of the scenario Cucumber Raita.}
\end{figure*}

\begin{figure*}[t]
    \centering
    \subfloat[]{
        \includegraphics[width=0.45\textwidth]{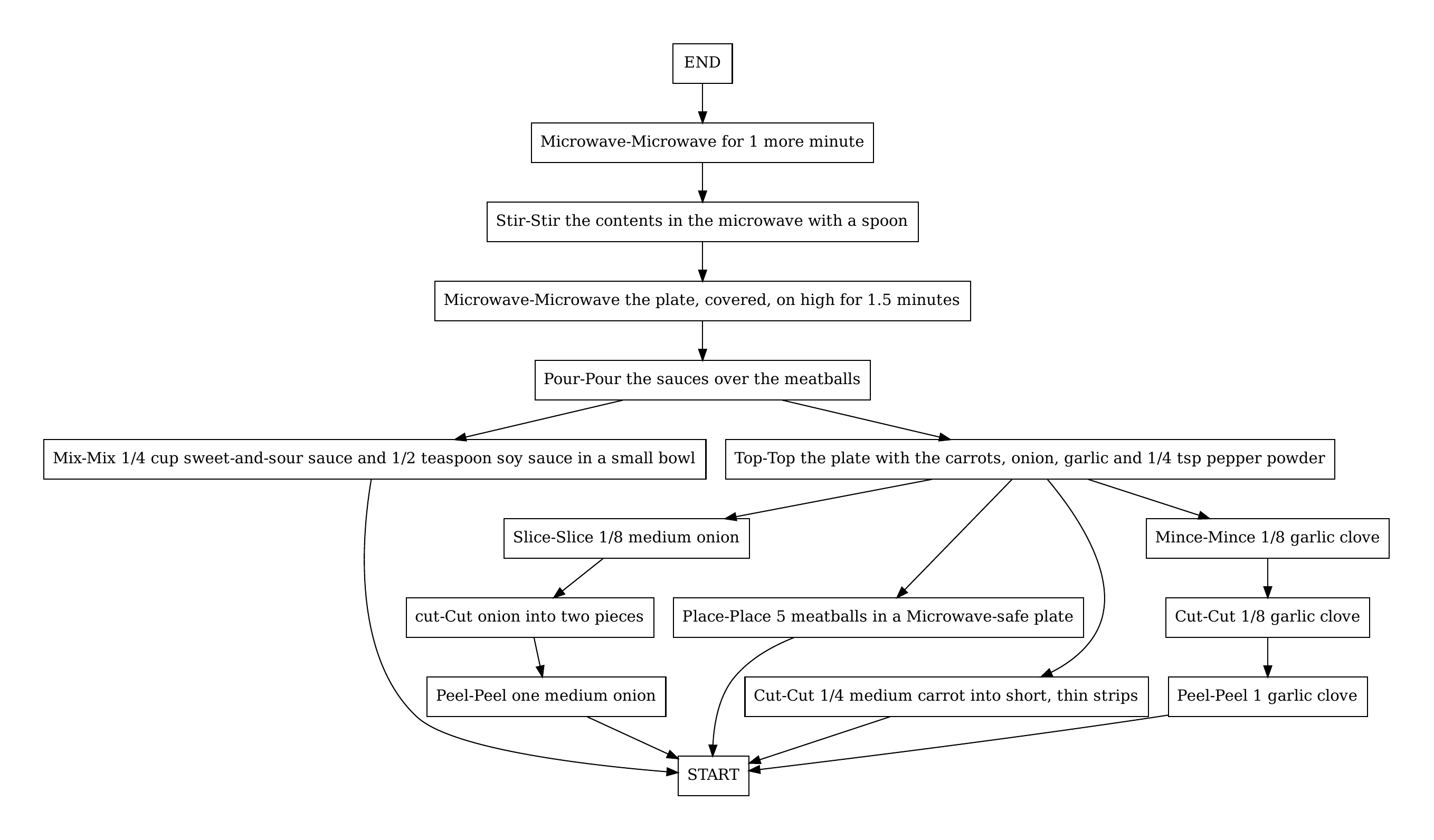}
    }
    \hfill
    \subfloat[]{
        \includegraphics[width=0.45\textwidth]{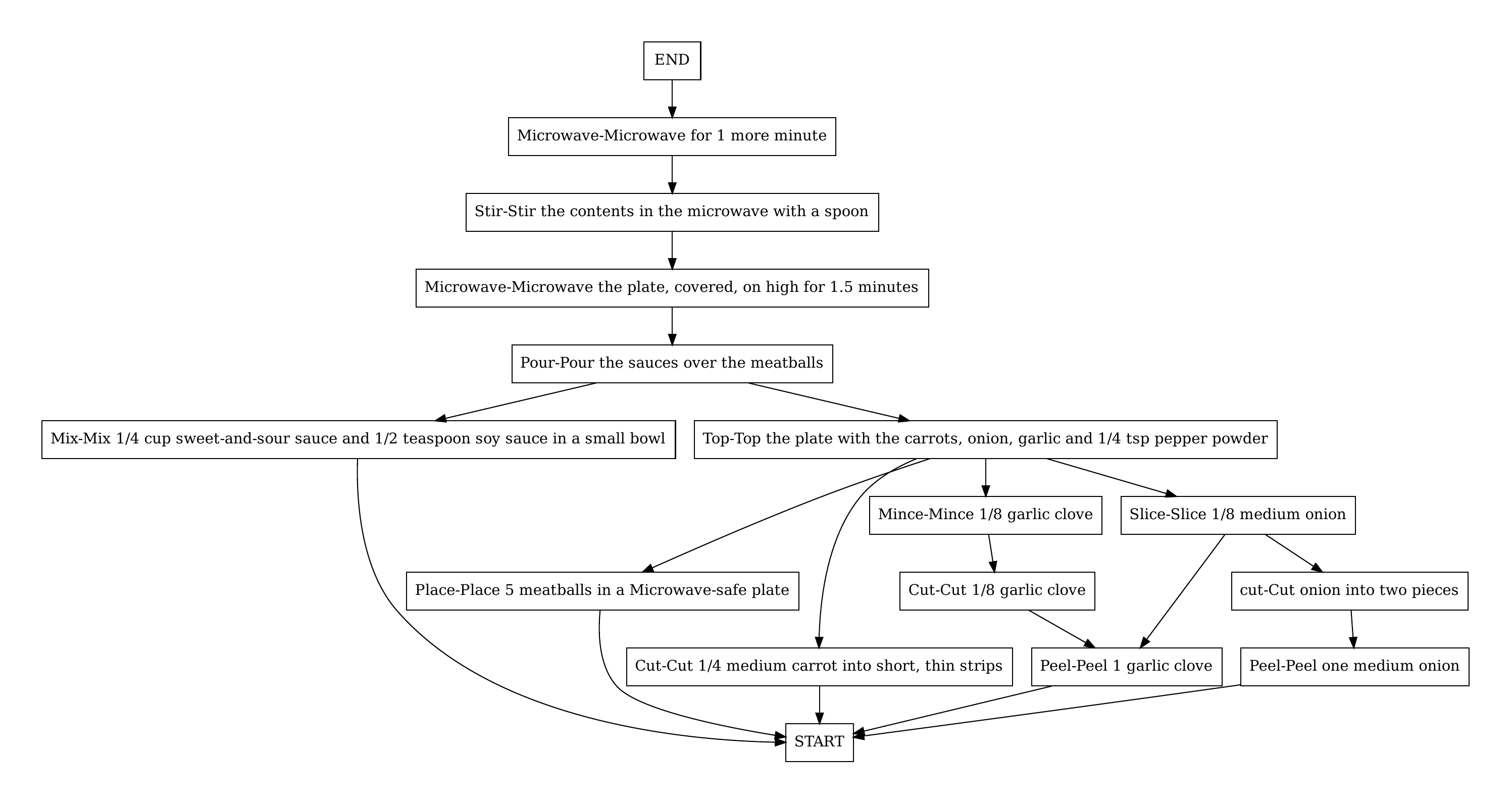}
    }
    \caption{(a) Ground truth task graph and (b) predicted task graph of the scenario Dressed Up Meatballs.}
\end{figure*}

\begin{figure*}[t]
    \centering
    \subfloat[]{
        \includegraphics[width=0.45\textwidth]{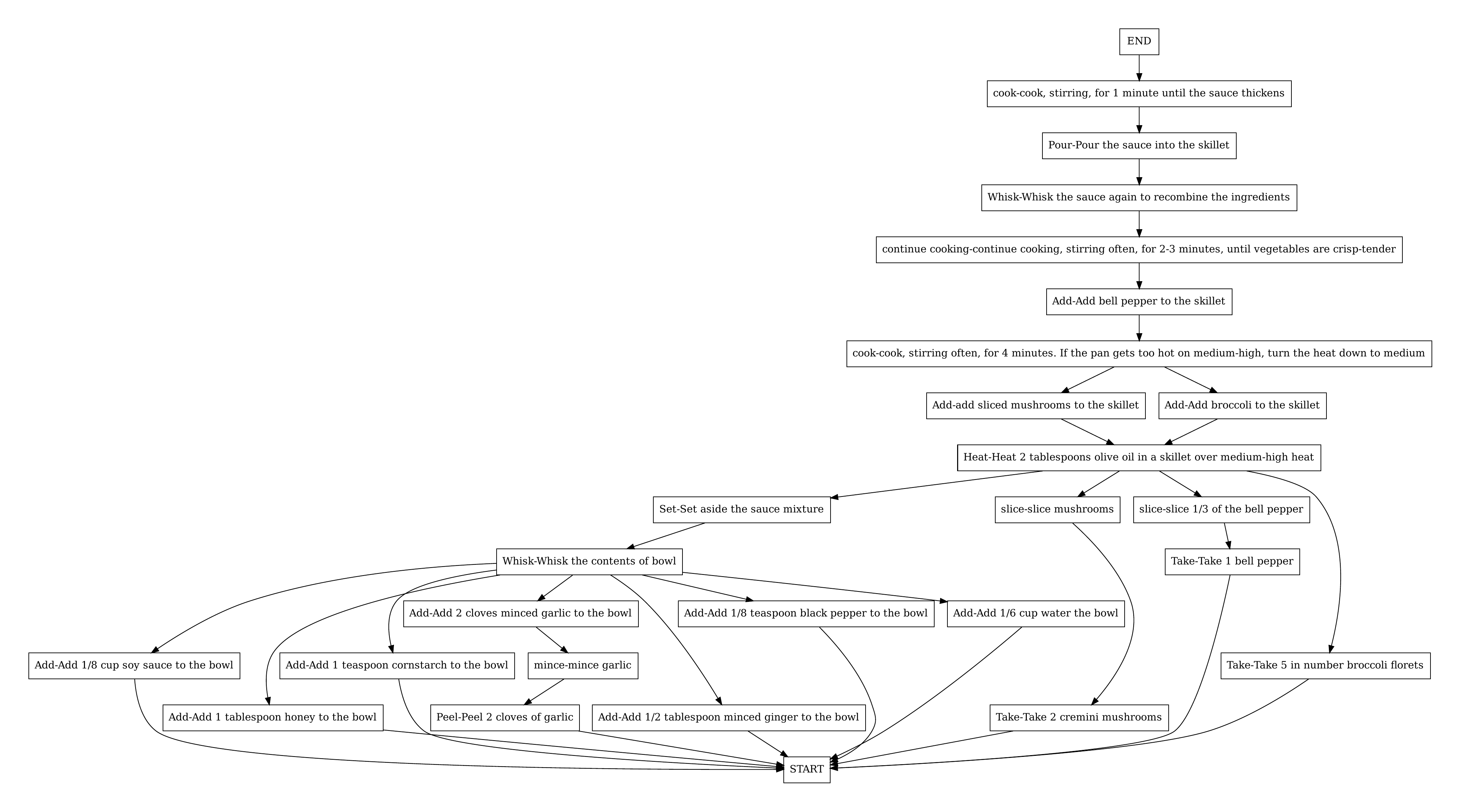}
    }
    \hfill
    \subfloat[]{
        \includegraphics[width=0.45\textwidth]{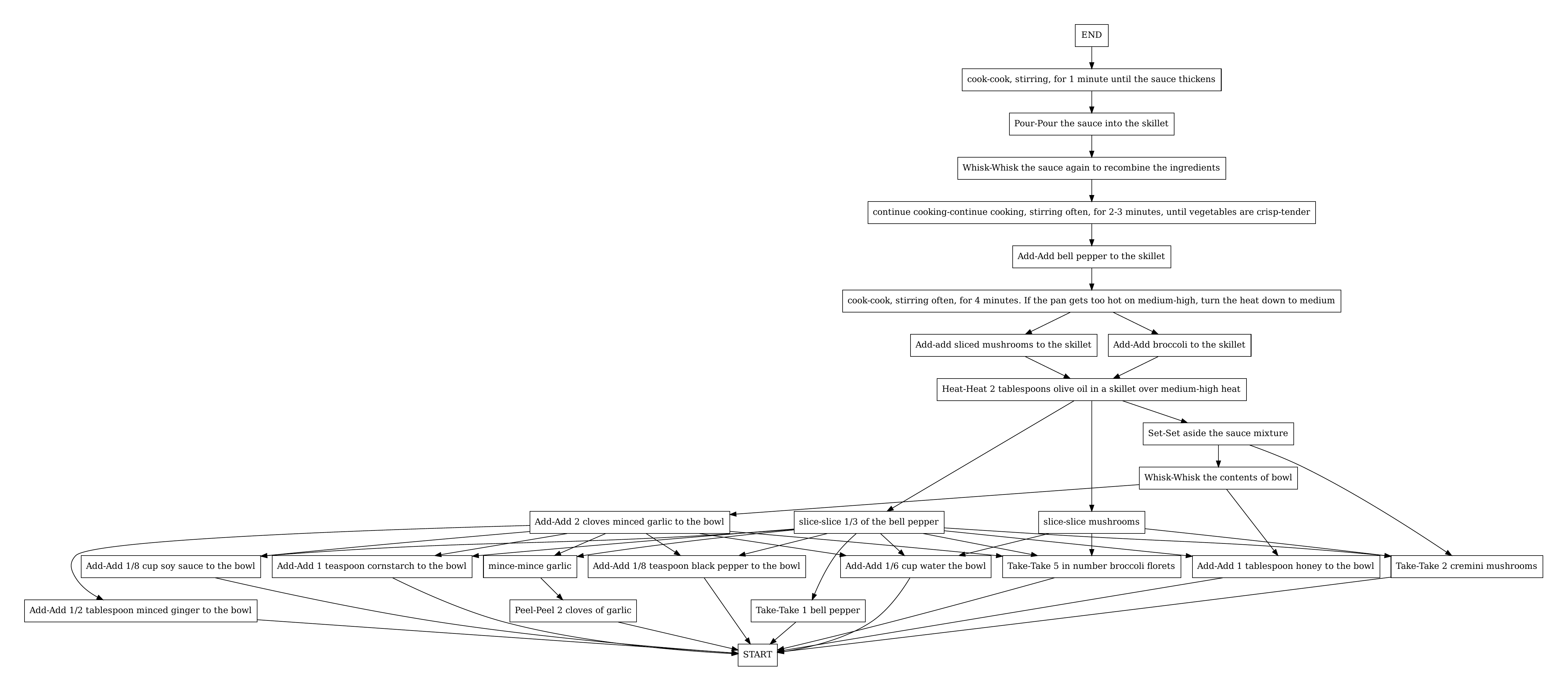}
    }
    \caption{(a) Ground truth task graph and (b) predicted task graph of the scenario Broccoli Stir Fry.}
\end{figure*}

\begin{figure*}[t]
    \centering
    \subfloat[]{
        \includegraphics[width=0.45\textwidth]{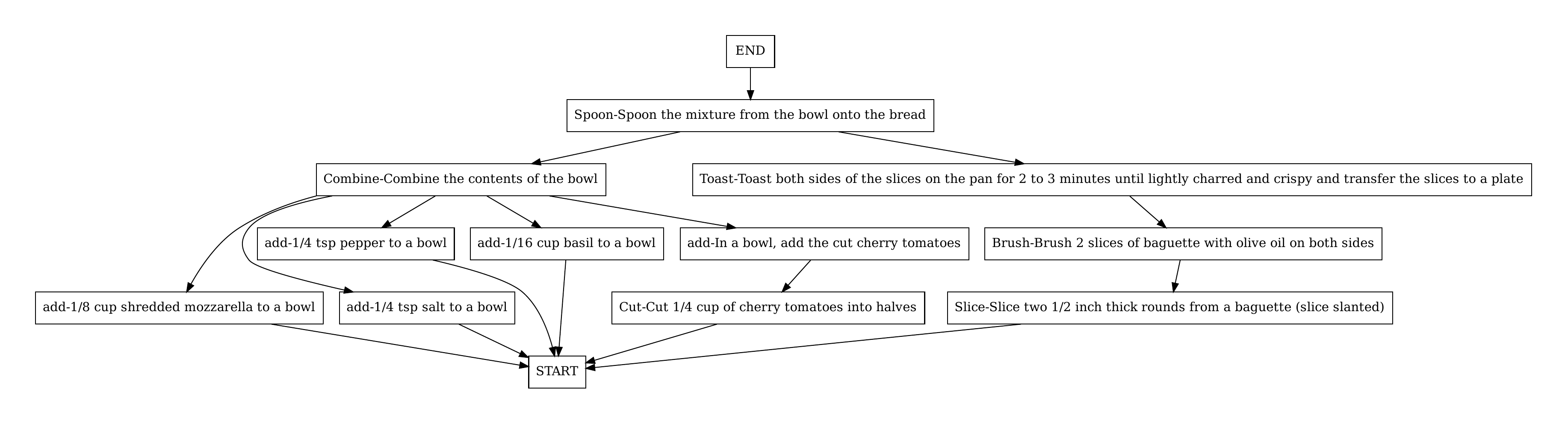}
    }
    \hfill
    \subfloat[]{
        \includegraphics[width=0.45\textwidth]{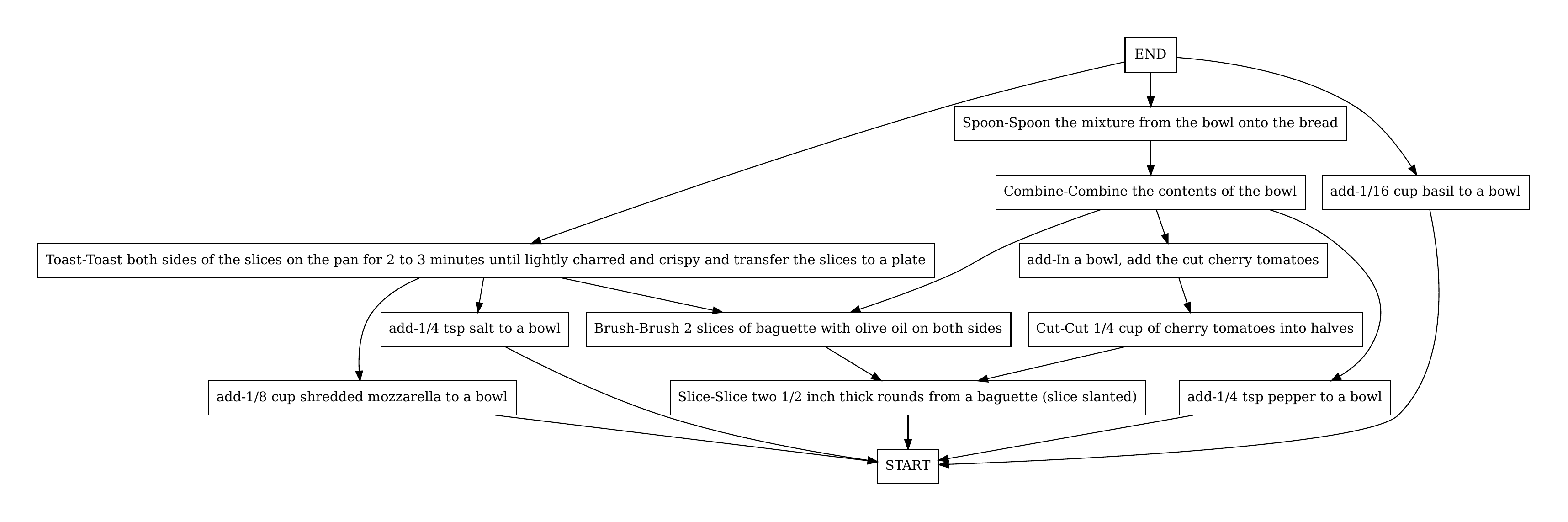}
    }
    \caption{(a) Ground truth task graph and (b) predicted task graph of the scenario Caprese Bruschetta.}
\end{figure*}

\begin{figure*}[t]
    \centering
    \subfloat[]{
        \includegraphics[width=0.45\textwidth]{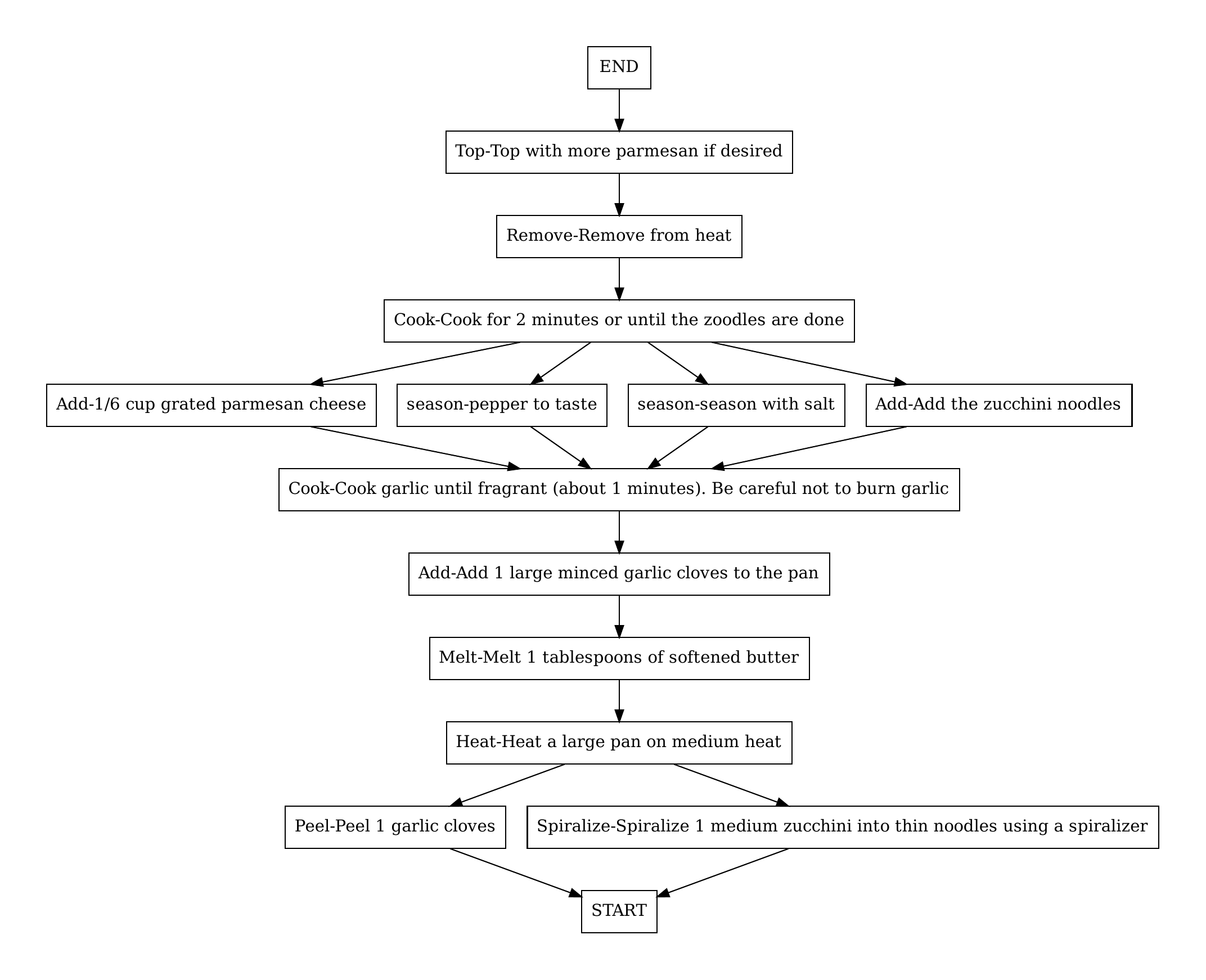}
    }
    \hfill
    \subfloat[]{
        \includegraphics[width=0.45\textwidth]{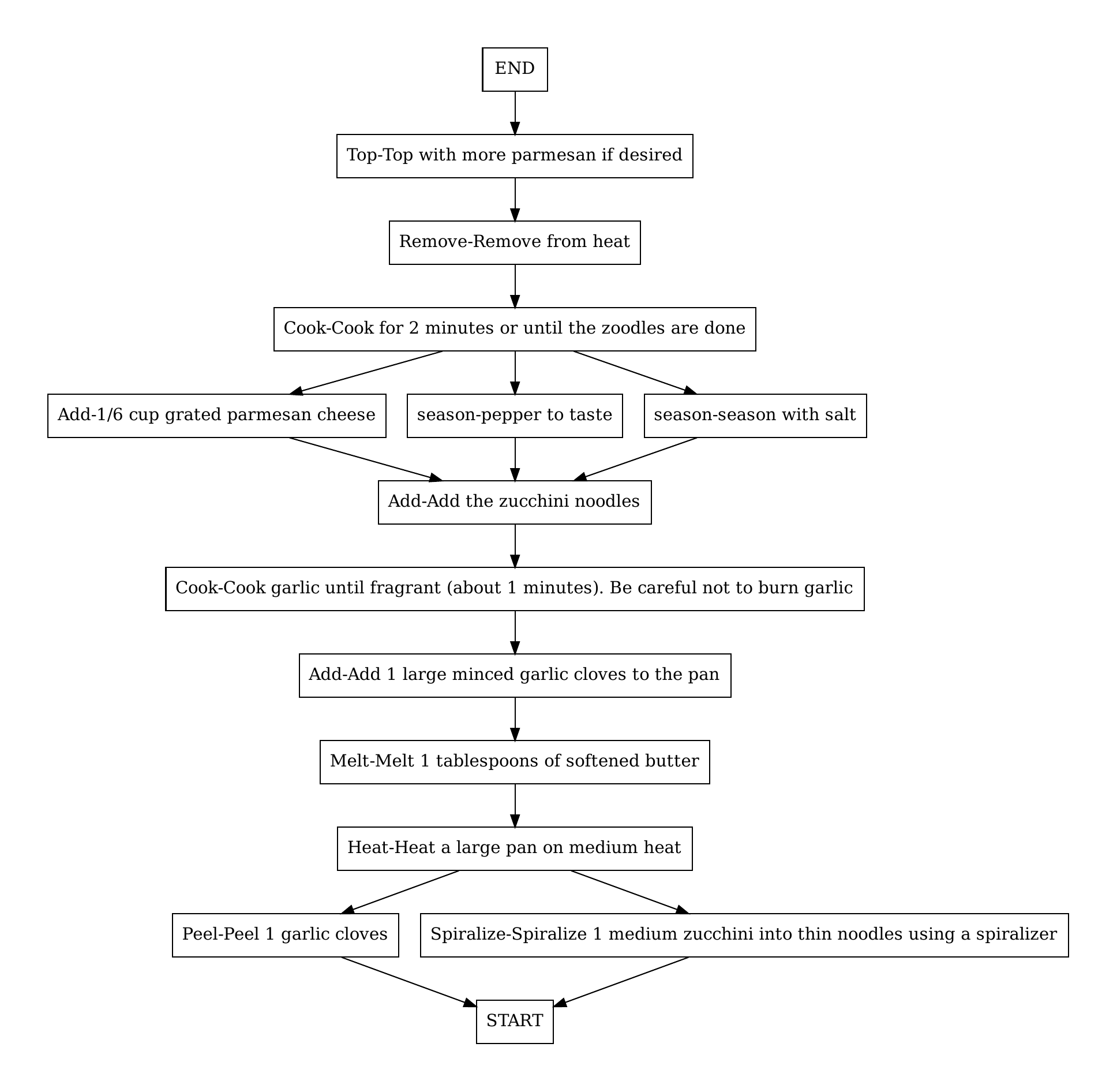}
    }
    \caption{(a) Ground truth task graph and (b) predicted task graph of the scenario Zoodles.}
\end{figure*}

\begin{figure*}[t]
    \centering
    \subfloat[]{
        \includegraphics[width=0.45\textwidth]{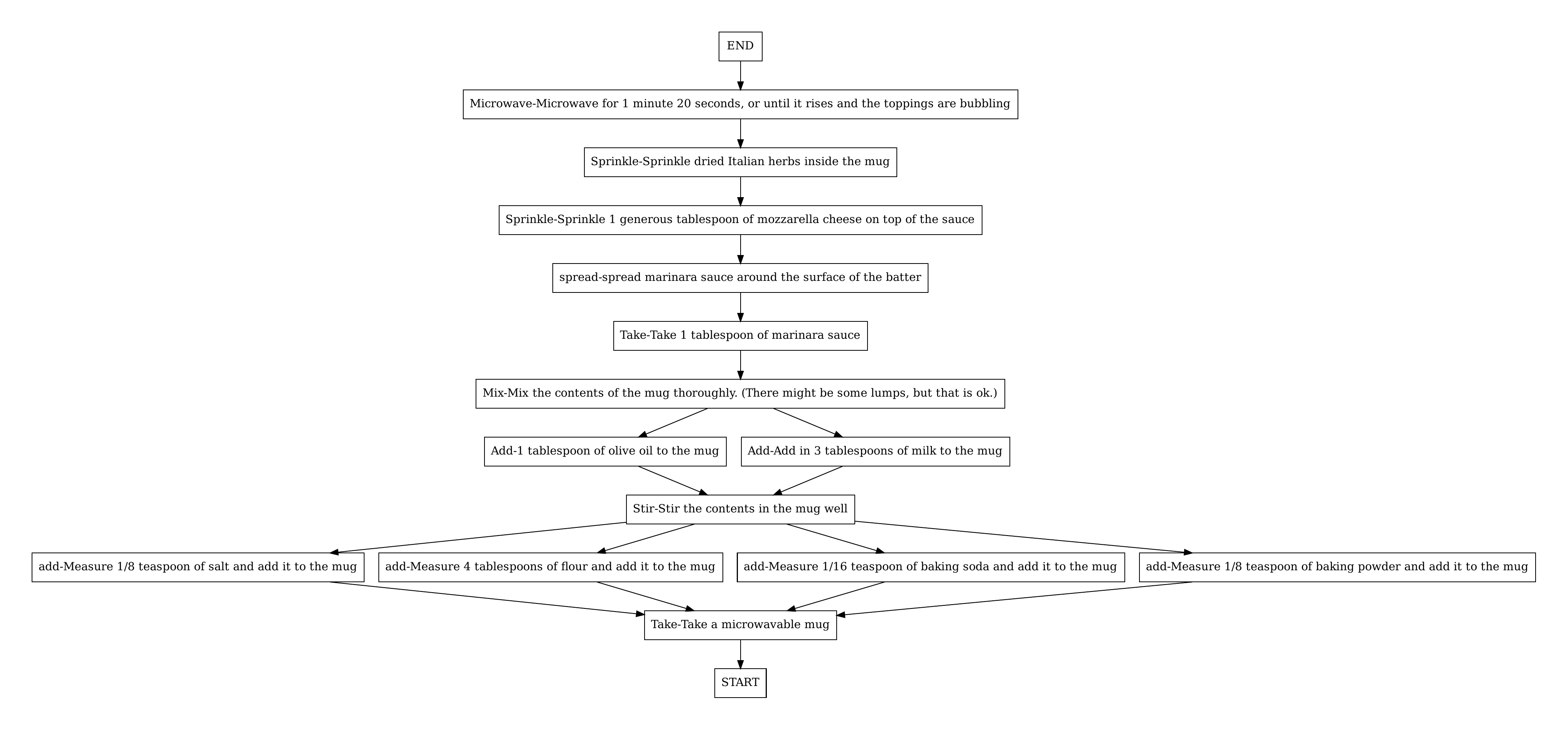}
    }
    \hfill
    \subfloat[]{
        \includegraphics[width=0.45\textwidth]{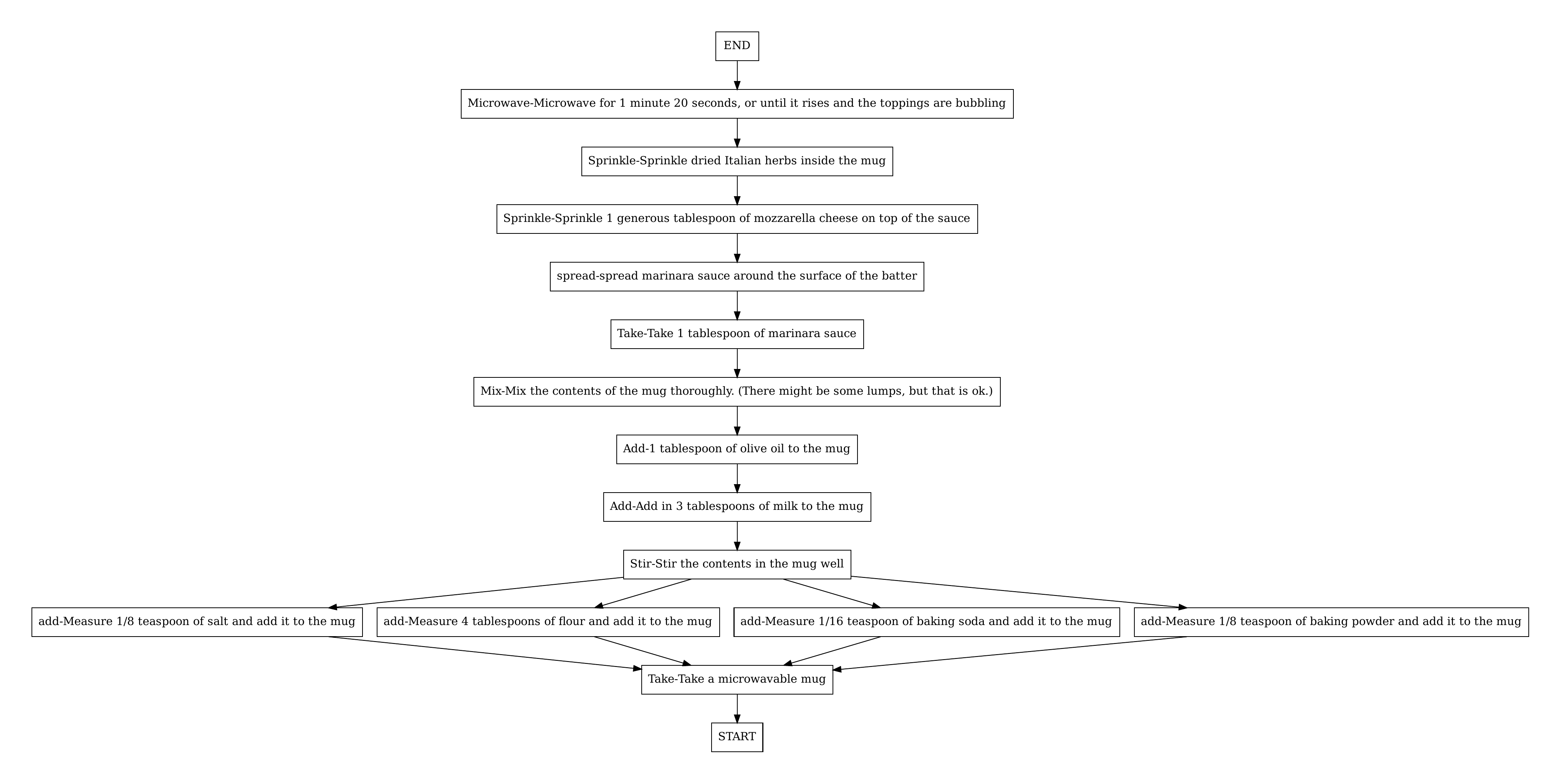}
    }
    \caption{(a) Ground truth task graph and (b) predicted task graph of the scenario Microwave Mug Pizza.}
\end{figure*}

\begin{figure*}[t]
    \centering
    \subfloat[]{
        \includegraphics[width=0.45\textwidth]{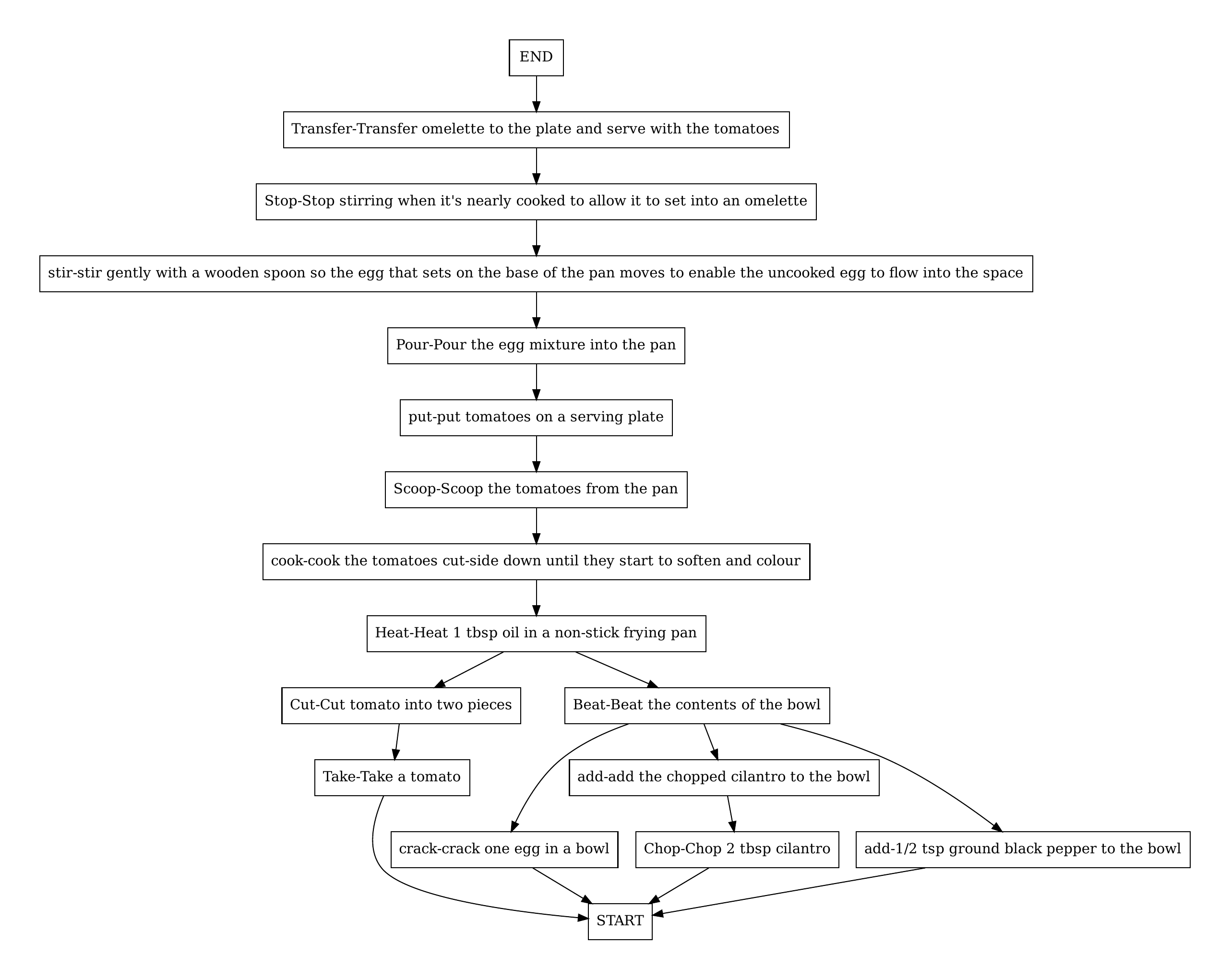}
    }
    \hfill
    \subfloat[]{
        \includegraphics[width=0.45\textwidth]{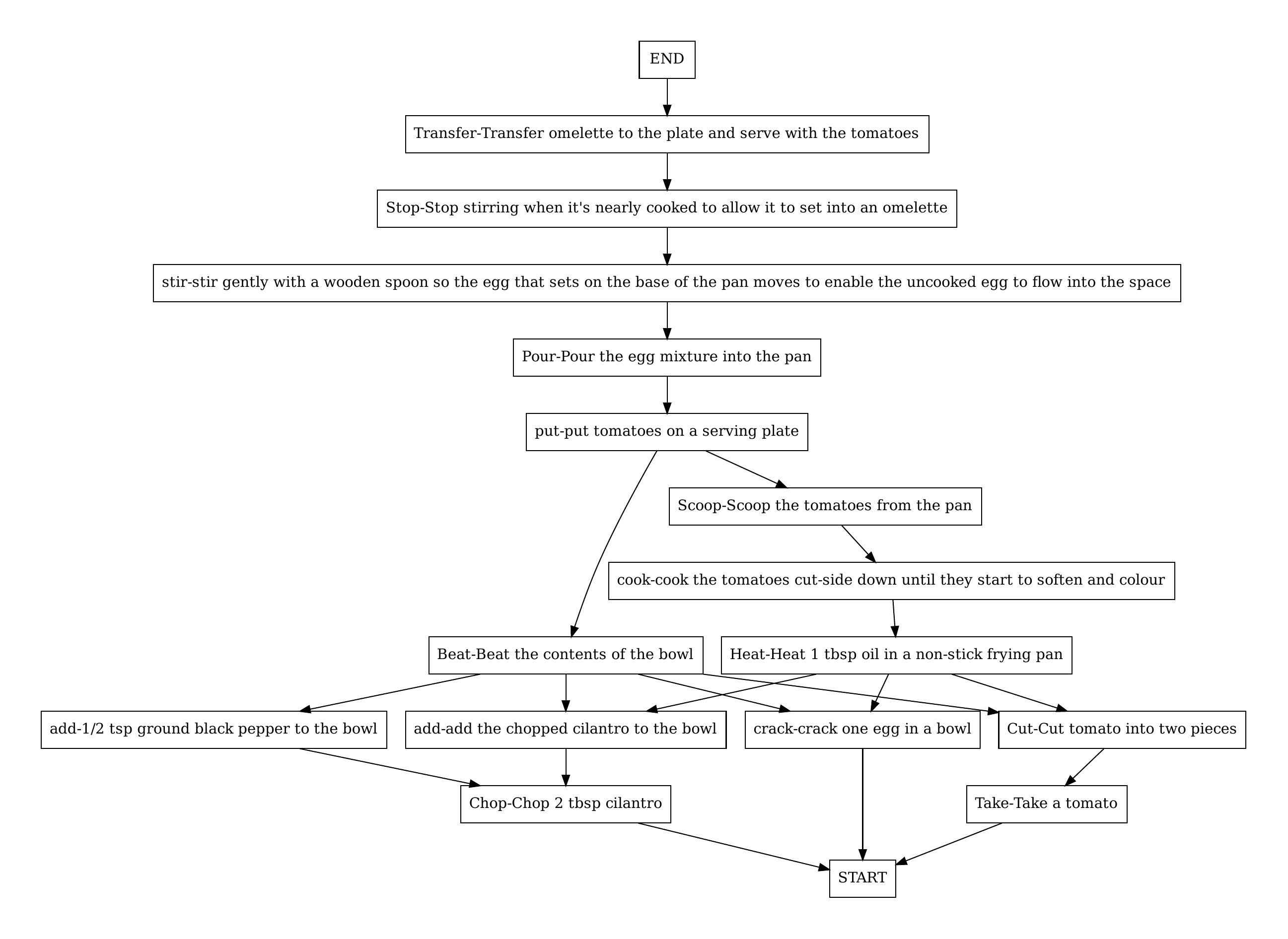}
    }
    \caption{(a) Ground truth task graph and (b) predicted task graph of the scenario Herb Omelet with Fried Tomatoes.}
\end{figure*}

\begin{figure*}[t]
    \centering
    \subfloat[]{
        \includegraphics[width=0.45\textwidth]{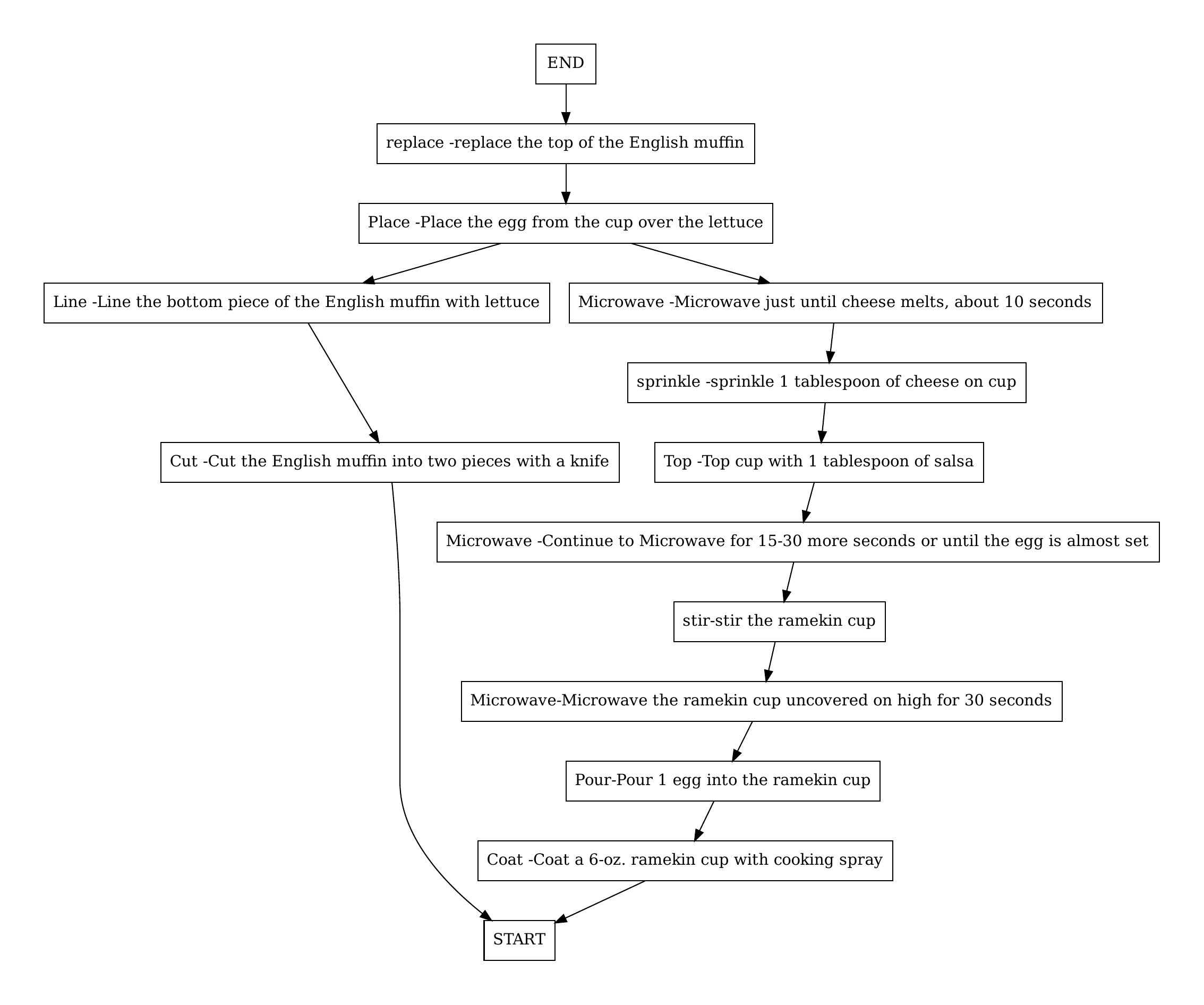}
    }
    \hfill
    \subfloat[]{
        \includegraphics[width=0.45\textwidth]{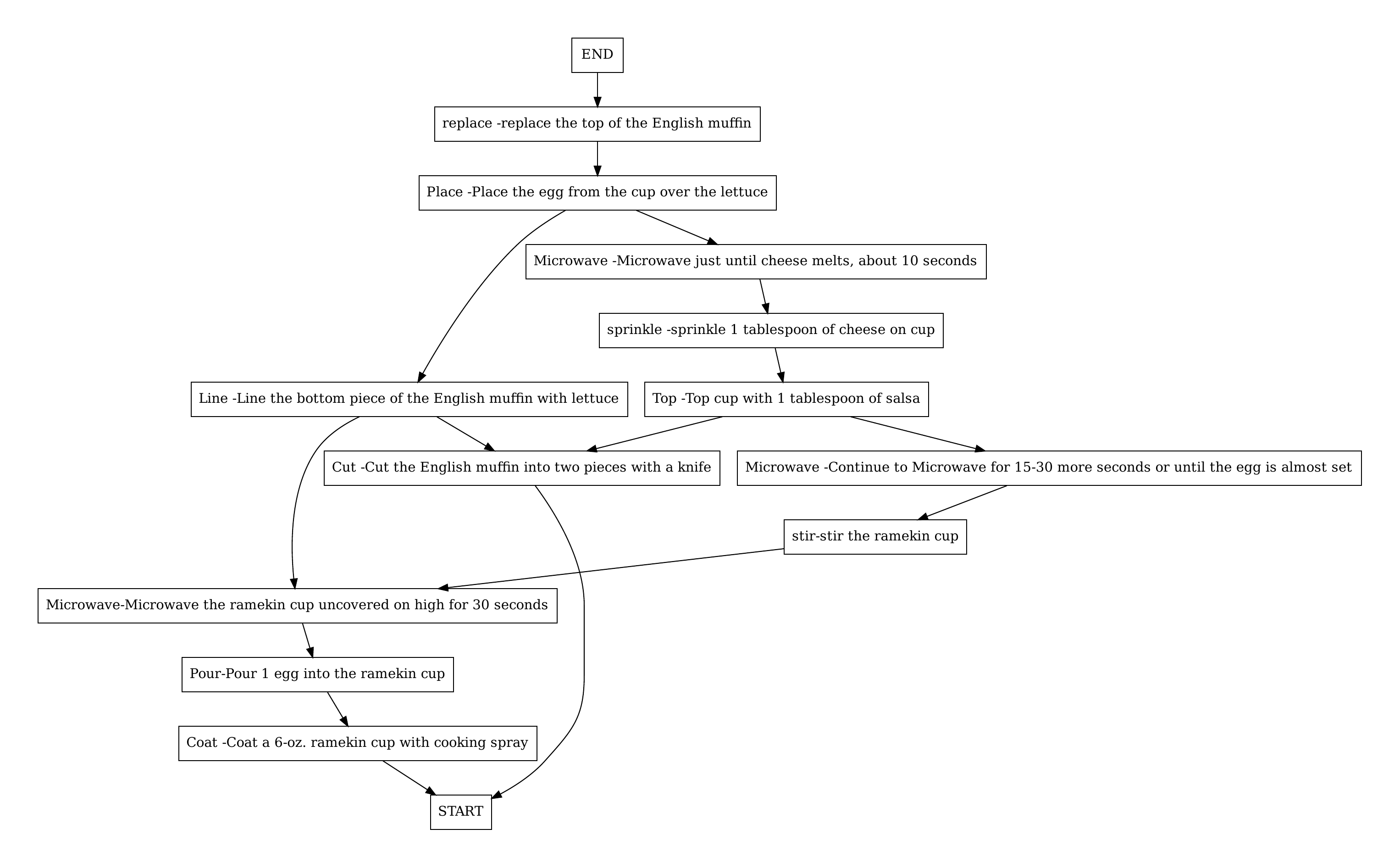}
    }
    \caption{(a) Ground truth task graph and (b) predicted task graph of the scenario Microwave Egg Sandwich.}
\end{figure*}

\begin{figure*}[t]
    \centering
    \subfloat[]{
        \includegraphics[width=0.45\textwidth]{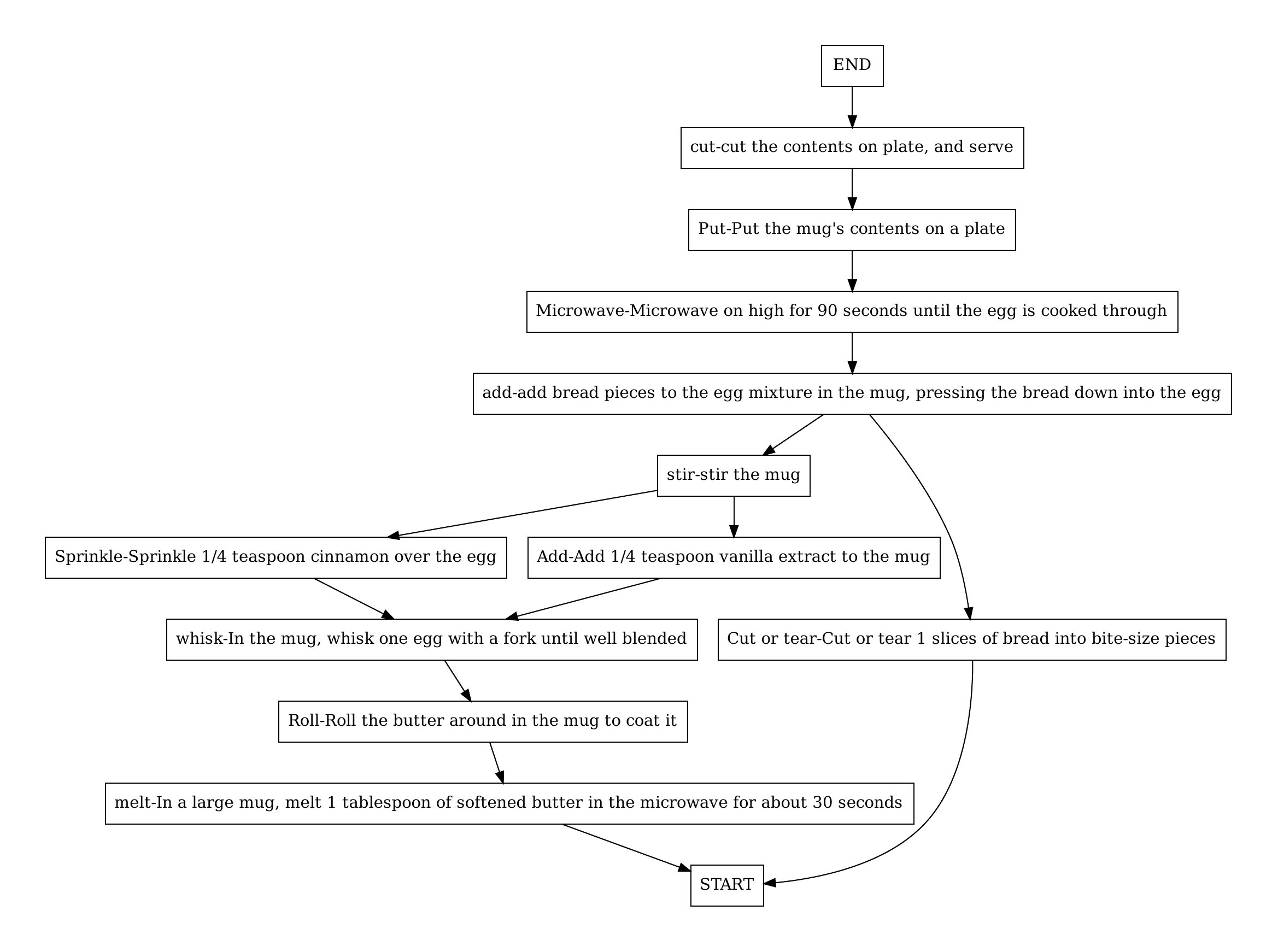}
    }
    \hfill
    \subfloat[]{
        \includegraphics[width=0.45\textwidth]{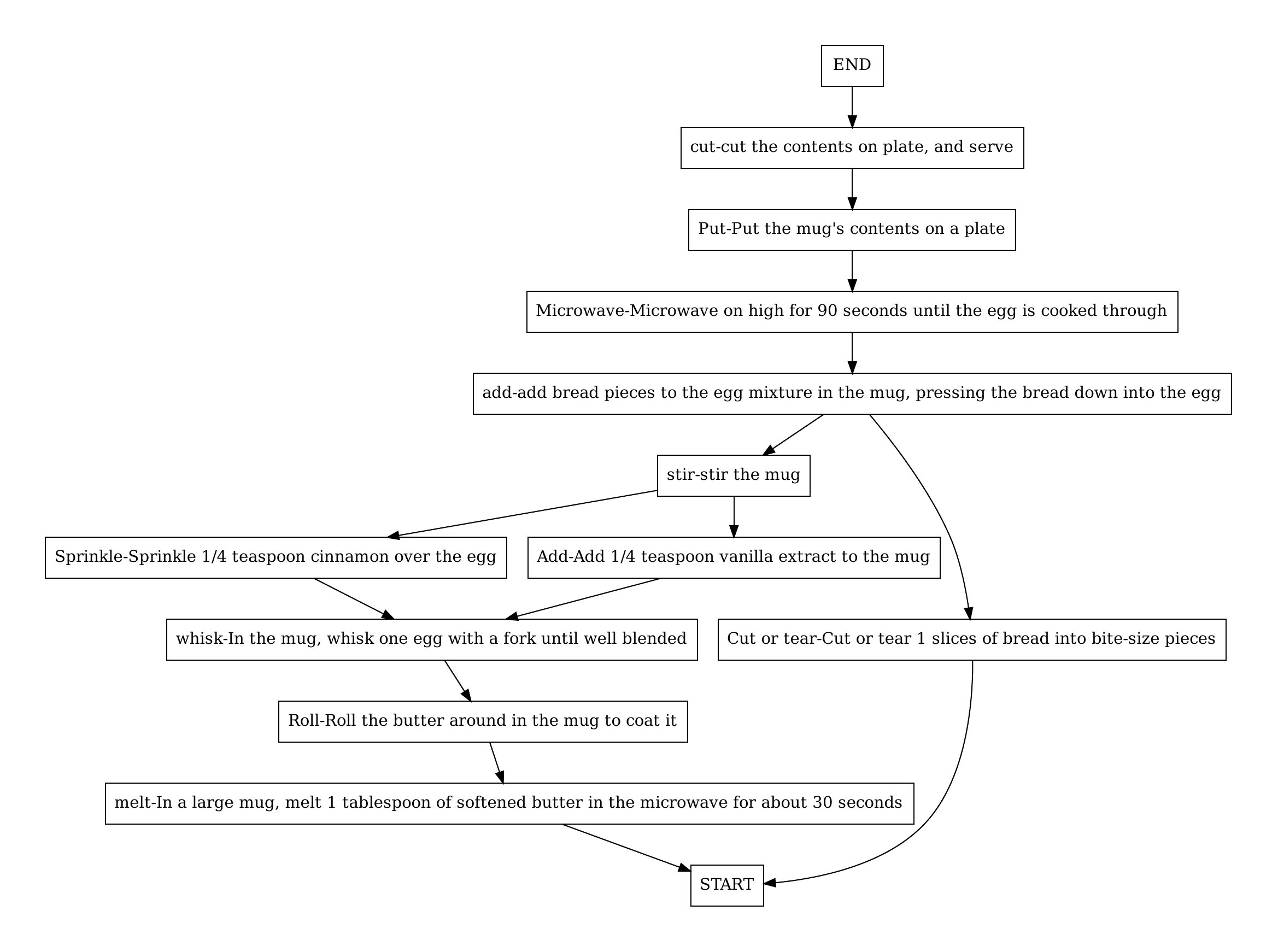}
    }
    \caption{(a) Ground truth task graph and (b) predicted task graph of the scenario Microwave French Toast.}
\end{figure*}

\begin{figure*}[t]
    \centering
    \subfloat[]{
        \includegraphics[width=0.45\textwidth]{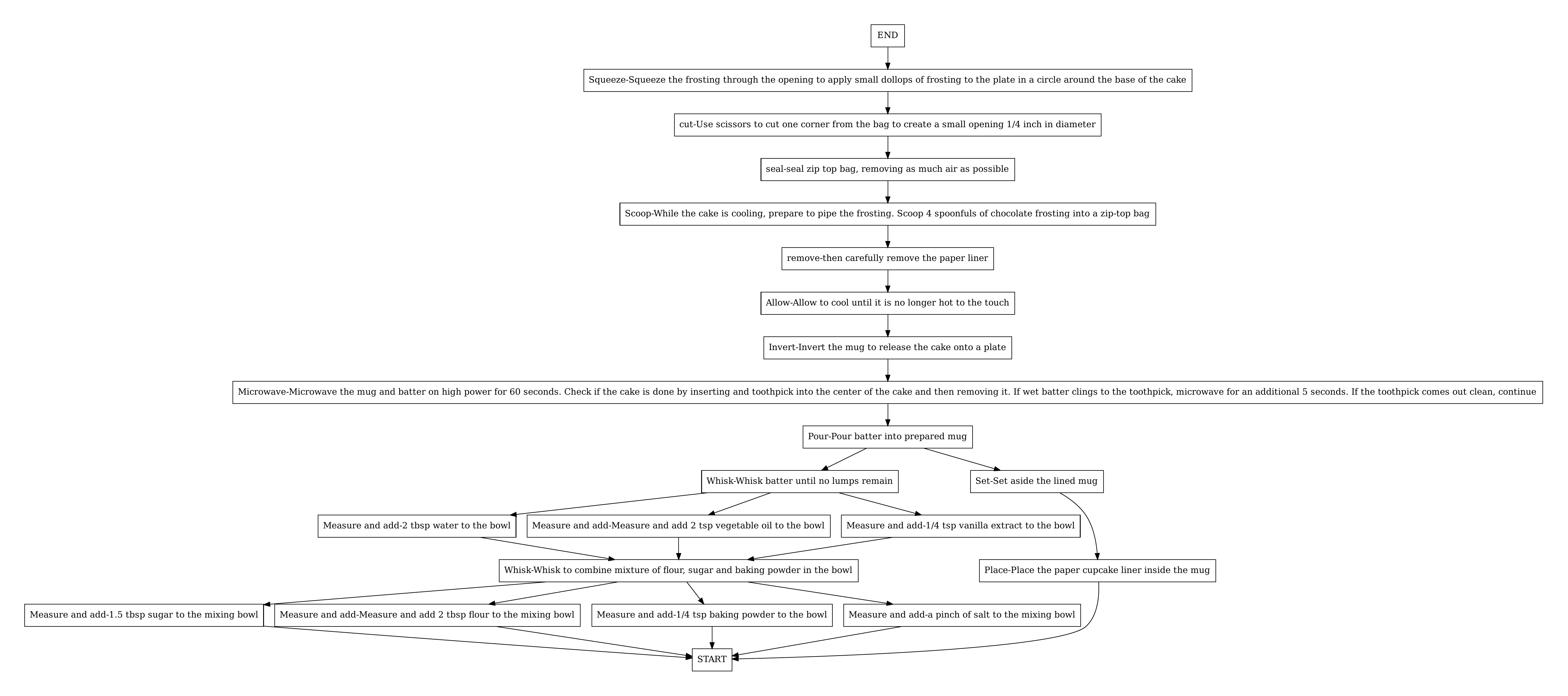}
    }
    \hfill
    \subfloat[]{
        \includegraphics[width=0.45\textwidth]{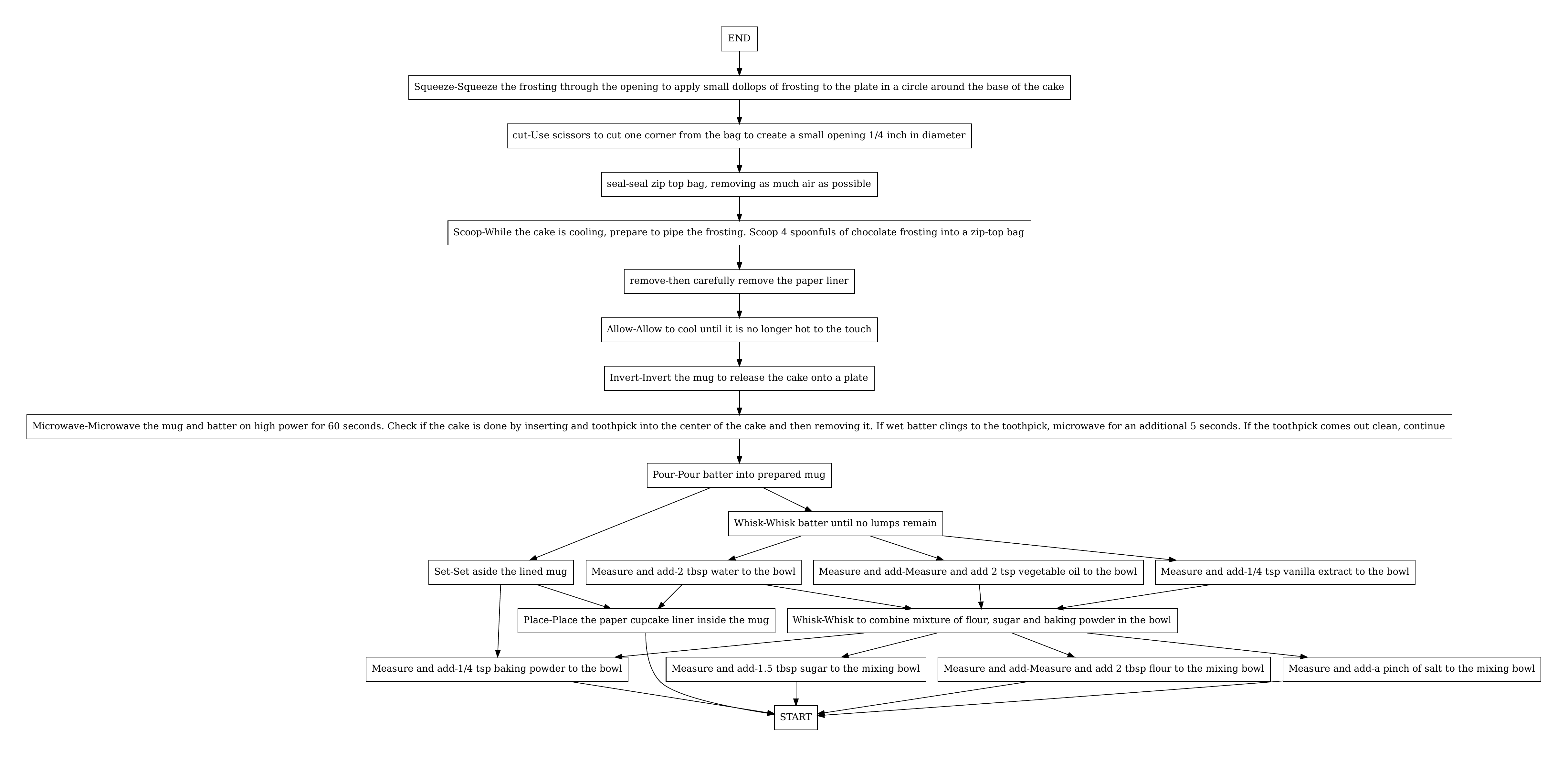}
    }
    \caption{(a) Ground truth task graph and (b) predicted task graph of the scenario Mug Cake.}
\end{figure*}

\begin{figure*}[t]
    \centering
    \subfloat[]{
        \includegraphics[width=0.45\textwidth]{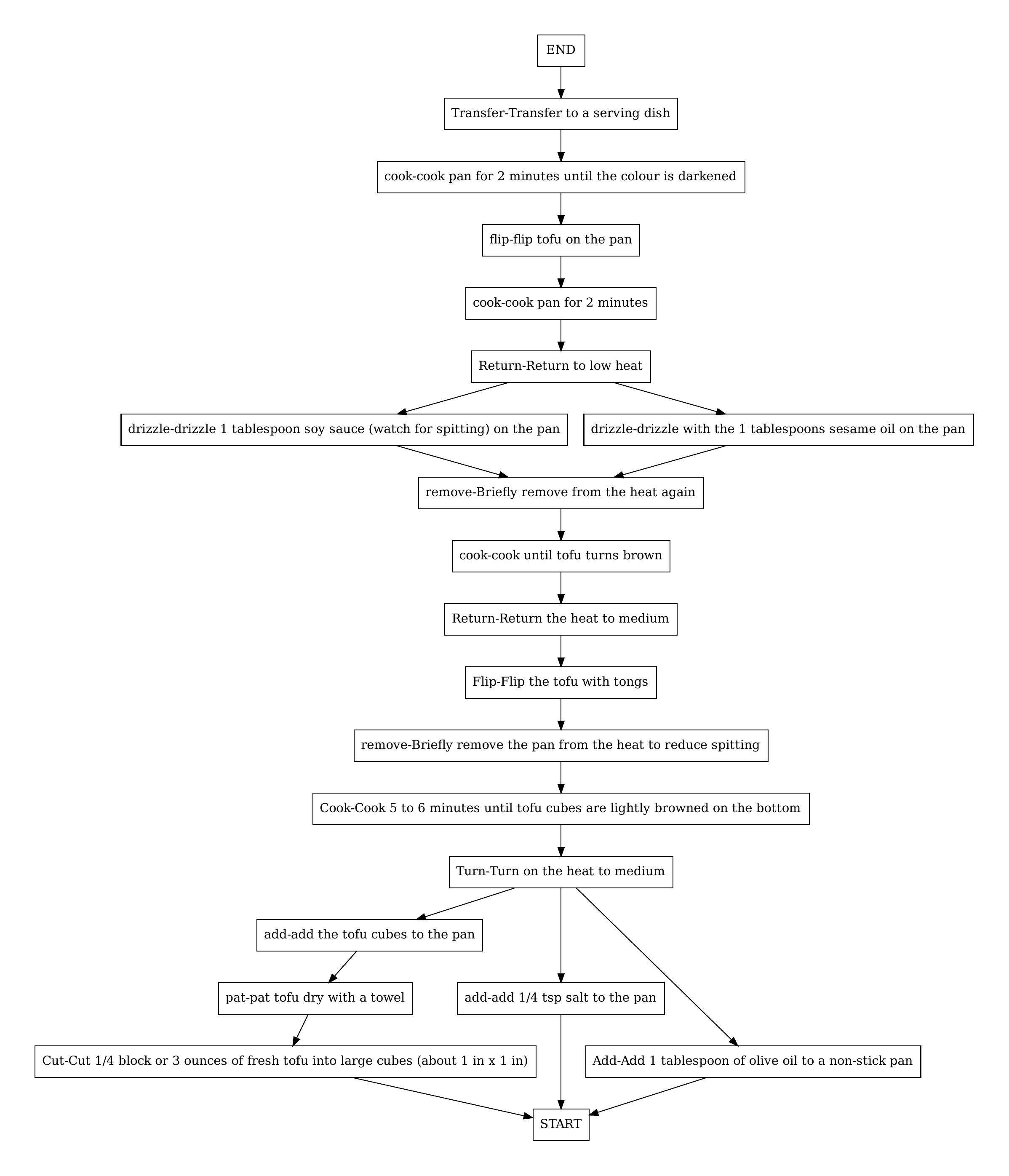}
    }
    \hfill
    \subfloat[]{
        \includegraphics[width=0.45\textwidth]{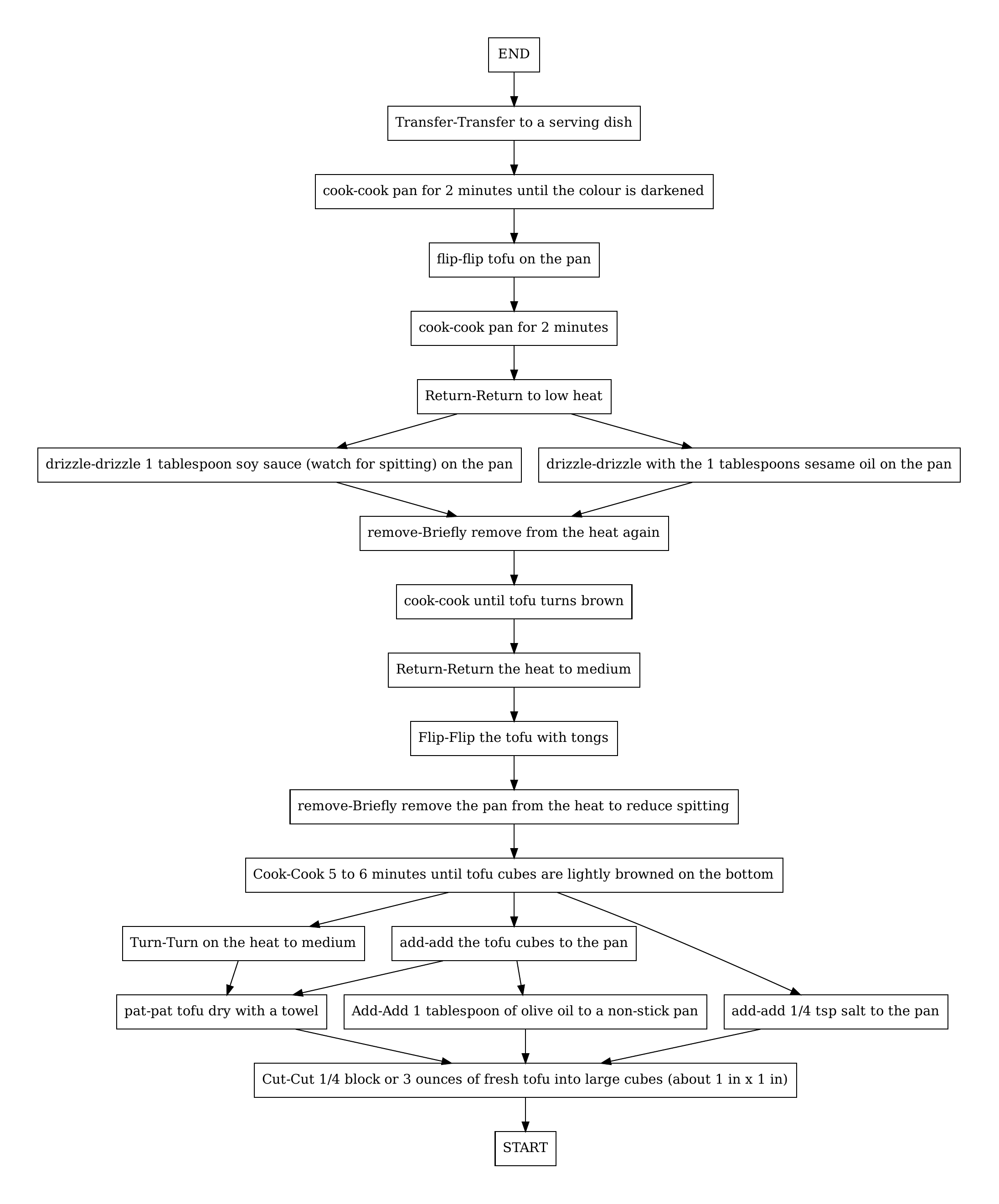}
    }
    \caption{(a) Ground truth task graph and (b) predicted task graph of the scenario Pan Fried Tofu.}
\end{figure*}

\begin{figure*}[t]
    \centering
    \subfloat[]{
        \includegraphics[width=0.45\textwidth]{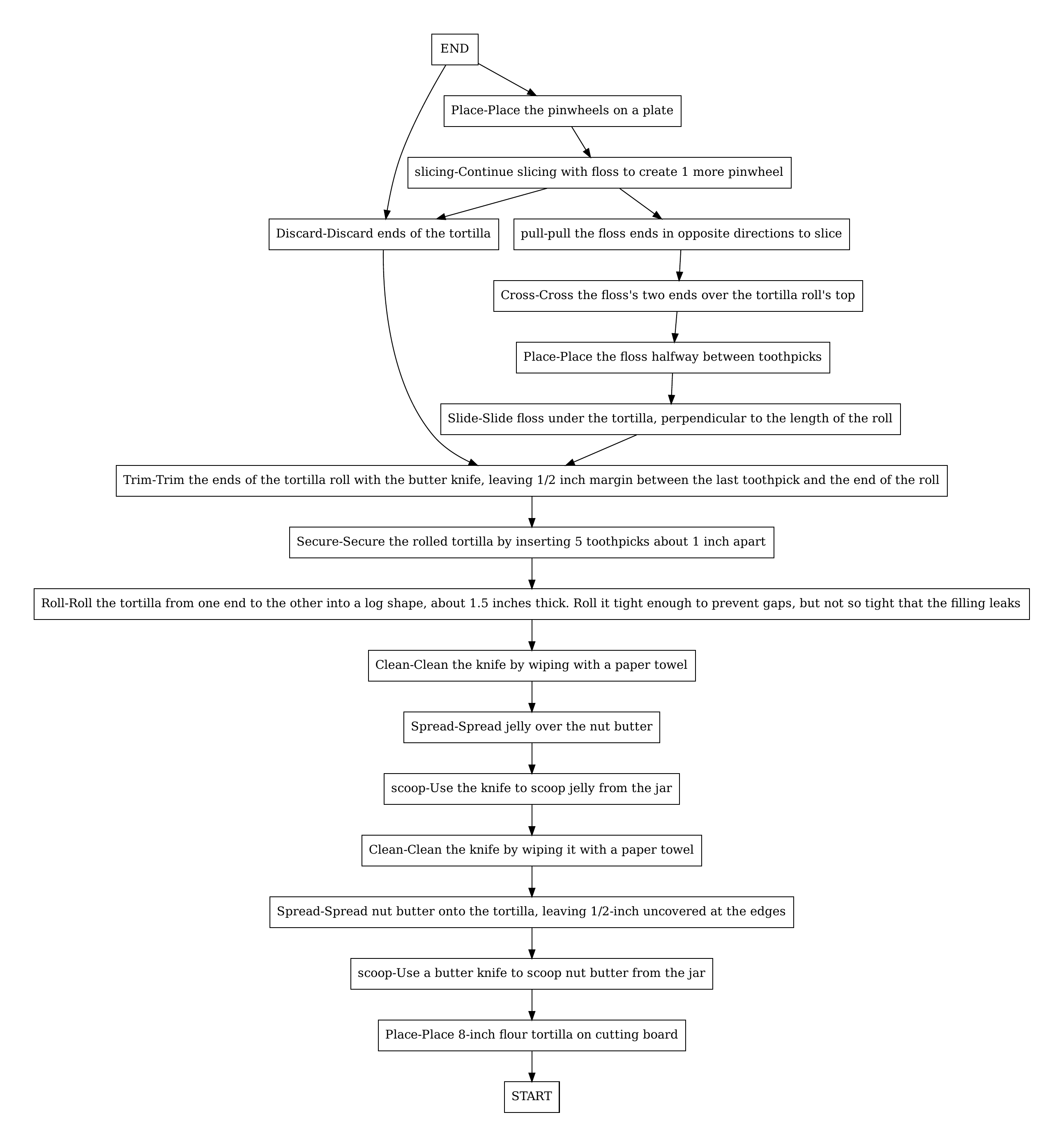}
    }
    \hfill
    \subfloat[]{
        \includegraphics[width=0.45\textwidth]{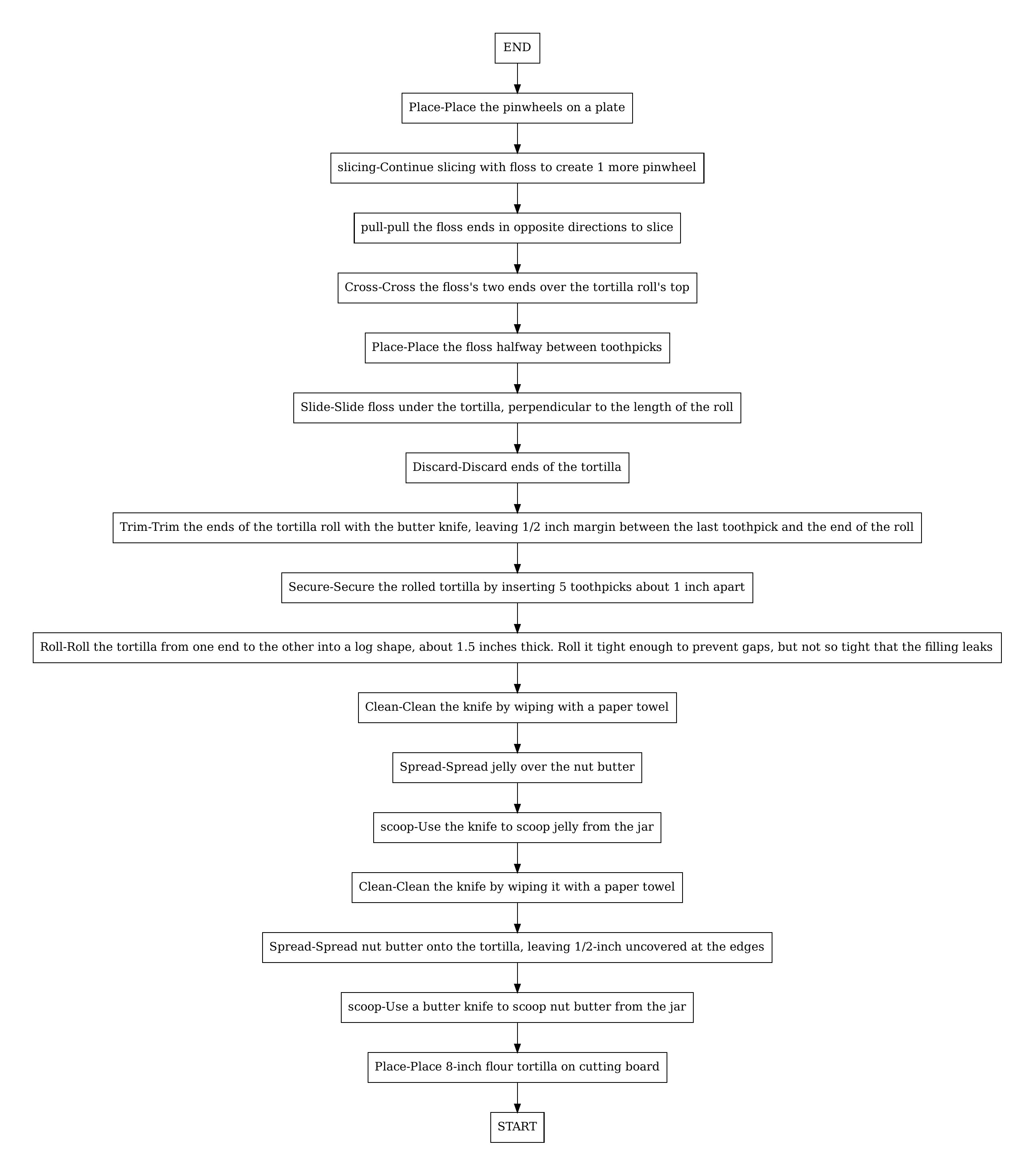}
    }
    \caption{(a) Ground truth task graph and (b) predicted task graph of the scenario Pinwheels.}
\end{figure*}

\begin{figure*}[t]
    \centering
    \subfloat[]{
        \includegraphics[width=0.45\textwidth]{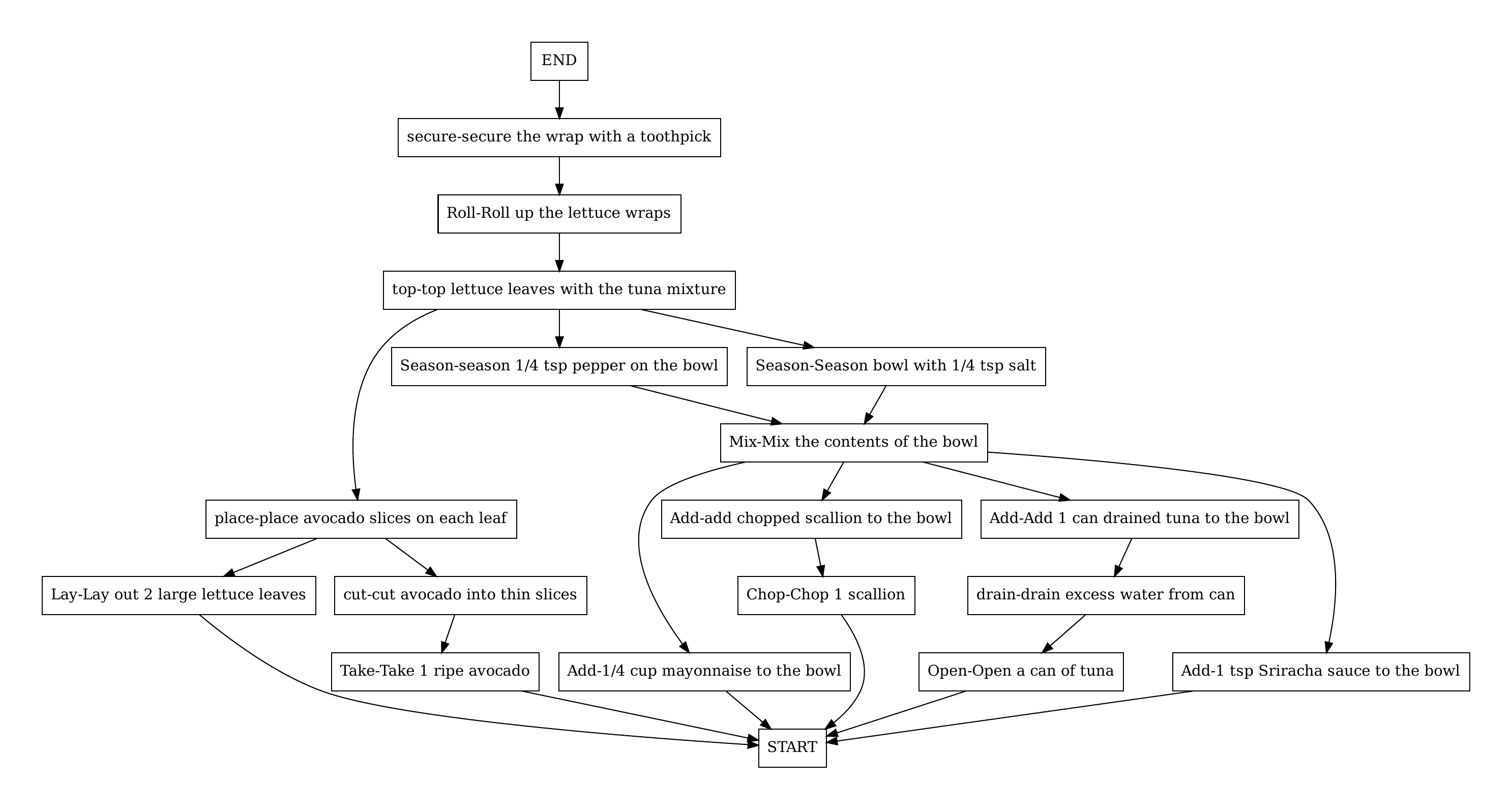}
    }
    \hfill
    \subfloat[]{
        \includegraphics[width=0.45\textwidth]{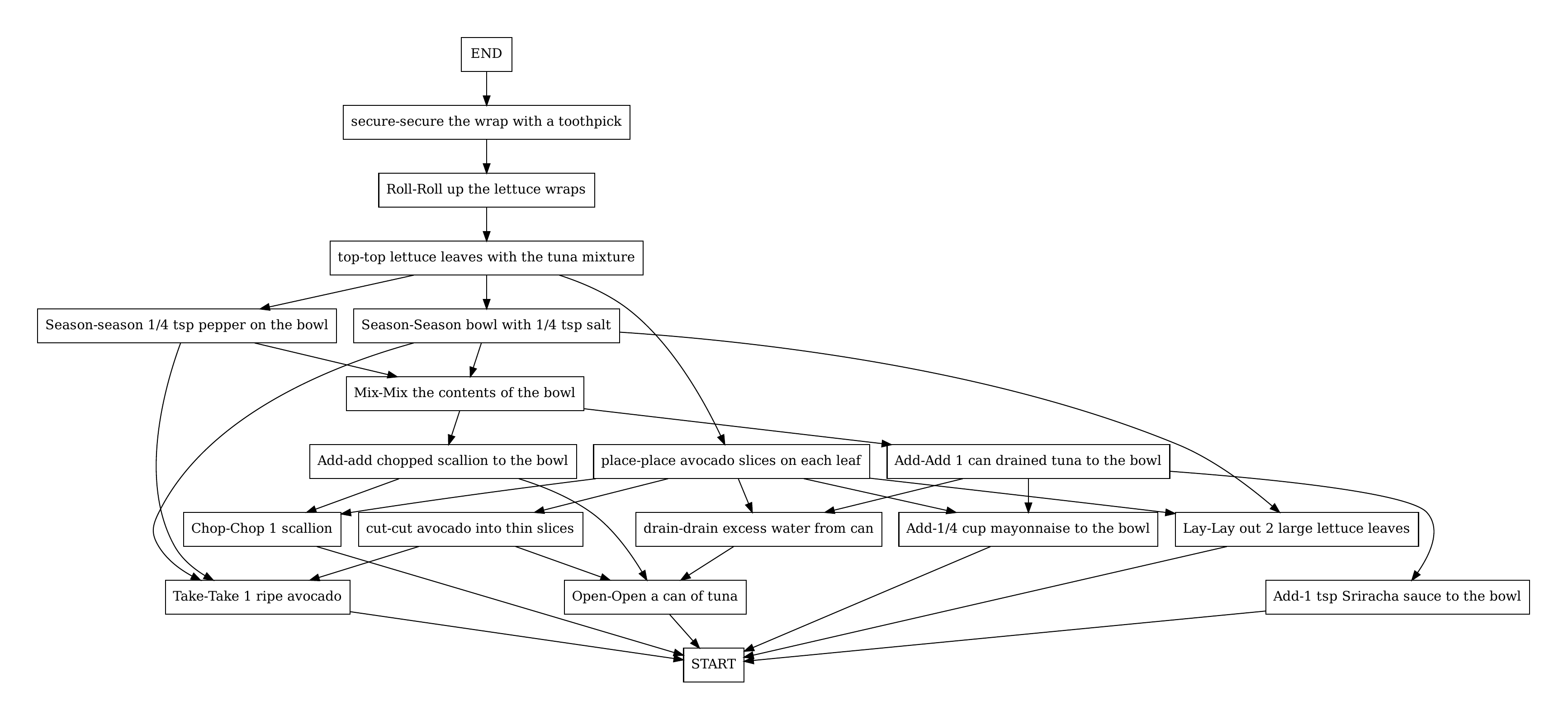}
    }
    \caption{(a) Ground truth task graph and (b) predicted task graph of the scenario Spicy Tuna Avocado Wraps.}
\end{figure*}

\begin{figure*}[t]
    \centering
    \subfloat[]{
        \includegraphics[width=0.45\textwidth]{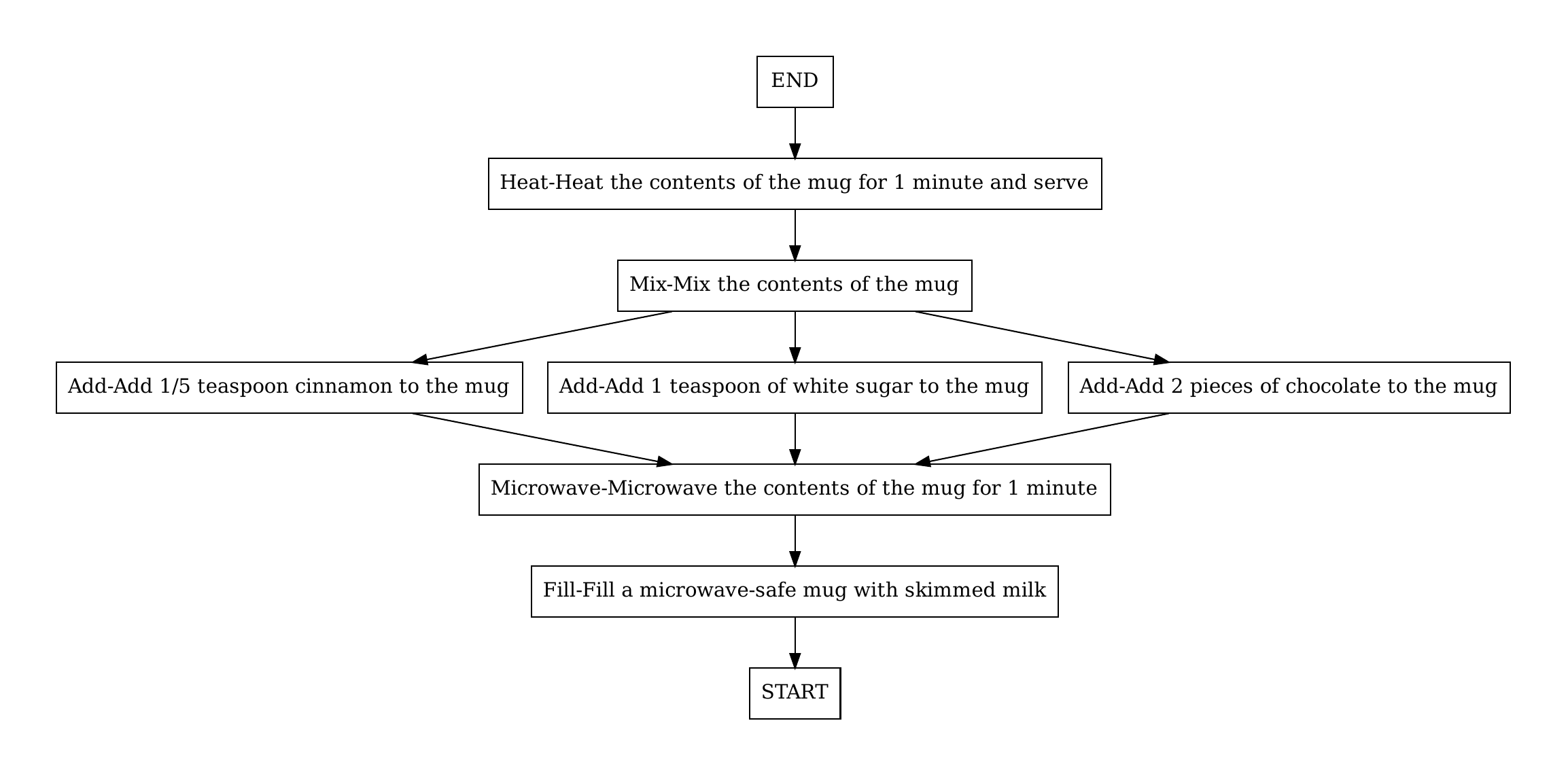}
    }
    \hfill
    \subfloat[]{
        \includegraphics[width=0.45\textwidth]{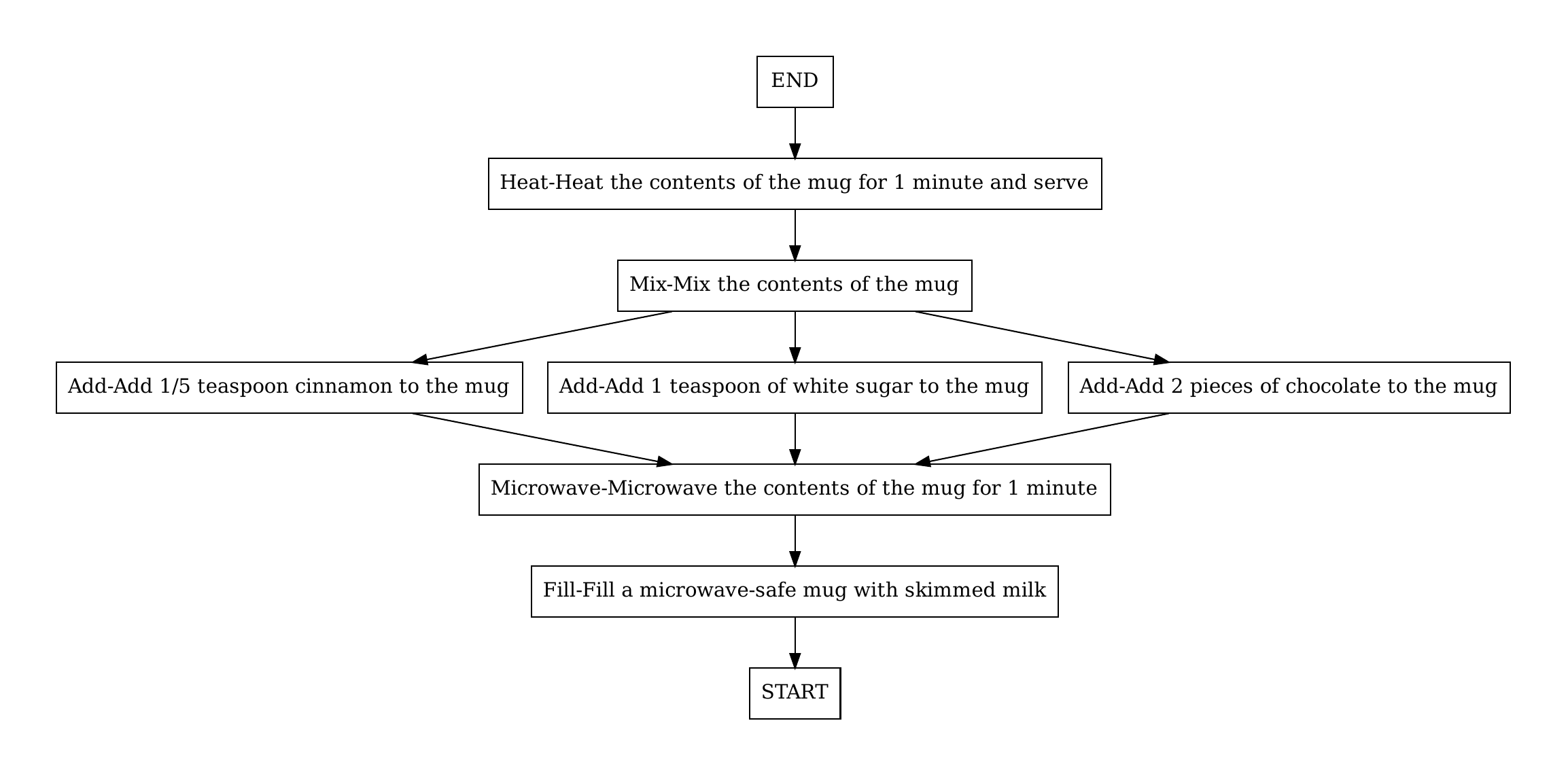}
    }
    \caption{(a) Ground truth task graph and (b) predicted task graph of the scenario Spiced Hot Chocolate.}
\end{figure*}

\begin{figure*}[t]
    \centering
    \subfloat[]{
        \includegraphics[width=0.45\textwidth]{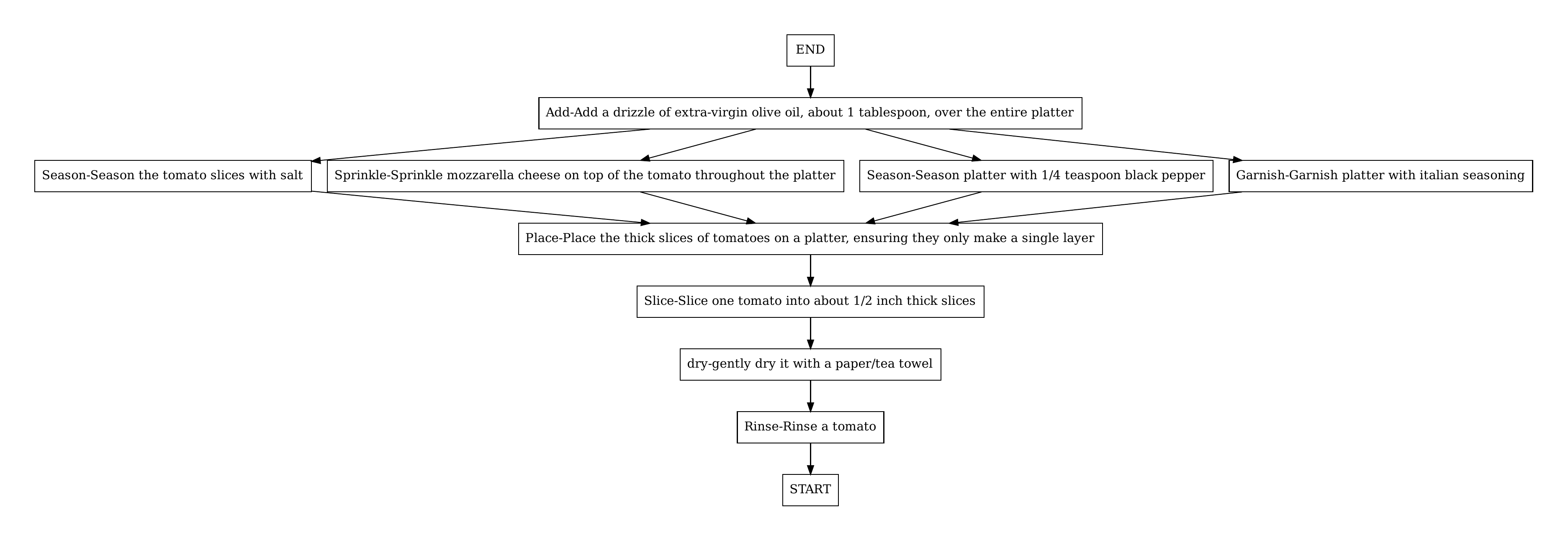}
    }
    \hfill
    \subfloat[]{
        \includegraphics[width=0.45\textwidth]{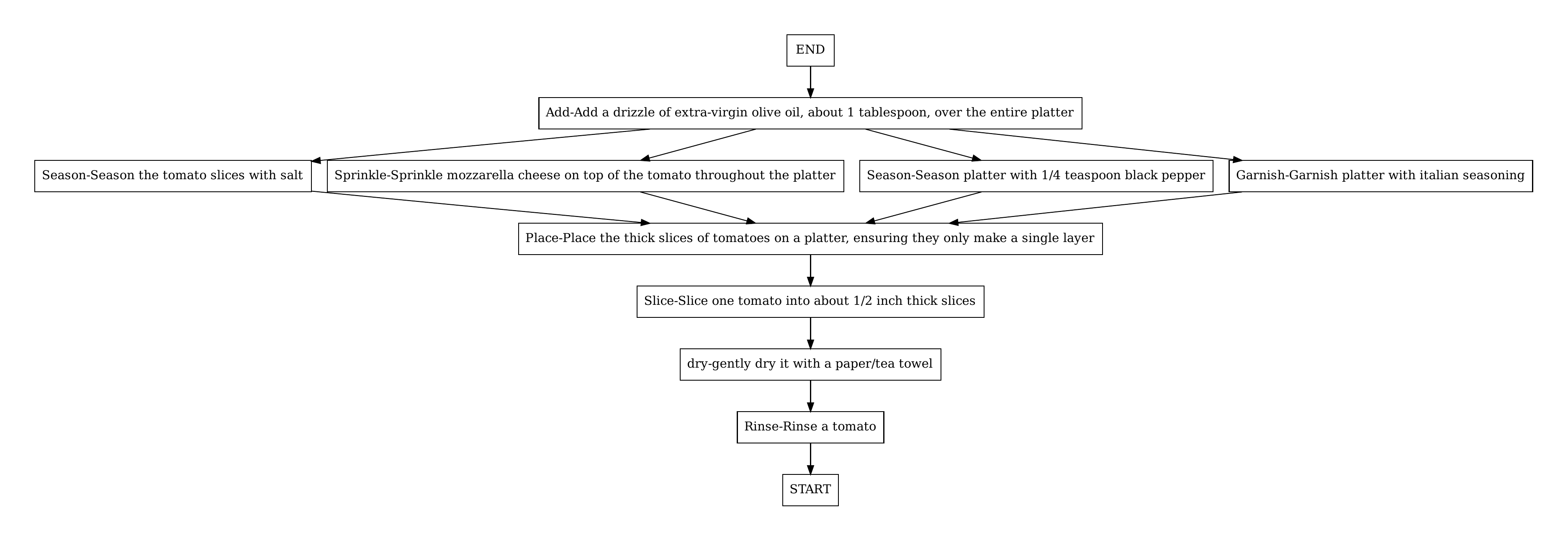}
    }
    \caption{(a) Ground truth task graph and (b) predicted task graph of the scenario Tomato Mozzarella Salad.}
\end{figure*}

\begin{figure*}[t]
    \centering
    \subfloat[]{
        \includegraphics[width=0.45\textwidth]{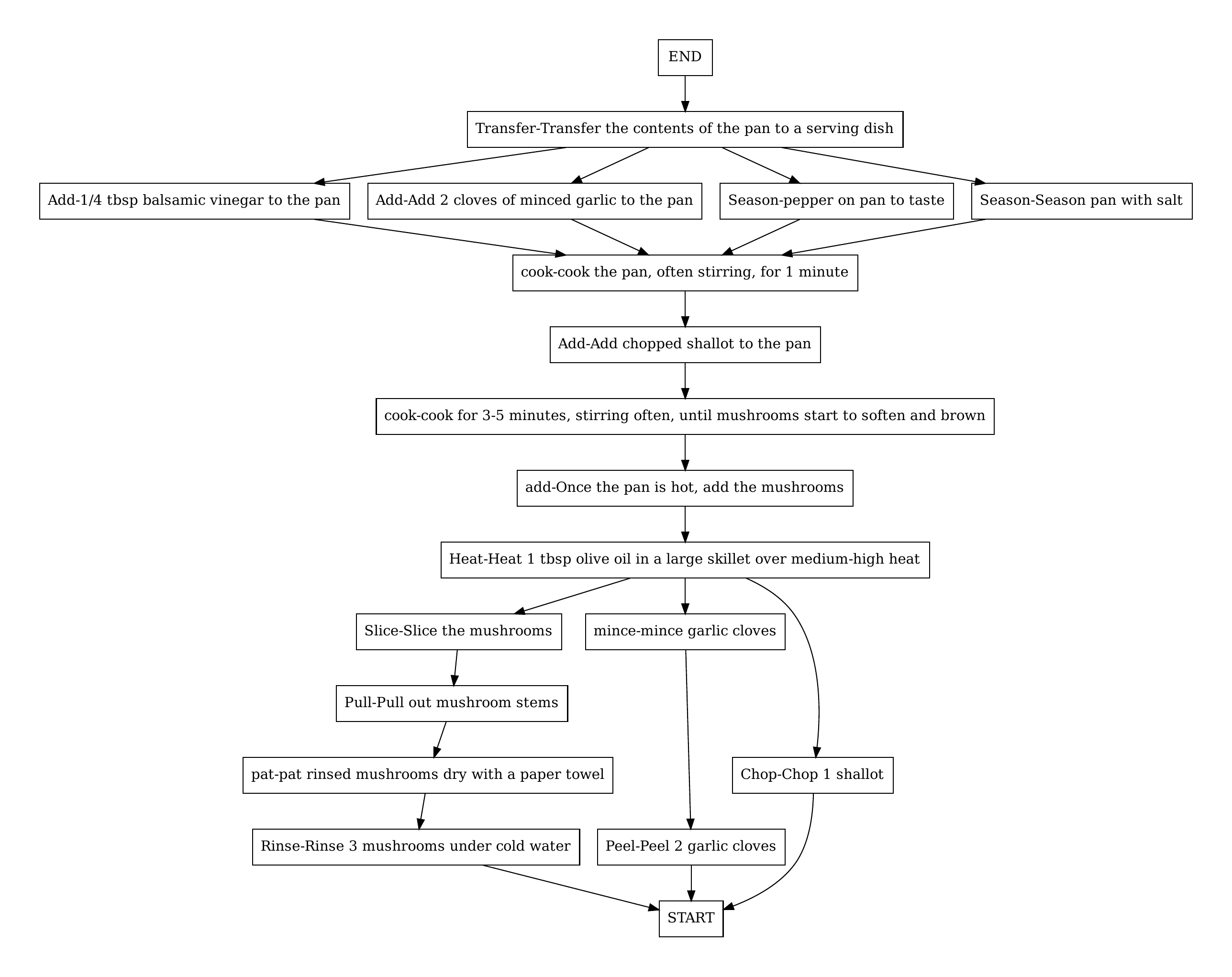}
    }
    \hfill
    \subfloat[]{
        \includegraphics[width=0.45\textwidth]{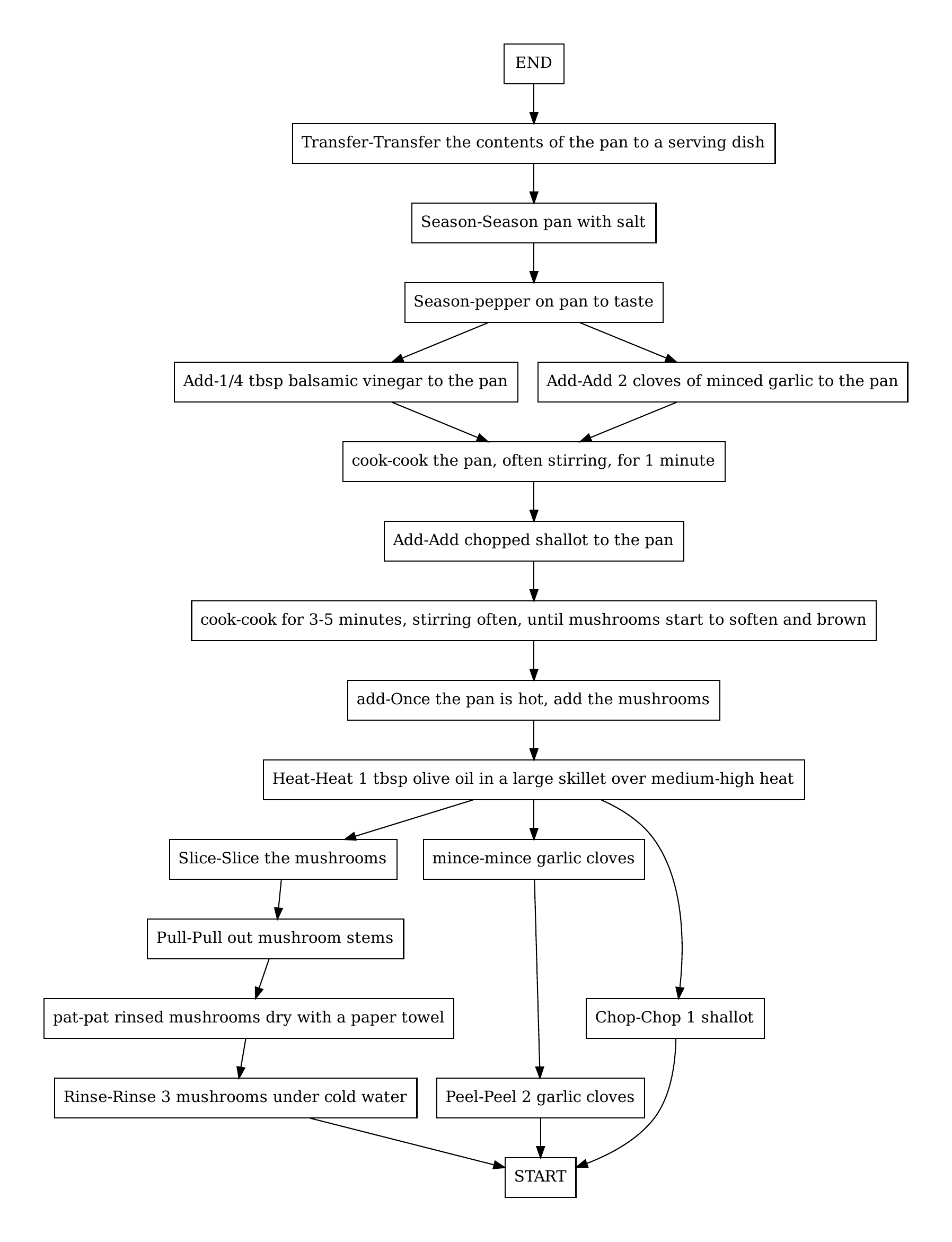}
    }
    \caption{(a) Ground truth task graph and (b) predicted task graph of the scenario Salted Mushrooms.}
\end{figure*}

\begin{figure*}[t]
    \centering
    \subfloat[]{
        \includegraphics[width=0.45\textwidth]{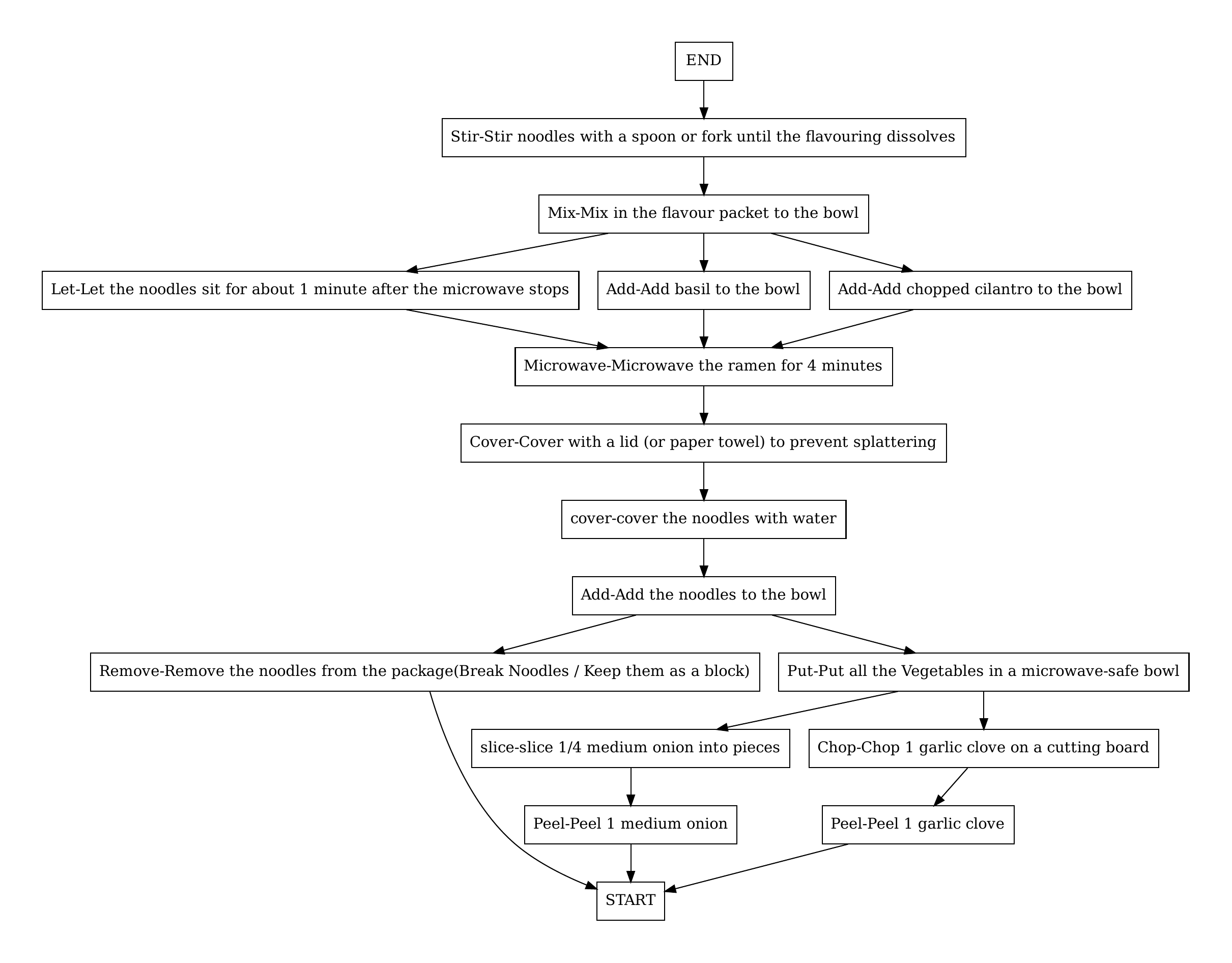}
    }
    \hfill
    \subfloat[]{
        \includegraphics[width=0.45\textwidth]{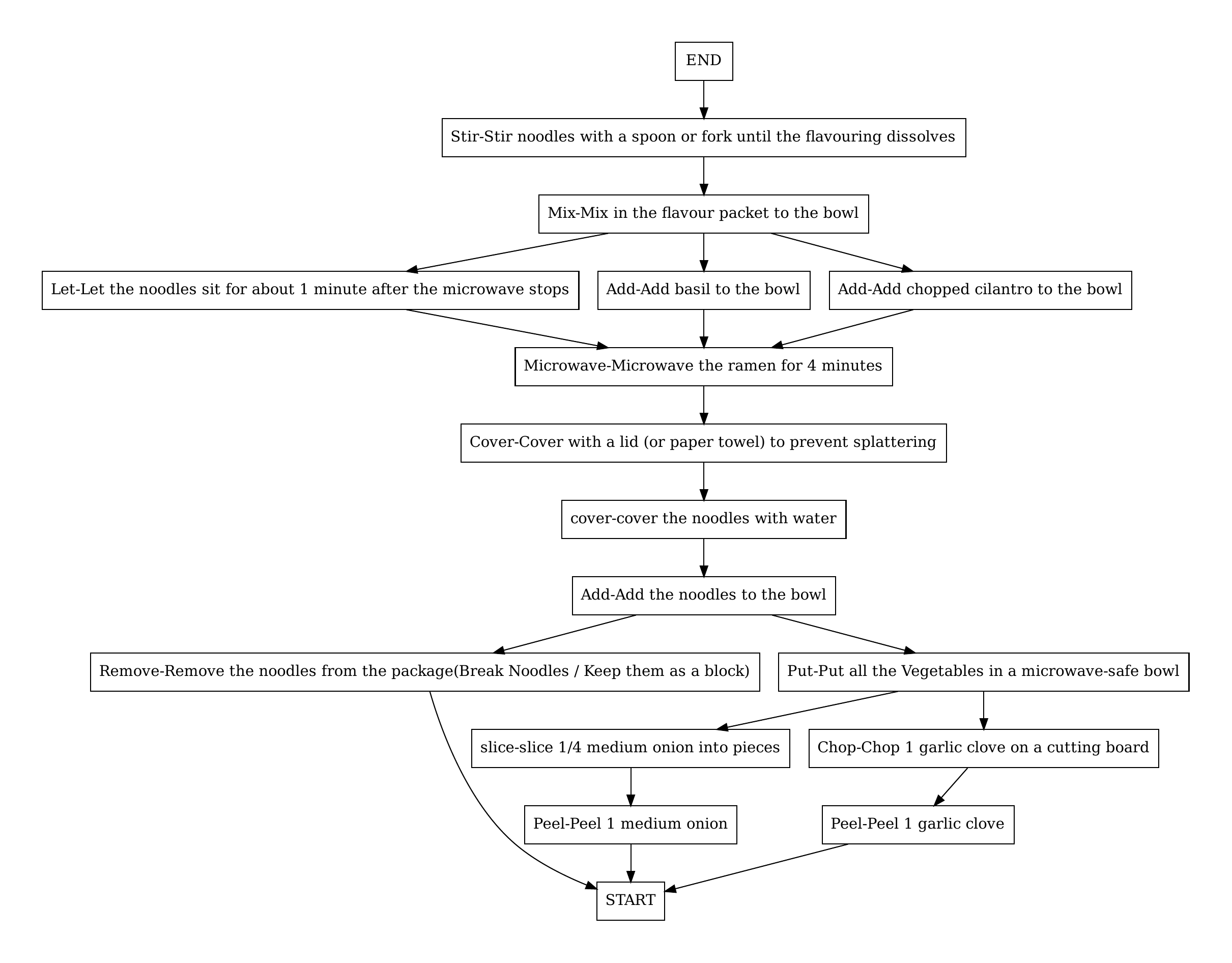}
    }
    \caption{(a) Ground truth task graph and (b) predicted task graph of the scenario Ramen.}
\end{figure*}

\begin{figure*}[t]
    \centering
    \subfloat[]{
        \includegraphics[width=0.45\textwidth]{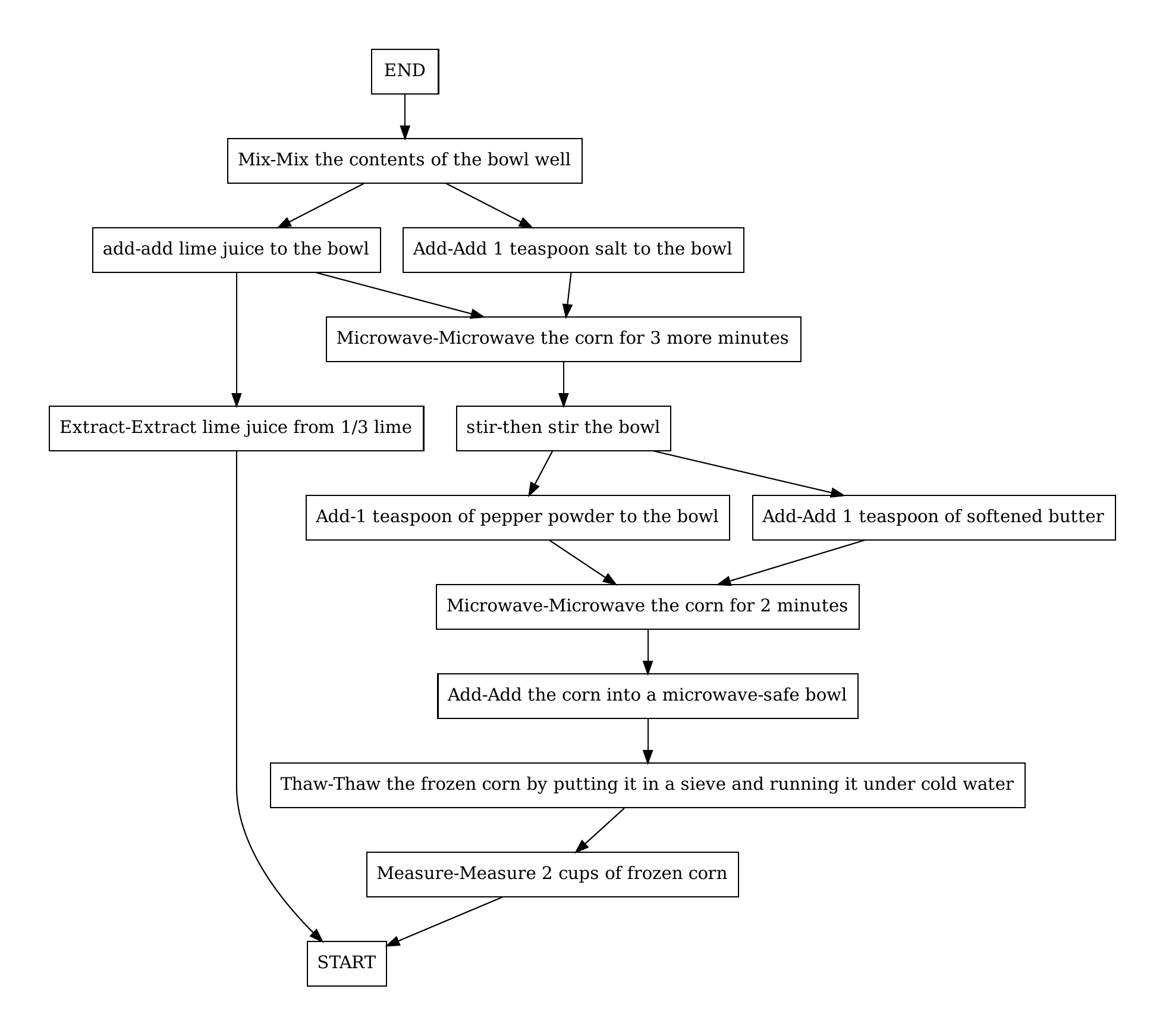}
    }
    \hfill
    \subfloat[]{
        \includegraphics[width=0.45\textwidth]{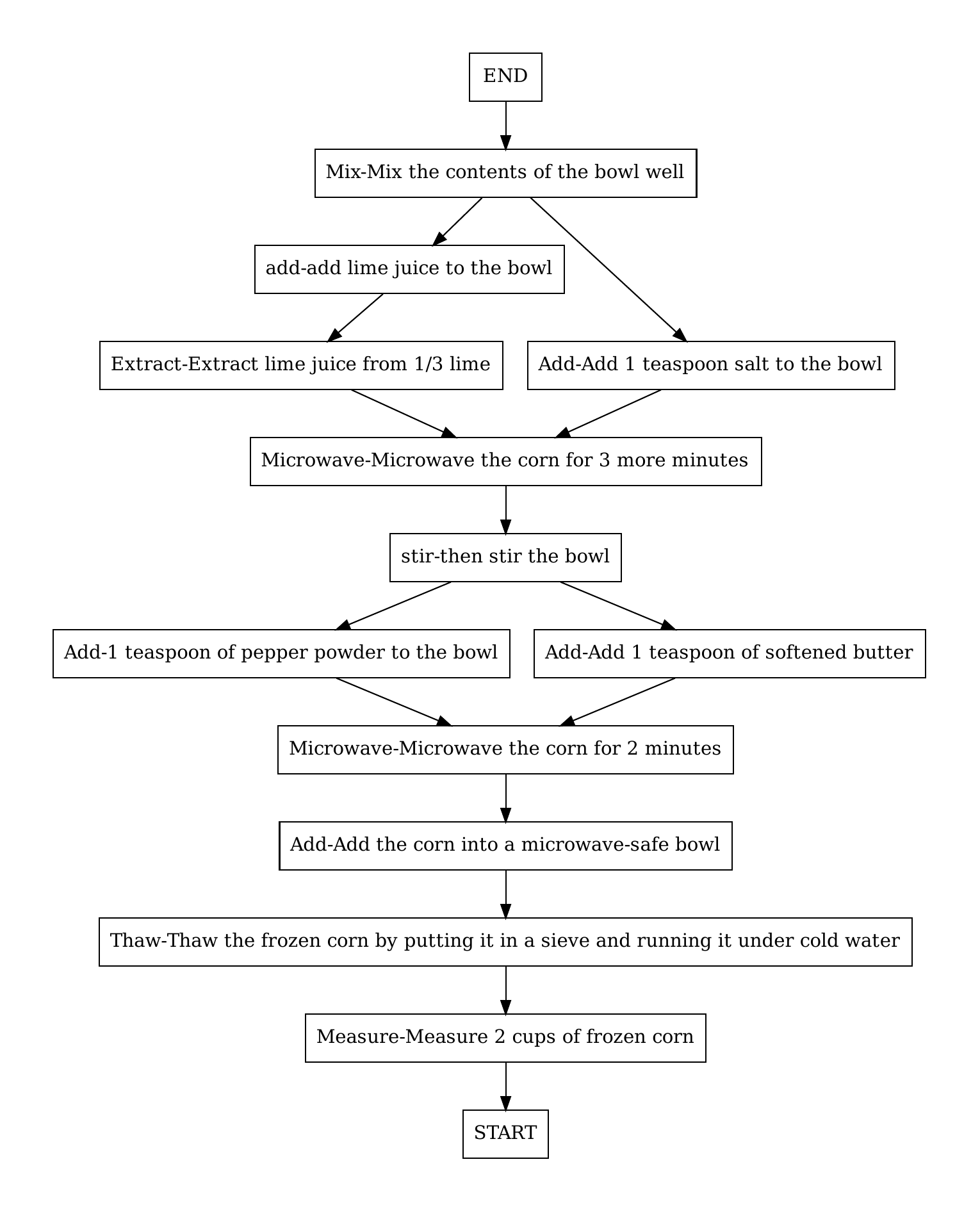}
    }
    \caption{(a) Ground truth task graph and (b) predicted task graph of the scenario Butter Corn Cup.}
\end{figure*}

\begin{figure*}[t]
    \centering
    \subfloat[]{
        \includegraphics[width=0.45\textwidth]{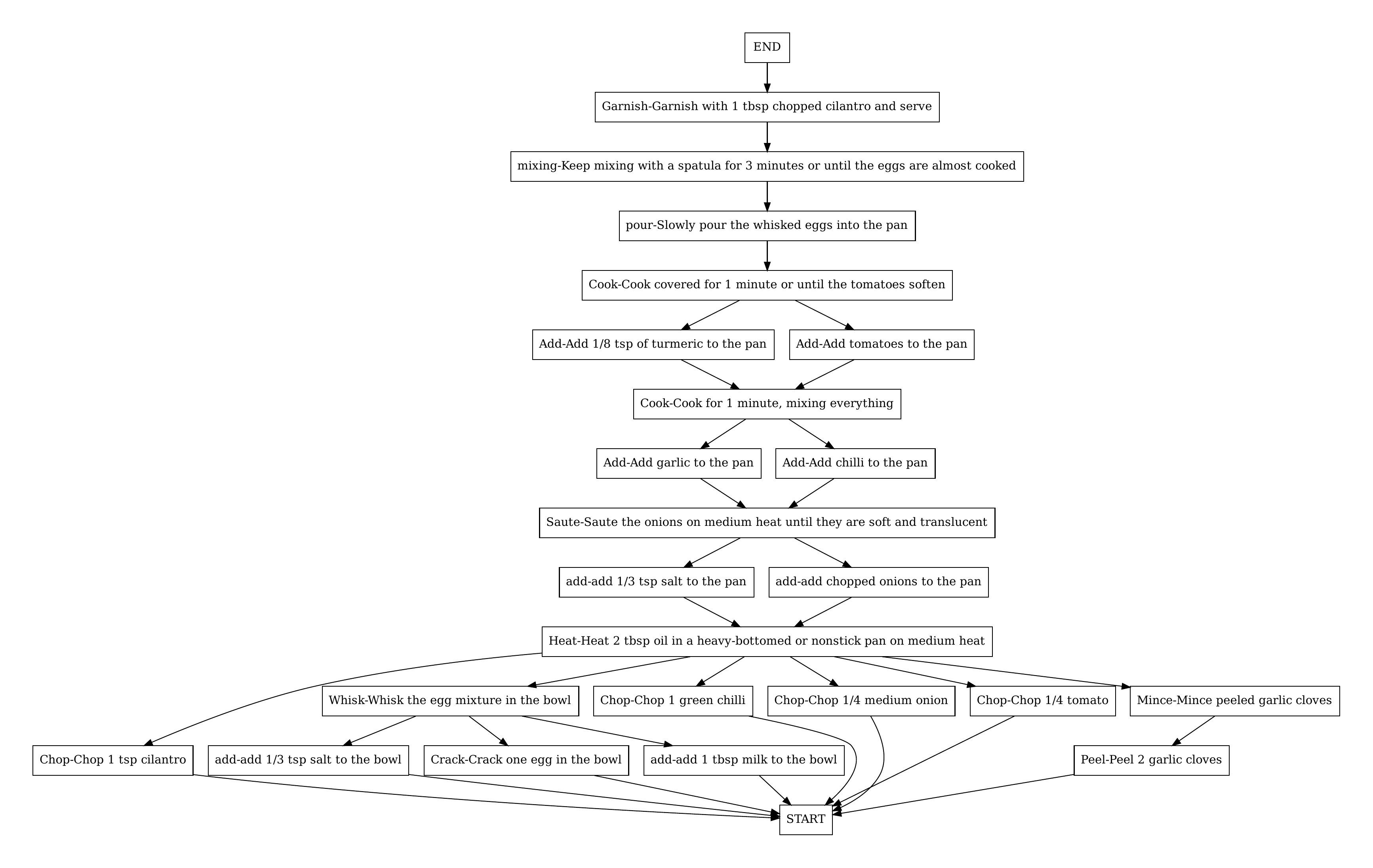}
    }
    \hfill
    \subfloat[]{
        \includegraphics[width=0.45\textwidth]{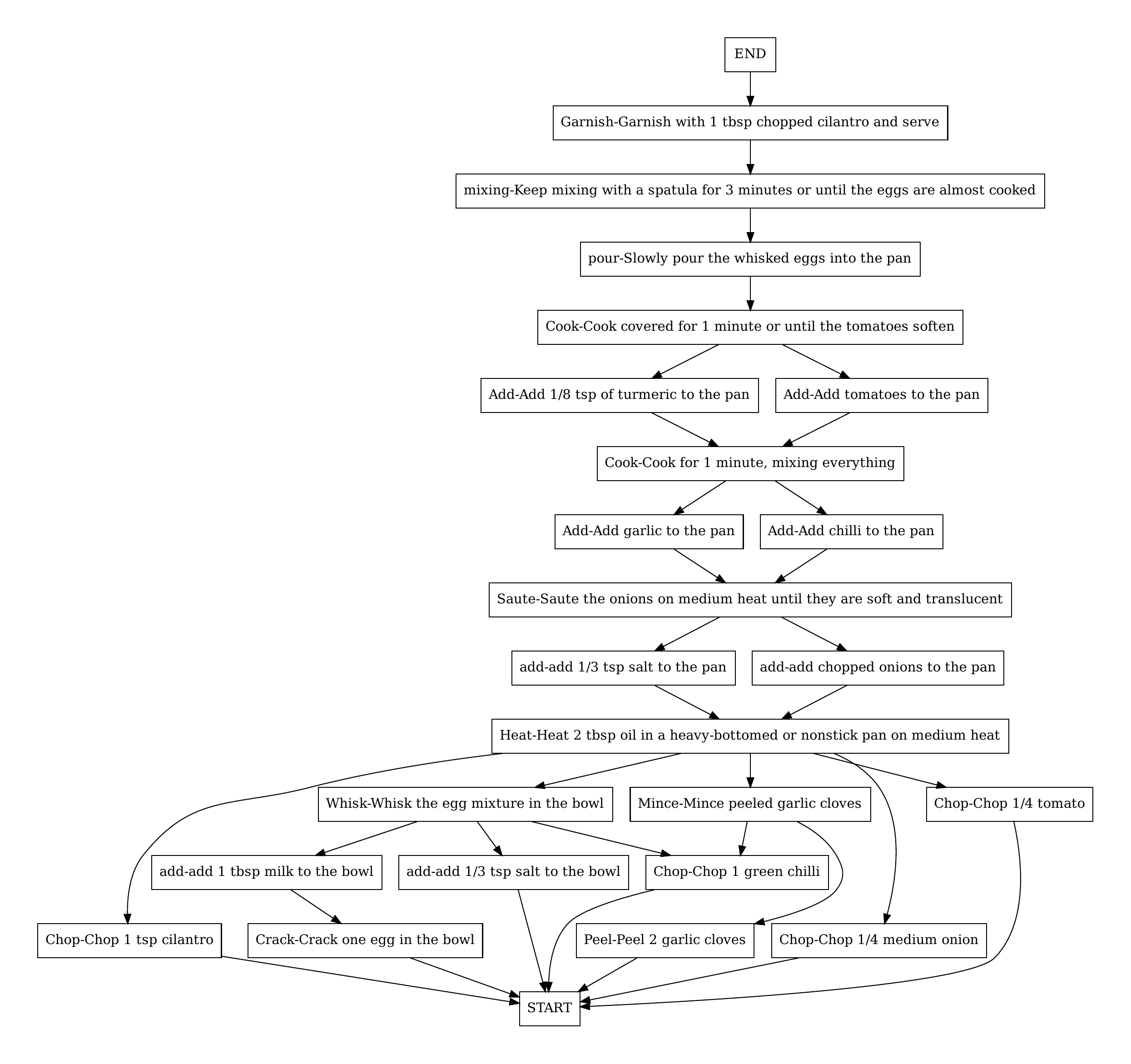}
    }
    \caption{(a) Ground truth task graph and (b) predicted task graph of the scenario Scrambled Eggs.}
\end{figure*}

\begin{figure*}[t]
    \centering
    \subfloat[]{
        \includegraphics[width=0.45\textwidth]{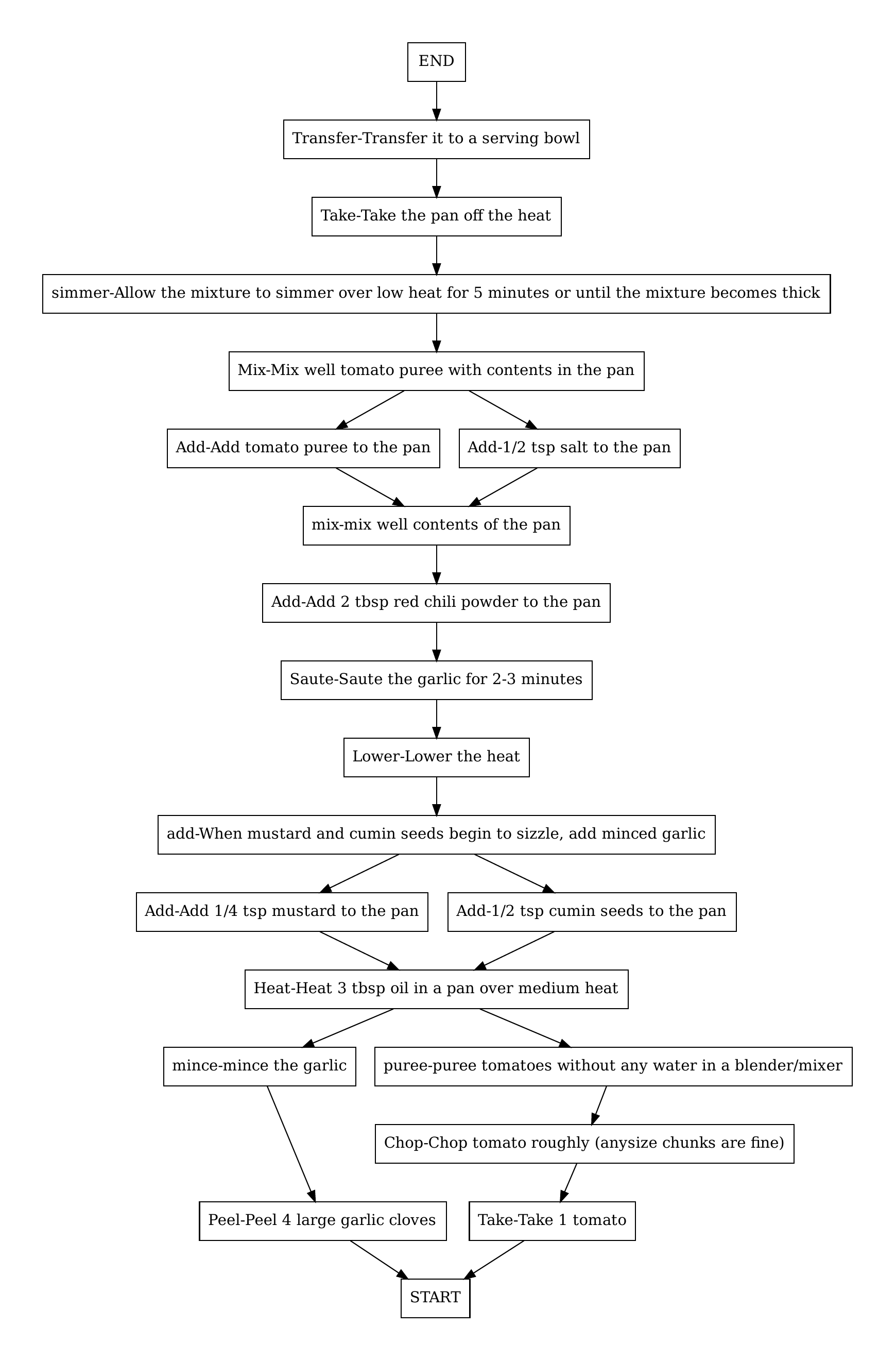}
    }
    \hfill
    \subfloat[]{
        \includegraphics[width=0.45\textwidth]{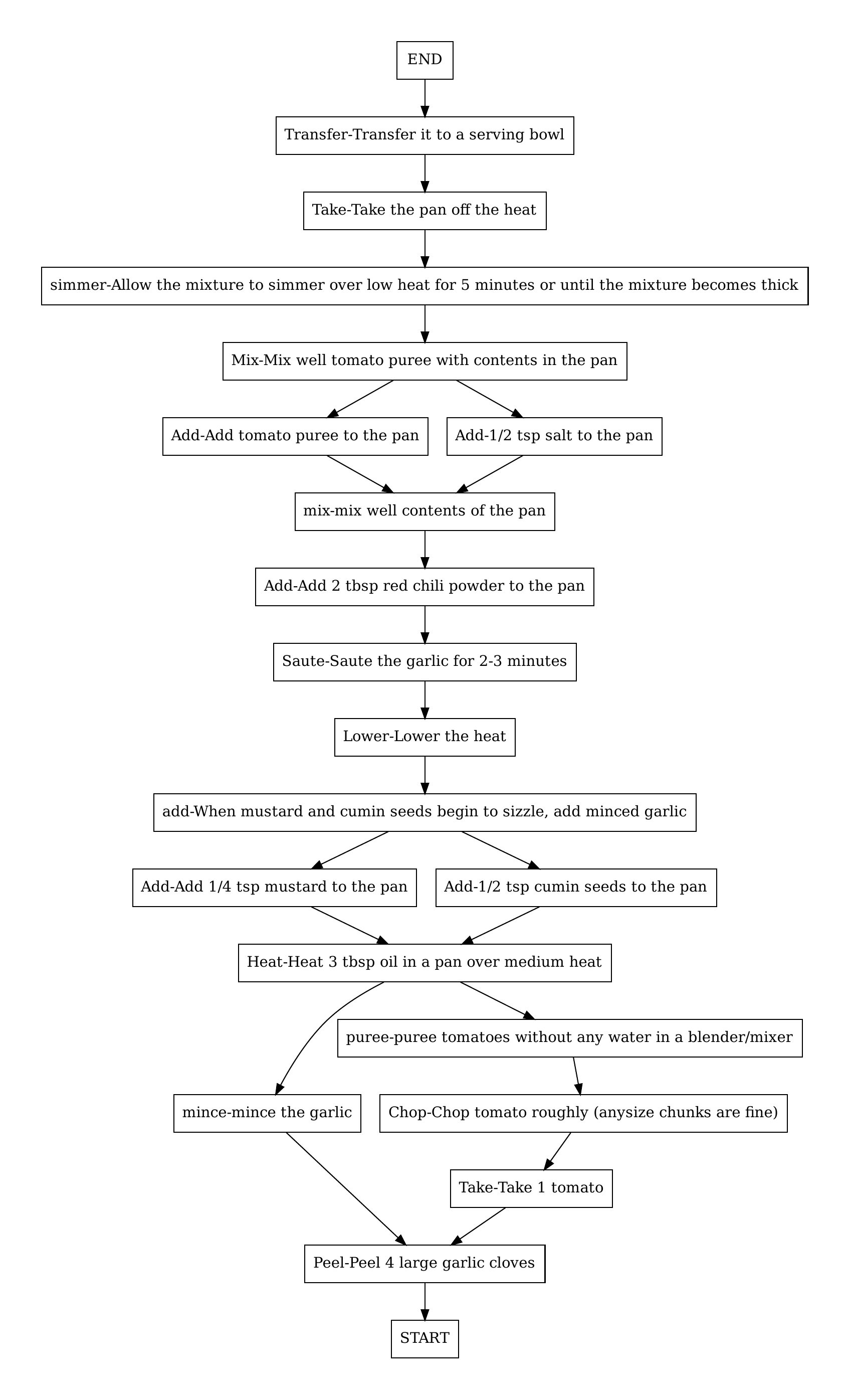}
    }
    \caption{(a) Ground truth task graph and (b) predicted task graph of the scenario Tomato Chutney.}
    \label{fig:qualitative24}
\end{figure*}

\begin{figure*}
    \centering
    \subfloat[MSGI]{
        \includegraphics[width=0.45\textwidth]{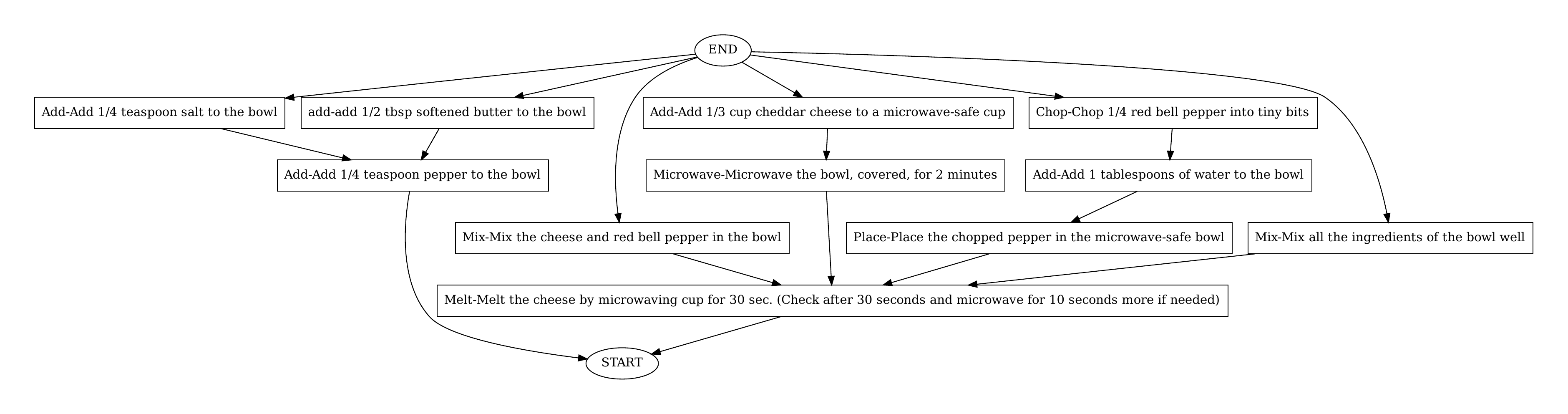}
    }
    \subfloat[Llama-3.1-405B-instruct]{
        \includegraphics[width=0.45\textwidth]{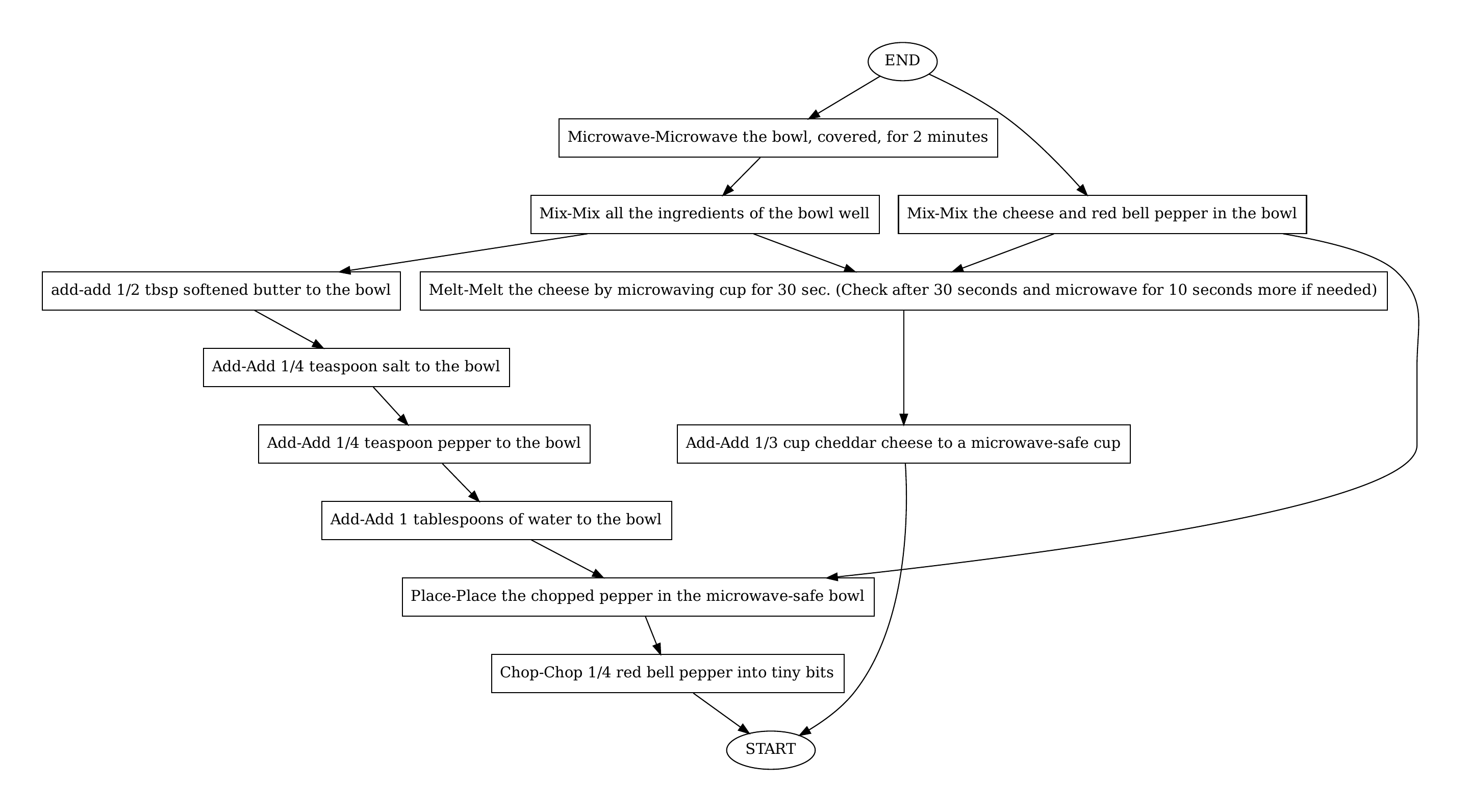}
    }\\
    \subfloat[Count-Based]{
        \includegraphics[width=0.45\textwidth]{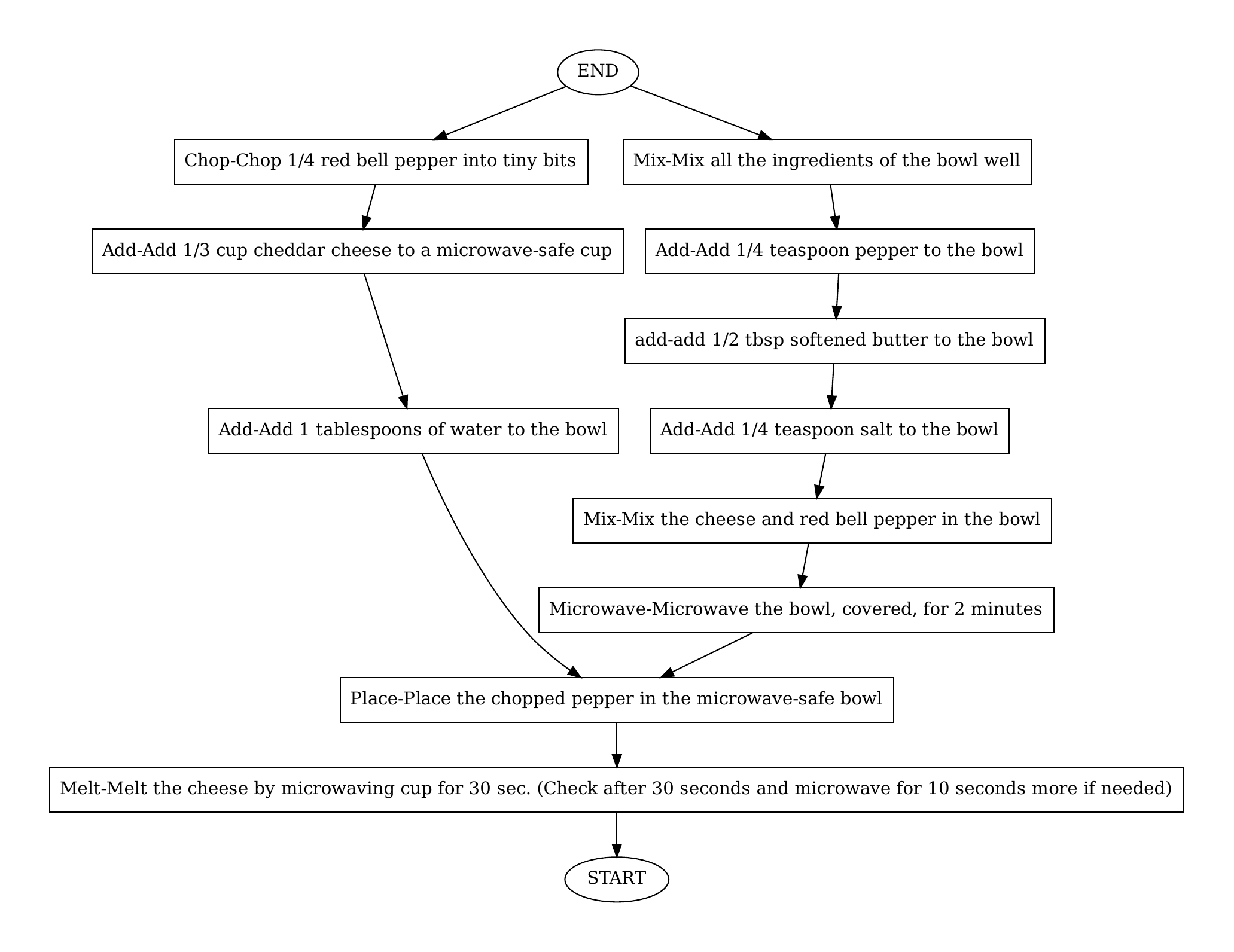}
    }
    \subfloat[MSG$^2$]{
        \includegraphics[width=0.45\textwidth]{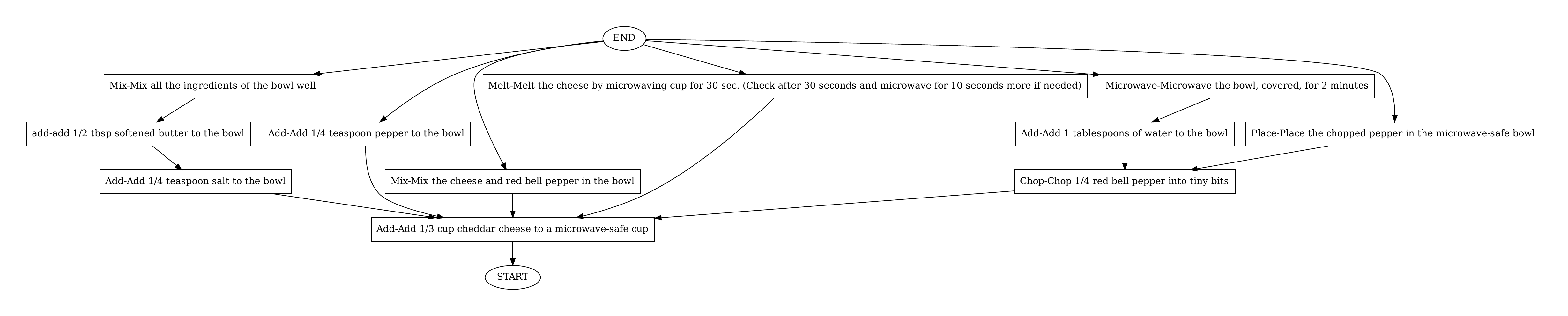}
    }

    \subfloat[TGT-text]{
        \includegraphics[width=0.45\textwidth]{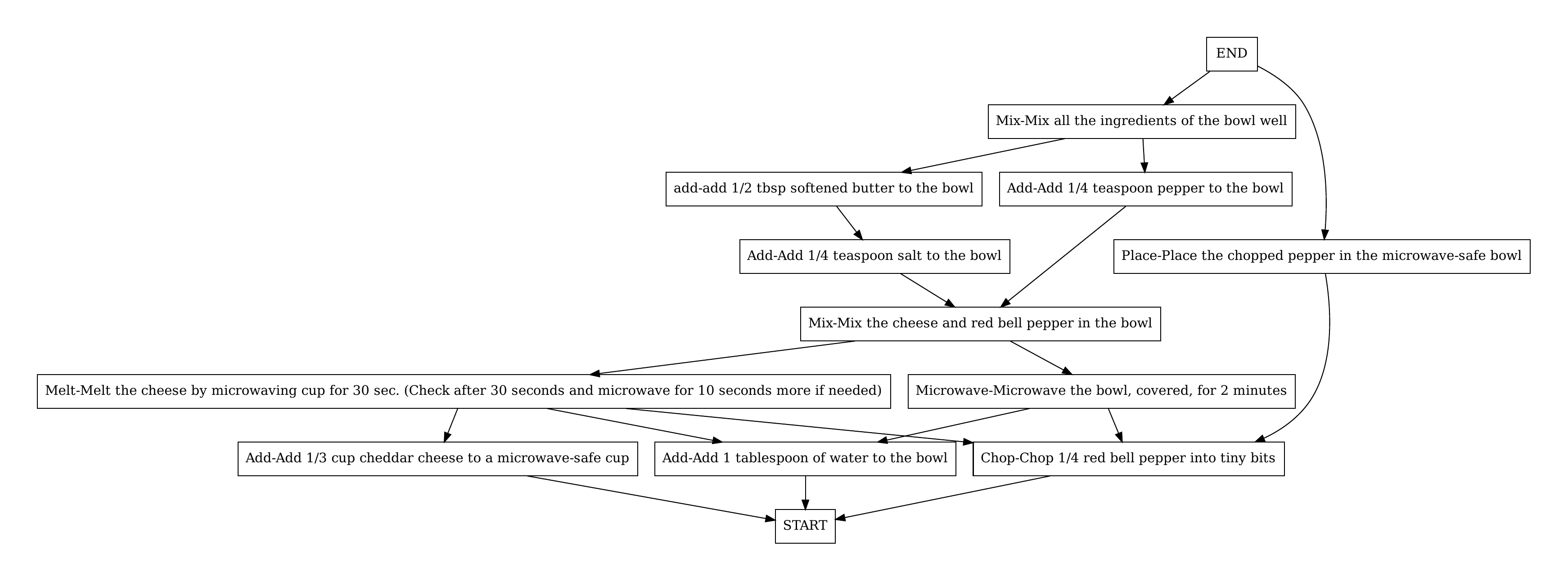}
    }
    \subfloat[DO]{
        \includegraphics[width=0.45\textwidth]{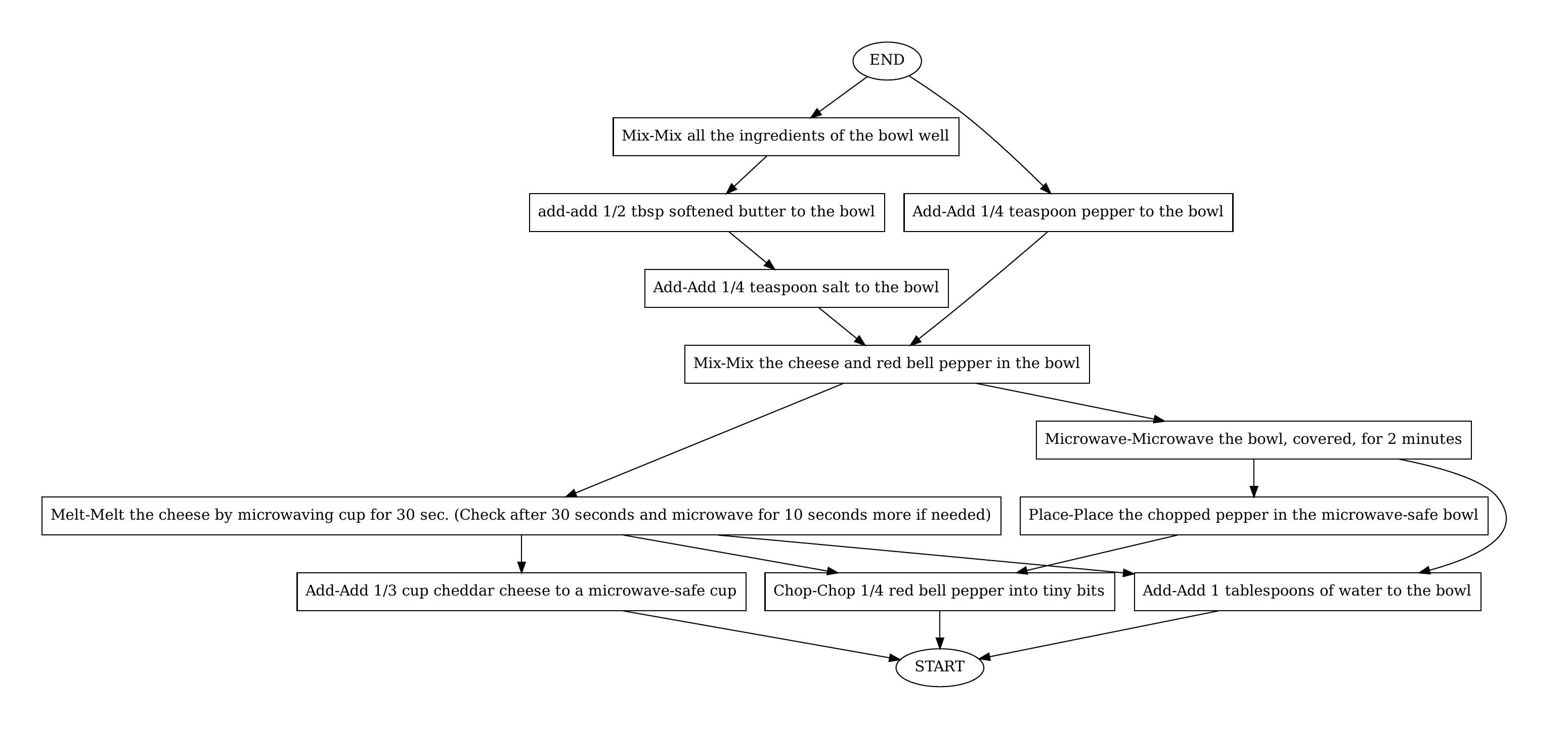}
    }\\
    \subfloat[GT]{
        \includegraphics[width=0.45\textwidth]{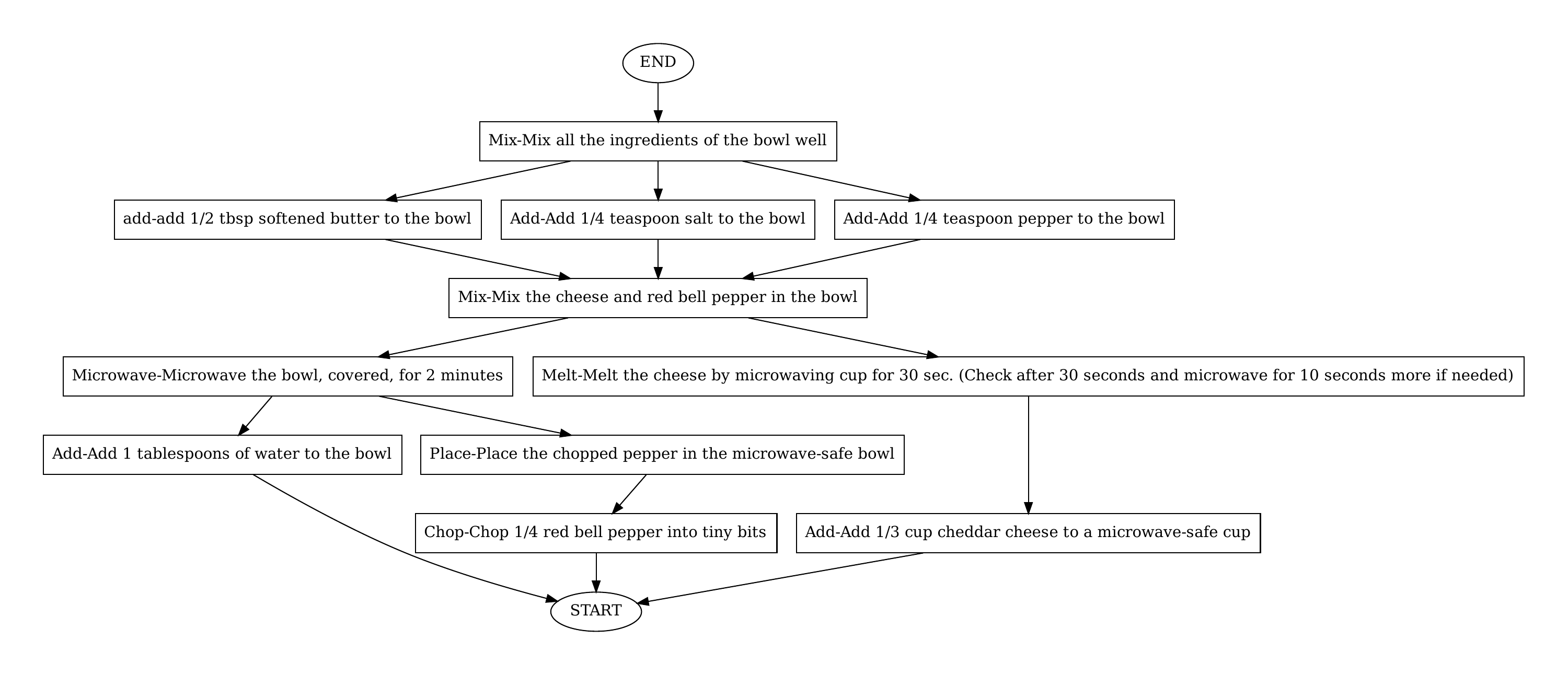}
    }
    \caption{Task graphs of the scenario ``Cheese Pimiento'' from CaptainCook4D generated with (a) MSGI, (b) Llama-3.1-405B-Instruct, (c) Count-Based, (d) MSG$^2$, (e) TGT using textual embedding, and (f) DO. (g) reports the ground truth.}
    \label{fig:comparison_captaincook4d}
\end{figure*}

\begin{figure*}
    \centering
    \subfloat[MSGI]{%
        \includegraphics[width=0.45\textwidth]{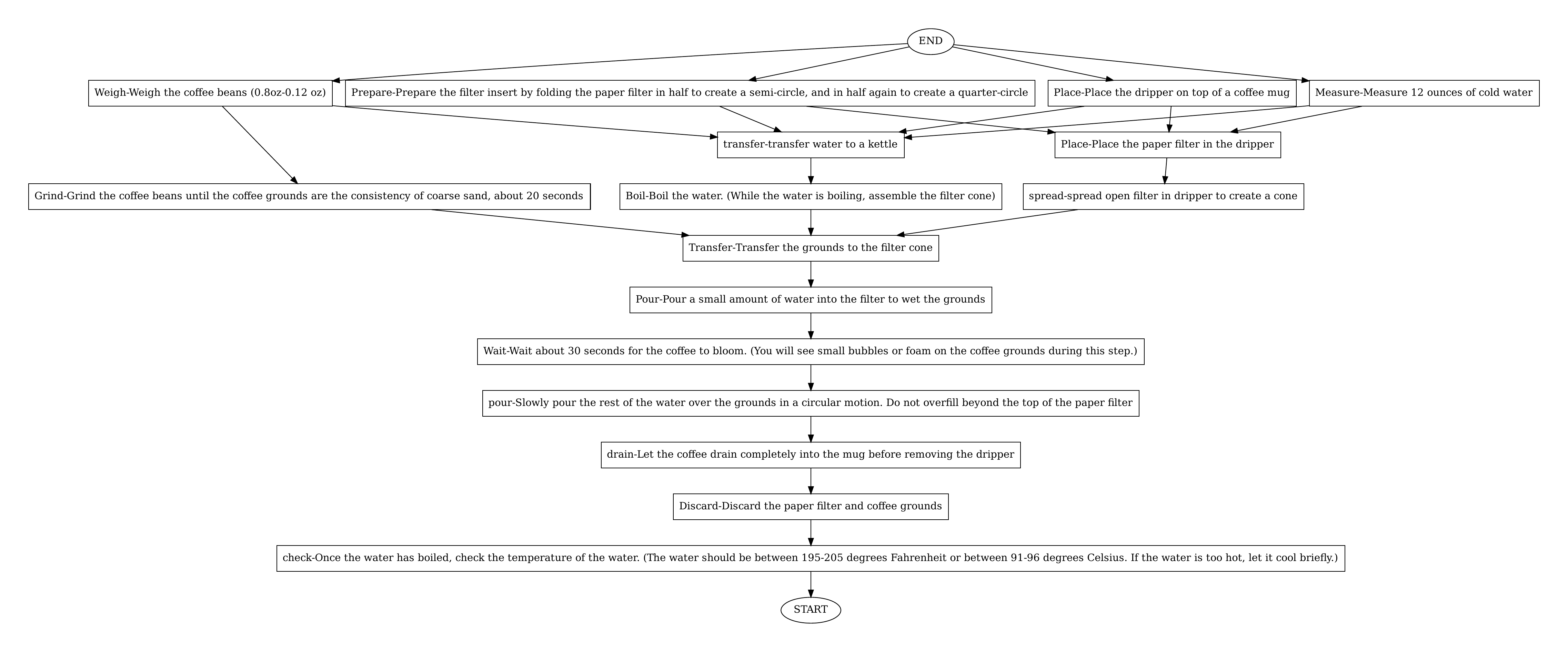}
    }
    \subfloat[Llama-3.1-405B-instruct]{%
        \includegraphics[width=0.45\textwidth]{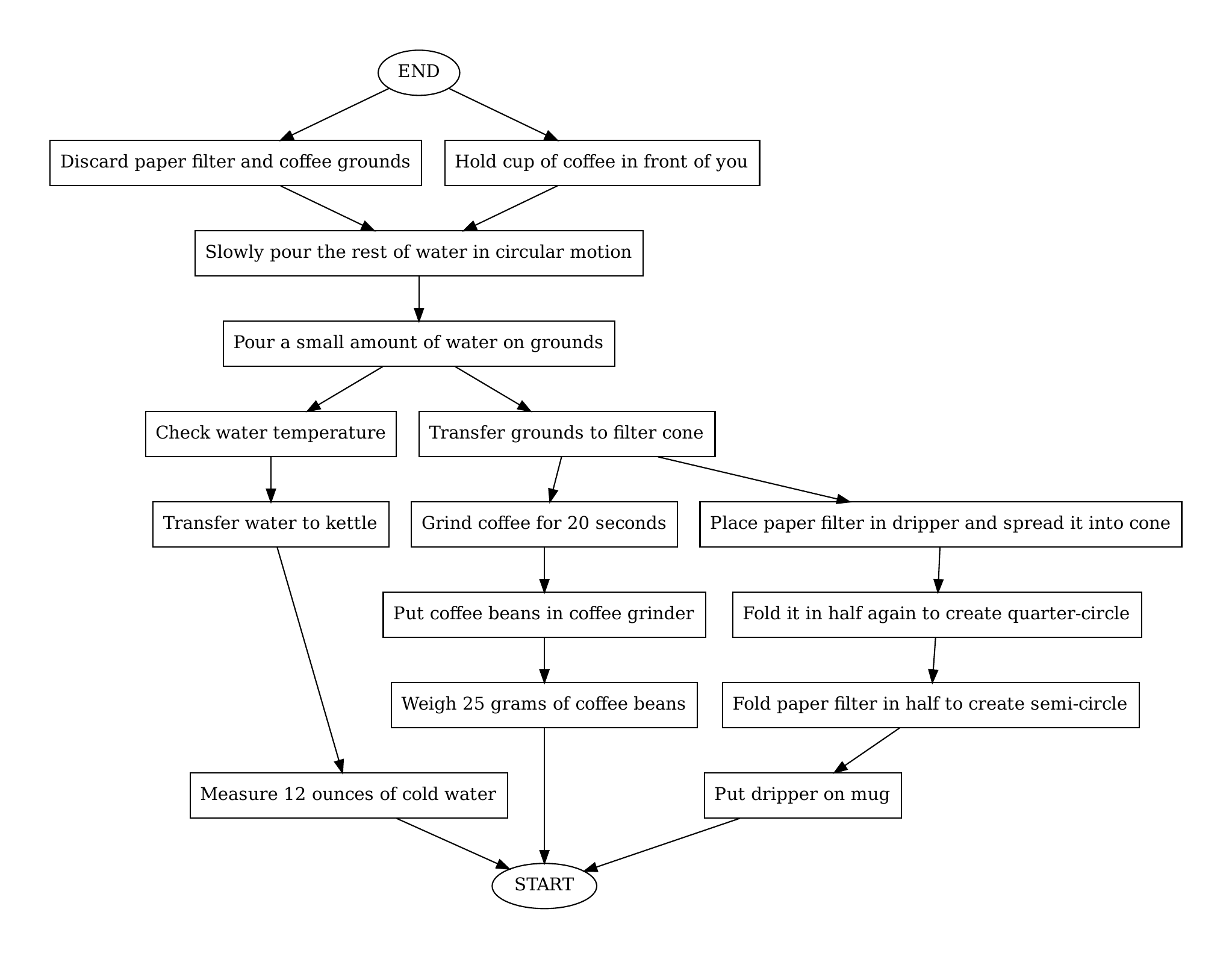}
    }\\
    \subfloat[Count-Based]{%
        \includegraphics[width=0.45\textwidth]{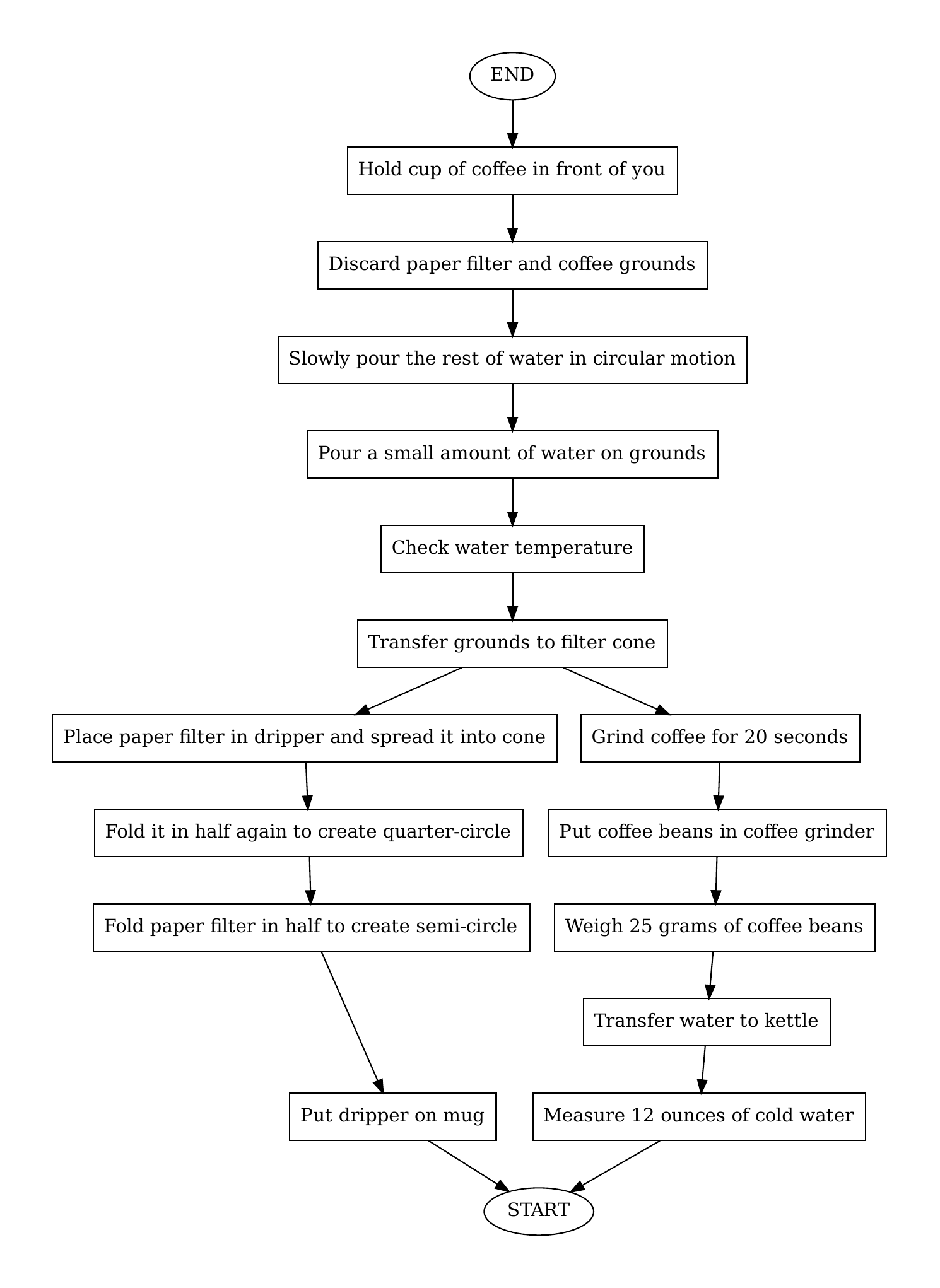}
    }
    \subfloat[MSG$^2$]{%
        \includegraphics[width=0.45\textwidth]{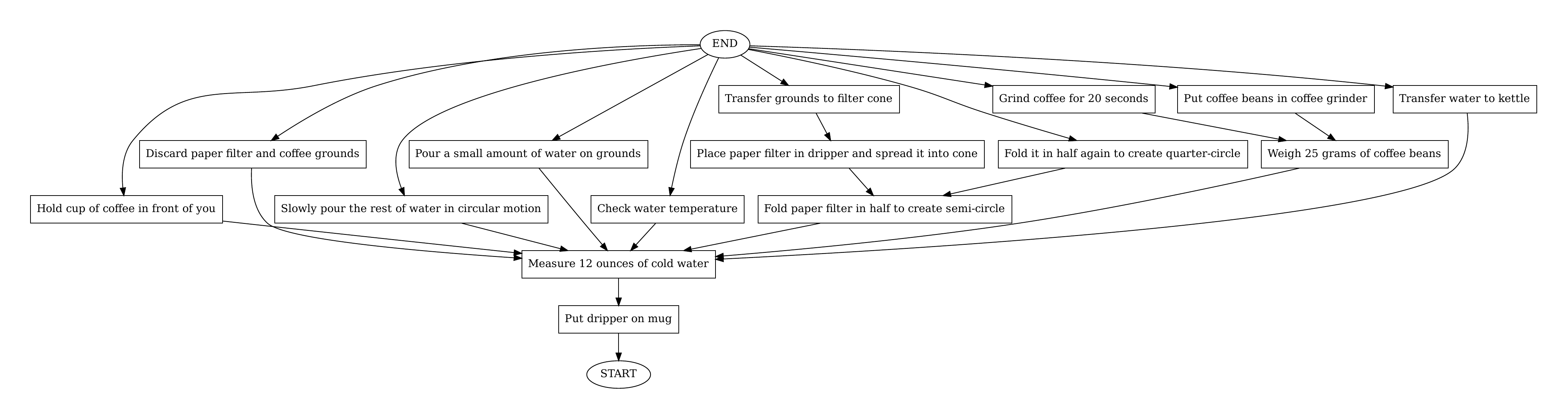}
    }
    \caption{Task graphs of the scenario ``Coffee'' from EgoPER generated with (a) MSGI, (b) Llama-3.1-405B-Instruct, (c) Count-Based, (d) MSG$^2$.}
\end{figure*}

\begin{figure*}\ContinuedFloat
    \centering
    \subfloat[TGT-text]{%
        \includegraphics[width=0.45\textwidth]{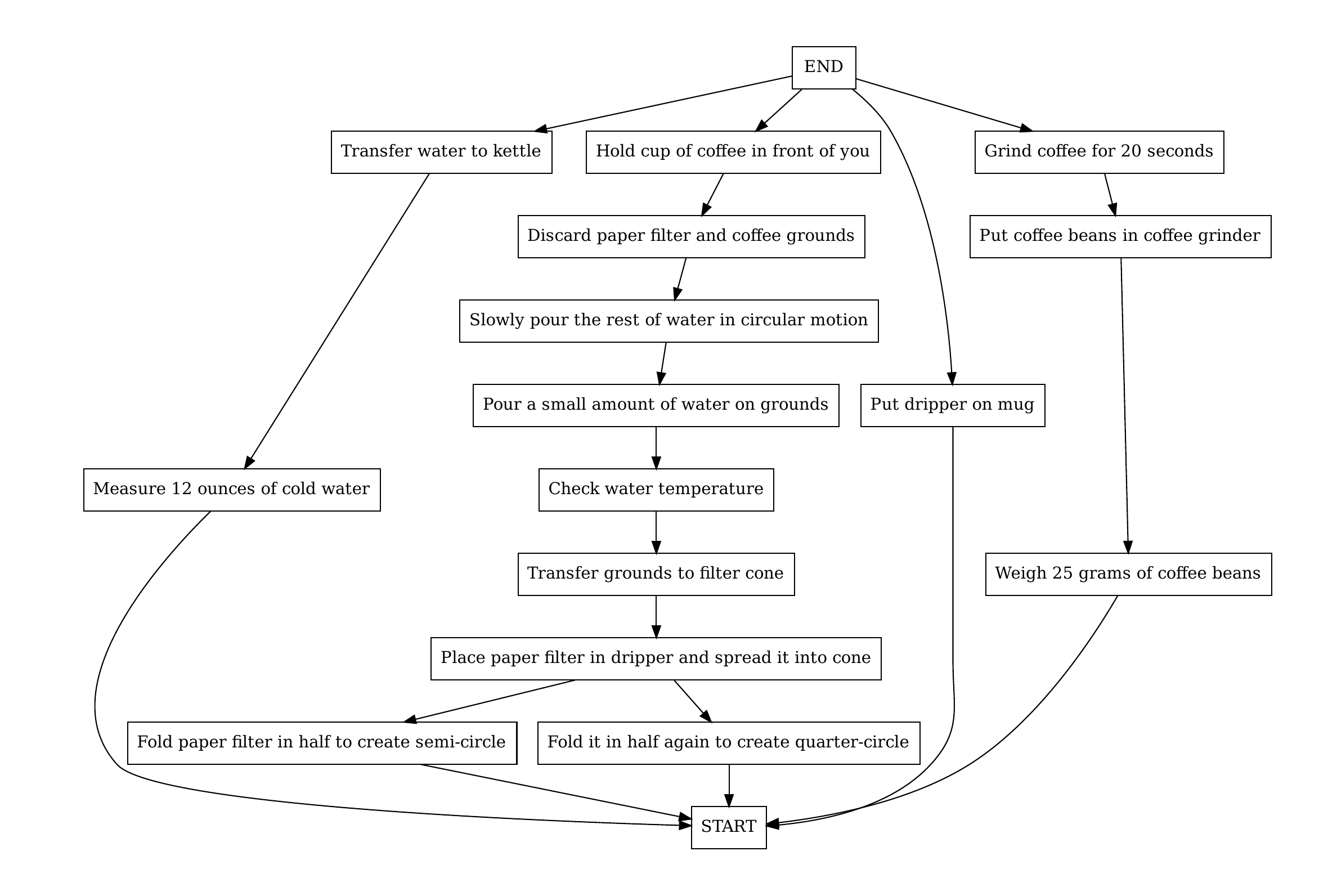}
    }
    \subfloat[DO]{%
        \includegraphics[width=0.45\textwidth]{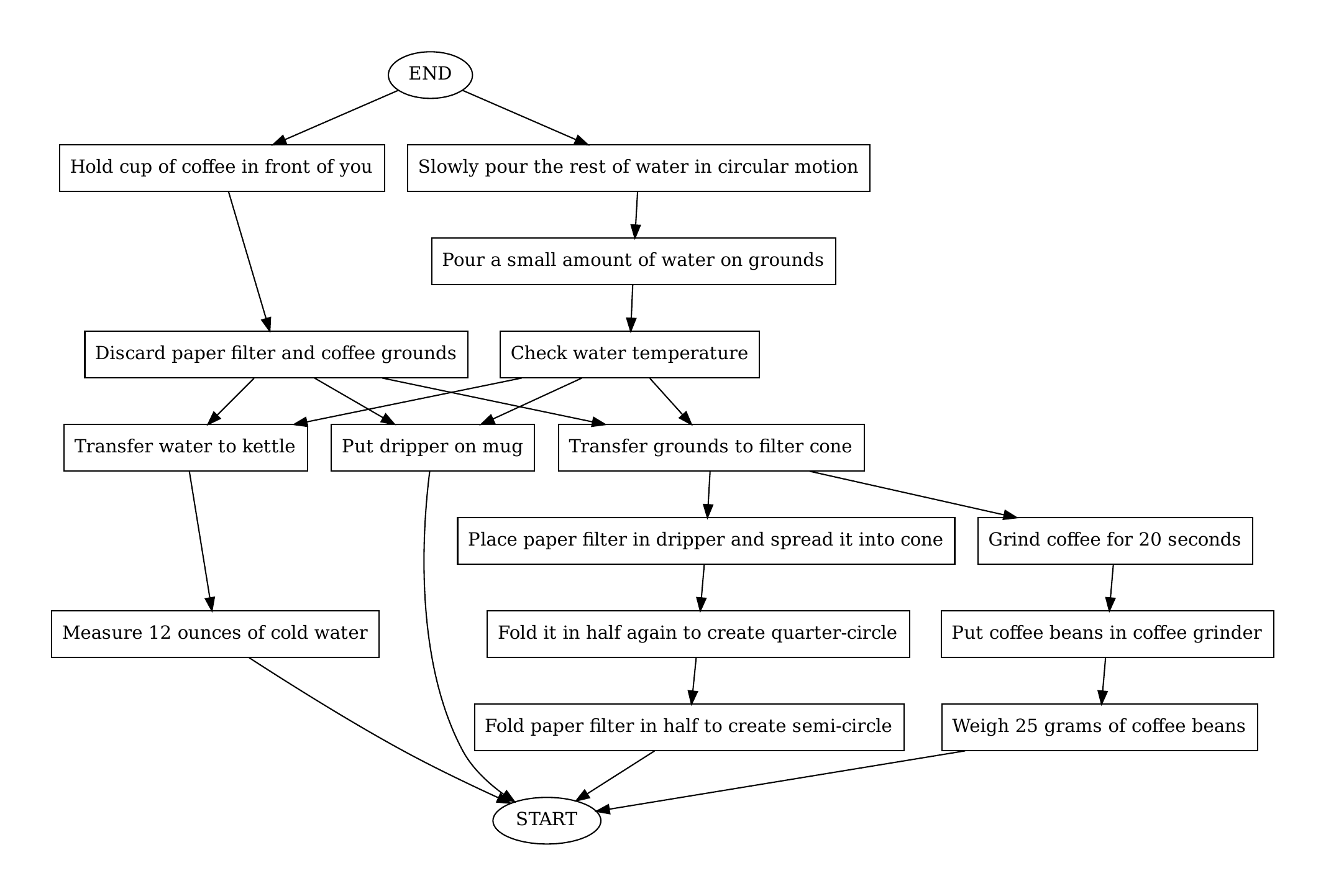}
    }\\
    \subfloat[GT]{%
        \includegraphics[width=0.45\textwidth]{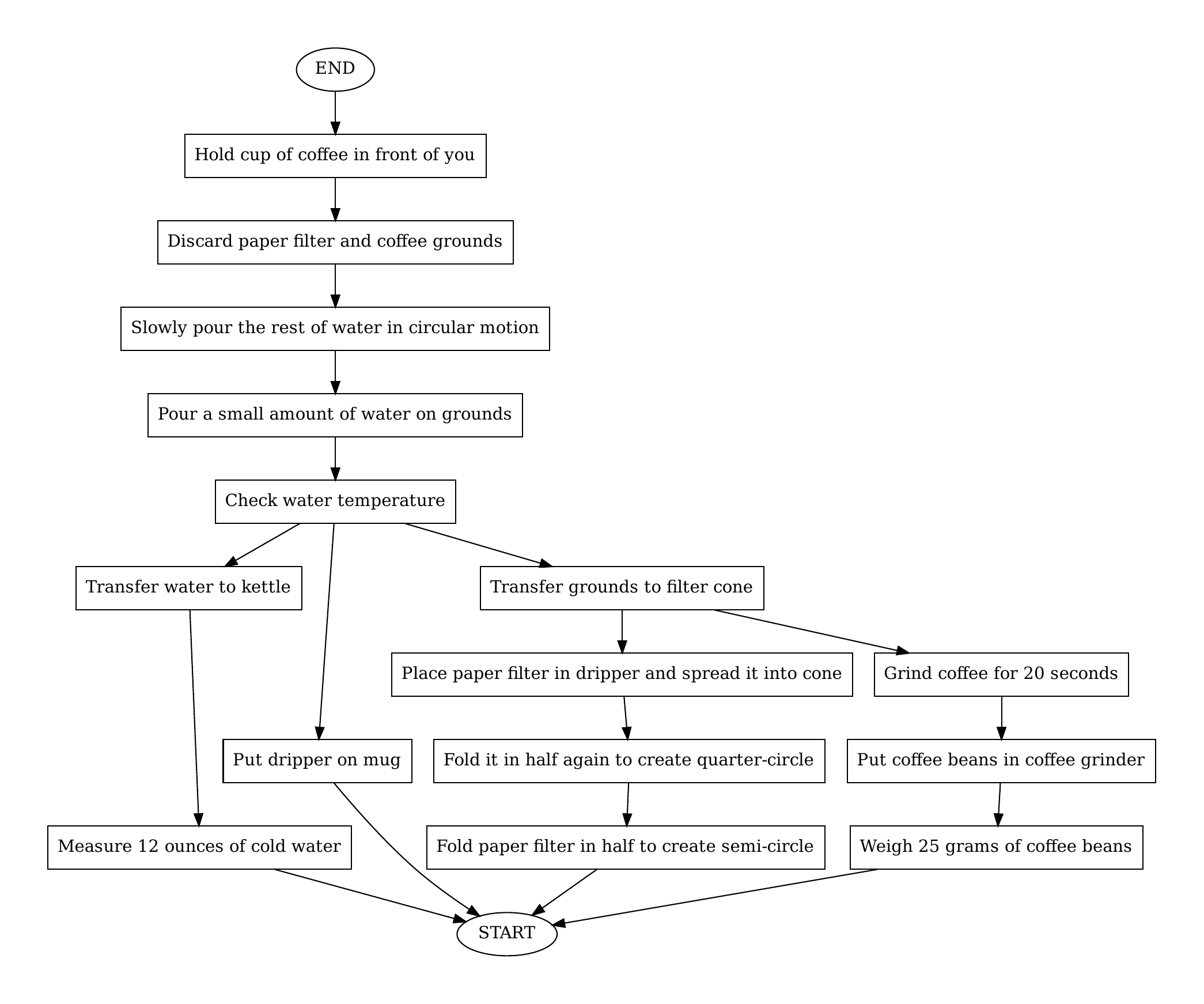}
    }
    \caption{Task graphs of the scenario ``Coffee'' from EgoPER generated with (e) TGT using textual embedding, and (f) DO. (g) reports the ground truth.}
    \label{fig:comparison_egoper}
\end{figure*}

\begin{figure*}
    \centering
    \subfloat[MSGI]{
        \includegraphics[width=0.45\textwidth]{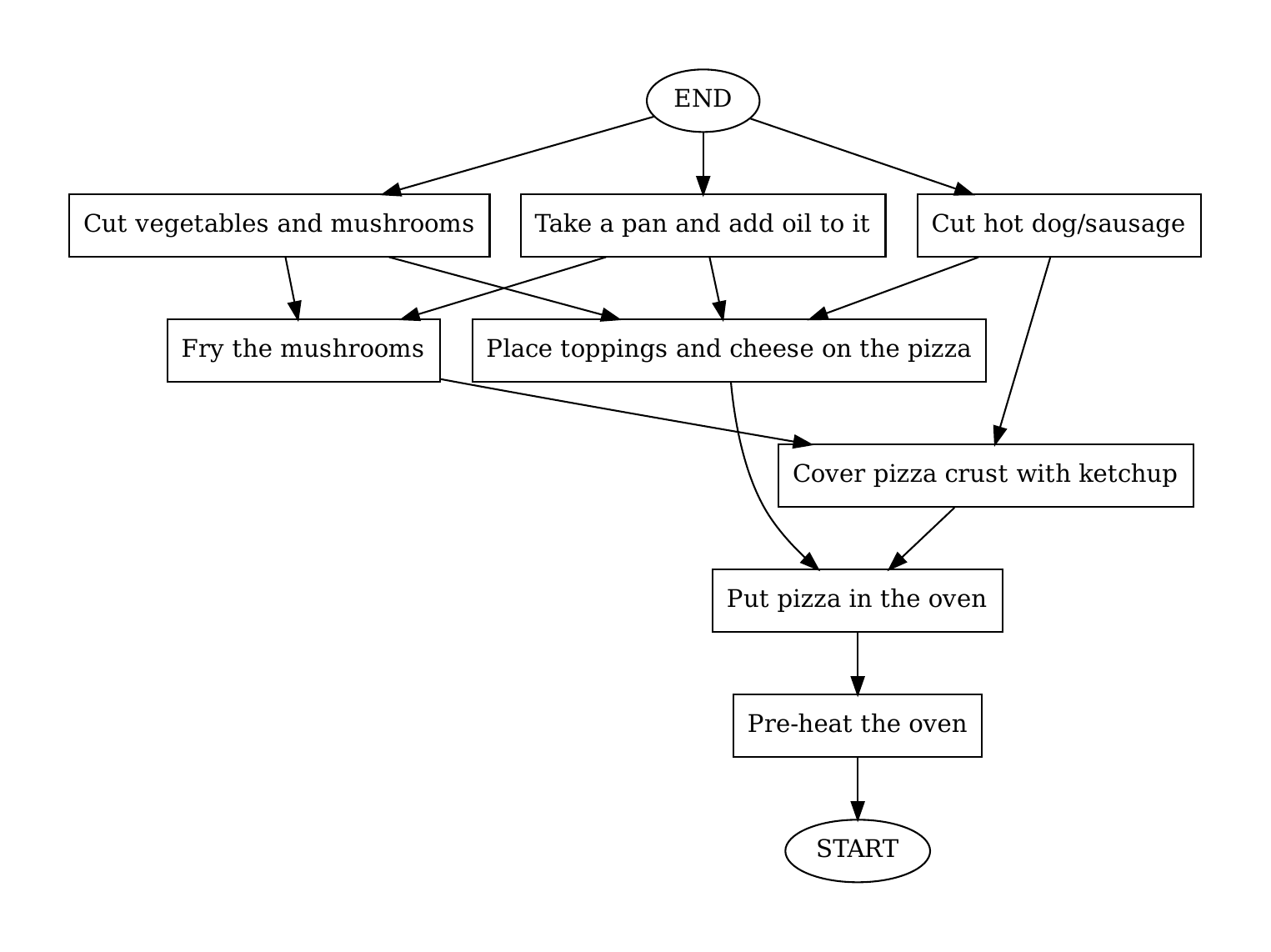}
    }
    \subfloat[Llama-3.1-405B-instruct]{
        \includegraphics[width=0.45\textwidth]{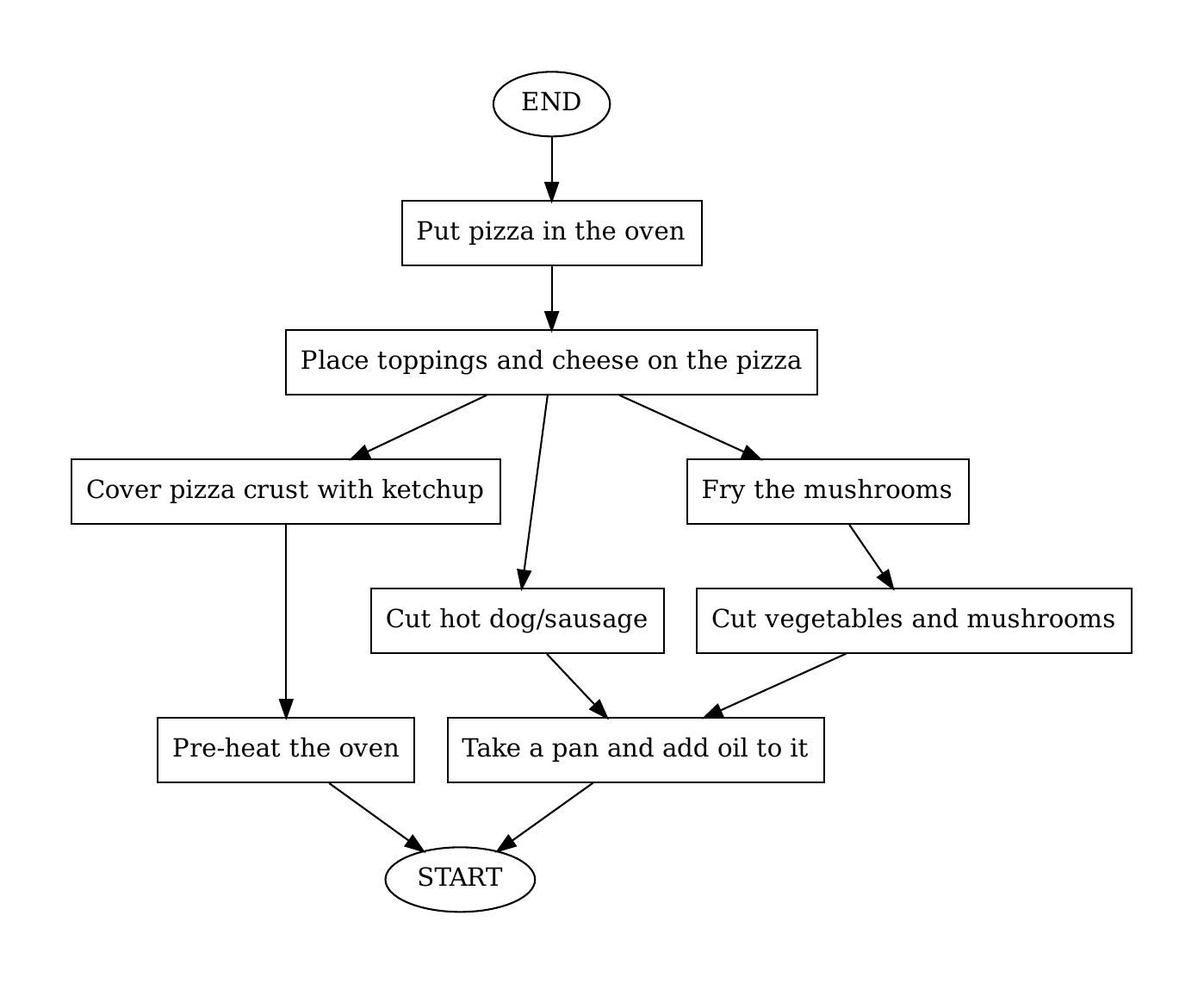}
    }\\
    \subfloat[Count-Based]{
        \includegraphics[width=0.45\textwidth]{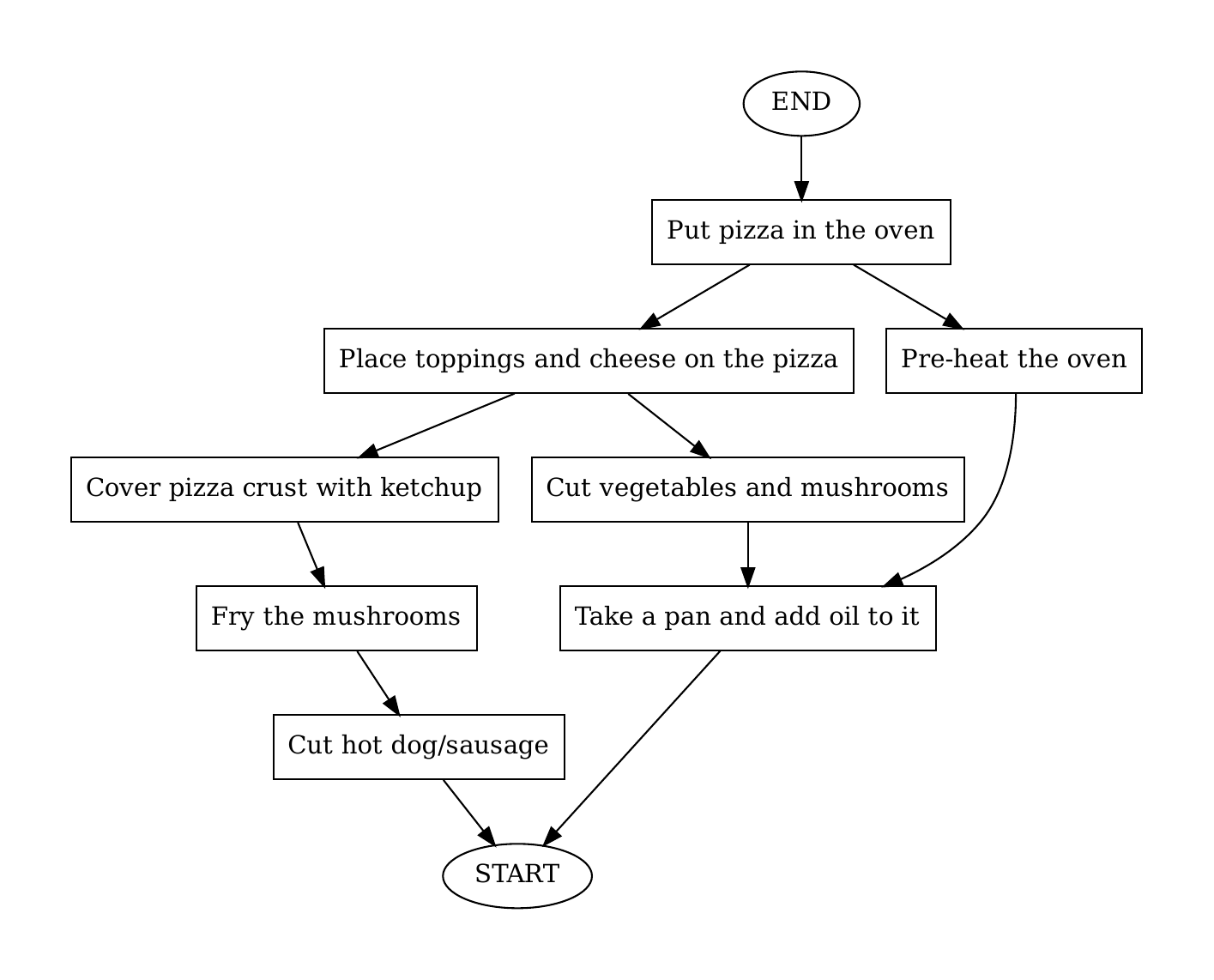}
    }
    \subfloat[MSG$^2$]{
        \includegraphics[width=0.45\textwidth]{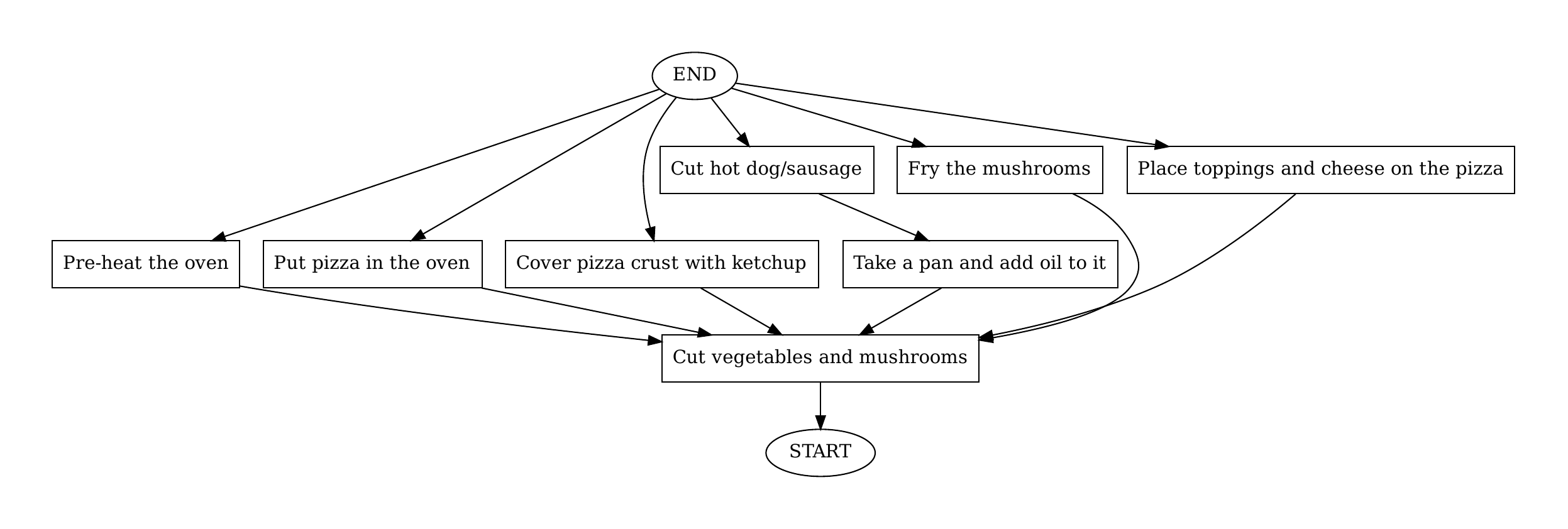}
    }
    \caption{Task graphs of the scenario ``EGTEA-Gaze+ Pizza'' from EgoProceL generated with (a) MSGI, (b) Llama-3.1-405B-Instruct, (c) Count-Based, (d) MSG$^2$.}
\end{figure*}

\begin{figure*}\ContinuedFloat
    \centering
    \subfloat[TGT-text]{
        \includegraphics[width=0.45\textwidth]{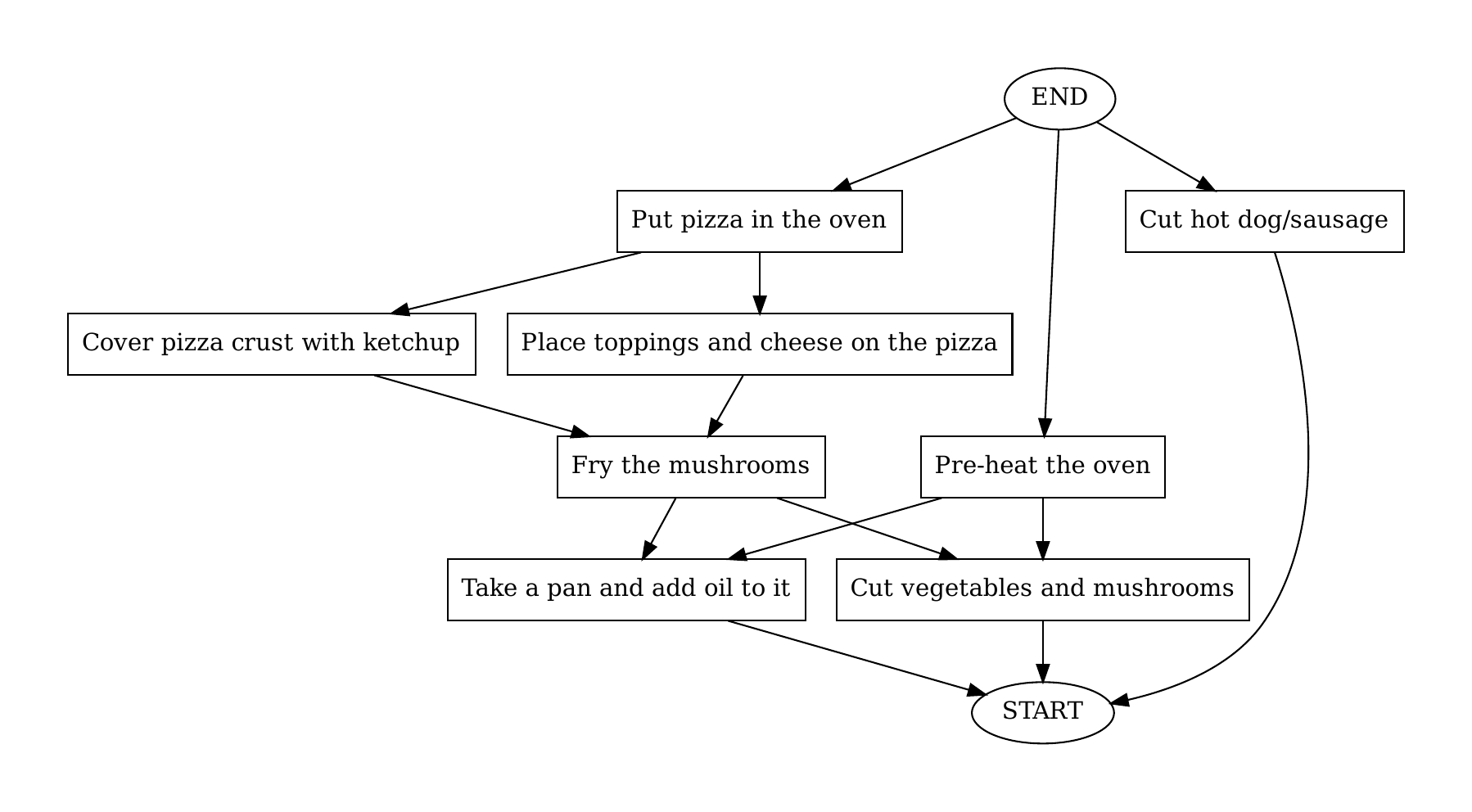}
    }
    \subfloat[DO]{
        \includegraphics[width=0.45\textwidth]{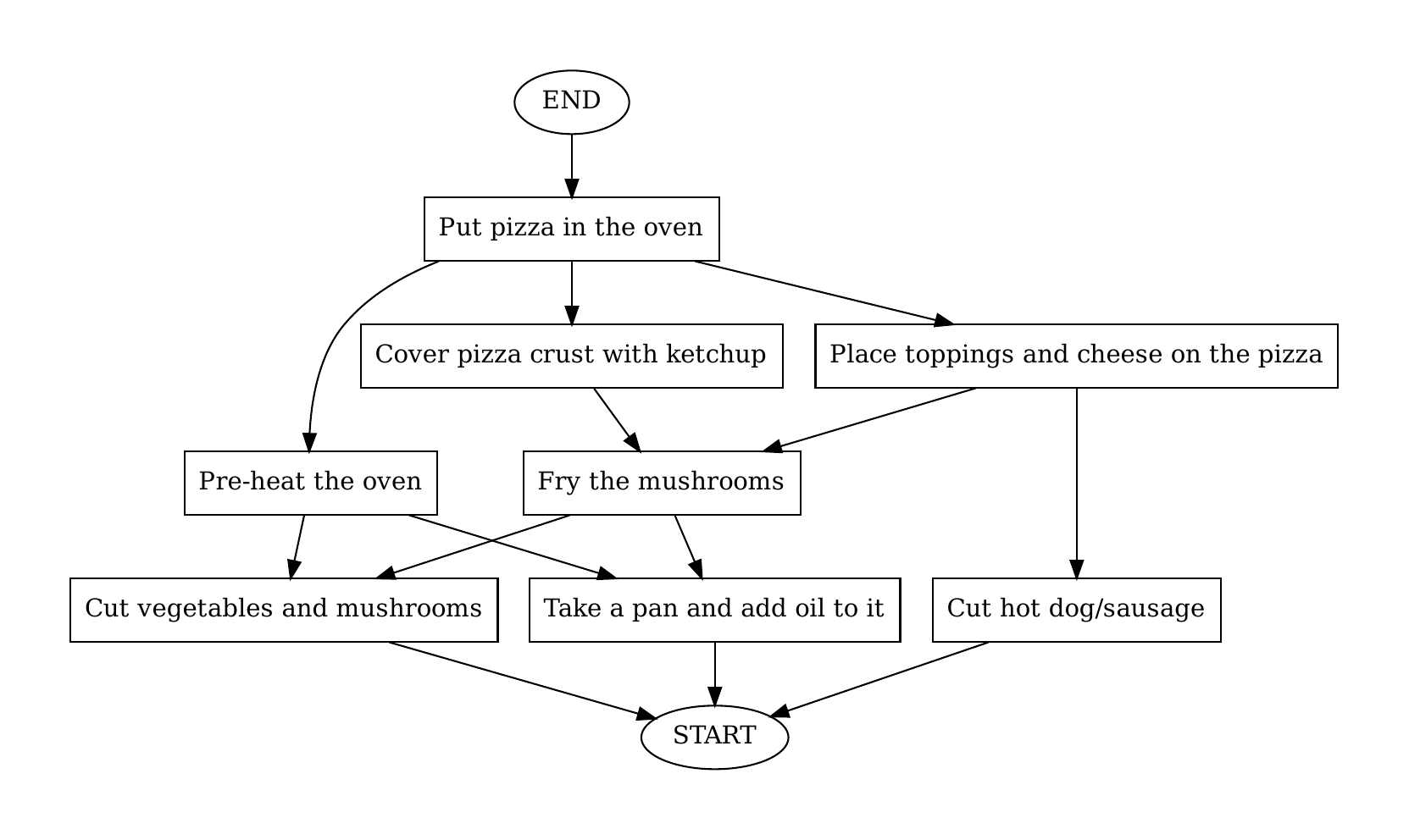}
    }\\
    \subfloat[GT]{
        \includegraphics[width=0.45\textwidth]{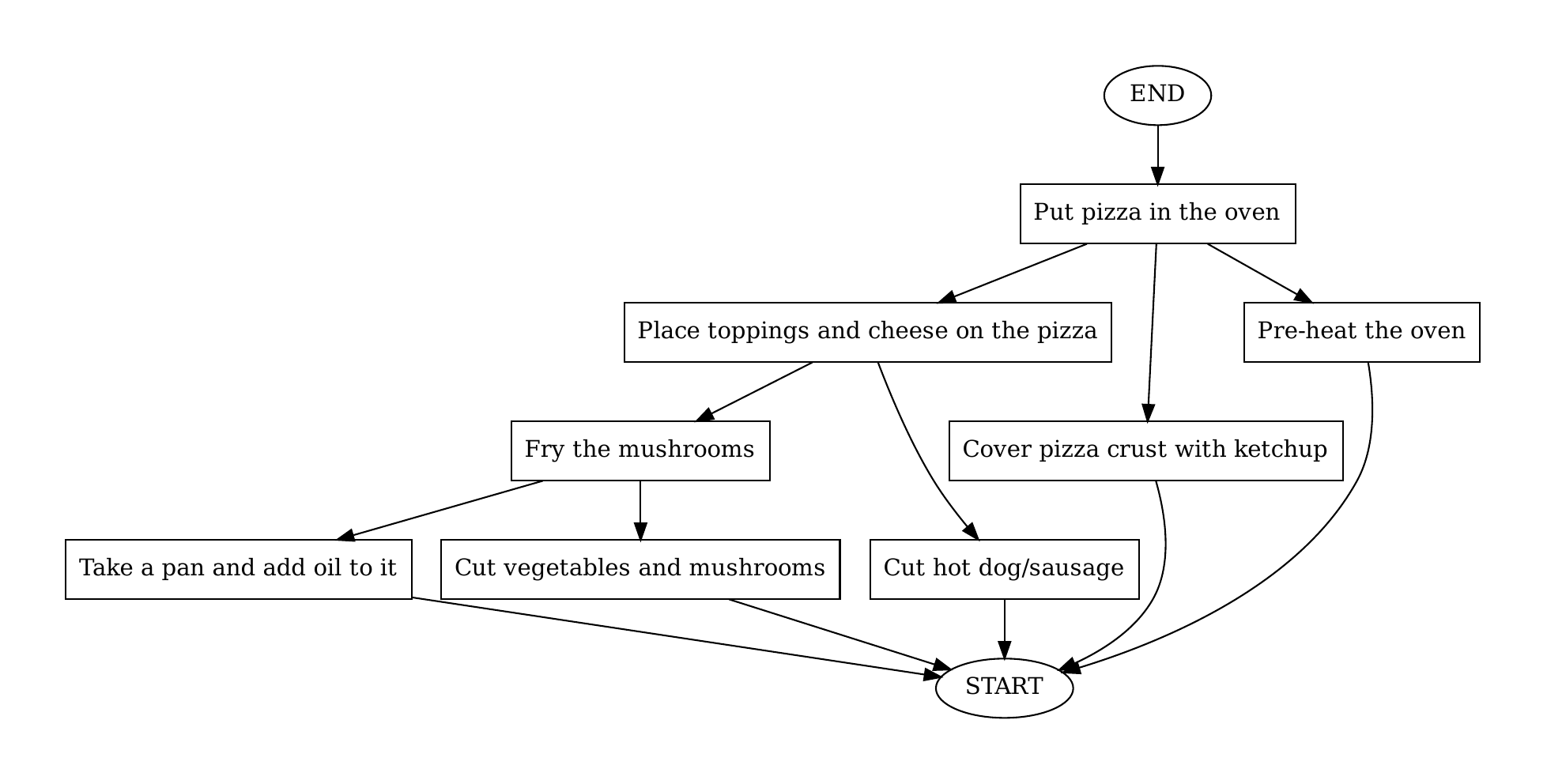}
    }
    \caption{Task graphs of the scenario ``EGTEA-Gaze+ Pizza'' from EgoProceL generated with (e) TGT using textual embedding, and (f) DO. (g) reports the ground truth.}
    \label{fig:comparison_egoprocel}
\end{figure*}

\begin{figure*}[t]
    \centering
    \subfloat[]{
        \includegraphics[width=0.45\textwidth]{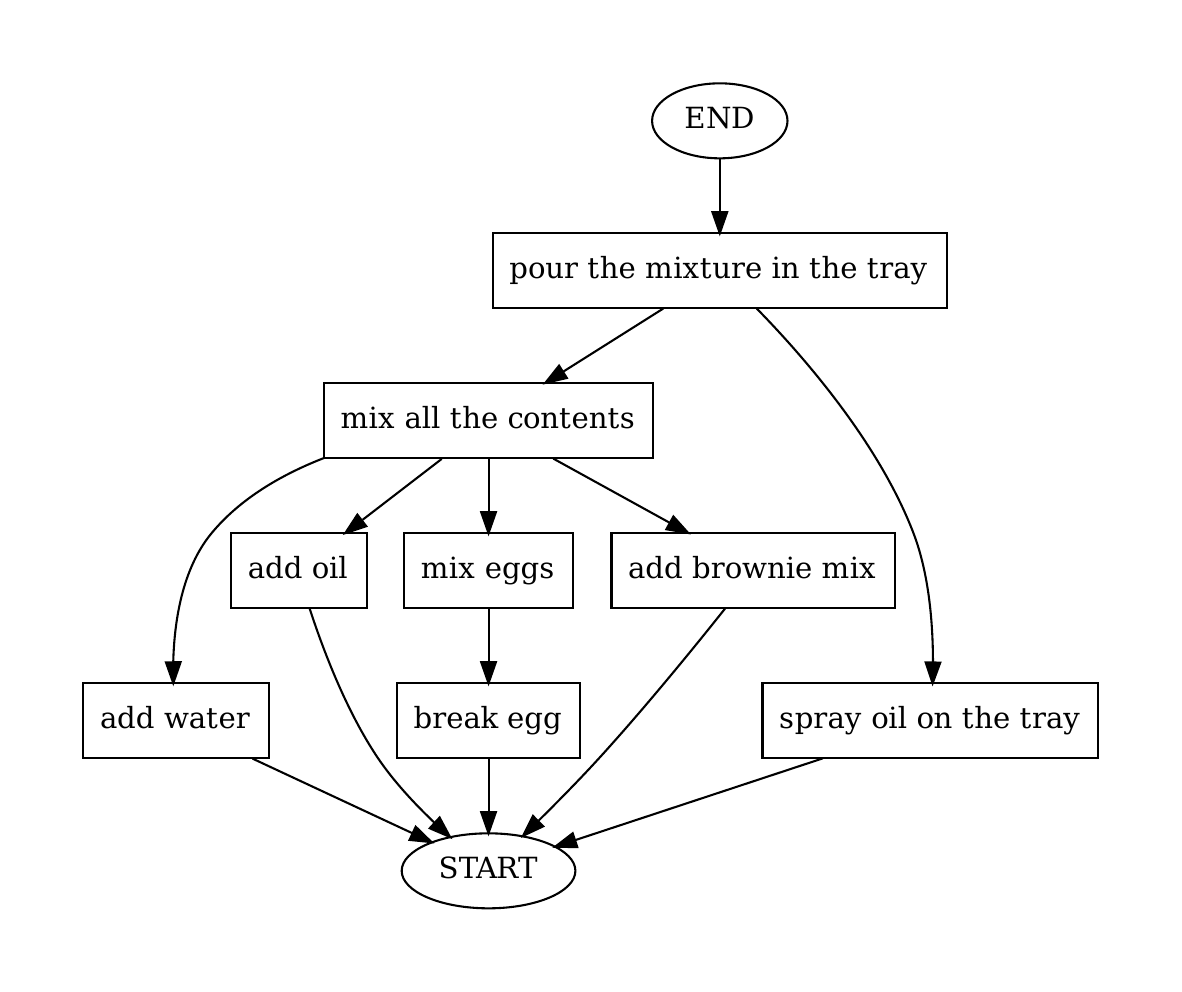}
    }
    \hfill
    \subfloat[]{
        \includegraphics[width=0.45\textwidth]{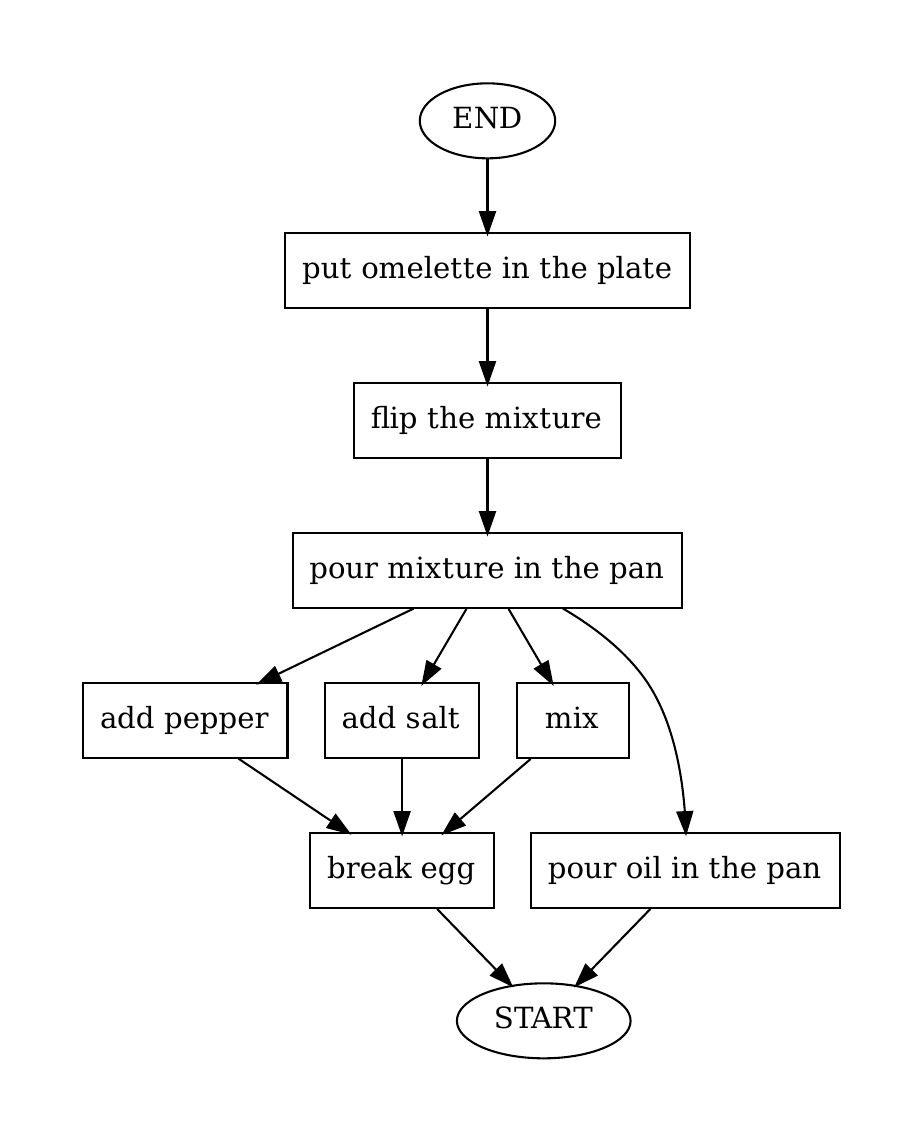}
    }
    \caption{Annotated (a) ``CMU-MMAC Brownie'' and (b) ``CMU-MMAC Eggs'' task graphs from EgoProceL.}
    \label{fig:annotated_start}
\end{figure*}

\begin{figure*}[t]
    \centering
    \subfloat[]{
        \includegraphics[width=0.45\textwidth]{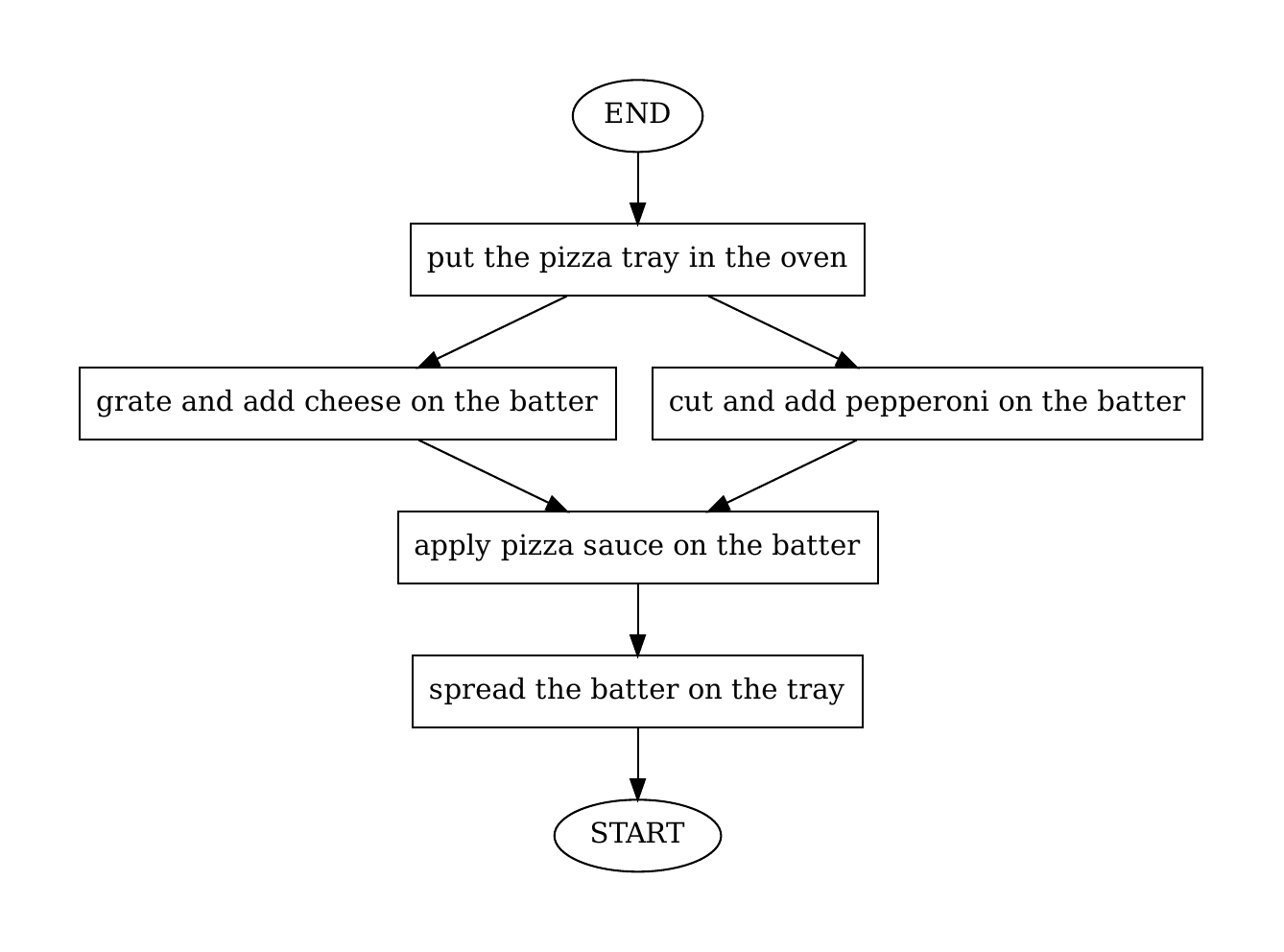}
    }
    \hfill
    \subfloat[]{
        \includegraphics[width=0.45\textwidth]{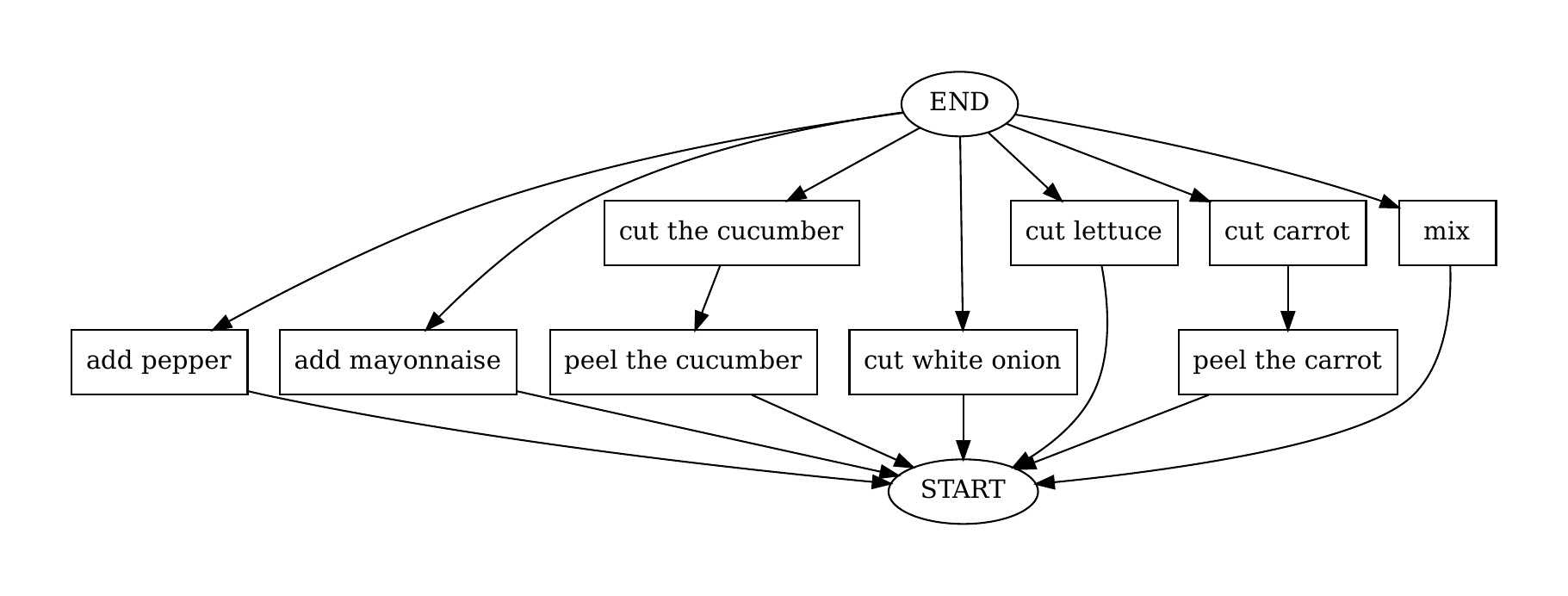}
    }
    \caption{Annotated (a) ``CMU-MMAC Pizza'' and (b) ``CMU-MMAC Salad'' task graphs from EgoProceL.}
\end{figure*}

\begin{figure*}[t]
    \centering
    \subfloat[]{
        \includegraphics[width=0.45\textwidth]{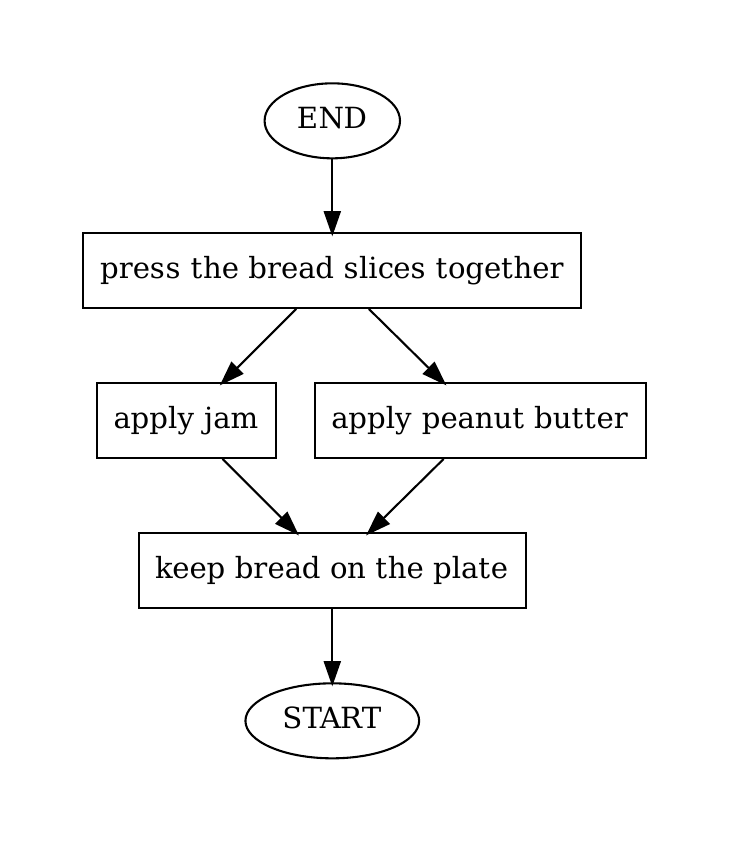}
    }
    \hfill
    \subfloat[]{
        \includegraphics[width=0.45\textwidth]{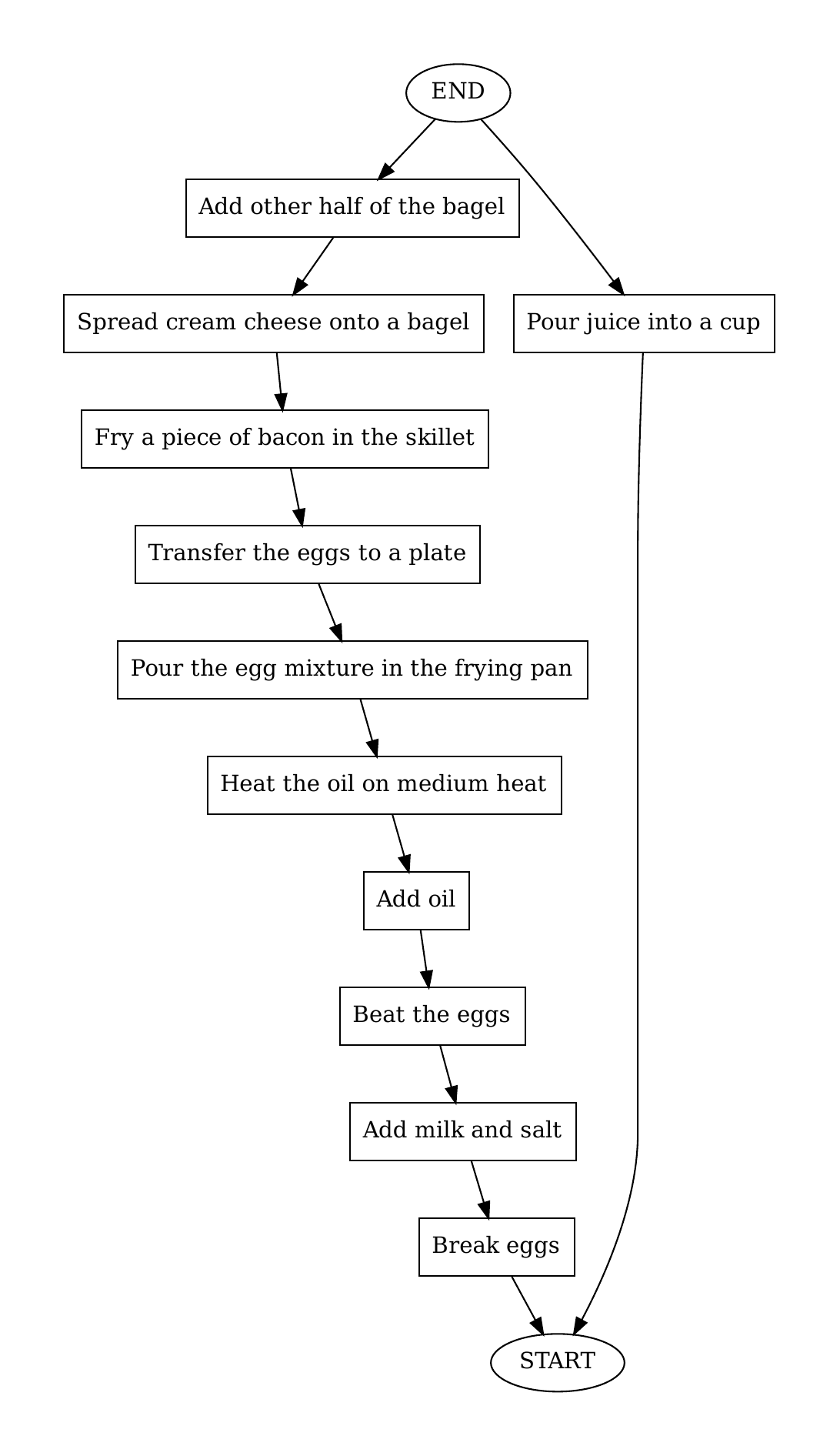}
    }
    \caption{Annotated (a) ``CMU-MMAC Sandwich'' and (b) ``EGTEA-Gaze+ Bacon and Eggs'' task graphs from EgoProceL.}
\end{figure*}

\begin{figure*}[t]
    \centering
    \subfloat[]{
        \includegraphics[width=0.45\textwidth]{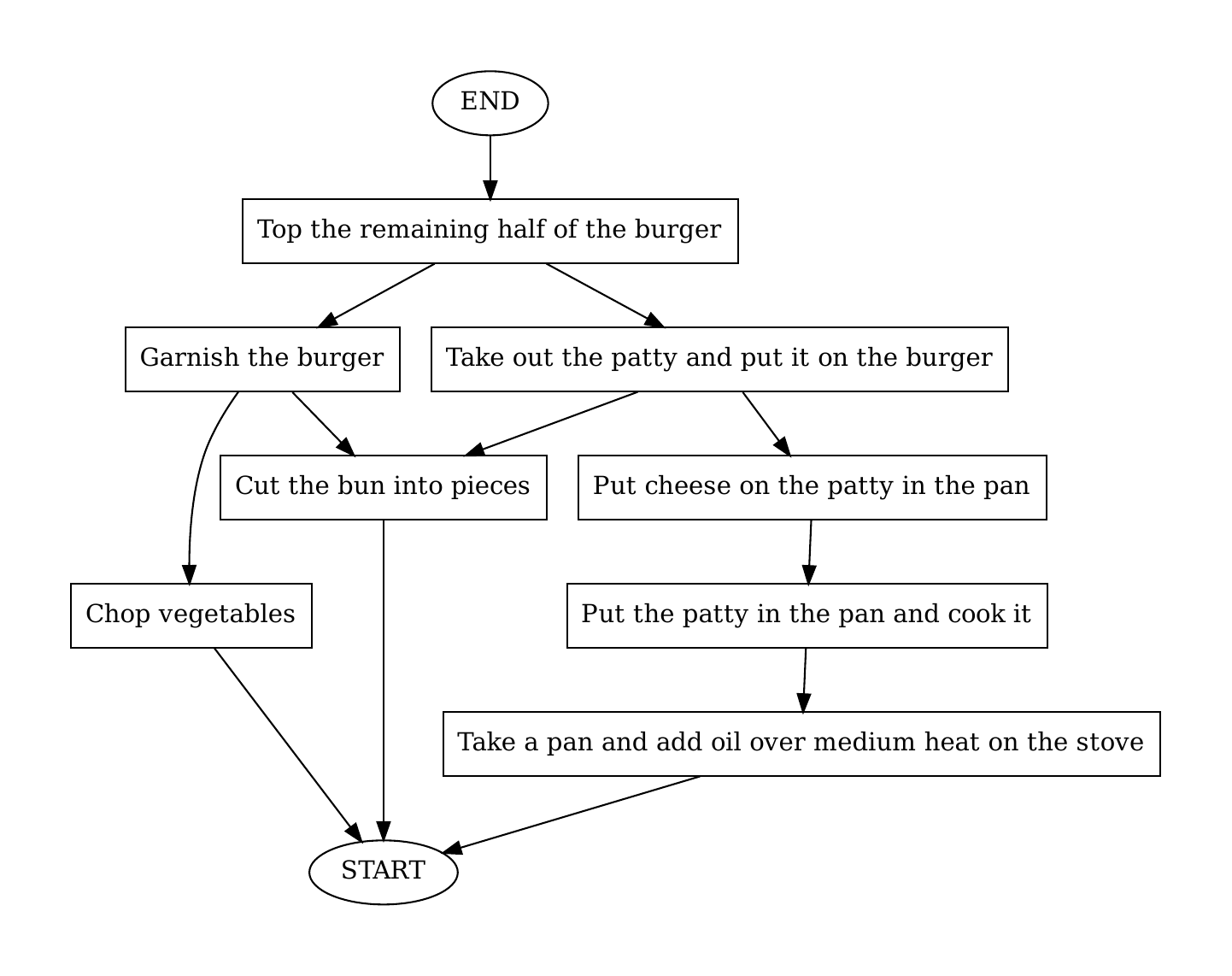}
    }
    \hfill
    \subfloat[]{
        \includegraphics[width=0.45\textwidth]{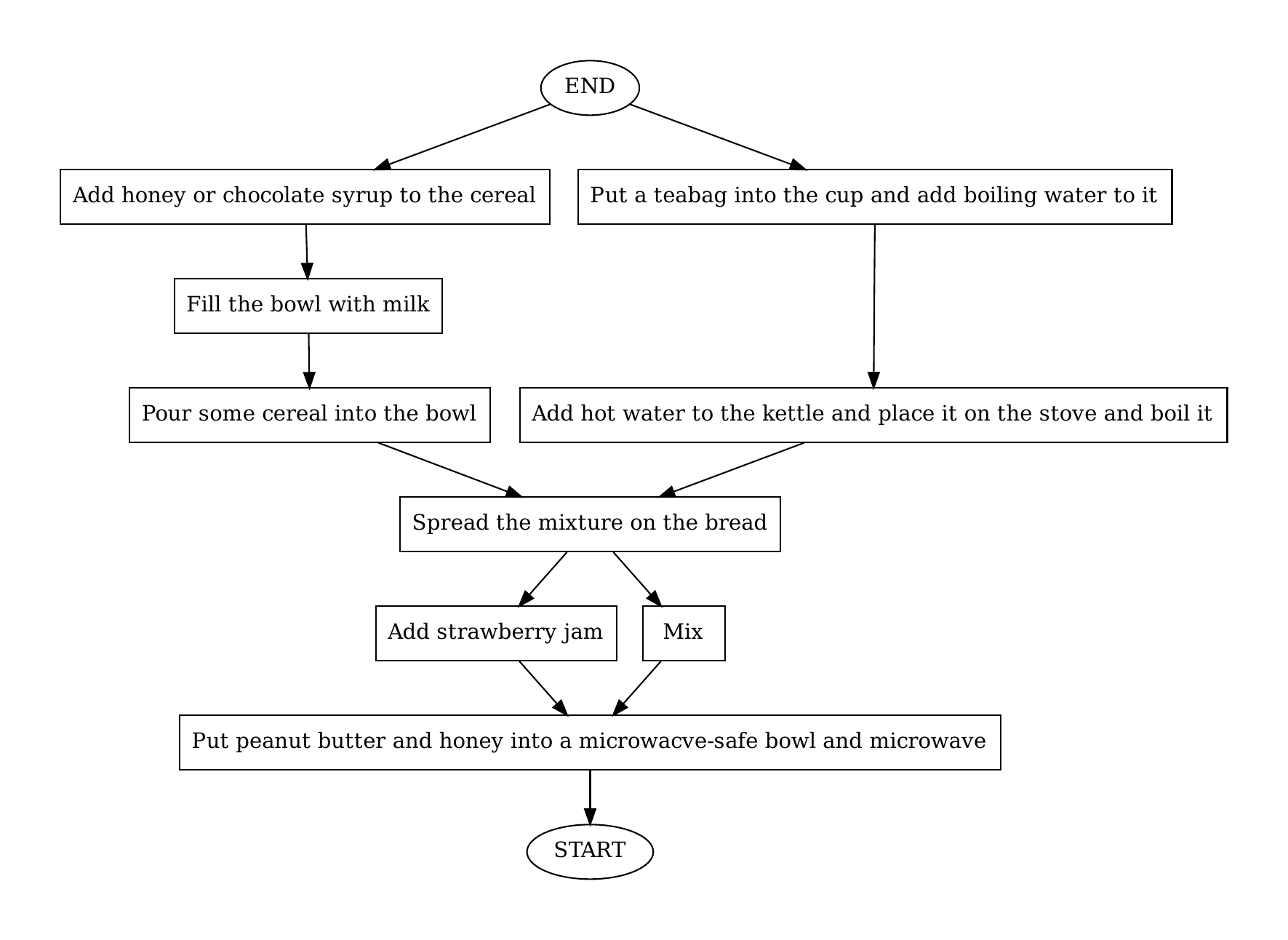}
    }
    \caption{Annotated (a) ``EGTEA-Gaze+ Cheeseburger'' and (b) ``EGTEA-Gaze+ Continental Breakfast'' task graphs from EgoProceL.}
\end{figure*}

\begin{figure*}[t]
    \centering
    \subfloat[]{
        \includegraphics[width=0.45\textwidth]{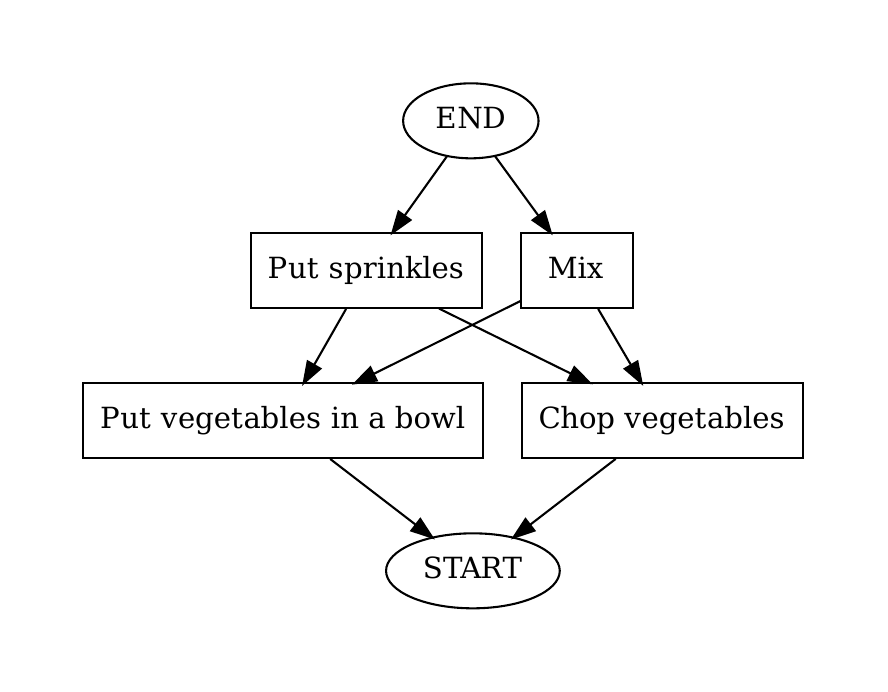}
    }
    \hfill
    \subfloat[]{
        \includegraphics[width=0.45\textwidth]{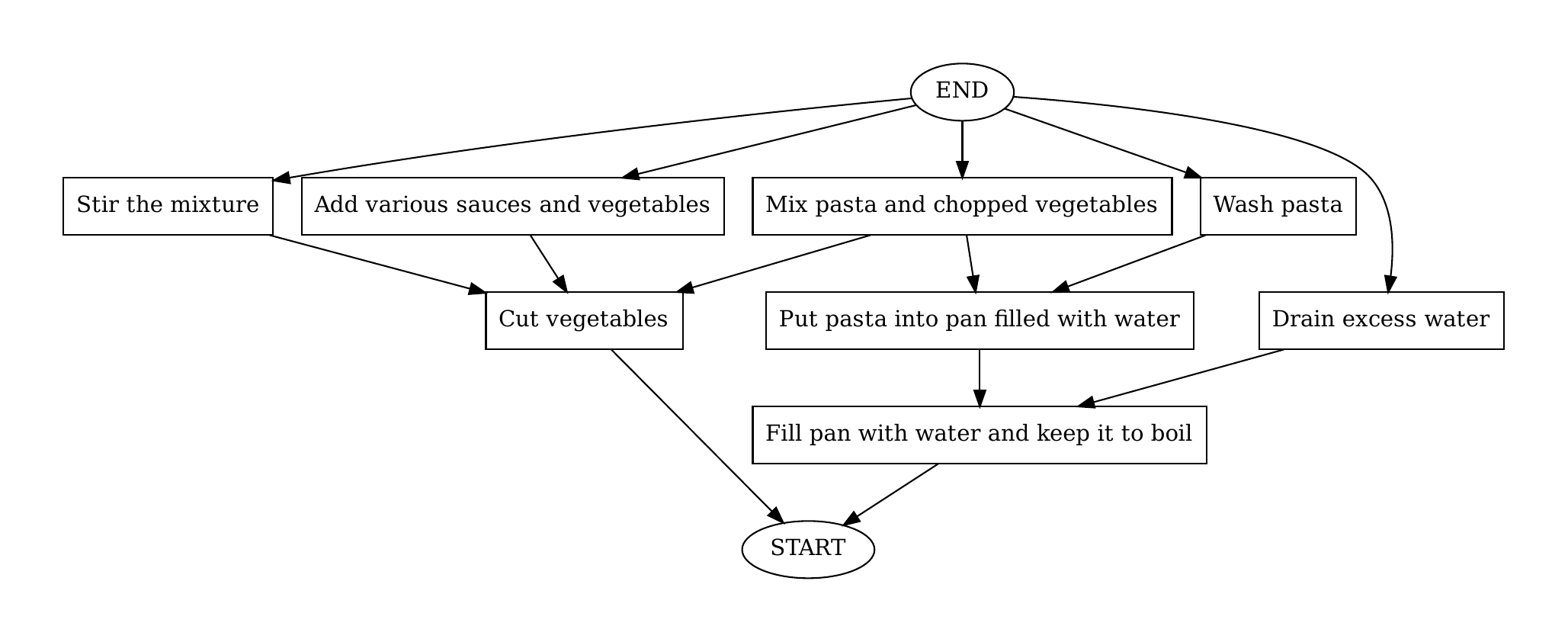}
    }
    \caption{Annotated (a) ``EGTEA-Gaze+ Greek Salad'' and (b) ``EGTEA-Gaze+ Pasta Salad'' task graphs from EgoProceL.}
\end{figure*}

\begin{figure*}[t]
    \centering
    \subfloat[]{
        \includegraphics[width=0.45\textwidth]{Qualitative/EgoProceL/GT_graph_egtea_gaze+_pizza.pdf}
    }
    \hfill
    \subfloat[]{
        \includegraphics[width=0.45\textwidth]{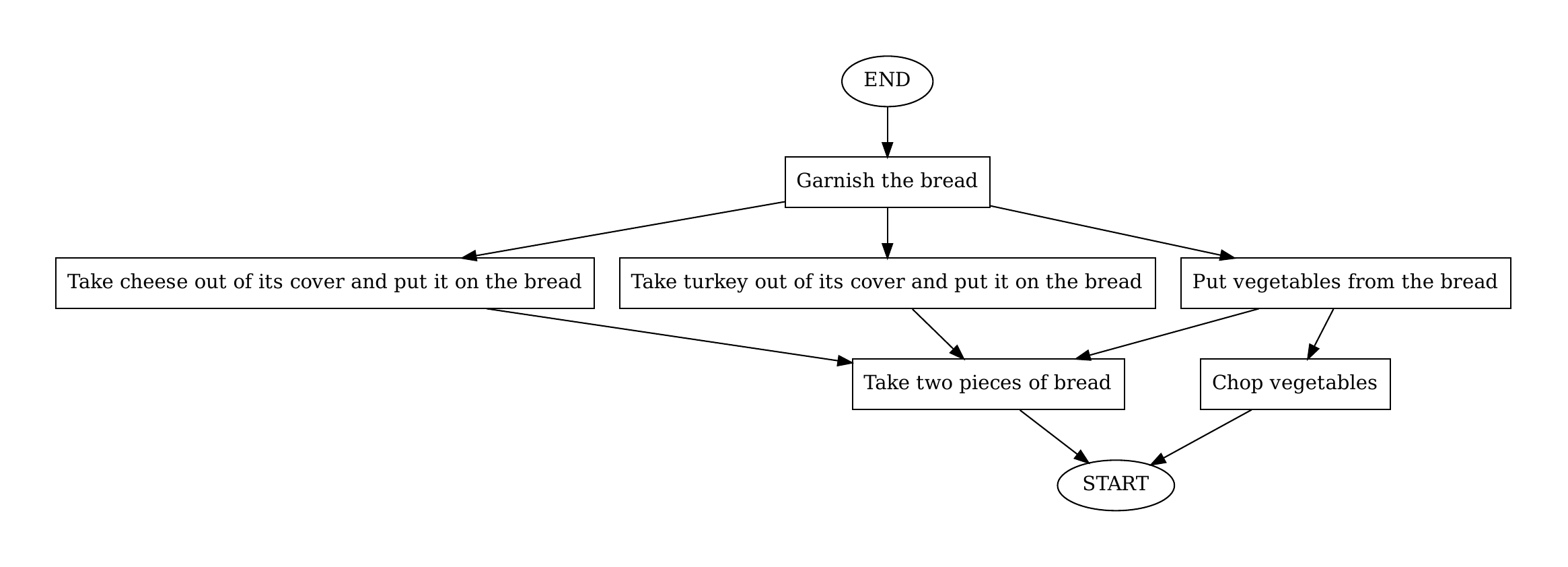}
    }
    \caption{Annotated (a) ``EGTEA-Gaze+ Pizza'' and (b) ``EGTEA-Gaze+ Turkey Sandwich'' task graphs from EgoProceL.}
\end{figure*}

\begin{figure*}[t]
    \centering
    \includegraphics[width=\textwidth]{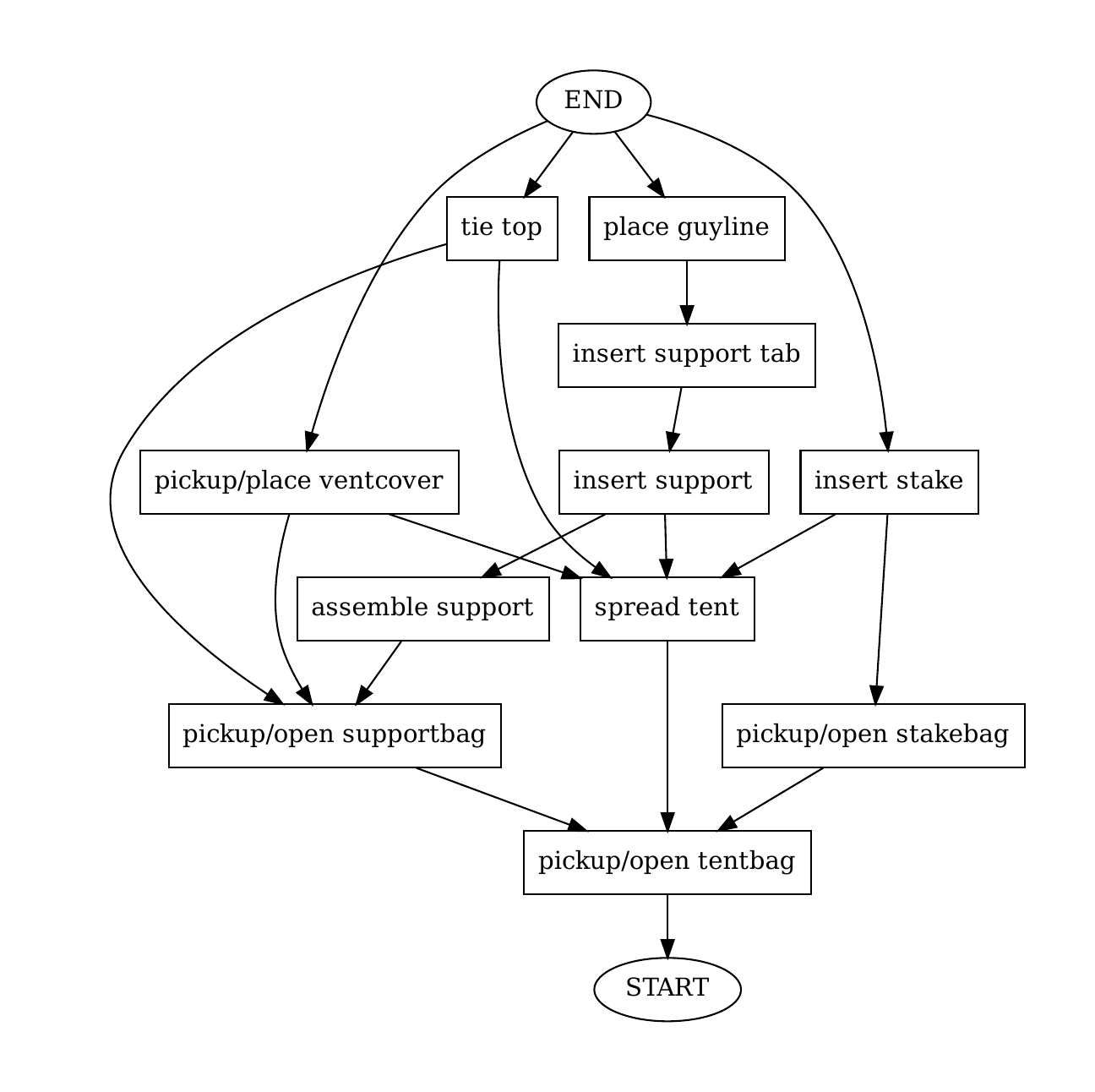}
    \caption{Annotated ``EPIC-Tents'' task graphs from EgoProceL.}
    \label{fig:annotated_end}
\end{figure*}

\end{document}